\documentclass[10pt]{article}

\newcommand{\arxiv}[1]{#1}
\newcommand{\iclr}[1]{}
\newcommand{\loose}{\looseness=-1}

\newif\ifparttwo
\parttwofalse
\newcommand{\parttwo}[1]{\ifparttwo#1\fi}

\usepackage[noend]{algpseudocode}

\PassOptionsToPackage{hypertexnames=false}{hyperref}  %
\arxiv{
\usepackage[letterpaper, left=.95in, right=.95in, top=.95in,bottom=.95in]{geometry}
  \usepackage{parskip}
  \usepackage[colorlinks=true, linkcolor=blue!70!black, citecolor=blue!70!black,urlcolor=black,breaklinks=true]{hyperref}
  \usepackage[dvipsnames]{xcolor}
}

\iclr{
  \usepackage{parskip}
    \usepackage[colorlinks=true, linkcolor=blue!70!black, citecolor=blue!70!black,urlcolor=black,breaklinks=true]{hyperref}
}

\usepackage{amsmath}
\usepackage{microtype}
\usepackage{hhline}

\usepackage{amsthm}
\usepackage{mathtools}
\usepackage{bbm}
\usepackage{amsfonts}
\usepackage{amssymb}
\usepackage[nameinlink,capitalize]{cleveref}

\usepackage{xargs}

\makeatletter
\newcommand{\neutralize}[1]{\expandafter\let\csname c@#1\endcsname\count@}
\makeatother

\usepackage{algorithm}

\arxiv{
\usepackage{natbib}
\bibliographystyle{plainnat}
\bibpunct{(}{)}{;}{a}{,}{,}
}

\usepackage{xpatch}

\usepackage{thm-restate}
\usepackage{autonum}
\declaretheorem[name=Theorem,parent=section]{theorem}
\declaretheorem[name=Lemma,parent=section]{lemma}
\declaretheorem[name=Assumption, parent=section]{assumption}
\declaretheorem[name=Definition, parent=section]{definition}
\declaretheorem[name=Condition, parent=section]{condition}
\declaretheorem[name=Corollary, parent=section]{corollary}

\declaretheorem[name=Remark, parent=section]{remark}
\declaretheorem[name=Proposition, parent=section]{proposition}

\newcommand{\pfref}[1]{Proof of \cref{#1}}

\renewcommand{\eqref}[1]{\texorpdfstring{\hyperref[#1]{(\ref*{#1})}}{(\ref*{#1})}}
\crefformat{equation}{#2Eq.\,(#1)#3}
\Crefformat{equation}{#2Eq.\,(#1)#3}
\Crefformat{figure}{#2Figure~#1#3}
\Crefformat{assumption}{#2Assumption~#1#3}
\Crefname{assumption}{Assumption}{Assumptions}
\crefname{fact}{Fact}{Facts}
\Crefformat{figure}{#2Figure #1#3}
\Crefformat{assumption}{#2Assumption #1#3}

\usepackage{crossreftools}
\makeatletter
\renewenvironment{proof}[1][Proof]%
{%
  \par\noindent{\bfseries\upshape {#1.}\ }%
}%
{\qed\newline}
\makeatother

\xpatchcmd{\proof}{\itshape}{\normalfont\proofnameformat}{}{}
\newcommand{\proofnameformat}{\bfseries}

\usepackage[utf8]{inputenc}
\usepackage[T1]{fontenc}
\usepackage{url}
\usepackage{booktabs}
\usepackage{nicefrac}
\usepackage{makecell}
\usepackage{enumitem}
\usepackage{breakcites}
\usepackage{mathrsfs}
\usepackage{bm}
\usepackage{accents}
\usepackage{verbatim}
\usepackage{multicol}
\usepackage{colortbl}
\usepackage{setspace}
\usepackage{transparent}
\usepackage{upgreek}
\usepackage{inconsolata}
\usepackage[scaled=.90]{helvet}
\usepackage{xspace}
\usepackage{xparse}
\usepackage{graphicx}
\usepackage{pdflscape}
\usepackage{pifont}
\usepackage{titletoc}
\usepackage{flafter}

\makeatletter
\let\OldStatex\Statex
\renewcommand{\Statex}[1][3]{%
  \setlength\@tempdima{\algorithmicindent}%
  \OldStatex\hskip\dimexpr#1\@tempdima\relax}
\makeatother

\DeclareFontFamily{U}{jkpmia}{}
\DeclareFontShape{U}{jkpmia}{m}{it}{<->s*jkpmia}{}
\DeclareFontShape{U}{jkpmia}{bx}{it}{<->s*jkpbmia}{}
\DeclareMathAlphabet{\mathfrak}{U}{jkpmia}{m}{it}
\SetMathAlphabet{\mathfrak}{bold}{U}{jkpmia}{bx}{it}

\usepackage{comment}

\DeclarePairedDelimiter{\abr}{\lvert}{\rvert} %
\DeclarePairedDelimiter{\sbr}{[}{]}
\DeclarePairedDelimiter{\cbr}{\{}{\}}
\DeclarePairedDelimiter{\rbr}{(}{)}
\DeclarePairedDelimiter{\nbr}{\|}{\|}

\DeclarePairedDelimiter{\inner}{\langle}{\rangle}

\let\Pr\undefined

\DeclareMathOperator{\En}{\mathbb{E}}
\DeclareMathOperator{\Var}{\mathbb{V}}

\DeclareMathOperator{\Pr}{\bbP}

\DeclareMathOperator*{\argmin}{arg\,min} %
\DeclareMathOperator*{\argmax}{arg\,max}

\newcommand{\wt}[1]{\widetilde{#1}}
\newcommand{\wh}[1]{\widehat{#1}}

\newcommand{\ol}[1]{\overline{#1}}

\def\ddefloop#1{\ifx\ddefloop#1\else\ddef{#1}\expandafter\ddefloop\fi}
\def\ddef#1{\expandafter\def\csname bb#1\endcsname{\ensuremath{\mathbb{#1}}}}
\ddefloop ABCDEFGHIJKLMNOPQRSTUVWXYZ\ddefloop
\def\ddefloop#1{\ifx\ddefloop#1\else\ddef{#1}\expandafter\ddefloop\fi}
\def\ddef#1{\expandafter\def\csname b#1\endcsname{\ensuremath{\mathbf{#1}}}}
\ddefloop ABCDEFGHIJKLMNOPQRSTUVWXYZ\ddefloop
\def\ddef#1{\expandafter\def\csname sf#1\endcsname{\ensuremath{\mathsf{#1}}}}
\ddefloop ABCDEFGHIJKLMNOPQRSTUVWXYZ\ddefloop
\def\ddef#1{\expandafter\def\csname c#1\endcsname{\ensuremath{\mathcal{#1}}}}
\ddefloop ABCDEFGHIJKLMNOPQRSTUVWXYZ\ddefloop
\def\ddef#1{\expandafter\def\csname h#1\endcsname{\ensuremath{\widehat{#1}}}}
\ddefloop ABCDEFGHIJKLMNOPQRSTUVWXYZ\ddefloop
\def\ddef#1{\expandafter\def\csname hc#1\endcsname{\ensuremath{\widehat{\mathcal{#1}}}}}
\ddefloop ABCDEFGHIJKLMNOPQRSTUVWXYZ\ddefloop
\def\ddef#1{\expandafter\def\csname t#1\endcsname{\ensuremath{\widetilde{#1}}}}
\ddefloop ABCDEFGHIJKLMNOPQRSTUVWXYZ\ddefloop
\def\ddef#1{\expandafter\def\csname tc#1\endcsname{\ensuremath{\widetilde{\mathcal{#1}}}}}
\ddefloop ABCDEFGHIJKLMNOPQRSTUVWXYZ\ddefloop
\def\ddefloop#1{\ifx\ddefloop#1\else\ddef{#1}\expandafter\ddefloop\fi}
\def\ddef#1{\expandafter\def\csname scr#1\endcsname{\ensuremath{\mathscr{#1}}}}
\ddefloop ABCDEFGHIJKLMNOPQRSTUVWXYZ\ddefloop

\newcommand{\veps}{\varepsilon}

\newcommand{\ldef}{\vcentcolon=}

\newcommand{\Dfeat}{D_\phi}
\newcommand{\Dstar}{D_\star}
\newcommand{\Bstar}{B_\star}
\newcommand{\Bfeat}{B_\phi}
\newcommand{\op}{\operatorname{op}}
\newcommand{\Sigmat}[1][t]{\Sigma\ind[#1]}

\newcommand{\pistar}{\pi^{\star}}
\newcommand{\thetastar}{\theta_\star}
\newcommand{\thetatarget}{\theta^\star_{\eta,t}} %
\newcommand{\piref}{\pi_{\sfref}}
\newcommand{\pirefstar}{\piref^\star}
\newcommand{\pibon}[1][N]{\pi^{\textsf{BoN}}_{#1}}

\newcommand{\thetaref}{\theta_{\sfref}}
\newcommand{\sfref}{\mathsf{base}}

\newcommand{\sfrefgen}{\mathsf{ref}} %
\newcommand{\pirefgen}{\pi_{\sfrefgen}}
\newcommand{\thetarefgen}{\theta_{\sfrefgen}}

\newcommand{\Jt}[1][t]{J_{\eta, {#1}}}
\newcommand{\Jtr}[1][t]{J_{\eta, {#1}}}
\newcommand{\Zt}[1][t]{Z_{#1}}

\newcommand{\pifin}{\pi_{\mathsf{post}}}
\newcommand{\pit}[1][t]{\pi\ind[#1]}
\newcommand{\pitp}[1][t]{\pi\ind[#1+1]}
\newcommand{\piT}{\pit[T]}
\newcommand{\thetat}[1][t]{\theta\ind[#1]}
\newcommand{\thetatp}[1][t]{\theta\ind[{#1}+1]}
\newcommand{\pt}[1][t]{p\ind[{#1}]}
\newcommand{\ptp}[1][t]{p\ind[{#1}+1]}

\newcommand{\pitk}[1][k]{\pit^{#1}}
\newcommand{\pitK}{\pitk[K]}
\newcommand{\pitkp}[1][k]{\pitk[{#1} +1]}

\newcommand{\cplus}{\tau_{+}}
\newcommand{\cminus}{\tau_{-}}
\newcommand{\Jtclip}[1][t]{J_{#1}^{\eta, \clip}}
\newcommand{\hJtclip}[1][t]{\wh{J}_{#1}^{\eta, \clip}}
\newcommand{\clip}{\mathsf{clip}}

\newcommand{\Lkimi}[1][t]{\Lt[{#1}]^{\kimi}}
\newcommand{\hLkimi}[1][t]{\hLt[{#1}]^{\kimi}}
\newcommand{\Ltkimib}[1][t]{\Lt[{#1}]^{\kimi,b}}
\newcommand{\hLtkimib}[1][t]{\hLt[{#1}]^{\kimi,b}}
\newcommand{\tLtkimib}[1][t]{\wt{L}_{#1}^{\kimi,b}}
\newcommand{\Lkiminorm}[1][t]{\Lt[{#1}]^{\mathrm{norm}}}
\newcommand{\hLkiminorm}[1][t]{\hLt[{#1}]^{\mathrm{norm}}}
\newcommand{\tLkiminorm}[1][t]{\widetilde L_{#1}^{\mathrm{norm}}}
\newcommand{\Lt}[1][t]{L_{#1}}
\newcommand{\hLt}[1][t]{\wh{L}_{#1}}

\newcommand{\thetatk}[1][k]{\thetat^{#1}}
\newcommand{\thetatkp}[1][k]{\thetatk[{#1} +1]}

\newcommand{\bt}[1][t]{b\ind[{#1}]}
\newcommand{\btstar}[1][t]{b\ind[{#1}]^\star}
\newcommand{\hbt}[1][G]{\wh{b}_t^{#1}}
\newcommand{\tbt}[1][G]{\wt{b}_t^{#1}}

\newcommand{\sigmatG}{\sigma_t^G}

\newcommand{\rebel}{\texttt{Rebel}\xspace}
\newcommand{\rebelon}{\texttt{Rebel-OnPolicy}\xspace}

\newcommand{\sfon}{\texttt{On}}
\newcommand{\Ltrebelon}[1][t]{\Lt[{#1}]^{\text{\ttfamily\rebel{}-\sfon}}}
\newcommand{\hLtrebelon}[1][t]{\hLt[{#1}]^{\text{\ttfamily\rebel{}-\sfon}}}

\newcommand{\Ystarx}{\cY^\star_x}
\newcommand{\piplust}[1][t]{\pi^\star_{\eta, {#1}}}

\newcommand{\Jr}{J_\eta}

\newcommand{\Cresp}[1][\sfref]{\cC_{#1}}

\newcommand{\Cpi}[2]{\cC_{#2}^{#1}}
\newcommand{\Cpibase}[1]{\Cpi{#1}{\sfref}}
\newcommand{\Cpit}[2][t]{\Cpi{#2}{#1}}

\newcommand{\covpres}{coverage-preserving\xspace}
\newcommand{\Covpres}{Coverage-preserving\xspace}
\newcommand{\CovPres}{Coverage-Preserving\xspace}

\newcommand{\ppolite}{\texttt{PPO-Lite}\xspace}
\newcommand{\ppolitenc}{\texttt{PPO-NoClip}\xspace}

\newcommand{\opglong}{\ppo with Reference Resets, without Clipping\xspace} %

\newcommand{\clippg}{\ppolite}
\newcommand{\clippglong}{\ppo with Reference Resets and Clipping\xspace} %
\newcommand{\Nsample}{N_{\mathtt{sample}}}
\newcommand{\Nstep}{N_{\mathtt{step}}}

\newcommand{\kimipg}{\texttt{KimiPO}\xspace}

\newcommand{\kiminorm}{\texttt{KimiPO-Norm}\xspace}
\newcommand{\kimipglong}{Kimi Policy Optimization\xspace}

\newcommand{\framework}{\texttt{APMD}\xspace}
\newcommand{\frameworklong}{Approximate Policy Mirror Descent\xspace}

\newcommand{\bonlong}{Best-of-N\xspace}

\newcommand{\dpo}{\texttt{DPO}\xspace}
\newcommand{\npg}{\texttt{NPG}\xspace}

\newcommand{\kimi}{\mathsf{Kimi}}
\newcommand{\reinforce}{\texttt{REINFORCE}\xspace}

\newcommand{\mse}{\mathsf{MSE}}
\newcommand{\mset}[1][t]{\mathsf{MSE}_{#1}}
\newcommand{\trpo}{\texttt{TRPO}\xspace}
\newcommand{\ppo}{\texttt{PPO}\xspace}
\newcommand{\grpo}{\texttt{GRPO}\xspace}

\newcommand{\refuel}{\texttt{REFUEL}\xspace}
\newcommand{\oapl}{\texttt{OAPL}\xspace}

\newcommand{\astarpo}{\texttt{A}$^\star$\texttt{-PO}\xspace}

\newcommand{\Rmax}{R_{\max}}
\newcommand{\Jstar}{J^\star}

\newcommand{\ein}{\veps_{\mathsf{in}}}
\newcommand{\einsq}{\ein^{\mathsf{Sq}}}
\newcommand{\vepsopt}{\veps_{\mathsf{opt}}}
\newcommand{\vepsstat}{\veps_{\mathsf{stat}}}

\newcommand{\para}[1]{\textbf{#1.}}
\newcommand{\poly}{\mathrm{poly}}

\newcommand{\comp}{\mathrm{comp}}

\newcommand{\ind}[1][t]{_{#1}}
\newcommand{\one}[1]{\bbI\sbr*{#1}}

\newcommand{\Dkl}[2]{D_{\mathsf{KL}}\rbr*{#1\,\|\,#2}}

\input{dylan}

\input{widebar}

\usepackage{xcolor}

\usepackage{tcolorbox}

\newtcolorbox{frameworkbox}[1]{
  title={#1},
  colback=white,
  colbacktitle=black!12,
  colframe=black!35,
  coltitle=black,
  fonttitle=\bfseries,
  boxrule=0pt,
  leftrule=2.5pt,
  rightrule=0pt,
  toprule=0pt,
  bottomrule=0pt,
  arc=0pt,
  outer arc=0pt,
  boxsep=0pt,
  left=8pt,
  right=8pt,
  top=6pt,
  bottom=6pt,
  lefttitle=8pt,
  righttitle=8pt,
  toptitle=4pt,
  bottomtitle=4pt,
  titlerule=0pt,
  before skip=10pt,
  after skip=10pt
}

\title{\huge A Theory of Policy Optimization in Post-Training: \\[.1em]
\huge Coverage Collapse and Reference Resets}

\author{}
\date{}

\begin{document}

\maketitle

\vspace{-4em}
\begin{center}
\large
\setlength{\tabcolsep}{10pt}
\begin{tabular}{@{}ccc@{}}
Audrey Huang\footnotemark[1]
& Nan Jiang\footnotemark[2]
& Sham M. Kakade\footnotemark[3] \\[1em]
Sadhika Malladi\footnotemark[4]
& Akshay Krishnamurthy\footnotemark[5]
& Dylan J. Foster\footnotemark[5]
\end{tabular}
\end{center}
\vspace{2em}

\footnotetext[1]{Kempner Institute at Harvard University and UIUC,
  \texttt{audrey\_huang@harvard.edu}}
\footnotetext[2]{UIUC, \texttt{nanjiang@illinois.edu}}
\footnotetext[3]{Kempner Institute at Harvard University,
  \texttt{sham@seas.harvard.edu}}
\footnotetext[4]{University of California, San Diego,
  \texttt{samalladi@ucsd.edu}. Most of this work was completed as a postdoc at Microsoft Research.}
\footnotetext[5]{Microsoft Research NYC,
  \texttt{\{akshaykr, dylanfoster\}@microsoft.com}}

\begin{abstract}
  Policy optimization plays a central role in language model post-training,
yet principled guidance for methodology remains limited.
We revisit the theoretical foundations of policy optimization to identify algorithms that learn near-optimal policies
using total samples and gradient steps governed by the base model's \emph{response coverage},
a measure of how readily it generates optimal responses.
We call such algorithms \emph{\covpres}.
This guarantee captures the basic goal of amplifying capabilities already accessible through sampling from the base model,
yet has remained elusive in policy optimization.

\begin{itemize}[leftmargin=*,itemsep=2pt,topsep=3pt]
\item \textbf{Coverage collapse.}
\reinforce, the basis for \ppo and \grpo, can drive the probability of optimal responses to near zero, even as reward improves.
This \emph{coverage collapse} stalls optimization, and convergence can require exponentially many steps despite having a base model with good coverage.
The failure arises even with exact gradients and persists under KL regularization to the base model.\loose

\item \textbf{Reference resets.}
We show that repeatedly \emph{resetting the KL reference} to the most recent policy mitigates coverage collapse.
We formalize this intervention in \frameworklong (\framework),
a meta-algorithm that approximately solves a sequence of KL-regularized RL subproblems.
\framework improves the efficiency of optimization by enabling
stronger regularization, which makes each subproblem easier to solve,
while reference resets still allow convergence to global optimality.
\item \textbf{Coverage-preserving algorithms.}
For softmax linear policies over pretrained features,
we use \framework to establish the first global convergence guarantees
showing that widely used families of post-training algorithms are \covpres.
These include stylized \ppo and \grpo variants with explicit KL regularization and reference resets,
and squared-loss algorithms that implement them implicitly,
including the methods used to train the Kimi models \citep{du2025kimi,team2025kimi}.
\end{itemize}
These results clarify how regularization, clipping, and objective choice
affect the sampling and optimization costs of post-training,
and provide a foundation for policy optimization methodology.\loose

\end{abstract}

\startcontents[main]

\section{Introduction}\label{sec:intro}
Policy optimization algorithms are the reinforcement learning (RL) methods of choice in practice,
with successes spanning dexterous robotic manipulation and control
\citep{kaufmann2023champion,radosavovic2024humanoid,luo2025precise,ren2025diffusion},
superhuman performance in strategic games \citep{silver2016mastering,vinyals2019grandmaster,openai2019dota,perolat2022stratego,baker2022vpt},
and, most notably,
frontier reasoning in language models (LMs) \citep{ouyang2022instructgpt,openai2024reasoning,guo2025deepseekr1,du2025kimi}.
Across these applications, \emph{post-training} has emerged as a successful paradigm
in which policy optimization uses reward feedback to finetune an already capable \emph{base model},
developed through large-scale supervised pretraining.
The base model provides a foundation for learning complex tasks,
while reward feedback offers a way to improve beyond the limits of available demonstrations.\loose

Despite these successes, principled guidance for policy optimization methodology remains limited~\citep{ilyas2020closer,ahmadian2024back},
while the range of interventions continues to grow~\citep{shao2024deepseekmath,yu2025dapo,liu2025understanding,yue2025vapo,minimax2025minimaxm1,zheng2025gspo}.
Tuning any given method requires extensive trial and error~\citep{huang2024implementation,eimer2023hyperparameters,liu2025tricks},
making it difficult to evaluate the effects of specific interventions~\citep{ouyang2022instructgpt,khatri2025scalerl,spangher2025rlhf}.
Even training recipes for leading open models show little consensus on core algorithmic choices, such as which objective to optimize~\citep{guo2025deepseekr1,team2025olmo3,zheng2025gspo,team2025kimi}.
Principled insights are therefore increasingly important for focusing limited resources
on promising interventions and hyperparameter regimes.

Theory has, in the past, provided key design principles for policy optimization methods.
\trpo~\citep{schulman2015trpo}, followed by \ppo~\citep{schulman2017ppo},
directly drew from the proximal updates and policy improvement guarantees established for early theoretical algorithms \citep{kakade2001naturalpg,kakade2002cpi}.
In post-training, however, the central goal is to learn a performant policy while making the most of the base model,
a goal that existing theory has left largely unaddressed.
Most guarantees either require a uniform initial policy,
sidestepping the role of the base model altogether~\citep{mei2020softmaxpg,mei2021nonuniform,lattimore2026lyapunov},
or cannot preclude poor performance even with a high-quality base model~\citep{agarwal2020theory,mei2021nonuniform,mei2023stochasticbandits,wang2020neuralpg}.
They therefore offer little guidance for understanding and designing policy optimization methods in post-training.

Motivated by this gap, we revisit the theory of policy optimization in post-training, and ask:
\begin{center}
  \emph{
    Which policy optimization algorithms effectively leverage the base model for learning?
  }
\end{center}
To capture this goal, we will seek algorithms
whose performance is governed by the quality of the base model.
This is especially important given
the substantial resources invested in developing pretrained base models with strong post-training performance~\citep{grattafiori2024llama,deepseekai2024deepseekv3}.
We formalize our question and summarize our contributions below.

\subsection{Contributions}

We ask which policy optimization algorithms learn near-optimal policies using total samples and gradient steps
governed by the base model's coverage over optimal responses,
a guarantee that we call \covpres (\cref{sec:problem}).
This captures the basic post-training goal of leveraging the base model for learning
and is achieved by methods such as \bonlong sampling (\cref{sec:background-coverage}),
but has not been established in policy optimization.
By designing algorithms that achieve this guarantee,
we recover key interventions in modern post-training
and help explain their practical efficacy.

\begin{enumerate}[leftmargin=*]
  \item \para{Coverage collapse with a fixed reference policy or no regularization (\cref{sec:lower-bound})}
  When initialized from a base model with good (constant) coverage,
  \reinforce may require exponentially many gradient steps to converge,
  even with KL regularization to the base model.
    The cause is \emph{coverage collapse}, in which coverage over optimal responses
  deteriorates even as reward improves,
  causing the gradient signal to vanish (\cref{fig:coverage-collapse-leading}).
  This failure is intrinsic to the RL objective---%
  it occurs with exact gradients and without parameter sharing---%
  motivating interventions that preserve coverage for efficient learning.

  \item \para{\frameworklong (\framework) preserves coverage (\cref{sec:framework})}
  We show that algorithms built around a single intervention---resetting the reference policy---are \covpres.
  We formalize this in \framework,
  a meta-algorithm that approximates mirror descent by
  iteratively setting the reference to the most recent policy,
  then running an inner optimization loop for a regularized RL subproblem.
  \framework algorithms can heavily regularize the learner to the reference,
  making each subproblem easy to solve.
  While the learner improves only locally,
  over successive resets it reaches global optimality (\cref{fig:apmd-reanchoring}).\loose

  \begin{figure}[t]
  \centering
  \includegraphics[width=\linewidth]{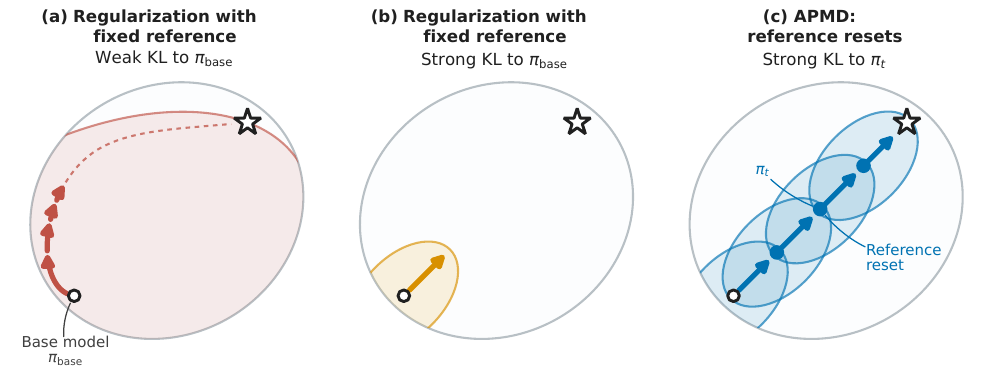}
  \caption{%
  \textbf{Overview of \framework{}:
  Continued progress through reference resets.}
  Shaded regions depict KL neighborhoods around the chosen reference policy.
  Each arrow marks the learner's progress after a fixed number of gradient steps.
  \textbf{(a)} With weak regularization to the base model, coverage can collapse and cause optimization to stall.
  \textbf{(b)} Strong regularization controls coverage and keeps the learner close to the reference policy,
  but far from optimality (marked \ding{73}).
  \textbf{(c)} \framework{} resets the reference to the most recent policy,
  reaching optimality through successive local improvements.}
  \label{fig:apmd-reanchoring}
\end{figure}

  \item \para{Applying \framework{}: \Covpres algorithms}
  \framework is a template for algorithm design and analysis,
  and can be combined with existing methods or directly applied to analyze them.
  We give end-to-end convergence guarantees for softmax linear policies,
  a model of learning with frozen pretrained features.\loose
  \begin{enumerate}[labelindent=-0.4em,leftmargin=*]
    \item
    \para{\ppo, \grpo, and variants (\cref{sec:pg-algo,sec:clip-pg})}
    We analyze variants of \ppo and \grpo with reference resets,
    as in ProRL \citep{liu2025prorl} and DeepSeek \citep{shao2024deepseekmath,guo2025deepseekr1}.
    In \ppolitenc (\cref{alg:offline-pg}), which does not clip the importance weights,
    regularization alone enables efficient learning by controlling the policy ratios (\cref{cor:pg-main}).
    We then add clipping to obtain \ppolite (\cref{alg:clip-pg}).
    Our analysis shows that clipping complements regularization by limiting the statistical cost of large importance weights,
    rather than directly improving optimization (\cref{cor:clippg-main}).\loose

    \item
    \para{Squared-loss algorithms (\cref{sec:regression})}
    These algorithms fit log policy ratios to rewards using squared-loss objectives.
    We first show that Kimi-style algorithms \citep{du2025kimi,team2025kimi},
    which shift the rewards by an estimated baseline function, are \covpres
    for two common baseline choices,
    the mean reward in \kimipg (\cref{cor:mean-kimi-main})
    and the normalization function in \kiminorm (\cref{cor:kimi-norm-regret}).
    Lastly, we show that \rebelon, an on-policy variant of \rebel
    that eliminates the baseline using paired responses,
    is also \covpres (\cref{cor:rebel-main}).
  \end{enumerate}
\end{enumerate}

\para{Implications of our theory (\cref{sec:conclusion})}
Our results show that optimization in a local neighborhood of the reference policy
makes improvement tractable,
whether regularization is explicitly used or implicit.
This role of the reference in anchoring optimization is distinct from its use in importance weighting or batch reuse.
Compared to fixed-reference (or unregularized) policy optimization,
algorithms that regularize the learner to a periodically updated reference
converge using
substantially fewer gradient steps and even fewer rounds of batch generation.

\subsection{Background: Policy Optimization in Post-Training}\label{sec:background-pg}

We study policy optimization in a contextual bandit setting motivated by RL post-training,
where reward feedback is used to finetune the weights of a pretrained base model.
A language model (LM) is a policy $\pi_\theta : \cX \rightarrow \Delta(\cY)$
that maps a prompt $x \in \cX$ to a distribution over complete responses, denoted $\cY$. 
Given a prompt distribution $\mu \in \Delta(\cX)$
and bounded reward function $r : \cX \times \cY \rightarrow [0, \Rmax]$,
the goal of policy optimization is to find\loose
\begin{align}
  \argmax_{\pi_\theta \in \Pi_\Theta}{} J(\pi_\theta),
  \quad\text{where}\quad
  J(\pi) \ldef{} \En_{x \sim \mu, y \sim \pi(\cdot \mid{} x)}\sbr*{r(x,y)}.
\end{align}
The objective $J(\pi)$ is the expected reward,
and $\Pi_\Theta = \{\pi_\theta : \theta \in \Theta\}$ denotes the policy class
induced by a fixed architecture with weights in $\Theta \subseteq \bbR^d$.
Reward feedback is assigned to a complete response $y$, the sequence of tokens generated for prompt $x$.
This abstraction captures RL for one-shot reasoning.

In post-training, policy optimization starts from a pretrained base model $\pit[0]=\piref=\pi_{\thetaref}\in\Pi_\Theta$
and produces the policy iterates $\{\pit=\pi_{\thetat}\}_{t=0}^T$ over $T$ iterations.
Objectives and update rules vary across algorithms, but we focus on first-order methods
that update $\pit$ to $\pitp$ using gradients of reward-based objectives.
Regardless of the algorithm, we evaluate the learned policy by the suboptimality of its expected reward.
For any $\veps \in (0, \Rmax]$,
we say that a policy $\pi$ is \emph{$\veps$-optimal} if $\Jstar-J(\pi)\leq\veps$,
where $\Jstar \ldef{} \max_{\pi : \cX \rightarrow \Delta(\cY)}{} J(\pi)$ is the maximum reward achievable by any conditional distribution.

\para{Example: \reinforce}
\reinforce{} \citep{williams1992simple,sutton1999policy} is a foundational policy optimization algorithm
that performs stochastic gradient ascent on the expected reward $J(\pi_\theta)$.
For a fixed learning rate $\alpha > 0$,
the parameter update with exact gradients is $\thetatp = \thetat + \alpha \nabla J(\pit)$,
where $\nabla$ denotes differentiation with respect to the policy parameters
and $\nabla J(\pit)$ is evaluated at $\theta=\thetat$,\loose
\begin{align}
  \label{eq:reinforce-gradient}
  \nabla J(\pit) = \En_{x \sim \mu, y \sim \pit(\cdot \mid{} x)}\sbr*{\nabla \log \pit(y\mid{}x) r(x,y)}.
\end{align}
In practice, \reinforce estimates $\nabla J(\pit)$ using responses sampled from $\pit$.
Modern policy gradient methods build on this update with additions such as
KL regularization to the base model \citep{ouyang2022instructgpt,shao2024deepseekmath},
estimated advantage functions,
and clipped importance weights for batch reuse \citep{schulman2017ppo,shao2024deepseekmath}.
In \cref{sec:lower-bound}, we study \reinforce to identify optimization challenges specific to RL.
These challenges motivate from first principles the design of modern post-training methods
such as \ppo and \grpo (\cref{sec:pg}).

\subsection{Background: Coverage and the Role of the Base Model}\label{sec:background-coverage}

Policy optimization in post-training begins from a base model,
and we measure the quality of the base model by its \emph{response coverage},
which quantifies how often it generates optimal responses.
\begin{definition}[Response coverage]\label{def:response-coverage}
  For any prompt $x \in \cX$, let
  $\Ystarx \ldef{} \cbr*{y \in \cY : r(x,y) = \max_{y'} r(x,y')}$
  denote the set of optimal responses.
  Then the response coverage of a policy $\pi$ is measured by
  \begin{align}
    \Cresp[\pi] \ldef{} \En_{x \sim \mu}\sbr*{\frac{1}{\pi(\Ystarx \mid{} x)}}.
  \end{align}
  For the base model,
  we use $\Cresp \equiv \Cresp[\piref]$.
\end{definition}
For a policy $\pi$ and prompt $x$, the expected number of generations
needed to observe an optimal response is $\pi(\Ystarx\mid{}x)^{-1}$,
and $\Cresp[\pi]$ averages this cost over prompts.
Smaller values of $\Cresp[\pi]$ indicate better response coverage,
meaning that fewer generations are needed to find an optimal response.
A central premise of post-training is that pretraining imbues the base model with favorable coverage (a moderate $\Cresp$).
This has helped explain the success of inference-time sampling \citep{brown2024monkeys, huang2025bestofn}
and supervised finetuning on filtered generations \citep{singh2024beyond,lee2025selfimproving,huang2024self}.

\para{Coverage and \bonlong}
The base model's response coverage governs the performance of \bonlong sampling,
a widely used inference-time method for improving response quality without training.
Given a prompt $x$, \bonlong draws $N$ independent responses from $\piref(\cdot\mid{}x)$
and returns the one with the highest reward $r(x,y)$.
It fails to return an optimal response only when none of the $N$ candidates is optimal,
an event with probability
$\En_{x}\sbr{(1 - \piref(\Ystarx \mid{} x))^N} \le \Cresp / N$.
Thus, by using $N=\wt\Theta(\Cresp/\veps)$ responses per prompt, \bonlong returns
an optimal response with probability at least $1-\veps$ \citep{huang2025bestofn,chen2025coverage}.\loose

\para{Coverage and policy optimization}
The \bonlong guarantee above suggests that policy optimization algorithms should learn near-optimal policies
using total samples controlled by $\Cresp$.
However, unlike \bonlong, which always samples from $\piref$,
in policy optimization responses are sampled from the learner, whose coverage evolves during training.
For example, with binary rewards $r(x,y)\in\{0,1\}$,
the estimated \reinforce gradient
$\nabla J(\pit) = \En_{\pit}\sbr*{\nabla \log \pit(y\mid{}x) r(x,y)}$
is zero unless the batch contains a correct response.
It is therefore the coverage of $\pit$ that governs convergence in policy optimization.
While the initial policy inherits the base model's coverage,
it is unclear whether this coverage, which supports the learning signal, persists during training.\loose

\section{Motivation: The Coverage Collapse Phenomenon}\label{sec:motivation}
We therefore ask whether policy optimization algorithms can learn efficiently when initialized from a high-quality base model, the basic goal of post-training.
We first show that \reinforce can suffer \emph{coverage collapse},
causing exponentially slow convergence despite favorable initial coverage
(\cref{sec:lower-bound}).
This failure motivates our central question in \cref{sec:problem},
where we ask which policy optimization methods learn near-optimal policies
with total samples and gradient steps controlled by the base model's coverage,
a guarantee we call \covpres.\loose

\subsection{Lower Bound: Coverage Collapse in Na\"ive Policy Optimization Methods}\label{sec:lower-bound}

\begin{figure}[t]
  \centering
  \setlength{\abovecaptionskip}{3pt}
  \includegraphics[width=0.95\linewidth]{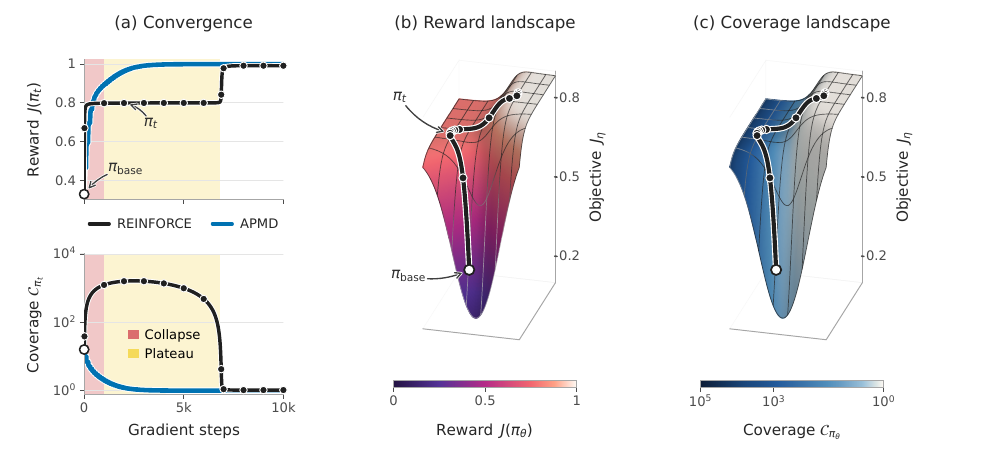}
  \input{sections/coverage-collapse-landscapes-caption.tex}
  \label{fig:coverage-collapse-leading}
  \vspace{-6pt}
\end{figure}

We now investigate whether \reinforce converges efficiently when initialized from a base model with good coverage.
Throughout this section, we use exact rather than estimated gradients
to isolate the effect of coverage on the dynamics of policy optimization.
We also allow KL regularization to the base model,
with regularization parameter $\eta \in (0, \infty]$, giving the objective
\begin{align}
  \Jr(\pi_\theta)
  \ldef{} J(\pi_\theta) - \frac{1}{\eta}\Dkl{\pi_\theta}{\piref},
  \qquad \text{where}\quad
  \Dkl{\pi}{\piref}
  = \En_{x\sim\mu,\,y\sim\pi(\cdot\mid{}x)}
  \sbr*{\log\frac{\pi(y\mid{}x)}{\piref(y\mid{}x)}},
  \label{eq:fixed-ref-objective}
\end{align}
where $\eta=\infty$ recovers the unregularized objective $J$.
Starting from $\pit[0]=\piref$,
\reinforce performs gradient ascent with a fixed learning rate $\alpha > 0$,
\begin{align}
  \thetatp = \thetat + \alpha\nabla\Jr(\pit),
  \qquad \text{where}\quad
  \nabla\Jr(\pit)
  = \En_{\pit}
  \sbr*{\nabla\log\pit(y\mid{}x)
    \rbr*{r(x,y)-\frac{1}{\eta}\log\frac{\pit(y\mid{}x)}{\piref(y\mid{}x)}}}.
  \label{eq:fixed-ref-update}
\end{align}

KL regularization is widely used in post-training to stabilize optimization
and help preserve pre-existing behavior \citep{shao2024deepseekmath,lambert2024tulu,liu2025prorl}.
Because it discourages deviations from the base model,
one might expect $\pit$ to stay close to $\piref$ in distribution
and retain the base model's coverage,
especially since improving performance involves assigning greater probability to higher-reward responses.\loose

However, \cref{thm:lower-bound-main} shows that
the policy's coverage can deteriorate until it is exponentially worse than the base model's,
causing convergence to be exponentially slow.
Favorable coverage in the base model therefore does not guarantee efficient convergence of \reinforce,
with or without KL regularization.

\begin{theorem}[Coverage collapse in \reinforce]
  \label{thm:lower-bound-main}
  Fix any $\veps \in (0, \frac{1}{64}]$.
  There exists a bandit instance with prompts $\cX = \{x\}$,
  responses $\cY$, a reward function $r$ with $\Rmax \le 1$,
  and a base model $\piref$ with $\Cresp \le 6$
  for which the following holds.
  Consider the tabular softmax policy class
  $\Pi_\Theta = \cbr{\pi_\theta : \pi_\theta(y\mid{}x)\propto\exp(\theta_y),\ \theta \in \bbR^{|\cY|}}$.
  Initialize $\pit[0]=\piref$ and run the \reinforce update in \cref{eq:fixed-ref-update}
  with any regularization parameter $\eta \in (0, \infty]$
  and any learning rate $0 < \alpha \le 1 / 5$.
  If the policy is $\veps$-optimal from iteration $T$ onward,
  then there exists $0 < t < T$ such that\loose
  \begin{align}
    T = \Omega \rbr*{\frac{\Cresp[\pit]}{\alpha \sqrt{\veps}}}
    \qquad\text{and}\qquad
    \Cresp[\pit] = \exp\rbr*{\Theta\rbr*{\frac{1}{\sqrt{\veps}}}}.
  \end{align}
\end{theorem}
\Cref{thm:lower-bound-main} shows that the convergence of \reinforce depends on the policy's coverage along the optimization path,
measured by $\Cresp[\pit]$, rather than just the base model's.
The problem is that, in the first phase of training,
$\Cresp[\pit]$ grows from $\Cresp \le 6$ to an exponentially large value---%
meaning that the optimal response has exponentially small probability---%
even though expected reward increases monotonically.
This \emph{coverage collapse} makes the gradient signal vanish,
and \reinforce requires exponentially many steps to regain coverage and converge.
This can be seen in the prolonged reward plateau of \cref{fig:coverage-collapse-leading}a (yellow).
Given the simplicity of our construction,
we expect coverage collapse to persist under the complex optimization landscapes and stochastic noise encountered in practice.\loose

\para{Coverage collapse: Nonconcavity in RL}
The dependence of convergence on coverage along the optimization path
results from the nonconcavity particular to RL (\cref{fig:coverage-collapse-leading}b).
The objective $J(\pi_\theta)$ is nonconcave for softmax policies,
even without parameter sharing (as in \cref{thm:lower-bound-main}).
Correspondingly, convergence analyses for \reinforce
use \L{}ojasiewicz-type inequalities of the form
\citep{mei2020softmaxpg,mei2021nonuniform,agarwal2020theory},
\begin{align}
  \Jstar - J(\pit) \leq c\cdot\nbr*{\pistar/\pit}_\infty\cdot\nbr*{\nabla J(\pit)}.
  \label{eq:gradient-domination}
\end{align}
This bound shows how near-stationarity can still imply near-optimality,
depending on a coefficient $\nbr*{\pistar/\pit}_\infty$ that measures $\pit$'s coverage of an optimal policy $\pistar$.
During coverage collapse this coefficient becomes exponentially large (\cref{fig:coverage-collapse-leading}c).
\Cref{eq:gradient-domination} then permits $\pit$ to be nearly stationary yet highly suboptimal---%
precisely the behavior exhibited by \cref{thm:lower-bound-main} and \cref{fig:coverage-collapse-leading}a (yellow).\loose

\para{Why regularization to $\piref$ fails}
Because KL regularization anchors the policy to the base model,
it must be weak enough ($\frac{1}{\eta} = O(\veps)$)
for the policy to improve substantially and achieve $\Jstar - J(\piT) \le \veps$.
This freedom also permits large deviations that reduce the probability of optimal responses,
which is the mechanism behind the lower bound in \cref{thm:lower-bound-main}.
There, KL regularization exerts a significant restoring force
only after coverage has degraded exponentially (see \cref{fig:apmd-reanchoring}a).\loose

\begin{remark}[Comparison to prior work]
To our knowledge, \cref{thm:lower-bound-main} is the first result isolating the
failure of \reinforce from coverage collapse,
given that we use tabular parameters and exact gradients.
Related results either
\emph{(i)} show asymptotic failure without characterizing coverage \citep{lin2025rethinking};
\emph{(ii)} have additional sources of hardness,
such as stochastic noise \citep{chung2021baselines,mei2021stochasticity,baudry2025stochastic,lattimore2026diffusion},
parameter interference \citep{schaul2019ray},
or multi-step dynamics \citep{li2021exponential}; or
\emph{(iii)} scale with the base model's coverage,
remaining mild when this coverage is constant \citep{agarwal2020theory,razin2024vanishing,mei2020gravitational}.
In contrast, our lower bound is exponential in $1/\sqrt{\veps}$ even when $\Cresp$ is constant.\loose
\end{remark}

\subsection{\texorpdfstring{Question: Which Policy Optimization Algorithms Are \CovPres?}{Question: Which Policy Optimization Algorithms Are \CovPres?}}\label{sec:problem}
In light of \cref{thm:lower-bound-main}, we ask which policy optimization methods
learn efficiently when initialized from a base model and avoid coverage collapse.
Concretely, this means that they output near-optimal policies
using $\Nsample$ samples (prompt-response pairs) and $\Nstep$ gradient steps,
with both quantities controlled by the response coverage $\Cresp$.
We call algorithms with this guarantee \emph{\covpres}, as formalized below.

\begin{frameworkbox}{Which policy optimization algorithms are \covpres?}
  For a given policy class
  $\Pi_\Theta$ with complexity $\comp\rbr*{\Pi_\Theta}$,
  an algorithm is \emph{\covpres} if,
  for every $\veps \in (0,\Rmax]$ and $\delta \in (0,1)$,
  it returns a policy $\pifin$ satisfying
  $\Jstar - J\rbr*{\pifin} \leq \veps$ with probability at least $1-\delta$,
  using total gradient steps $\Nstep(\veps,\delta)$ and total samples $\Nsample(\veps,\delta)$ bounded as follows:
  \begin{align}
    \Nstep\rbr*{\veps,\delta}
      &\leq
      \poly\rbr*{
        \Cresp,\,
        \comp\rbr*{\Pi_\Theta},\,
        \veps^{-1},\,
        \log\rbr*{1/\delta}
      }, \\
    \Nsample\rbr*{\veps,\delta}
      &\leq
      \poly\rbr*{
        \Cresp,\,
        \comp\rbr*{\Pi_\Theta},\,
        \veps^{-1},\,
        \log\rbr*{1/\delta}
      }.
      \label{eq:question-response}
  \end{align}
\end{frameworkbox}

For \covpres algorithms,
the cost of learning is controlled by the base model's coverage.
The guarantee therefore formalizes the basic post-training goal of amplifying high-reward responses from the base model,
sometimes called ``sharpening''\footnote{\hypertarget{Hfootnote.\theHfootnote}{}
The difference is that ``sharpening'' refers generally to a policy's distribution becoming narrower,
which can degrade its coverage over optimal responses and its performance,
whereas we aim to preserve this coverage.
}
\citep{huang2024self,zhao2025echo,yue2025rlbeyond}.
Analogous sample complexity guarantees already hold for \bonlong sampling (\cref{sec:background-coverage})
and SFT on reward-filtered samples from $\piref$ \citep{huang2024self,liu2026taskcentric}.
Indeed, in the construction of \cref{thm:lower-bound-main}, where $\Cresp = O(1)$,
\bonlong produces an optimal response using only $O(1)$ generations,
whereas \reinforce requires exponentially many gradient steps to output a near-optimal policy due to coverage collapse.
This failure motivates interventions that preserve the base model's coverage throughout training (hence the name \covpres).

\para{Focus: Online policy optimization}
We focus on policy optimization algorithms that train on samples generated by the learner, as in \ppo, \grpo, and other widely used post-training methods.
We seek algorithms that are \covpres without additional samples from the base model or other fixed distributions with favorable coverage,
as standard procedures do not use them.

\para{Theoretical testbed: Softmax linear policies}
While many of our results apply to general policy parameterizations,
our end-to-end complexity analyses work with \emph{softmax linear policy classes},
defined below.
We explicitly invoke \cref{asm:softmax-linear} in results that require it.

\begin{assumption}[Softmax linear policies]
  \label{asm:softmax-linear}
  For a fixed feature map
  $\phi : \cX \times \cY \rightarrow \bbR^d$, the policy class is\loose
  \begin{align}
    \Pi_\Theta \ldef{} \cbr*{\pi_\theta : \theta \in \bbR^d},
    \qquad
    \pi_\theta(y\mid{}x)
    =
    \frac{\exp\rbr*{\phi(x,y)^\top\theta}}
    {\sum_{y'\in\cY}\exp\rbr*{\phi(x,y')^\top\theta}}.
  \end{align}
  Here, $y \in \cY$ denotes a complete response consisting of a sequence of tokens.
  The reward function is linear in the same features,
  meaning there exists some $\thetastar \in \bbR^d$ such that
  \begin{align}
    r(x,y)=\phi(x,y)^\top\thetastar,
    \qquad
    \forall(x,y)\in\cX\times\cY.
  \end{align}
  We write
  $\Bfeat\ldef{}\max_{x\in\cX,\,y\in\cY}\nbr*{\phi(x,y)}_2$ and
  $\Bstar\ldef{}\nbr*{\thetastar}_2$.
\end{assumption}

\Cref{asm:softmax-linear} is the simplest function approximation setting for policies,
where shared parameters allow learning from observed samples to transfer to unseen prompts.
It models post-training as finetuning weights over the base model's frozen features $\phi$.
For softmax linear policies, the cross-entropy loss used in pretraining is convex \citep{chen2025coverage},
whereas the expected reward $J(\pi_\theta)$ is nonconcave in $\theta$ \citep{agarwal2020theory,mei2020softmaxpg}.
The nonconcavity is therefore particular to the RL objective rather than the parameterization itself,
and is reflected in the coverage collapse exhibited by \cref{thm:lower-bound-main},
where the policy class is a special case of \cref{asm:softmax-linear} with one-hot features $\phi$.

\parttwo{\part{Main Results: A Unifying Framework for Policy Optimization}\label{part:main}}
\section{Reference Resets Preserve Coverage: \framework Framework}\label{sec:pg}
We now show that algorithms built around a single intervention---%
resetting the reference policy used for regularization---%
are \covpres in the sense of \cref{eq:question-response}.
Recall from \cref{sec:lower-bound} that KL regularization to the base model does not prevent coverage collapse,
as regularization must be weak enough for the learner to improve substantially (\cref{fig:apmd-reanchoring}a).
Stronger regularization would mitigate loss of coverage but bias the learner toward the base model,
limiting potential improvement (\cref{fig:apmd-reanchoring}b).
When the reference is periodically reset to the most recent policy, however,
the learner can keep improving and reach optimality (\cref{fig:apmd-reanchoring}c).

We formalize this intervention as a meta-algorithm, \frameworklong{} (\framework),
which solves a sequence of RL subproblems regularized to the most recent policy.
As long as each RL subproblem is solved accurately enough,
the policies preserve the base model's coverage throughout training and converge to optimality (\cref{sec:framework}).
We then instantiate \framework with stylized versions of \ppo and \grpo,
without (\cref{sec:pg-algo}) and with (\cref{sec:clip-pg}) clipping.
In \cref{sec:regression}, we study a second family of \framework algorithms based on squared-loss objectives,
including those used to train the Kimi models \citep{team2025kimi,du2025kimi}.

\subsection{Framework: Approximate Policy Mirror Descent}\label{sec:framework}

\frameworklong (\framework), defined below, starts from the base model, $\pit[0] = \piref$.
At each outer iteration $t$, it first \emph{resets the reference}---%
the policy to which regularization is anchored---%
to the most recent policy $\pit$.
This sets the optimization target to
$\piplust \propto \pit e^{\eta r}$, the \emph{exact mirror descent update} to $\pit$ under KL regularization.\footnote{\hypertarget{Hfootnote.\theHfootnote}{}%
As in the policy mirror descent literature \citep{shani2020adaptive,xiao2022rates,lan2023pmd},
we say ``descent'' for reward maximization.}
\framework then trains the next policy $\pitp$ to approximate $\piplust$
using an algorithm-specific inner loop.\loose

\begin{frameworkbox}{Meta-algorithm: \frameworklong (\framework)}
  \textbf{Require:} base model $\piref$, regularization parameter $\eta > 0$, and outer iterations $T \ge 1$.\par
  Initialize $\pit[0] = \piref$. For each $0 \le t < T$, do:
  \begin{enumerate}
      \item Reset the reference policy to $\pit$ and define the optimization target,
      \begin{align}
        \piplust
        \ldef{}
        \argmax_{\pi : \cX \rightarrow \Delta(\cY)}{} \En_{\pi}\sbr*{r(x,y)}
        - \frac{1}{\eta} \Dkl{\pi}{\pit}
        ,
        \quad\text{where  }
        \piplust(y\mid{}x) \propto \pit(y\mid{}x) e^{\eta r(x,y)}
        .
        \label{eq:framework}
      \end{align}
      \item Learn $\pitp \approx \piplust$ by running an inner optimization loop for an algorithm-specific RL subproblem.
    \end{enumerate}
    \textbf{Return:} policy iterates $\cbr*{\pit}_{t=1}^T$.
\end{frameworkbox}

All \framework algorithms perform the reference reset in Step~1 and share the optimization target $\piplust$.
They differ in the RL subproblem and inner optimization procedure used for Step~2.
For example, the algorithm in \cref{sec:pg-algo} uses gradient ascent on the KL-regularized expected reward in \cref{eq:framework}.
The algorithms in \cref{sec:regression} instead minimize squared-loss objectives to approximate the same target.

If every RL subproblem were solved exactly,
so that $\pitp = \piplust$ for all $t$,
the policy iterates would follow the mirror descent path, satisfying (\cref{app:population-mirror-descent})
\begin{align}
  \pit(y\mid x) \propto \piref(y\mid x)e^{t\eta r(x,y)},
  \quad\text{and}\quad
  \Jstar-J(\pit) \le \frac{\log\Cresp}{\eta t},
  \qquad\forall 0 < t \le T
  .
  \label{eq:mirror-descent}
\end{align}
The policies converge at the sublinear \emph{mirror descent rate} above,
which depends on the base model only through $\log \Cresp$.
Moreover, both expected reward and coverage improve monotonically in $t$;
each iterate $\pit$ draws optimal responses at least as often as the base model,
preserving useful learning signals and avoiding the coverage collapse
under fixed-reference policy optimization in \cref{thm:lower-bound-main}.
The challenge is maintaining these properties when $\pitp$ only approximates $\piplust$,
given that sampling and optimization errors compound across iterations.\loose

The following theorem, a central result of the paper,
shows that the final policy $\piT$ retains the mirror descent rate in \cref{eq:mirror-descent},
provided each $\pitp$ is sufficiently close to $\piplust$ in KL divergence.
The learned policies also \emph{preserve coverage},
meaning that, for most prompts,
every iterate generates an optimal response at least half as often as the base model.\footnote{\hypertarget{Hfootnote.\theHfootnote}{}
``Preserves coverage'' describes learned policies, while ``coverage-preserving'' refers to algorithms satisfying \cref{eq:question-response}.
}
\begin{theorem}[\framework preserves coverage]
  \label{thm:kl-regret-main}
  Given any $\eta>0$ and any sequence of policies
  $\cbr*{\pit}_{t=0}^T$ with $\pit[0]=\piref$ and $\pit>0$,
  suppose that
  \begin{align}
    \Dkl{\pitp}{\piplust}
    \leq
    \ein \ldef{}
    \frac{1}{64\Cresp T^2\log(eT)},
    \qquad
    \forall 0\leq t<T.
    \label{eq:framework-kl-tolerance}
  \end{align}
  Then the final policy converges at the mirror descent rate,
  \begin{align}
    \Jstar-J(\piT)
    \leq
    \frac{\log\rbr*{2\Cresp}}{\eta T},
    \label{eq:framework-convergence}
  \end{align}
  and all policies \emph{preserve coverage}:
  with probability at least $1 - \frac{1}{2\sqrt{\log(eT)}}$ over $x \sim \mu$,
  \begin{align}
    \pit(\Ystarx \mid{} x) \ge \frac{1}{2} \piref(\Ystarx \mid{} x),
    \qquad
    \forall 0 \le t \le T
    ,
  \end{align}
  where $\Ystarx$ is the set of optimal responses for the prompt $x$ (\cref{def:response-coverage}).
\end{theorem}

\Cref{thm:kl-regret-main} provides a criterion for identifying \covpres \framework algorithms.
It applies to any policy sequence $\cbr*{\pit}_{t=0}^T$ satisfying the \emph{inner KL tolerance} $\ein$ in \cref{eq:framework-kl-tolerance}.
For any $\eta>0$, the final policy is $\veps$-optimal when\loose
\begin{align}
  T
  &=
  \left\lceil
  \frac{\log\rbr*{2\Cresp}}{\eta\veps}
  \right\rceil,
  &
  \Dkl{\pitp}{\piplust}
  &\leq
  \ein =
  \wt O\rbr*{
    \frac{\eta^2\veps^2}{\Cresp}
  },
  \quad \forall 0\leq t<T.
  \label{eq:kl-conditions}
\end{align}
Thus, for an \framework algorithm to be \covpres, the inner optimization loop
should meet the KL tolerance above using a polynomial number of samples and gradient steps,
for a suitable choice of $\eta$.
Following this approach, we will show that stylized variants of \ppo and \grpo
are \covpres in \cref{sec:pg-algo,sec:clip-pg}.\loose

\para{Intuition: \framework algorithms are \covpres}
Across our results, strong regularization to the reference---%
typically with $\eta = \Theta(1)$, compared to $\eta = \Omega(1 / \veps)$ when the reference is fixed (\cref{sec:lower-bound})---%
is key to obtaining \covpres guarantees for \framework algorithms.
This sets the target $\piplust$ close to the reference $\pit$
and stabilizes the inner loop,
making each RL subproblem relatively easy to solve (\cref{fig:apmd-reanchoring}b).
Correspondingly, as we will see in \cref{sec:pg-algo},
\framework algorithms converge using exponentially fewer gradient steps than \reinforce
in the construction of \cref{thm:lower-bound-main}, illustrated in \cref{fig:coverage-collapse-leading}a.

Reference resets in \framework make global optimality possible under strong regularization.
Each $\pitp \approx \piplust$ is only a local improvement to $\pit$,
but these improvements accumulate over successive resets (\cref{fig:apmd-reanchoring}c).
\Cref{eq:kl-conditions} makes this precise,
showing that the same performance can be achieved under stronger regularization
by proportionally increasing the number of outer iterations $T$.
This role of reference resets echoes empirical observations from \citet{liu2025prorl},
who report that the KL penalty to a fixed reference can dominate the loss and slow policy updates,
and periodically reset the reference to counteract this effect.

\subsection{Instantiating \framework: \texorpdfstring{Importance-Weighted Policy Optimization}{Importance-Weighted Policy Optimization} (\ppolitenc)}
\label{sec:pg-algo}

\begin{algorithm}[t]
  \caption{\opglong (\ppolitenc)}
  \label{alg:offline-pg}
  \begin{algorithmic}[1]
    \Require \begingroup\raggedright
    \mbox{Base model $\piref$},
    \mbox{outer iterations $T$}, \mbox{regularization parameter $\eta>0$},
    \mbox{prompt batch size $N$}, \mbox{inner steps $K$}, \mbox{and learning rate $\alpha$}.\par
    \endgroup
    \State Initialize $\pit[0]=\piref=\pi_{\theta_0}$.
    \For{$t=0,1,\ldots,T-1$}
      \State Draw a batch
      $\cD_t=\cbr*{(x_i,y_i,r_i)}_{i=1}^N$, where
      $x_i\overset{\mathrm{iid}}{\sim}\mu$, $y_i\sim\pit(\cdot\mid{}x_i)$, and
      $r_i=r(x_i,y_i)$.
      \State Initialize $\theta_t^0=\thetat$ and $\pi_t^0=\pit$.\label{line:inner-loop-start}
      \For{$k=0,1,\ldots,K-1$}
        \State Compute the gradient estimate
        \begin{align}
          \wh{\nabla\Jtr}(\pi_t^k)
          =
          \frac{1}{N}
          \sum_{i=1}^N
            \frac{\pi_t^k(y_i\mid{}x_i)}{\pit(y_i\mid{}x_i)}
            \nabla \log \pi_t^k(y_i\mid{}x_i)
            \rbr*{
              r_i
              -
              \frac{1}{\eta}
              \log
              \frac{\pi_t^k(y_i\mid{}x_i)}{\pit(y_i\mid{}x_i)}
            } .
            \label{eq:alg-offline-pg-gradient}
        \end{align}
        \State Update\label{line:inner-loop-end}
        $\theta_t^{k+1}=\theta_t^k+\alpha{\wh{\nabla\Jtr}}(\pi_t^k)$ and
        $\pi_t^{k+1}=\pi_{\theta_t^{k+1}}$.
      \EndFor
      \State Set $\thetatp=\theta_t^K$ and $\pitp=\pi_t^K$.
    \EndFor
    \State \Return $\pifin = \pit[T]$.
  \end{algorithmic}
\end{algorithm}

\ppo and \grpo are the most recognizable policy optimization algorithms in post-training,
yet the individual roles of key design elements---%
importance-weighted gradients, clipping, and KL regularization---%
are not well understood.
To disentangle their effects, we analyze stylized variants using \framework.
We begin with \ppolitenc,
which uses KL regularization with reference resets but omits clipping.
We then add clipping in \cref{sec:clip-pg} to obtain \ppolite.
This approach reveals how KL regularization controls the optimization landscape,
while clipping controls the statistical cost of importance weighting.

In each outer iteration $t$, \ppolitenc aims to maximize the KL-regularized objective with reference $\pit$,
\begin{align}
  \Jtr(\pi) \ldef{} \En_{\pi}\sbr*{r(x,y)}
  - \frac{1}{\eta}\Dkl{\pi}{\pit}.
  \label{eq:pg-objective}
\end{align}
This objective is uniquely maximized by $\piplust$ (\cref{eq:framework}),
so \ppolitenc implements \framework.
Note that we use observed rewards rather than advantage estimates,
as our focus is on the effects of regularization and clipping.

To optimize this objective, the inner loop of \ppolitenc performs $K$ steps of full-batch gradient ascent, starting from $\pitk[0]=\pit$.
It reuses a pre-collected batch of $N$ prompt-response pairs from $\pit$ to estimate these gradients,
following standard practice in \ppo and \grpo to reduce generation costs \citep{schulman2017ppo,shao2024deepseekmath}.
At inner step $k$, the estimated gradient uses importance weights $\pitk(y\mid{}x)/\pit(y\mid{}x)$
to correct for distribution shift between the learner $\pitk$ and the sampling policy $\pit$
(\cref{eq:alg-offline-pg-gradient}).
This procedure is equivalent to gradient ascent on the importance-weighted formulation of \cref{eq:pg-objective} below,
which resembles the \ppo objective without clipping
and with regularization to $\pit$.

\begin{align}
  \Jtr(\pi_\theta)
  = \En_{\pit}\sbr*{
    \frac{\pi_\theta(y\mid{}x)}{\pit(y\mid{}x)}
    \rbr*{r(x,y)-\frac{1}{\eta}\log\frac{\pi_\theta(y\mid{}x)}{\pit(y\mid{}x)}}}.
  \label{eq:offline-pg-objective}
\end{align}
Finally, \ppolitenc sets $\pitp=\pitk[K]$,
which becomes the reference policy, sampling policy, and starting point for the next outer iteration.

The following result shows that \ppolitenc is \covpres under \cref{asm:softmax-linear}.

\begin{corollary}[\ppolitenc is \covpres]
  \label{cor:pg-main}
  Suppose that \cref{asm:softmax-linear} holds, and let $B \ldef B_\phi\Bstar$.
  In \cref{alg:offline-pg}, set $\eta = 1 / B$.
  Then, for any $\veps\in(0,\Rmax]$ and $\delta\in(0,1)$,
  there exist choices of $\alpha$, $N$, and $K$ (see \cref{cor:noclip-regret}),
  with $T$ specified by \cref{eq:kl-conditions},
  such that \ppolitenc returns a policy $\pifin$ satisfying $\Jstar-J(\pifin)\le\veps$
  with probability at least $1-\delta$, using
  \begin{align}
    \Nstep(\veps, \delta)
    = \wt O\rbr*{\frac{\Cresp B^3}{\veps^3}}
    \qquad\text{and}\qquad
    \Nsample(\veps, \delta)
    = \wt O\rbr*{\frac{\Cresp B^3\rbr*{d+\log(1/\delta)}}{\veps^3}}.
    \label{eq:complexity-pg-main}
  \end{align}
  Here, $\Nstep=KT$ and $\Nsample=NT$.
\end{corollary}
In the lower bound construction (\cref{thm:lower-bound-main}), where $\Cresp B^3 = O(1)$,
\cref{cor:pg-main} gives $\Nstep = \wt O(\veps^{-3})$,
while \reinforce on the fixed-reference objective requires
$\exp(\Omega(1/\sqrt{\veps}))$ steps to converge even with exact gradients.
\Cref{fig:coverage-collapse-leading}a illustrates this performance gap with exact gradients for both methods.
The difference is that \ppolitenc uses strong regularization, $\eta = 1 / B$,
thereby avoiding coverage collapse.
The proof sketch below makes clear how optimization benefits from this regularization.

\paragraph{Overview of analysis for \ppolitenc (\cref{cor:pg-main})}
By \cref{thm:kl-regret-main},
it suffices to show that each inner loop returns a policy with $\Dkl{\pitp}{\piplust}\le\ein$.
Maximizing $\Jtr$ (\cref{eq:pg-objective}) directly controls this KL divergence,
which is proportional to the suboptimality of $\pitp$ on the RL subproblem (\cref{lem:noclip-objective-kl}),
\begin{align}
  \frac{1}{\eta}\Dkl{\pi}{\piplust} = \Jtr(\piplust)-\Jtr(\pi).
\end{align}
We therefore bound the batch size $N$ and gradient steps $K$
needed for $\pitp$ to be $\ein/\eta$-suboptimal.

\emph{(Optimization benefits of regularization)}
Although $\Jtr$ is nonconcave in $\theta$,
a local form of concavity holds at each inner iterate $\pitk$
(\cref{lem:noclip-population-geometry}),
\begin{align}
  \Jtr(\piplust)-\Jtr(\pitk)
  \le
  \frac{1}{2}
  \nbr*{\frac{\piplust}{\pitk}}_{\infty}
  \inner*{\nabla\Jtr(\pitk), \thetat - \thetatk + \eta \thetastar}.
  \label{eq:pg-concave}
\end{align}
Like the \L{}ojasiewicz inequality in \cref{eq:gradient-domination},
this bound weakens when $\pitk$ covers the target poorly
and the ratio $\piplust/\pitk$ is large.
This allows for suboptimal policies that are also near-stationary,
consistent with the training plateaus caused by coverage collapse in \cref{sec:lower-bound}.
Convergence also depends on a second policy ratio,
the importance weight $\pitk/\pit$,
which, when large, worsens smoothness and amplifies gradient noise.

KL regularization controls both ratios under \cref{asm:softmax-linear},
and we show that employing strong regularization keeps all inner iterates
close both to the reference and sampling policy $\pit$ and to the target $\piplust$,
in the sense that\loose

\begin{align}
  \frac{\pitk(y\mid{}x)}{\pit(y\mid{}x)}
  \le e^{8\eta B}
  \quad\text{and}\quad
  \frac{\piplust(y\mid{}x)}{\pitk(y\mid{}x)}
  \le e^{4\eta B},
  \quad \forall k\in[K].
  \label{eq:policy-ratios}
\end{align}
With $\eta=1/B$, both ratios are bounded by absolute constants
throughout the inner optimization loop,
giving uniform control over gradient smoothness and the local concavity in \cref{eq:pg-concave},
as illustrated by the optimization landscape of APMD in \cref{fig:reference-reanchoring-landscapes}c.

\emph{(Inner loop convergence)}
Adapting standard gradient ascent analysis shows it suffices to take
\begin{align}
  K &= \wt O\rbr*{\frac{1}{\ein}},
  &
  N &= \wt O\rbr*{\frac{d+\log(T/\delta)}{\ein}}.
  \label{eq:ppo-inner-costs}
\end{align}
With $T$ and $\ein$ as in \cref{eq:kl-conditions} for $\eta = 1 / B$,
this gives the total gradient steps $\Nstep(\veps, \delta) = KT$
and samples $\Nsample(\veps, \delta) = NT$ stated in \cref{cor:pg-main}.
The full proof appears in \cref{app:noclip-global};
our analyses of other algorithms will follow the template sketched here.

\para{On-policy \ppolitenc}
KL regularization to the reference improves convergence
by controlling the distribution shift (\cref{eq:policy-ratios})
that governs the optimization landscape (\cref{eq:pg-concave}).
This role of the reference policy---anchoring optimization to a local region where it is well-behaved---%
is distinct from its use in the importance weight.
To see this distinction, consider the on-policy variant of \ppolitenc,
which is also of independent interest given recent empirical work on on-policy training
\citep{xiang2026exprl,marek2026score}.
This variant also regularizes to $\pit$,
but draws a fresh batch from $\pitk$ at each inner step to estimate the on-policy gradient,
\begin{align}
  \nabla\Jtr(\pitk)
  =
  \En_{\pitk}\sbr*{
    \nabla\log\pitk(y\mid x)
    \rbr*{
      r(x,y)-\frac{1}{\eta}\log\frac{\pitk(y\mid x)}{\pit(y\mid x)}
    }
  }.
\end{align}
This removes importance weights,
but the objective $\Jtr$ remains the same
and the local concavity bound in \cref{eq:pg-concave} still applies.
Strong regularization to $\pit$ therefore remains key for convergence of the inner loop.
Our analysis readily adapts to establish that on-policy \ppolitenc is \covpres under \cref{asm:softmax-linear}.\loose

\subsection{Instantiating \framework: \texorpdfstring{{A Stylized Variant of PPO with Clipping}}{A Stylized Variant of PPO with Clipping} (\clippg)}
\label{sec:clip-pg}

\begin{algorithm}[t]
  \caption{\clippglong (\clippg)}
  \label{alg:clip-pg}
  \begin{algorithmic}[1]
    \Require \begingroup\raggedright
    \mbox{Base model $\piref$},
    \mbox{outer iterations $T$}, \mbox{regularization parameter $\eta>0$},
    \mbox{prompt batch size $N$}, \mbox{clipping thresholds $\cminus$ and $\cplus$},
    \mbox{and inner optimization tolerance $\vepsopt$}.\par
    \endgroup
    \State Initialize $\pit[0]=\piref$.
    \For{$t=0,1,\ldots,T-1$}
      \State Draw a batch
      $\cD_t=\cbr*{(x_i,y_i,r_i)}_{i=1}^N$, where
      $x_i\overset{\mathrm{iid}}{\sim}\mu$, $y_i\sim\pit(\cdot\mid{}x_i)$, and
      $r_i=r(x_i,y_i)$.
      \State Define the empirical clipped objective
      \begin{align}
        \hJtclip(\pi)
        \ldef{}
        \frac{1}{N}\sum_{i=1}^N
        \min\cbr*{
          \frac{\pi(y_i\mid{}x_i)}{\pit(y_i\mid{}x_i)}\, r_i,
          \sbr*{\frac{\pi(y_i\mid{}x_i)}{\pit(y_i\mid{}x_i)}}_{1-\cminus}^{1+\cplus}\, r_i
        }
        -\frac{1}{\eta}\Dkl{\pi}{\pit}.
        \label{eq:clip-pg-empirical-objective}
      \end{align}
      \State Find $\pitp=\pi_\theta\in\Pi_\Theta$ satisfying\label{line:clip-pg-erm}
      \begin{align}
        \hJtclip(\pitp)
        \geq
        \sup_{\pi\in\Pi_\Theta}{}\hJtclip(\pi)
        - \vepsopt
        .
        \label{eq:clip-pg-erm}
      \end{align}
    \EndFor
    \State \Return $\pifin=\pit[T]$.
  \end{algorithmic}
\end{algorithm}

For softmax linear policies, we showed that
KL regularization keeps policy ratios controlled throughout training.
This may not hold for more complex nonlinear policies,
and to limit their effects we now add clipping to the importance weights.
Clipping is a core component of \ppo and \grpo widely used in practice,
but its benefits are not well understood theoretically.
We establish a concrete statistical benefit by obtaining sample complexity guarantees for general policy parameterizations
without requiring bounded policy ratios throughout training.

\ppolite clips the importance weights
in the \ppolitenc objective (\cref{eq:offline-pg-objective}),
using separate lower and upper thresholds, $1-\cminus$ and $1+\cplus$.
We denote the clipping operator by
$[z]_{1-\cminus}^{1+\cplus} \ldef{}\min\cbr*{1+\cplus,\max\cbr*{1-\cminus,z}}$.
Writing $w_t^\pi(x,y)\ldef{}\pi(y\mid{}x)/\pit(y\mid{}x)$ for the importance weight,
we define the clipped objective as
\begin{align}
  \Jtclip(\pi)
  \ldef{}
  \En_{\pit}\sbr*{
    \min\cbr*{
      w_t^\pi(x,y) r(x,y),
      [w_t^\pi(x,y)]_{1-\cminus}^{1+\cplus} r(x,y)
    }
  }
  -\frac{1}{\eta}\Dkl{\pi}{\pit}.
  \label{eq:clip-pg-objective}
\end{align}
\Cref{eq:clip-pg-objective} uses the same clipping scheme as \ppo \citep{schulman2017ppo},
taking the minimum of the clipped and unclipped importance-weighted rewards.
It also allows for the asymmetric clipping widely adopted in RL post-training
\citep{yu2025dapo,zheng2025gspo,yue2025vapo,liu2025prorl}.
Like \ppolitenc, \ppolite (\cref{alg:clip-pg}) optimizes $\Jtclip$
using a pre-collected batch of $N$ prompt-response pairs from $\pit$ at each outer iteration $t$.
We leave the inner optimization loop abstract (Line~\ref{line:clip-pg-erm});
it returns a policy $\pitp$ that is approximately optimal for $\hJtclip$,
the empirical version of $\Jtclip$, up to optimization error $\vepsopt$.

The following result bounds the samples needed for \ppolite to learn an $\veps$-optimal policy,
provided each inner loop meets the prescribed optimization tolerance.
In this sense, the sample complexity of \ppolite is \covpres.
\begin{corollary}[\ppolite is \covpres (\protect\cref{cor:clip-global}, informal)]
  \label{cor:clippg-main}
  Suppose that $\Pi_\Theta$ is finite,%
  \footnote{\hypertarget{Hfootnote.\theHfootnote}{}
  The finiteness assumption can be relaxed using standard complexity measures, such as covering numbers.}
  $r(x,y)\in[-\Rmax,\Rmax]$,
  and $\piplust\in\Pi_\Theta$ for every $t$.
  In \cref{alg:clip-pg}, set $\eta=1/\Rmax$
  and the clipping thresholds to $\cminus=1-e^{-2\eta\Rmax}$ and $\cplus=e^{2\eta\Rmax}-1$.
  Then, for any $\veps\in(0,2\Rmax]$ and $\delta\in(0,1)$,
  set $T$ and $\ein$ as specified by \cref{eq:kl-conditions}.
  If each inner loop satisfies \cref{eq:clip-pg-erm} with optimization tolerance
  $\vepsopt\le\ein/(8\eta)$,
  then \ppolite returns a policy $\pifin$ satisfying $\Jstar-J(\pifin)\le\veps$
  with probability at least $1-\delta$, using
  \begin{align}
    \Nsample(\veps,\delta)
    = NT = \wt O\rbr*{
      \frac{\Cresp\Rmax^3\rbr*{\log|\Pi_\Theta|+\log(1/\delta)}}{\veps^3}
    }.
  \end{align}
\end{corollary}

\para{Clipping thresholds \& KL regularization}
Our guarantee relies on both regularization and clipping.
The strength of regularization defines how close the target $\piplust$ is to the reference $\pit$,
with $\piplust/\pit\in[e^{-2\eta\Rmax},e^{2\eta\Rmax}]$.
\cref{cor:clippg-main} chooses the narrowest clipping interval guaranteed to leave
$\piplust$ unclipped,
which therefore remains a maximizer of the clipped objective.
Although clipping was introduced in \ppo as an alternative to KL regularization \citep{schulman2017ppo},
here it complements rather than replaces regularization.
Regularization defines the optimization target itself and sets the difficulty of the problem,
while clipping on its own does not directly provide the same control.

\para{Benefits and costs of clipping}
To understand the benefits of clipping, let us compare
\cref{cor:clippg-main} with the guarantee for \ppolitenc in \cref{cor:pg-main},
which holds for softmax linear policies.
There, sufficiently strong KL regularization keeps policy ratios bounded by a constant throughout training.
If clipping were added using similar thresholds to \cref{cor:clippg-main}
it would bring no theoretical benefits, as the policy ratios would remain unclipped.
Indeed, \cref{cor:clip-softmax}, our end-to-end analysis for \ppolite under \cref{asm:softmax-linear}, recovers \cref{cor:pg-main}.\loose

In contrast, \cref{cor:clip-pg} applies to \emph{any} final policy
obtained by approximately maximizing the empirical objective $\hJtclip$ at each outer iteration,
regardless of the policy parameterization, the inner optimization procedure,
or how large the importance weights become.
Clipping controls the contribution of importance-weighted rewards,
so the required batch size depends on the preset clipping thresholds
rather than the magnitude of the policy ratios during training.
To our knowledge, this benefit of clipping has not been isolated in previous theory.
We note that these statistical benefits come with potentially higher optimization costs
as the objective is non-smooth at clipping thresholds.
We therefore do not analyze specific optimization procedures for the inner loop when clipping may be active,
leaving it to future work.

\begin{remark}[Connection to practice]
  \ppolite resembles practical methods that periodically reset the reference policy
  and add KL regularization to a clipped importance-weighted objective,
  including the DeepSeek training procedures \citep{shao2024deepseekmath,guo2025deepseekr1}
  and ProRL and its variants \citep{liu2025prorl,hu2025prorlv2,hu2026brorl}.
  Our analysis offers theoretical insight into this practice
  and suggests tuning the clipping thresholds together with the regularization strength,
  where stronger regularization may improve the optimization landscape.
  In particular, \cref{cor:clip-global} sets the clipping interval to $[e^{-2\eta\Rmax}, e^{2\eta\Rmax}]$,
  so stronger regularization calls for tighter clipping.

\end{remark}

\para{Additional results for \cref{sec:pg}: General performance target}
  \Cref{thm:general-target-comparator} in \cref{app:general-kl} extends \cref{thm:kl-regret-main}
  to provide performance guarantees against any comparator policy $\pi$.
  In particular, it bounds $J(\pi)-J(\piT)$ under an inner KL tolerance $\ein$ that depends on the base model's coverage of $\pi$.
  Combined with the analyses of \ppolitenc or \ppolite, this gives guarantees on samples and/or gradient steps (see \cref{eq:general-target-noclip-complexity})
  for attaining reward at least $J(\pi)-\veps$.

\section{\mbox{Squared-Loss Algorithms Preserve Coverage: Kimi and Rebel}}
\label{sec:regression}
In this section, we study a second family of algorithms that instead minimize \emph{squared-loss objectives}
to approximate the \framework target $\piplust$.
These objectives are derived from the identity below,
obtained by rearranging the target's closed form $\piplust(y\mid{}x) \propto \pit(y\mid{}x) e^{\eta r(x,y)}$.
\begin{align}
  \frac{1}{\eta}\log\frac{\piplust(y\mid{}x)}{\pit(y\mid{}x)} = r(x,y) - \btstar(x)
  \qquad
  \forall (x,y).
  \label{eq:kkt}
\end{align}
The reference $\pit$ now appears in the denominator of the log policy ratio,
and $\btstar(x) = \frac{1}{\eta}\log\En_{y'\sim\pit(\cdot\mid{}x)}\sbr{e^{\eta r(x,y')}}$
is the normalization function that ensures $\piplust(\cdot\mid{}x)$ is a valid distribution.
We can therefore approximate $\piplust$ by fitting the log-ratios of a policy
to the shifted rewards $r(x,y)-\btstar(x)$.
Take, for example, the squared-loss objective\loose
\begin{align}
  L_t(\pi_\theta)
  \ldef{}
  \En_{\pit}\sbr*{
    \rbr*{
      \frac{1}{\eta}\log\frac{\pi_\theta(y\mid{}x)}{\pit(y\mid{}x)}
      -r(x,y)+\btstar(x)
    }^2
  }.
  \label{eq:squared-loss-example}
\end{align}
This objective averages the squared residuals of \cref{eq:kkt} over $\pit$,
and can therefore be optimized without importance weights.
It is uniquely minimized by $\piplust$ and therefore implements \framework.
Although the objective has no explicit KL penalty,
it implicitly regularizes to $\pit$ in the log policy ratio.

\para{Overview}
We analyze squared-loss algorithms with objectives based on \cref{eq:squared-loss-example},
which differ in how they handle the unknown normalization function $\btstar(x)$.
Kimi-style objectives estimate a proxy baseline, while pairwise objectives eliminate $\btstar$ using paired responses.
We first establish conditions on the squared-loss error that ensure convergence under \framework (\cref{sec:mse}),
then use this to show that three practical algorithms are \covpres.\loose

\para{Kimi-style objectives: An estimated baseline (\cref{sec:kimi})}
Kimi-style objectives replace $\btstar(x)$ in \cref{eq:squared-loss-example}
with a \emph{baseline} $\bt(x)$, typically estimated using samples from $\pit$.
This gives the objective\loose
\begin{align}
  \Lkimi(\pi_\theta)
  \ldef{}
  \En_{\pit}\sbr*{
    \rbr*{
      \frac{1}{\eta}\log\frac{\pi_\theta(y\mid{}x)}{\pit(y\mid{}x)}
      -r(x,y)+\bt(x)
    }^2
  }.
  \label{eq:kimi-objective-intro}
\end{align}
The earliest uses of this formulation were, to our knowledge,
in Kimi K1.5 \citep{du2025kimi} and Kimi K2 \citep{team2025kimi}.
It has since been adopted in algorithms such as
\oapl \citep{ritter2026llms} and KARL \citep{chang2026karl}. %
These methods differ in their choice of baseline, which generally biases the learned policies away from $\piplust$.
We use \framework to analyze the two most common choices,
characterizing the additional samples and gradient steps needed to control this bias.
\begin{itemize}
\item \kimipg (\cref{sec:kimi}) uses a \emph{mean baseline} $\bt(x) = r(x,\pit)$,
the expected reward under $\pit$ \citep{du2025kimi,team2025kimi}.
The bias arises from how much $r(x,\pit)$ underestimates $\btstar$,
and we show that stronger regularization controls this error.

\item \kiminorm (\cref{app:kimi-norm}) uses the \emph{normalizer baseline},
which estimates $\btstar$ using responses from $\pit$
\citep{du2025kimi,cohen2025soft,ritter2026llms,chang2026karl}.
This estimator is biased but consistent, and we instead use additional responses to control its error.
\end{itemize}

\para{Pairwise objectives: Removing the baseline with paired responses (\cref{sec:rebel})}
Pairwise objectives eliminate $\btstar(x)$ by comparing two responses to the same prompt,
an approach popularized by \dpo \citep{rafailov2023dpo}.
We analyze \rebelon, a variant of \rebel \citep{gao2024rebel} tailored to online RL with observed rewards.
It minimizes the following objective with both responses drawn from $\pit$,
\begin{align}
  \Ltrebelon(\pi_\theta)
  \ldef{}
  \En_{x\sim\mu,\,y,y'\sim\pit(\cdot\mid{}x)}\sbr*{
    \rbr*{
      \frac{1}{\eta}\log\frac{\pi_\theta(y\mid{}x)}{\pit(y\mid{}x)}
      -\frac{1}{\eta}\log\frac{\pi_\theta(y'\mid{}x)}{\pit(y'\mid{}x)}
      -\rbr*{r(x,y)-r(x,y')}
    }^2
  }.
  \label{eq:rebel-objective-intro}
\end{align}
$\Ltrebelon$ is unbiased with respect to $\piplust$,
making \rebelon a theoretically clean instance of \framework.\loose

\para{Natural policy gradient}
\npg is a foundational policy optimization method with a squared-loss formulation
closely related to the objectives studied here.
We discuss these connections in \cref{app:npg}.

\subsection{Specializing \framework: A Mean-Squared Error Criterion}\label{sec:mse}

The algorithms in \cref{sec:pg} approximate the \framework target $\piplust$ in KL divergence.
For squared-loss algorithms, which fit log policy ratios rather than distributions,
we instead use the \emph{mean-squared error} ($\mse$),
defined below in terms of log-ratios centered under $\pit$.
Centering removes response-independent normalization terms that are handled differently across squared-loss algorithms,
yielding a common criterion for analysis.
For a fixed iteration $t$ and reference policy $\pit$, let
\begin{align}
  l_{t,\eta}^\pi(x,y)
  \ldef{}
  \frac{1}{\eta}\log\frac{\pi(y\mid{}x)}{\pit(y\mid{}x)}
  -
  \En_{y'\sim\pit(\cdot\mid{}x)}\sbr*{
    \frac{1}{\eta}\log\frac{\pi(y'\mid{}x)}{\pit(y'\mid{}x)}
  }
\end{align}
denote the centered log-ratios of a candidate policy $\pi$.
The mean-squared error is then
\begin{align}
  \mset(\pi)
  \ldef{}
  \En_{\pit}\sbr*{
    \rbr*{l_{t,\eta}^\pi(x,y) - (r(x,y)-r(x,\pit))}^2
  }.
  \label{eq:mse-definition}
\end{align}
The $\mse$ compares the centered log-ratios of $\pi$ with those of the target $\piplust$.
Indeed, centering \cref{eq:kkt} under $\pit$ gives $l_{t,\eta}^{\piplust}(x,y)=r(x,y)-r(x,\pit)$,
so $\mset(\pi) = \En_{\pit}\sbr{(l_{t,\eta}^\pi(x,y) - l_{t,\eta}^{\piplust}(x,y))^2}$.
Thus, $\piplust$ uniquely minimizes $\mset$ with zero error,
and $\mset(\pi)$ measures how well $\pi$ approximates $\piplust$.

\begin{algorithm}[t]
  \caption{\kimipglong (\kimipg)}
  \label{alg:kimi-main}
  \begin{algorithmic}[1]
    \Require \begingroup\raggedright
    \mbox{Base model $\piref$},
    \mbox{outer iterations $T$}, \mbox{regularization parameter $\eta>0$},
    \mbox{prompt batch size $N$}, \mbox{response group size $G$},
    \mbox{inner steps $K$}, \mbox{and learning rate $\alpha$}.\par
    \endgroup
    \State Initialize $\pit[0]=\piref=\pi_{\theta_0}$.
    \For{$t=0,1,\ldots,T-1$}
      \State Draw a batch\label{line:sampling-kimi-main}
      $\cD_t=\cbr*{(x_i,y_i,r_i,\bar r_i)}_{i=1}^N$, where
      $x_i\overset{\mathrm{iid}}{\sim}\mu$ and $y_i,y'_{i,1},\ldots,y'_{i,G}\sim\pit(\cdot\mid{}x_i)$ independently.
      \Statex[1] Set $r_i=r(x_i,y_i)$ and estimate the mean baseline by
      $\bar r_i=\frac1G\sum_{j=1}^G r(x_i,y'_{i,j})$.
      \State Define the empirical objective\label{line:objective-kimi-main}
      \begin{align}
        \hLkimi(\pi_\theta)
        \ldef{}
        \frac{1}{N}\sum_{i=1}^N
        \rbr*{
          \frac{1}{\eta}\log\frac{\pi_\theta(y_i\mid x_i)}
          {\pit(y_i\mid x_i)}
          -\rbr*{r_i-\bar r_i}
        }^2.
      \end{align}
      \State Initialize $\theta_t^0=\thetat$ and $\pi_t^0=\pit$.\label{line:inner-start-kimi-main}
      \For{$k=0,1,\ldots,K-1$}
        \State Compute the empirical gradient $\nabla\hLkimi(\pi_t^k)$
        (e.g., via backpropagation).
        \State Update
        $\theta_t^{k+1}=\theta_t^k
        -\alpha\nabla\hLkimi(\pi_t^k)$ and
        $\pi_t^{k+1}=\pi_{\theta_t^{k+1}}$.
      \EndFor
      \State Set $\thetatp=\theta_t^K$ and $\pitp=\pi_t^K$.\label{line:inner-end-kimi-main}
    \EndFor
    \State \Return $\pifin=\pit[T]$.
  \end{algorithmic}
\end{algorithm}

Later, we will show how the Kimi-style (\cref{eq:kimi-objective-intro}) and pairwise objectives (\cref{eq:rebel-objective-intro})
control $\mset(\pi)$ and yield \covpres algorithms.
These results rely on the following theorem, a counterpart to \cref{thm:kl-regret-main},
that ensures convergence and coverage preservation
when $\mset(\pitp)$ is sufficiently small at every iteration.
\begin{theorem}[\framework: Mean-squared error criterion]
  \label{thm:mse-regret-main}
  Given any $\eta>0$ and any sequence of policies
  $\cbr*{\pit}_{t=0}^T$ with $\pit[0]=\piref$ and $\pit>0$,
  suppose that
  \begin{align}
    \mset(\pitp)
    \leq
    \frac{1}{4\eta^2T^2\rbr*{\Cresp\log(eT)+e^{\eta S}}},
    \qquad
    \forall 0\leq t<T,
    \label{eq:framework-mse-tolerance}
  \end{align}
  where $S\ldef{}\max_{0\leq t<T}\max_{x,y}
  \abr*{l_{t,\eta}^{\pitp}(x,y)-(r(x,y)-r(x,\pit))}$.
  Then
  \mbox{$\Jstar-J(\pit[T])\leq\frac{\log\rbr*{2\Cresp}}{\eta T}$},
  and with probability at least
  $1-\frac{7}{8\sqrt{\log(eT)}}$ over $x\sim\mu$,
  $ \pitp(\Ystarx\mid{}x) \geq \frac{1}{2}\piref(\Ystarx\mid{}x)$ for all $0 \leq t < T$.
\end{theorem}
Because $\mset$ is measured under the sampling policy $\pit$,
squared-loss algorithms can drive it down using a pre-collected batch without importance weights.
However, the inner MSE tolerance shrinks with the term $e^{\eta S}$ in its denominator (\cref{eq:framework-mse-tolerance}),
where $S$ bounds the magnitude of the residual error.
This cost arises from distribution shift when translating small $\mset(\pitp)$
into guarantees for the learned policy $\pitp$.
In our results, sufficiently strong regularization ensures $e^{\eta S}=O(\Cresp)$,
making this term relatively harmless.

We now apply this criterion to squared-loss algorithms, starting with the Kimi-style objectives.\loose

\subsection{Instantiating \framework: Kimi-Style Policy Optimization Algorithms}
\label{sec:kimi}

We first study the Kimi-style objective (\cref{eq:kimi-objective-intro})
with the \emph{mean baseline} $\bt(x)=r(x,\pit)$ used in Kimi K2
\citep{du2025kimi,team2025kimi}.
Here $r(x,\pit)=\En_{y'\sim\pit(\cdot\mid{}x)}\sbr*{r(x,y')}$
is the expected reward under $\pit$ for prompt $x$.
The resulting algorithm, \kimipg (\cref{alg:kimi-main}),
uses the following objective in each outer iteration $t$,
\begin{align}
  \Lkimi(\pi_\theta)
  =
  \En_{\pit}\sbr*{
    \rbr*{
      \frac{1}{\eta}\log\frac{\pi_\theta(y\mid{}x)}{\pit(y\mid{}x)}
      -r(x,y)+r(x,\pit)
    }^2
  }.
  \label{eq:kimi-mean-objective}
\end{align}
To minimize this objective, \kimipg uses a pre-collected batch of $N$ prompts
and $G+1$ responses per prompt from $\pit$.
For each prompt, it estimates the mean baseline by averaging the rewards of $G$ responses
(Line~\ref{line:sampling-kimi-main}),
then combines this estimate with the remaining response to form the empirical objective $\hLkimi$
(Line~\ref{line:objective-kimi-main}).
Starting from $\pit$, the inner loop takes $K$ full-batch gradient descent steps on $\hLkimi$
and sets $\pitp=\pitk[K]$.%
\footnote{\hypertarget{Hfootnote.\theHfootnote}{}
This procedure differs from Kimi~\citep{du2025kimi,team2025kimi}
only in replacing Kimi K2's Muon optimizer with gradient descent
and using separate responses for baseline estimation and gradient computation.}

The following result shows that \kimipg is \covpres under \cref{asm:softmax-linear}.
\begin{corollary}[\kimipg is \covpres]
  \label{cor:mean-kimi-main}
  Suppose that \cref{asm:softmax-linear} holds, and let $B \ldef \Bfeat\Bstar$.
  For any $\veps\in(0,\Rmax]$ and $\delta\in(0,1)$,
  in \cref{alg:kimi-main}, set $G=1$ and
  $\eta=\wt\Theta\rbr*{\sqrt{\veps}/(\Cresp^{1/4}B^{3/2})}$.
  Then there exist choices of $\alpha$, $N$, and $K$ (see \cref{cor:mean-kimi-geometric-regret}),
  with $T$ specified by \cref{eq:kl-conditions},
  such that \kimipg returns a policy $\pifin$ satisfying $\Jstar-J(\pifin)\le\veps$
  with probability at least $1-\delta$, using
  \begin{align}
    \Nstep(\veps,\delta)
    = \wt O\rbr*{\frac{\Cresp^{5/4}B^{7/2}}{\veps^{7/2}}}
    \qquad\text{and}\qquad
    \Nsample(\veps,\delta)
    = \wt O\rbr*{\frac{\Cresp^{5/4}B^{7/2}\rbr*{d+\log(1/\delta)}}{\veps^{7/2}}}.
    \label{eq:complexity-kimi-main}
  \end{align}
  Here, $\Nstep=KT$ and $\Nsample=N(G+1)T$.
\end{corollary}
To our knowledge, \cref{cor:mean-kimi-main} gives the first end-to-end global convergence guarantee for a Kimi-style algorithm.
As in \cref{sec:pg}, this guarantee relies on reference resets,
which \kimipg implements by updating $\pit$ in the log policy ratio.
  This design choice is employed by both Kimi K1.5 and Kimi K2 \citep{du2025kimi,team2025kimi},
  as well as specific instantiations of \oapl \citep{ritter2026llms} and KARL \citep{chang2026karl},
  all of which can be viewed as \framework algorithms.
  By contrast, \astarpo \citep{brantley2026accelerating} and SPO \citep{cohen2025soft}
  retain a fixed reference policy in the denominator of the log-ratio and do not implement \framework.\loose

We highlight specific features of this result and our analysis below.

\para{Bias from the baseline function}
Our bounds on samples and gradient steps scale as $\wt O(\veps^{-7/2})$ for \kimipg,
compared with $\wt O(\veps^{-3})$ for \ppolitenc (\cref{cor:pg-main}).
This extra dependence on $\veps$ arises because \cref{cor:mean-kimi-main} uses stronger regularization
($\eta = \wt \Theta(\sqrt{\veps})$ rather than $\eta = 1/B$)
and thereby requires more outer iterations.
\kimipg needs stronger regularization to control the bias introduced by the mean baseline,
as the following decomposition shows (\cref{lem:mean-kimi-objective-decomposition}),\loose
\begin{align}
  \Lkimi(\pi)
  =
  \mset(\pi)
  + \frac{1}{\eta^2} \En_{x\sim\mu}\sbr*{\Dkl{\pit(\cdot\mid{}x)}{\pi(\cdot\mid{}x)}^2}.
  \label{eq:kimi-mean-decomposition-main}
\end{align}
The first term is the MSE from \cref{thm:mse-regret-main} that \kimipg must control.
The second favors policies close to $\pit$,
consistent with the implicit regularization identified by \citet{xu2026approximation},
so even exact minimization of $\Lkimi$ may not recover the target $\piplust$.
We show that this bias increases $\mset$ by $O(\eta^4B^6)$ (\cref{lem:mean-kimi-geometric-loss-comparison})---%
an error that is controlled by stronger regularization.
Indeed, \cref{cor:mean-kimi-main} holds for any $G \ge 1$,
and choosing $\eta = \wt\Theta(\sqrt{\veps})$ keeps the bias small enough to meet the MSE criterion in \cref{thm:mse-regret-main}.\loose

\para{Optimization landscape of Kimi-style objectives}
Although the mean baseline introduces bias, it makes the \kimipg objective globally convex in $\theta$
under \cref{asm:softmax-linear} (\cref{lem:mean-kimi-global-curvature}).
By contrast, the \ppolitenc objective is unbiased
but satisfies only a local form of concavity around $\pit$ (\cref{eq:pg-concave}).
The curvature of Kimi-style objectives depends on the baseline choice (\cref{lem:kimi-local-curvature});
the \kiminorm objective studied below is likewise only locally convex.

\para{Additional result: \kiminorm estimates $\btstar$}
In \cref{app:kimi-norm}, we analyze \kiminorm (\cref{alg:kimi-norm}),
which optimizes the Kimi-style objective with baseline $\bt = \btstar$.
As in prior work \citep{du2025kimi,cohen2025soft,ritter2026llms,chang2026karl},
the empirical objective uses a plug-in estimate of $\btstar(x)$
computed over $G$ responses sampled from $\pit$,
  \begin{align}
    \widehat b_t(x)
    \ldef{}
    \frac{1}{\eta}\log
    \rbr*{\frac{1}{G}\sum_{j=1}^G e^{\eta r(x,y'_j)}}.
  \end{align}
This estimate is consistent but biased,
and increasing $G$ is what controls the resulting error (\cref{lem:kimi-norm-estimation-bias}).
\Cref{cor:kimi-norm-regret} shows that \kiminorm is \covpres under \cref{asm:softmax-linear}
with $G=\wt\Theta(1/\veps)$ and $\eta=\Theta(1/B)$.
Compared with \kimipg, which takes $\eta=\wt\Theta(\sqrt{\veps})$ (\cref{cor:mean-kimi-main}),
\kiminorm uses significantly weaker regularization but more responses for accurate baseline estimation.
This saves a factor of $1/\sqrt{\veps}$ in $\Nstep$
but increases $\Nsample$ by the same factor (full comparison in \cref{tab:kimi-baseline-comparison}).

\subsection{Instantiating \framework: Pairwise Squared-Loss Algorithms (\rebelon)}
\label{sec:rebel}

Finally, we study \emph{pairwise squared-loss algorithms},
which do not require baseline functions.
Our analysis focuses on \rebelon, a variant of \rebel \citep{gao2024rebel}
that draws two independent responses from $\pit$ for each prompt.
At outer iteration $t$, the objective is
\begin{align}
  \Ltrebelon(\pi_\theta)
  \ldef{}
  \En_{x\sim\mu,\,y,y'\sim\pit(\cdot\mid{}x)}\sbr*{
    \rbr*{
      \frac{1}{\eta}\log\frac{\pi_\theta(y\mid{}x)}{\pit(y\mid{}x)}
      -\frac{1}{\eta}\log\frac{\pi_\theta(y'\mid{}x)}{\pit(y'\mid{}x)}
      -\rbr*{r(x,y)-r(x,y')}
    }^2
  }.
  \label{eq:rebel-objective}
\end{align}
To optimize this objective, \rebelon (\cref{alg:dpo}) draws a batch of $N$ prompts
and two independent responses per prompt from $\pit$.
It then takes $K$ full-batch gradient descent steps on the empirical objective starting from $\pit$,
and sets $\pitp=\pitk[K]$.

The following result shows that \rebelon is \covpres under \cref{asm:softmax-linear}.

\begin{corollary}[\rebelon is \covpres]
  \label{cor:rebel-main}
  Suppose that \cref{asm:softmax-linear} holds, and let
  $B \ldef{} \Bfeat\Bstar$.
  In \cref{alg:dpo}, set $\eta=\log(2\Cresp)/(2B)$.
  Then, for any $\veps\in(0,\Rmax]$ and $\delta\in(0,1)$,
  there exist choices of $\alpha$, $N$, and $K$ (see \cref{cor:dpo-regret}),
  with $T$ specified by \cref{eq:kl-conditions},
  such that \rebelon returns a policy $\pifin$
  satisfying $\Jstar-J(\pifin)\le\veps$
  with probability at least $1-\delta$, using
  \begin{align}
    \Nstep(\veps,\delta)
    = \wt O\rbr*{\frac{\Cresp B^3}{\veps^3}}
    \qquad\text{and}\qquad
    \Nsample(\veps,\delta)
    = \wt O\rbr*{\frac{\Cresp B^3}{\veps^3}}.
    \label{eq:rebel-complexity}
  \end{align}
  Here, $\Nstep=KT$ and $\Nsample=2NT$.
  $\Nsample$ depends only logarithmically on $d$ and $1/\delta$.
\end{corollary}

\begin{remark}[Relation to \rebel]
\citet{gao2024rebel} allow the second response to be sampled from a general distribution, such as $\piref$,
though they use $\pit$ for their natural language experiments.
Their analysis assumes that the sampling distribution supplies the necessary coverage,
so it does not by itself show that \rebelon is \covpres.
The key to our result is showing that $\pit$ itself provides coverage
by preserving the base model's coverage throughout training.
\end{remark}

\para{\rebelon is unbiased}
The \rebelon objective directly controls the mean-squared error, $\Ltrebelon(\pi)=2\mset(\pi)$ (\cref{lem:rebel-mse}).
It is therefore uniquely minimized by the target $\piplust$ and unbiased.
\rebelon circumvents the baseline estimation in Kimi-style algorithms
by taking the difference of residuals for paired responses,
thereby canceling the shared normalization function $\btstar(x)$.
\rebelon is also a particularly clean instance of \framework under \cref{asm:softmax-linear},
because every empirical objective is a globally convex quadratic in $\theta$,
on which $\piplust$ has zero loss (\cref{lem:dpo-convexity}).

\para{Comparison: Pairwise algorithms with a fixed reference policy}
  Reference resets distinguish \rebelon from pairwise methods that use a fixed reference,
  such as \dpo \citep{rafailov2023dpo}, IPO \citep{azar2024ipo},
  and their online variants \citep{calandriello2024online}.
  For example, online \dpo draws response pairs from $\pit$
  but retains $\piref$ in the denominator of the log-ratio.
  Consider its squared-loss formulation for observed rewards,
  \begin{align}
    L_t^{\texttt{Online-DPO}}(\pi_\theta)
    \ldef{}
    \En_{x\sim\mu,\,y,y'\sim\pit(\cdot\mid{}x)}\sbr*{
      \rbr*{
        \frac{1}{\eta}\log\frac{\pi_\theta(y\mid{}x)}{\piref(y\mid{}x)}
        -\frac{1}{\eta}\log\frac{\pi_\theta(y'\mid{}x)}{\piref(y'\mid{}x)}
        -\rbr*{r(x,y)-r(x,y')}
      }^2
    }.
    \label{eq:online-dpo-objective}
  \end{align}
  This objective differs from $\Ltrebelon$
  only in using $\piref$ instead of $\pit$ as the reference policy.
Since it has the same solution as the KL-regularized objective with fixed reference $\piref$ in \cref{eq:fixed-ref-objective},
  it is prone to the same coverage collapse highlighted by our lower bound (\cref{thm:lower-bound-main}).

\section{Conclusion}\label{sec:conclusion}
We develop a theory of policy optimization in post-training,
where the goal is to learn efficiently from a capable base model.
We formalize this goal by asking which algorithms are \covpres,
with learning costs governed by the base model's response coverage,
a criterion that proves discerning for algorithm design.
Basic methods such as \reinforce can suffer coverage collapse,
converging exponentially slowly despite a base model with good coverage.
\framework, a meta-algorithm built around reference resets, addresses this failure and yields \covpres algorithms.
We apply it to establish global convergence of
stylized \ppo and \grpo variants and squared-loss methods, including Kimi-style algorithms.

Our theory provides insights that apply to policy optimization more generally.
\begin{itemize}[leftmargin=*,itemsep=2pt,topsep=3pt]
\item \para{Optimization benefits of regularization to the reference}
Optimization in RL is challenging because the gradient signal depends on the learner's distribution,
whose coverage can deteriorate during training and thereby slow convergence to a given target.
If we instead aim to optimize the policy in the proximity of a reference policy,
coverage is controlled and
local improvement becomes relatively inexpensive to realize.
Updating the reference is what translates these local improvements into global optimality over successive resets.
The reference policy therefore serves to anchor optimization,
a role distinct from its use in importance weighting or batch reuse.
In \framework, KL divergence defines the proximal region around the reference,
but other interventions or divergences may play a similar role.

These optimization benefits also translate to improved sample efficiency, even if the batch is not reused in the inner loop.
With reuse, algorithms adopting the \framework structure can improve more with a given batch in the inner loop,
and may therefore require significantly fewer generation rounds.
This is of particular importance in RL, where sampling can be the computational bottleneck~\citep{qin2026seer,he2026rhymerl,hu2025openrlhf}.\loose

\item \para{Implications for practical methodology}
These insights also provide principles underpinning practical methodology.
They suggest that the inner loop should be viewed as a local improvement step on the reference.
This step is more stable the more strongly it is regularized to the reference
and should be solved to a sufficient level to mitigate error compounding.
In contrast, it is common to use 1--4 gradient steps for the inner loop~\citep{shao2024deepseekmath,hu2025openreasonerzero,liu2025prorl,lambert2024tulu}
and to either omit regularization~\citep{yu2025dapo,hu2025openreasonerzero,team2025olmo3}
or treat it similarly to fixed-reference regularization~\citep{ouyang2022instructgpt,lambert2024tulu}.
Our theory also suggests that hyperparameters such as the clipping thresholds, number of inner steps, and batch size
depend on the regularization parameter and should be tuned together.

\item \para{Design choices within \framework}
Finally, while we unify different post-training algorithms under \framework,
we have also highlighted differences in the behavior of their objectives and in their sampling and optimization costs,
which may be consequential in practice and merit empirical study.
More broadly, \framework offers a starting point for designing and analyzing algorithms
as new practical considerations arise.
\end{itemize}

\para{Limitations and future work}
Our end-to-end guarantees are limited to softmax linear policies,
which we view as a useful but simplified model of learning with pretrained features.
While we expect the core insights to be relevant more broadly,
richer policy parameterizations may introduce challenges beyond the scope of our framework.\loose

In addition,
our theory focuses on the one-shot reasoning setting~\citep{lambert2024tulu,guo2025deepseekr1}
and treats each response as a single unit.
In reality, responses are token sequences that vary in length and structure,
and policy ratios over full responses can be large.
These practical considerations can significantly affect training dynamics~\citep{zheng2025gspo,tang2025pitfalls,shah2025comedy,yu2025dapo,liu2025understanding},
and how they interact with our theory is an interesting open question.
Beyond the one-shot setting, our results serve as a basis for understanding policy optimization in multi-step reasoning,
such as agentic tasks that involve many rounds of interaction with an environment~\citep{team2025kimi,chang2026karl},
and extending them is a natural next step.

Lastly, response coverage is only one basic notion of base model quality,
and our central question addresses what we view as the first-order consideration.
Ideally, post-training would do more than amplify existing behaviors~\citep{zhao2025echo,yue2025rlbeyond}.
Understanding when this is possible~\citep{liu2025prorl,yuan2025fx}
and which properties of the base model matter~\citep{chen2025coverage,foster2025good}
is an important direction for future work.

\para{Statement on AI usage}
The core results of this paper,
including coverage collapse (\cref{thm:lower-bound-main})
and coverage preservation under \framework (\cref{thm:kl-regret-main}),
were developed by the authors.
AI models (GPT-5.5 and GPT-5.6) were used to greatly speed up the exploration
of how to instantiate these results for specific algorithms.
AI tools (Claude and GPT-6) were also used for writing assistance, including drafting, editing, and generating figures.
The authors take full responsibility for all content in the final version of the paper,
including its results, proofs, and writing.

\subsection*{Acknowledgments}
We thank Alex Damian and Wen Sun for illuminating discussions that helped shape this paper.
This work was enabled in part by a gift from the Chan Zuckerberg Initiative Foundation to establish the Kempner Institute for the Study of Natural and Artificial Intelligence.
Part of this research was performed while AH was visiting the Institute for Mathematical and Statistical Innovation (IMSI), which is supported by the National Science Foundation (Grant No. DMS-2425650),
and while visiting the Simons Institute for the Theory of Computing.
AH gratefully acknowledges their support.

\clearpage
\bibliography{refs}

\clearpage
\appendix
\crefalias{section}{appendix}
\crefalias{subsection}{appendix}
\crefalias{subsubsection}{appendix}
\stopcontents[main]
\startcontents[appendix]
\clearpage
\section*{Appendix Contents}
\printcontents[appendix]{}{0}[2]{}
\clearpage

\section{Related Work}\label{app:related-work}
\para{Convergence theory for policy gradient methods}
A large body of work establishes global convergence of policy gradient methods,
despite the nonconcavity of the expected reward.
This includes results for tabular and softmax linear policies with exact gradients
\citep{agarwal2020theory,mei2020softmaxpg,bhandari2024global},
natural policy gradient (\npg) \citep{kakade2001naturalpg,agarwal2020theory,cen2022fast},
and policy mirror descent \citep{shani2020adaptive,lan2023pmd,xiao2022rates},
with extensions to log-linear and neural policies and to stochastic gradients
\citep{yuan2023linearnpg,cayci2024convergence,wang2020neuralpg,liu2019neuraltrpo,klein2024beyond,mei2023stochasticbandits,mei2024smallsteps,asad2025fast,lattimore2026lyapunov}.

These guarantees differ in how they account for initialization.
Some specialize to a uniform initial policy \citep{agarwal2020theory,mei2020softmaxpg,lan2023pmd},
while others depend on properties of the policy iterates, such as the smallest
probability assigned to an optimal action during training
\citep{mei2020softmaxpg,wang2020neuralpg,klein2024beyond,asad2025fast,cayci2024convergence,li2023inherently}.
Such quantities need not remain favorable even when the base model has good
coverage (\cref{thm:lower-bound-main}).
Exact mirror descent can instead depend on the initial policy through its KL
divergence to an optimal policy \citep{xiao2022rates}; related bounds hold for
log-linear \npg with compatible function approximation \citep{yuan2023linearnpg}.

Inexact mirror-descent analyses also give sample-complexity guarantees.
\citet{xiao2022rates} uses a generative model to control value-estimation error,
while \citet{shani2020adaptive} use uniform action sampling and a restart
distribution that covers the states visited by an optimal policy.
\citet{lan2023pmd} analyzes both generative sampling and single-trajectory
sampling under exploration conditions, as well as approximate subproblem solutions.
\citet{yuan2023linearnpg} give complexity bounds for solving the compatible regression
problems under concentrability assumptions and lower bounds on feature covariance
or Fisher matrices.
Our focus is the dependence on the base model's response coverage.
In the contextual-bandit setting, our KL and MSE criteria ensure that approximate
updates preserve this coverage (\cref{thm:kl-regret-main,thm:mse-regret-main}),
and our algorithm analyses bound the total samples and gradient steps needed
to implement them (\cref{cor:pg-main,cor:mean-kimi-main,cor:rebel-main}).

\para{Analogy to first-order convex optimization}\label{app:analogy-convex-optimization}
The \covpres criterion is motivated by the dependence on initialization in
classical convex optimization guarantees.
For an $L$-smooth convex function $f$, Euclidean gradient descent with learning rate
$1/L$ satisfies $f(x_T)-f(x_\star)\le L\nbr{x_0-x_\star}_2^2/(2T)$,
where $x_\star$ is a minimizer \citep{bubeck2015convex}.
This bound depends on the initial distance to an optimum rather than on
uncontrolled properties of later iterates.
The exact \framework guarantee in \cref{eq:mirror-descent} has a similar form,
with $\log\Cresp$ measuring the quality of the initialization.
Our question is whether implementable policy optimization methods retain a
favorable dependence on the base model after accounting for both sampling and
numerical optimization.
\Cref{thm:lower-bound-main} shows why this does not follow from good initialization alone.

\para{Lower bounds for policy gradient methods}
\citet{mei2020gravitational} show that softmax policy gradient with exact gradients
can remain near a suboptimal deterministic policy for a number of gradient steps
inversely proportional to the initial probability of the optimal action and the reward gap.
This bound grows as the initial coverage worsens.
In contrast, \cref{thm:lower-bound-main} starts with constant coverage ($\Cresp\le6$),
but coverage deteriorates while reward increases.
Attaining and maintaining $\veps$-optimality then requires a number of gradient steps
exponential in $1/\sqrt{\veps}$, with or without KL regularization to the base model.
\Cref{sec:lower-bound} compares this result to other policy-gradient lower bounds.

Other results identify limitations of exploration from a weak base model.
\citet{foster2025good} prove coverage-dependent lower bounds on sampling-oracle
complexity for KL-regularized alignment, distinguishing the number of sampling-oracle
calls from the number of reward queries.
\citet{mousavi2026post} establish reward-query lower bounds for post-training
linear autoregressive models, with the difficulty governed by the base model's
likelihood of correct responses.
In reasoning-tree models, \citet{liu2025uft} prove an exponential query lower bound
in the tree depth without hints and a polynomial guarantee with hints.
\citet{bu2025curriculum} establish an exponential-to-polynomial sample-complexity
separation between a direct gradient update and staged curriculum updates.
\citet{huang2026edge} study a different barrier in compositional reasoning,
showing that long horizons can produce exponentially small policy-gradient signals
and that training on a mixture of difficulties can enable progress from easier
to harder tasks.
Our construction instead isolates a failure of exact-gradient optimization
in a tabular bandit despite good initial coverage.

\para{Coverage in post-training}
The coverage of the base model governs the success of several post-training
and inference-time methods.
For \bonlong sampling, the chance of finding a high-reward response depends
on the probability that the base model assigns to such responses
\citep{brown2024monkeys,beirami2025theoretical,huang2025bestofn}.
\citet{huang2024self} formalize self-improvement as \emph{sharpening} the model
toward high-scoring responses and give coverage-dependent guarantees for
supervised finetuning on selected generations, as used in rejection-sampling
methods \citep{zelikman2022star,singh2024beyond,dong2023raft,xiong2025minimalist}.
\citet{chen2025coverage} explain why pretraining can provide this favorable
initialization, showing that next-token prediction implicitly promotes coverage.

Reference-policy quality also enters statistical analyses of RL post-training.
\citet{zhao2024sharp} give coverage-dependent sample bounds for KL-regularized
contextual bandits and RLHF, while the bounds for $Q\sharp$ depend on the
variance of returns under the reference policy \citep{zhou2025qsharp}.
The analyses of \rebel and \astarpo likewise relate policy performance to
regression accuracy or online-learning regret, with coverage conditions on the
sampling or reference policy \citep{gao2024rebel,brantley2026accelerating}.
These guarantees use optimization or online-learning oracles, or assume a
bound on regression error.
Our algorithm analyses additionally bound the number of gradient steps
needed to solve the subproblems and show how response coverage is preserved
throughout training.

\para{Trust regions and mirror descent}
Limiting how far each update moves the policy is a classical principle in
policy optimization.
Conservative policy iteration \citep{kakade2002cpi} and \trpo
\citep{schulman2015trpo} relate policy improvement to a surrogate objective
and the distance between consecutive policies.
\ppo \citep{schulman2017ppo} replaces an explicit trust-region constraint
with a clipped surrogate or an adaptive KL penalty.
\citet{neu2017unified} interpret entropy-regularized methods, including an exact
version of \trpo, through mirror descent and dual averaging,
and \citet{geist2019regularized} develop a theory of regularized MDPs that
includes these connections.

The KL specialization of the tabular update analyzed by
\citet{shani2020adaptive} maximizes expected advantage minus a KL penalty
toward the current policy; subsequent work sharpens policy mirror descent
rates \citep{lan2023pmd,xiao2022rates}.
Closest to our algorithms, on-policy mirror descent policy optimization
\citep[MDPO;][]{tomar2022mdpo} approximately solves each KL-regularized
subproblem using several gradient steps on an importance-weighted objective
computed from current-policy samples.
In contextual bandits, this resembles \ppolitenc (\cref{sec:pg-algo}).
For \ppolitenc, our analysis bounds the number of samples and gradient steps under
softmax linear parameterization and proves coverage preservation.
\citet{lan2023pmd} uses the name approximate policy mirror descent for a
variant with a diminishing initial-policy KL perturbation of the reward and update.
In our usage, ``approximate'' refers to inexact solutions of the KL-regularized
subproblems; Lan also separately studies inexact subproblem solutions.

\para{Reference resets in practice}
KL regularization toward a fixed reference, typically the initial model,
is commonly used to limit reward-model overoptimization
\citep{stiennon2020learning,ouyang2022instructgpt,gao2023scaling}.
Several training recipes instead update the reference during training.
DeepSeekMath resets it when retraining the reward model
\citep{shao2024deepseekmath}, while ProRL and DeepSeek-R1 use periodic resets
\citep{liu2025prorl,guo2025deepseekr1}.
EMA Policy Gradient uses an exponential moving average of the policy
parameters as its reference and studies both its stability and empirical
benefits \citep{zhang2026ema}.
Our analysis studies resetting the reference to the current policy at each outer iteration.
It shows how this choice allows strong regularization to preserve coverage
without stalling progress (\cref{thm:kl-regret-main}).

\para{Training without KL regularization}
Successful reasoning pipelines such as DAPO and Open-Reasoner-Zero omit the
KL penalty and use clipped policy-gradient objectives
\citep{yu2025dapo,hu2025openreasonerzero}.
Other interventions promote exploration through forward-KL regularization
\citep{nguyen2025boundary,deng2025unlocking}, entropy-aware token updates
\citep{cui2025entropy,wang2025beyond}, or protection of low-probability tokens
\citep{huang2025lowprobability}.
Our clipping analysis identifies a statistical benefit of clipping.
In \ppolite (\cref{sec:clip-pg}), clipping controls the contribution of large
importance weights to the empirical objective, allowing sample-efficient
learning without a uniform bound on policy ratios.
This need not keep consecutive policies close.
Empirical studies find heavy-tailed gradients in \ppo
\citep{garg2021heavytailedppo} and violations of the nominal ratio limits
\citep{engstrom2020implementation}; alternatives include early stopping
\citep{sun2022maynotneedppo} and sequence-level clipping \citep{zheng2025gspo}.
Our guarantee for \ppolite still uses KL regularization toward the current
policy. Whether clipping alone can ensure coverage preservation remains open.

\para{Squared-loss objectives}
Kimi K1.5 and K2 regress the log-ratio between the new and current policies
onto rewards after subtracting a baseline \citep{du2025kimi,team2025kimi}.
K1.5 considers both a sampled log-partition estimate and the sample mean reward,
while K2 uses the sample mean.
Sampled log-partition estimates also appear in \astarpo, \oapl, KARL,
and SPO \citep{brantley2026accelerating,ritter2026llms,chang2026karl,cohen2025soft}.
\citet{xu2026approximation} characterize the implicit regularization induced
by the mean baseline and report greater training stability than with a sampled
log-partition baseline in their math-reasoning experiments.
Our analysis in \cref{app:kimi-shared,app:kimi,app:kimi-norm} quantifies how
baseline choice affects objective bias, curvature, and the sample and gradient step
complexities under softmax linear parameterization.

These methods also differ in the reference policy used in the log-ratio.
Kimi K1.5 and K2 use the current policy, while \oapl and KARL
periodically replace the reference with the learned policy.
The original formulations of \astarpo and SPO instead retain a fixed reference.
Related single-response regression methods DRO \citep{richemond2024offline}
and QRPO \citep{matrenok2025quantile} also retain a fixed reference;
DRO learns a value baseline, whereas QRPO uses quantile rewards
to obtain an analytic partition function.
Thus the use of online samples alone does not determine whether an algorithm
implements reference resets.

Pairwise regression eliminates the baseline by fitting differences of log
policy ratios to differences in rewards.
\rebel \citep{gao2024rebel} and its multi-turn extension \refuel
\citep{gao2025regressing} take these ratios with respect to the current policy.
Related preference losses in \dpo and IPO
\citep{rafailov2023dpo,azar2024ipo} use a fixed reference.
The online IPO formulation of \citet{calandriello2024online} retains
this reference in the loss while sampling fresh responses from the current policy.
By contrast, DNO \citep{rosset2024dno} uses the current policy as the
reference for each batch of updates.
This distinction separates fitting a target defined by a fixed reference
from the successive policy-improvement targets analyzed in \cref{sec:rebel}.

\section{Technical Tools: Exact Mirror Descent and Gradient Methods}\label{app:technical-tools}
This section collects standard optimization results that our analysis builds on.
\begin{itemize}
  \item \cref{app:population-gd} includes standard gradient descent convergence guarantees for convex and smooth objectives.
  \item \cref{app:population-pg} proves convergence of exact policy gradient ascent
  for the KL-regularized objective under \cref{asm:softmax-linear}.
  Although the expected reward $J(\pi_\theta)$ is nonconcave in $\theta$,
  forms of local concavity (\cref{lem:pg-local-convex}) and local smoothness (\cref{lem:pg-local-smooth}) hold.
  With these properties, the rest of the proof closely follows the gradient descent analysis.
  \item \cref{app:population-mirror-descent} establishes convergence of exact policy mirror descent in RL,
  which provides the basis for the \frameworklong framework results in \cref{sec:framework}.
  \item \cref{app:softmax-linear} collects basic properties of softmax linear policies.
\end{itemize}

\para{Notation}
Throughout this section, we use $\alpha > 0$ to denote the learning rate,
and we always use exact gradients.

\subsection{Exact Gradient Descent}\label{app:population-gd}

This section develops standard gradient descent analysis for smooth and convex functions.
In this subsection, $\cX$ denotes the optimization domain.
We consider the optimization problem of finding
\begin{align}
  x_\star\in\argmin_{x\in\cX} f(x),
\end{align}

and provide convergence analysis for population gradient descent with a fixed learning rate $\alpha>0$,
when the function $f : \cX \rightarrow \bbR$ is convex and smooth.
For the remainder of this subsection, we assume that
$\cX\subseteq\bbR^d$ is a convex set with $x_\star \in \cX$,
and $f$ is differentiable, convex, and smooth.

\begin{definition}[Convex function]
  A function $f:\cX\to\bbR$ is convex if, for every $x,y\in\cX$ and
  $\lambda\in[0,1]$, it holds that
  $ f\rbr*{\lambda x+(1-\lambda)y} \le \lambda f(x)+(1-\lambda)f(y) $.
  When $f$ is differentiable, this is equivalent to
  \begin{align}
    f(y)
    \ge
    f(x)+\inner*{\nabla f(x),y-x},
    \qquad x,y\in\cX.
  \end{align}
\end{definition}

\begin{definition}[Smooth function]
  A differentiable function $f:\cX\to\bbR$ is $L$-smooth with respect
  to the Euclidean norm if $L>0$ and
  \begin{align}
    \nbr*{\nabla f(x)-\nabla f(y)}_2
    \le
    L\nbr*{x-y}_2,
    \qquad x,y\in\cX.
  \end{align}
  Equivalently, $L$-smoothness gives the quadratic upper bound
  \begin{align}
    f(y)
    \le
    f(x)
    +
    \inner*{\nabla f(x),y-x}
    +
    \frac{L}{2}\nbr*{y-x}_2^2,
    \qquad x,y\in\cX.
  \end{align}
\end{definition}

\para{Population gradient descent}
Given an initial point $x_0\in\cX$ and a learning rate $\alpha > 0$,
run population gradient descent for $T\ge1$ gradient steps using the update

\begin{equation}\label{eq:population-gd-update}
  x_{t+1}
  =
  x_t-\alpha\nabla f(x_t),
  \qquad t=0,\ldots,T-1.
\end{equation}

Assume that the iterates remain in $\cX$.

\begin{theorem}[Gradient descent convergence]
  \label{thm:population-gd-convergence}
  Suppose that $f$ is convex and $L$-smooth.
  Fix $T\ge1$ and a learning
  rate $0<\alpha\le 1/L$.
  Then the iterates in \cref{eq:population-gd-update} satisfy
  \begin{align}
    f(x_T)-f(x_\star)
    \le
    \frac{\nbr*{x_0-x_\star}_2^2}{2\alpha T}.
  \end{align}
  In particular, choosing $\alpha=1/L$ gives
  $ f(x_T)-f(x_\star) \le \frac{L\nbr*{x_0-x_\star}_2^2}{2T}$.
\end{theorem}

The central inequality to the proof is the following one-step progress bound,

\begin{align}
    f(x_{t})-f(x_\star)
    \le
    \frac{
      \nbr*{x_t-x_\star}_2^2
      -\nbr*{x_{t+1}-x_\star}_2^2
    }{2\alpha}
    +
    f(x_t) - f(x_{t+1})
    ,
\end{align}

equivalent to \cref{eq:population-gd-progress-2} when rearranged.
In words, an update to $x_t$ must either cause
the objective to decrease by at least half the current suboptimality,
or the squared Euclidean distance to decrease by at least $\alpha$ times the current suboptimality.
The inequality combines convexity,
which states that parameter distance decreases after taking an $\alpha$-step in the $-\nabla f(x_t)$ direction (\cref{eq:population-gd-progress-1}),
subject to a curvature correction that grows with $\nbr*{\nabla f(x_t)}_2^2$,
and smoothness (\cref{eq:population-gd-progress-3}),
which guarantees the existence of a learning rate $\alpha$ that decreases the objective value proportionally to $\nbr*{\nabla f(x_t)}_2^2$.

\begin{proof}[\pfref{thm:population-gd-convergence}]
  Write $g_t\ldef{}\nabla f(x_t)$.
  Summing the second inequality in \cref{lem:population-gd-progress}
  for all $0\le t<T$ gives
  \begin{align}
    \sum_{t=0}^{T-1}\rbr*{f(x_{t+1})-f(x_\star)}
    &\le
    \frac{
      \nbr*{x_0-x_\star}_2^2
      -
      \nbr*{x_T-x_\star}_2^2
    }{2\alpha}
    \\
    &\le
    \frac{\nbr*{x_0-x_\star}_2^2}{2\alpha}.
  \end{align}
  Moreover, \cref{lem:population-gd-descent} and $\alpha\le1/L$ imply
  $f(x_{t+1})\le f(x_t)$ for every $0\le t<T$.  Therefore,
  \begin{align}
    T\rbr*{f(x_T)-f(x_\star)}
    &\le
    \sum_{t=0}^{T-1}\rbr*{f(x_{t+1})-f(x_\star)}
    \le
    \frac{\nbr*{x_0-x_\star}_2^2}{2\alpha}.
  \end{align}
  Dividing by $T$ proves the first claim.  The second follows by setting
  $\alpha=1/L$.
\end{proof}

\begin{lemma}[One-step progress]
  \label{lem:population-gd-progress}
  Suppose that $f:\cX\to\bbR$ is convex.
  Then the iterates of \cref{eq:population-gd-update}
  for all $0 \le t < T$ satisfy,
  \begin{align}
    \label{eq:population-gd-progress-1}
    f(x_{t})-f(x_\star)
    \le
    \inner*{g_t, x_t - x_\star}
    =
    \frac{
      \nbr*{x_t-x_\star}_2^2
      -\nbr*{x_{t+1}-x_\star}_2^2
    }{2\alpha}
    +
    \frac{\alpha}{2} \nbr*{\nabla f(x_t)}_2^2
    .
  \end{align}
  If, additionally, $f$ is $L$-smooth
  and $\alpha \le \frac{1}{L}$,
  \begin{align}
    \label{eq:population-gd-progress-2}
    f(x_{t+1})-f(x_\star)
    \le
    \frac{
      \nbr*{x_t-x_\star}_2^2
      -\nbr*{x_{t+1}-x_\star}_2^2
    }{2\alpha}
    .
  \end{align}
\end{lemma}

\begin{proof}[\pfref{lem:population-gd-progress}]
  Write $g_t\ldef{}\nabla f(x_t)$.
  Expanding the update in \cref{eq:population-gd-update} gives
  \begin{align}
    \nbr*{x_{t+1}-x_\star}_2^2
    &=
    \nbr*{x_t-x_\star-\alpha g_t}_2^2
    \\
    &=
    \nbr*{x_t-x_\star}_2^2
    -
    2\alpha\inner*{g_t,x_t-x_\star}
    +
    \alpha^2\nbr*{g_t}_2^2.
  \end{align}
  Rearranging this identity and using convexity yields
  \begin{align}
    f(x_t)-f(x_\star)
    &\le
    \frac{
      \nbr*{x_t-x_\star}_2^2
      -
      \nbr*{x_{t+1}-x_\star}_2^2
    }{2\alpha}
    +
    \frac{\alpha}{2}\nbr*{g_t}_2^2,
  \end{align}
  which proves the first claim.

  Now suppose additionally that $f$ is $L$-smooth and
  $\alpha\le1/L$.  By \cref{lem:population-gd-descent} and the preceding bound,
  \begin{align}
    f(x_{t+1})-f(x_\star)
    &=
    f(x_t)-f(x_\star)+f(x_{t+1})-f(x_t)
    \\
    &\le
    \frac{
      \nbr*{x_t-x_\star}_2^2
      -
      \nbr*{x_{t+1}-x_\star}_2^2
    }{2\alpha}
    +
    \frac{\alpha}{2}\nbr*{g_t}_2^2
    -
    \alpha\rbr*{1-\frac{L\alpha}{2}}\nbr*{g_t}_2^2
    \\
    &=
    \frac{
      \nbr*{x_t-x_\star}_2^2
      -
      \nbr*{x_{t+1}-x_\star}_2^2
    }{2\alpha}
    -
    \frac{\alpha}{2}\rbr*{1-L\alpha}\nbr*{g_t}_2^2
    \\
    &\le
    \frac{
      \nbr*{x_t-x_\star}_2^2
      -
      \nbr*{x_{t+1}-x_\star}_2^2
    }{2\alpha},
  \end{align}
  where the last inequality uses $\alpha\le1/L$.
\end{proof}

\begin{lemma}[Descent lemma]
  \label{lem:population-gd-descent}
  Suppose that $f:\cX\to\bbR$ is $L$-smooth.
  Then for all $0 \le t < T$, the iterates of \cref{eq:population-gd-update} satisfy
  \begin{align}
    f(x_{t+1})-f(x_t)
    \le
    -\alpha\rbr*{1-\frac{L\alpha}{2}}
    \nbr*{\nabla f(x_t)}_2^2.
    \label{eq:population-gd-progress-3}
  \end{align}
\end{lemma}

\begin{proof}[\pfref{lem:population-gd-descent}]
  Applying $L$-smoothness with $x=x_t$ and $y=x_{t+1}$ gives
  \begin{align}
    f(x_{t+1})-f(x_t)
    \le
    \inner*{\nabla f(x_t),x_{t+1}-x_t}
    +
    \frac{L}{2}\nbr*{x_{t+1}-x_t}_2^2.
  \end{align}
  Substituting
  $x_{t+1}-x_t=-\alpha\nabla f(x_t)$ from
  \cref{eq:population-gd-update} yields
  \begin{align}
    f(x_{t+1})-f(x_t)
    &\le
    -\alpha\nbr*{\nabla f(x_t)}_2^2
    +
    \frac{L\alpha^2}{2}\nbr*{\nabla f(x_t)}_2^2
    \\
    &=
    -\alpha\rbr*{1-\frac{L\alpha}{2}}
    \nbr*{\nabla f(x_t)}_2^2.
  \end{align}
\end{proof}

\subsection{Exact Policy Gradient Ascent}\label{app:population-pg}
We analyze the convergence of gradient ascent on the regularized policy gradient objective over softmax linear policies.
Assume we are in the (non-contextual) bandit setting and \cref{asm:softmax-linear} holds,
and define the feature diameter
\begin{align}
  \Dfeat \ldef{} \max_{y,y' \in \cY} \nbr*{\phi(y) - \phi(y')}_2.
\end{align}

Fix a regularization parameter $\eta > 0$ and a reference policy $\pirefgen=\pi_{\thetarefgen}\in\Pi_\Theta$,
and define
\begin{align}
  \Jr(\pi)
  \ldef{}
  \En_{y\sim\pi}\sbr*{r(y)} - \frac{1}{\eta}\Dkl{\pi}{\pirefgen}.
\end{align}
Here $\pirefgen$ is a general reference policy, so $\Jr$ specializes to the fixed-reference objective
in \cref{eq:fixed-ref-objective} when $\pirefgen=\piref$ and to the objective $\Jtr$ in \cref{eq:pg-objective} when $\pirefgen$ is the most recent policy.
We consider the optimization problem of finding
\begin{align}
  \pi_\eta^\star = \argmax_{\pi_\theta \in \Pi_\Theta}{} \Jr(\pi_\theta).
\end{align}
Under \cref{asm:softmax-linear}, $\pi_\eta^\star=\pi_{\theta_\eta^\star}$ with
$\theta_\eta^\star\ldef{}\thetarefgen+\eta\thetastar$ (\cref{lem:softmax-target-realizability}).

\para{Policy gradient}
Given a learning rate $\alpha>0$ and an integer $T\ge1$, initialize
$\theta_0=\thetarefgen$ and run
\begin{align}
  \theta_{t+1}
  =
  \theta_t+\alpha \nabla\Jr(\pi_{\theta_t}),
  \qquad t=0,\ldots,T-1.
  \label{eq:population-pg-update}
\end{align}
The algorithm returns $\pi_{\theta_T}$.
Defining $\Sigma_\theta \ldef{} \En_{\pi_\theta}\sbr*{\widetilde\phi_\theta(y) \widetilde\phi_\theta(y)^\top}$,
where $\widetilde\phi_\theta(y) \ldef{} \phi(y) - \En_{y' \sim \pi_\theta}\sbr*{\phi(y')}$,
the gradient is given by
\begin{align}
  \nabla\Jr(\pi_\theta)
    =
    \frac{1}{\eta}
    \Sigma_\theta
    \rbr*{\theta_\eta^\star-\theta}.
\end{align}

The objective $\Jr$ is not globally concave, and its global smoothness constant can grow as the minimum response probability under the reference policy decreases.
However, if the parameters remain in the ball
\begin{align}
  \mathcal B_R\ldef{}\cbr*{\theta \in \bbR^d : \nbr*{\theta - \theta_\eta^\star}_2 \le R},
\end{align}
then $\Jr$ satisfies local notions of concavity (\cref{lem:pg-local-convex}) and smoothness (\cref{lem:pg-local-smooth}).
The guarantees in both these inequalities become looser as $R$ grows.

\begin{lemma}[Single-policy concavity]
  \label{lem:pg-local-convex}
  For any $\theta \in \bbR^d$,
  \begin{align}
    \Jr(\pi_\eta^\star)-\Jr(\pi_\theta)
    &\le
      \frac{1}{2} \nbr*{\frac{\pi_\eta^\star}{\pi_\theta}}_\infty
      \underbrace{
        \inner*{ \nabla\Jr(\pi_\theta),\theta_\eta^\star-\theta }
      }_{= \frac{1}{\eta}\nbr*{\theta_\eta^\star - \theta}_{\Sigma_\theta}^2}
      .
  \end{align}

  Moreover, for $R>0$ define
  \begin{align}
    W_R
    \ldef{}
    e^{\Dfeat R}.
    \label{eq:pg-local-convex-W}
  \end{align}
  If $\theta\in\mathcal B_R$, then
  \begin{align}
    \Jr(\pi_\eta^\star)-\Jr(\pi_\theta)
    &\le
    \frac{W_R}{2}
      \inner*{ \nabla\Jr(\pi_\theta),\theta_\eta^\star-\theta }
    .
  \end{align}

\end{lemma}

\begin{lemma}[Local smoothness]
  \label{lem:pg-local-smooth}
  Fix $R>0$ and define
  \begin{align}
    L_R
    \ldef{}
    \frac{\Dfeat^2}{\eta}\rbr*{1+\Dfeat R}.
  \end{align}
  Then $\theta\mapsto\Jr(\pi_\theta)$ is $L_R$-smooth on $\mathcal B_R$.  In particular, for
  any $\theta,\theta'\in\mathcal B_R$,
  \begin{align}
    \Jr(\pi_{\theta'})
    &\ge
    \Jr(\pi_\theta)
    +
    \inner*{\nabla\Jr(\pi_\theta),\theta'-\theta}
    -
    \frac{L_R}{2}\nbr*{\theta'-\theta}_2^2.
  \end{align}
\end{lemma}

With these properties, the proof follows using essentially the same steps as in the convex analysis of \cref{thm:population-gd-convergence} (with concavity in place of convexity, since we maximize $\Jr$).
It is crucial that the iterates remain inside the ball,
and there this follows purely from the population policy gradient update for softmax linear policies,
not the local concavity.
(Linearity here gives us extra structure; for general functions concavity is needed for this to hold.)
The factors $L_R$ and $W_R$ carry into the final guarantee,
and both grow with $R$.
Because $R$ is an analytical construct we can set $R = \eta\nbr*{\thetastar}_2$.

\begin{theorem}[Policy gradient convergence]
  \label{thm:population-pg}
    Fix $R\ge\eta\nbr*{\thetastar}_2$ and $T\ge1$, and let $L_R$ and
    $W_R$ be defined as in
    \cref{lem:pg-local-smooth,lem:pg-local-convex}, respectively.
    Let $\cbr*{\theta_t}_{t=0}^T$ follow \cref{eq:population-pg-update} with
    learning rate $ 0<\alpha\le\frac{1}{L_R} $.
    Then the iterates remain localized and the objective is nondecreasing, that is,
    for every $0\le t\le T$,
    \begin{align}
      \nbr*{\theta_t-\theta_\eta^\star}_2
      \le
      \eta\nbr*{\thetastar}_2
      \le R,
      \qquad\text{and for every $0\le t<T$,}\qquad
      \Jr(\pi_{\theta_{t+1}}) \ge \Jr(\pi_{\theta_t}).
    \end{align}
    Moreover, the final iterate satisfies
    \begin{align}
      \Jr(\pi_\eta^\star)-\Jr(\pi_{\theta_T})
      &\le
      \frac{e^{\Dfeat R}\,\eta^2\nbr*{\thetastar}_2^2}
      {2\alpha\rbr*{2-\alpha\Dfeat^2/\eta}T}.
    \end{align}
    In particular, choosing $\alpha = \frac{1}{L_R}$ gives
    \begin{align}
      \Jr(\pi_\eta^\star)-\Jr(\pi_{\theta_T})
      &\le
      \frac{W_R L_R\,\eta^2\nbr*{\thetastar}_2^2}{2T}
      .
    \end{align}
    Setting $R = \eta\Bstar$, where $\Bstar=\nbr*{\thetastar}_2$, and using $\Dstar \ldef{} \Dfeat \Bstar$,
    \begin{align}
      \Jr(\pi_\eta^\star)-\Jr(\pi_{\theta_T})
      &\le
      \frac{\eta e^{\eta \Dstar} \Dstar^2 (1 + \eta \Dstar)}{2T}
      .
    \end{align}
\end{theorem}

\begin{proof}[\pfref{thm:population-pg}]
    Since $L_R=\frac{\Dfeat^2}{\eta}\rbr*{1+\Dfeat R}$,
    the condition on $\alpha$ implies that
    $ 0<\alpha\Dfeat^2/\eta<2 $,
    so the RHS of the claim is well-behaved.

    We first prove localization by induction.  The base case follows from
    $\theta_0=\thetarefgen$, since $\theta_\eta^\star-\thetarefgen=\eta\thetastar$.  Suppose that
    $\nbr*{\theta_t-\theta_\eta^\star}_2\le\eta\nbr*{\thetastar}_2$.  By the gradient
    identity and $\Sigma_{\theta_t}\succeq0$,
    \begin{align}
      \inner*{\nabla\Jr(\pi_{\theta_t}),\theta_\eta^\star-\theta_t}
      =
      \frac{1}{\eta}\nbr*{\theta_\eta^\star-\theta_t}_{\Sigma_{\theta_t}}^2
      \ge0.
    \end{align}
    Applying \cref{lem:population-pg-progress} and using the positivity of
    $\alpha\rbr*{2-\alpha\Dfeat^2/\eta}$ therefore gives
    \begin{align}
      \nbr*{\theta_{t+1}-\theta_\eta^\star}_2^2
      \le
      \nbr*{\theta_t-\theta_\eta^\star}_2^2
      \le
      \eta^2\nbr*{\thetastar}_2^2.
    \end{align}
    We can now apply \cref{lem:population-pg-ascent} at every
    iteration.  Since $\alpha L_R\le1$, it yields
    \begin{align}
      \Jr(\pi_{\theta_{t+1}})-\Jr(\pi_{\theta_t})
      &\ge
      \alpha\rbr*{1-\frac{\alpha L_R}{2}}
      \nbr*{\nabla\Jr(\pi_{\theta_t})}_2^2
      \\
      &\ge
      \frac{\alpha}{2}
      \nbr*{\nabla\Jr(\pi_{\theta_t})}_2^2
      \ge0.
    \end{align}
    This proves monotonicity.
    Finally, summing
    \cref{lem:population-pg-progress} for every $0\le t<T$
    and telescoping gives
    \begin{align}
      \sum_{t=0}^{T-1}
      \rbr*{\Jr(\pi_\eta^\star)-\Jr(\pi_{\theta_t})}
      &\le
      \frac{W_R}
      {2\alpha\rbr*{2-\alpha\Dfeat^2/\eta}}
      \sum_{t=0}^{T-1}
      \rbr*{
        \nbr*{\theta_t-\theta_\eta^\star}_2^2
        -
        \nbr*{\theta_{t+1}-\theta_\eta^\star}_2^2
      }
      \\
      &\le
      \frac{W_R\,\eta^2\nbr*{\thetastar}_2^2}
      {2\alpha\rbr*{2-\alpha\Dfeat^2/\eta}}.
    \end{align}
    Combining the last two displays and dividing by $T$ proves the final
    guarantee.
\end{proof}

\begin{lemma}[One-step population PG progress]
  \label{lem:population-pg-progress}
  Let $\cbr*{\theta_t}_{t=0}^T$ follow
  \cref{eq:population-pg-update} with
  $0<\alpha<2\eta/\Dfeat^2$.
  Then, for every $0\le t<T$,
  \begin{align}
    \inner*{\nabla\Jr(\pi_{\theta_t}),\theta_\eta^\star-\theta_t}
    &\le
    \frac{
        \nbr*{\theta_t-\theta_\eta^\star}_2^2 - \nbr*{\theta_{t+1}-\theta_\eta^\star}_2^2
      }{
        \alpha\rbr*{2-\alpha\Dfeat^2/\eta}
      }
      .
  \end{align}
  If, additionally, $\theta_t\in\mathcal B_R$ for some $R>0$, then
  applying \cref{lem:pg-local-convex} gives
  \begin{align}
    \Jr(\pi_\eta^\star)-\Jr(\pi_{\theta_t})
    &\le
    \frac{W_R}{2}
    \cdot
    \frac{
        \nbr*{\theta_t-\theta_\eta^\star}_2^2 - \nbr*{\theta_{t+1}-\theta_\eta^\star}_2^2
      }{
        \alpha\rbr*{2-\alpha\Dfeat^2/\eta}
      }
    .
  \end{align}
\end{lemma}

\begin{proof}[\pfref{lem:population-pg-progress}]
    Fix $0\le t<T$.  Expanding the update gives
    \begin{align}
      \nbr*{\theta_{t+1}-\theta_\eta^\star}_2^2
      &=
      \nbr*{\theta_t-\theta_\eta^\star}_2^2
      -
      2\alpha\inner*{\nabla\Jr(\pi_{\theta_t}),\theta_\eta^\star - \theta_t}
      +
      \alpha^2\nbr*{\nabla\Jr(\pi_{\theta_t})}_2^2.
    \end{align}
    Since $\nabla\Jr(\pi_{\theta_t})=\frac{1}{\eta}\Sigma_{\theta_t}\rbr*{\theta_\eta^\star-\theta_t}$ and
    $0\preceq\Sigma_{\theta_t}\preceq\Dfeat^2 I$, we have
    \begin{align}
      \nbr*{\nabla\Jr(\pi_{\theta_t})}_2^2
      &=
      \frac{1}{\eta^2}\rbr*{\theta_\eta^\star-\theta_t}^\top\Sigma_{\theta_t}^2\rbr*{\theta_\eta^\star-\theta_t}
      \\
      &\le
      \frac{\Dfeat^2}{\eta^2}\rbr*{\theta_\eta^\star-\theta_t}^\top\Sigma_{\theta_t}\rbr*{\theta_\eta^\star-\theta_t}
      \\
      &=
      \frac{\Dfeat^2}{\eta}
      \inner*{\nabla\Jr(\pi_{\theta_t}),\theta_\eta^\star-\theta_t}.
    \end{align}
    Therefore,
    \begin{align}
      \nbr*{\theta_t-\theta_\eta^\star}_2^2
      -
      \nbr*{\theta_{t+1}-\theta_\eta^\star}_2^2
      \ge
      \alpha\rbr*{2-\alpha\Dfeat^2/\eta}
      \inner*{\nabla\Jr(\pi_{\theta_t}),\theta_\eta^\star-\theta_t},
    \end{align}
    and rearranging proves the first claim.

    Now suppose $\theta_t\in\mathcal B_R$.
    Using the local concavity in \cref{lem:pg-local-convex},
    \begin{align}
      \Jr(\pi_\eta^\star)-\Jr(\pi_{\theta_t})
      &\le
      \frac{W_R}{2}
      \inner*{\nabla\Jr(\pi_{\theta_t}),\theta_\eta^\star-\theta_t}.
    \end{align}
\end{proof}

\begin{lemma}[Population PG ascent]
  \label{lem:population-pg-ascent}
  Fix $R>0$ and let $L_R$ be defined as in
  \cref{lem:pg-local-smooth}.
  Let $\cbr*{\theta_t}_{t=0}^T$ follow
  \cref{eq:population-pg-update},
  and suppose that $\theta_t\in\mathcal B_R$ for all $0 \le t \le T$.
  Then, for every $0\le t<T$,
  \begin{align}
    \Jr(\pi_{\theta_{t+1}})-\Jr(\pi_{\theta_t})
    \ge
    \alpha\rbr*{1-\frac{\alpha L_R}{2}}
    \nbr*{\nabla\Jr(\pi_{\theta_t})}_2^2.
  \end{align}
\end{lemma}

\begin{proof}[\pfref{lem:population-pg-ascent}]

    Fix $0\le t<T$.  By assumption, both $\theta_t$ and $\theta_{t+1}$ belong to the ball $\mathcal B_R$.
    Since this ball is convex, the line segment joining the two iterates also
    lies in the ball, so the local smoothness inequality in
    \cref{lem:pg-local-smooth} applies and gives

    \begin{align}
      \Jr(\pi_{\theta_{t+1}})-\Jr(\pi_{\theta_t})
      &\ge
      \inner*{\nabla\Jr(\pi_{\theta_t}),\theta_{t+1}-\theta_t}
      -
      \frac{L_R}{2}\nbr*{\theta_{t+1}-\theta_t}_2^2.
    \end{align}
    Substituting
    $\theta_{t+1}-\theta_t=\alpha\nabla\Jr(\pi_{\theta_t})$ from
    \cref{eq:population-pg-update} yields
    \begin{align}
      \Jr(\pi_{\theta_{t+1}})-\Jr(\pi_{\theta_t})
      &\ge
      \alpha\nbr*{\nabla\Jr(\pi_{\theta_t})}_2^2
      -
      \frac{\alpha^2L_R}{2}\nbr*{\nabla\Jr(\pi_{\theta_t})}_2^2
      \\
      &=
      \alpha\rbr*{1-\frac{\alpha L_R}{2}}
      \nbr*{\nabla\Jr(\pi_{\theta_t})}_2^2,
    \end{align}
    as claimed.
\end{proof}

\begin{proof}[\pfref{lem:pg-local-convex}]
  Fix $\theta$ and write
  \begin{align}
    \delta_\theta(y)
    \ldef{}
    \eta r(y)-\phi(y)^\top\rbr*{\theta-\thetarefgen}
    =
    \phi(y)^\top\rbr*{\theta_\eta^\star-\theta}.
  \end{align}
  For $s\in[0,1]$, let $\pi_s(y)\propto\pi_\theta(y)e^{s\delta_\theta(y)}$ be the exponential tilt of $\pi_\theta$ by
  $s\delta_\theta$, so that $\pi_1=\pi_\eta^\star$.
  We can write
  \begin{align}
    \Jr(\pi_\eta^\star)-\Jr(\pi_\theta)
    &=
    \frac{1}{\eta}
    \Dkl{\pi_\theta}{\pi_\eta^\star}
    =
    \frac{1}{\eta}
    \int_0^1
      \rbr*{1-s}
      \Var_{y\sim\pi_s}\rbr*{\delta_\theta(y)}
    \,ds.
  \end{align}
  By a change of measure and since $\nbr*{\pi_s/\pi_\theta}_\infty$ is nondecreasing in $s$,
  \begin{align}
    \Var_{y\sim\pi_s}\rbr*{\delta_\theta(y)}
    \le
    \nbr*{\frac{\pi_s}{\pi_\theta}}_\infty
    \Var_{y\sim\pi_\theta}\rbr*{\delta_\theta(y)}
    \le
    \nbr*{\frac{\pi_\eta^\star}{\pi_\theta}}_\infty
    \Var_{y\sim\pi_\theta}\rbr*{\delta_\theta(y)}.
  \end{align}
  Combining the above two displays and integrating,
  \begin{align}
    \Jr(\pi_\eta^\star)-\Jr(\pi_\theta)
    \le
    \frac{1}{2\eta}
    \nbr*{\frac{\pi_\eta^\star}{\pi_\theta}}_\infty
    \Var_{y\sim\pi_\theta}\rbr*{\delta_\theta(y)}.
  \end{align}
  Since 
  \begin{align}
    \Var_{y\sim\pi_\theta}\rbr*{\delta_\theta(y)}
    =
    \nbr*{\theta_\eta^\star - \theta}_{\Sigma_\theta}^2
    =
    \eta \inner*{\nabla\Jr(\pi_\theta), \theta_\eta^\star - \theta},
  \end{align}
  this proves the first inequality.
  Finally, $\operatorname{range}(\delta_\theta)\le\Dfeat R$ implies
  \begin{align}
    \nbr*{\frac{\pi_\eta^\star}{\pi_\theta}}_\infty
    \le
    e^{\Dfeat R},
  \end{align}
  proving the second inequality.
\end{proof}

\begin{proof}[\pfref{lem:pg-local-smooth}]
  For every $h\in\bbR^d$,
  \begin{align}
    h^\top\nabla^2\Jr(\pi_\theta)h
    &=
    -\frac{1}{\eta}
    \En_{y\sim\pi_\theta}
    \sbr*{\rbr*{h^\top\widetilde\phi_\theta(y)}^2}
    \\
    &\quad+
    \frac{1}{\eta}
    \En_{y\sim\pi_\theta}
    \sbr*{
      \rbr*{h^\top\widetilde\phi_\theta(y)}^2
      \rbr*{\theta_\eta^\star-\theta}^\top\widetilde\phi_\theta(y)
    }.
  \end{align}
  The definitions of the ball and $\Dfeat$ imply
  \begin{align}
    \abr*{\rbr*{\theta_\eta^\star-\theta}^\top\widetilde\phi_\theta(y)}
    &\le
    \Dfeat R,
    &
    \abr*{h^\top\widetilde\phi_\theta(y)}
    &\le
    \Dfeat\nbr*{h}_2.
  \end{align}
  Therefore,
  \begin{align}
    \abr*{h^\top\nabla^2\Jr(\pi_\theta)h}
    &\le
    \frac{1}{\eta}\rbr*{1+\Dfeat R}
    \En_{y\sim\pi_\theta}
    \sbr*{\rbr*{h^\top\widetilde\phi_\theta(y)}^2}
    \\
    &\le
    L_R\nbr*{h}_2^2,
  \end{align}
  and $\nbr*{\nabla^2\Jr(\pi_\theta)}_{\op}\le L_R$ throughout the ball.
  To conclude, fix $\theta,\theta'\in\mathcal B_R$.  Since the ball is convex,
  the Hessian bound holds at $\theta+s(\theta'-\theta)$ for every $s\in[0,1]$.
  Therefore,
  \begin{align}
    \nbr*{\nabla\Jr(\pi_{\theta'})-\nabla\Jr(\pi_\theta)}_2
    &=
    \nbr*{
      \int_0^1 \nabla^2\Jr\rbr*{\pi_{\theta+s(\theta'-\theta)}}\rbr*{\theta'-\theta}\,ds
    }_2
    \\
    &\le
    L_R\nbr*{\theta'-\theta}_2.
  \end{align}
  Similarly, Taylor's theorem with integral remainder gives
  \begin{align}
    &\Jr(\pi_{\theta'})-\Jr(\pi_\theta)-\inner*{\nabla\Jr(\pi_\theta),\theta'-\theta}
    \\
    &\quad=
    \int_0^1
      \rbr*{1-s}\rbr*{\theta'-\theta}^\top
      \nabla^2\Jr\rbr*{\pi_{\theta+s(\theta'-\theta)}}\rbr*{\theta'-\theta}
    \,ds
    \\
    &\quad\ge
    -L_R\nbr*{\theta'-\theta}_2^2\int_0^1\rbr*{1-s}\,ds
    =
    -\frac{L_R}{2}\nbr*{\theta'-\theta}_2^2.
  \end{align}
  This proves both claims.
\end{proof}

\subsection{Exact Mirror Descent}\label{app:population-mirror-descent}

We analyze exact KL-regularized mirror descent for contextual bandits.
These are the updates obtained when every RL subproblem of \framework is solved exactly.

\para{Exact mirror descent}
Fix $\eta>0$ and initialize $\pit[0]=\piref>0$.
For every integer $t\geq0$, update
\begin{align}
  \pitp(\cdot\mid{}x)
  =\piplust(\cdot\mid{}x)
  =\argmax_{\pi(\cdot\mid{}x)\in\Delta(\cY)}
  \cbr*{J_x(\pi)-\frac1\eta\Dkl{\pi(\cdot\mid{}x)}{\pit(\cdot\mid{}x)}},
  \qquad x\in\cX.
  \label{eq:exact-apmd-optimization}
\end{align}
\Cref{lem:exact-apmd-closed-form} gives the closed form of this update and of the resulting policy sequence.
We write $J_x(\pi)\ldef{}\En_{y\sim\pi(\cdot\mid{}x)}\sbr*{r(x,y)}$ and $r_x^\star\ldef{}\max_{y\in\cY}r(x,y)$,
and define the partition function for the rewards centered by the mean baseline $r(x,\pit)=J_x(\pit)$ by
\begin{align}
  \Zt(x)\ldef{}\En_{y\sim\pit(\cdot\mid{}x)}
  \sbr*{e^{\eta(r(x,y)-r(x,\pit))}}.
  \label{eq:exact-apmd-partition}
\end{align}
Subtracting $r(x,\pit)$ from every reward on prompt $x$ leaves the exact update unchanged.

\begin{theorem}[Convergence of exact mirror descent]
  \label{thm:exact-apmd-rate}
  \label{prop:exact-apmd-rate}
  Let $\pt[0](x)\ldef{}\piref(\Ystarx\mid{}x)$ denote the probability of an optimal response under the base policy.
  For the exact mirror descent iterates in \cref{eq:exact-apmd-optimization},
  for every $t > 0$, it holds that
  \begin{align}
    \Jstar-J(\pit)
    \leq\frac{\En_{x\sim\mu}\sbr*{\log\rbr*{1/\pt[0](x)}}}{\eta t}
    =\frac{\Dkl{\pistar}{\piref}}{\eta t}
    \leq\frac{\log\Cresp}{\eta t},
    \label{eq:exact-apmd-rate}
  \end{align}
  where $\pistar(y\mid{}x) = \frac{\piref(y\mid{}x) \one{y \in \Ystarx}}{\piref(\Ystarx\mid{}x)}$
  is the optimal policy induced by conditioning $\piref$ on optimal responses.
\end{theorem}

\begin{proof}[\pfref{thm:exact-apmd-rate}]
  For this proof and the supporting lemmas below, write
  $\pt(x)\ldef{}\pit(\Ystarx\mid{}x)$ for $t\geq0$.
  Fix a prompt $x$.
  By \cref{lem:exact-apmd-coverage}, the probability $\pt(x)$ of an optimal response
  is nondecreasing in $t$, so $\pt(x)\geq\pt[0](x)>0$ for every $t\geq0$.

  Fix an iteration $t\geq1$.
  Applying \cref{lem:exact-apmd-one-step} at $s=0,\ldots,t-1$ and summing gives
  \begin{align}
    \eta\sum_{s=1}^{t}\rbr*{r_x^\star-J_x(\pit[s])}
    &\leq\sum_{s=0}^{t-1}\rbr*{\log\ptp[s](x)-\log\pt[s](x)}
    \\
    &=\log\pt(x)-\log\pt[0](x)
    \leq\log\frac1{\pt[0](x)}.
  \end{align}
  Since $J_x(\pit[s])$ is nondecreasing in $s$ by \cref{lem:exact-apmd-improvement},
  \begin{align}
    t\rbr*{r_x^\star-J_x(\pit)}
    \leq\sum_{s=1}^{t}\rbr*{r_x^\star-J_x(\pit[s])}.
  \end{align}
  Combining these inequalities gives
  \begin{align}
    r_x^\star-J_x(\pit)
    \leq\frac{\log\rbr*{1/\pt[0](x)}}{\eta t}.
    \label{eq:exact-md-one-step}
  \end{align}
  Taking expectations over prompts gives
  \begin{align}
    \Jstar-J(\pit)
    &\leq\frac{\En_{x\sim\mu}\sbr*{\log\rbr*{1/\pt[0](x)}}}{\eta t}
    .
  \end{align}
  The equality in the theorem statement follows from the definition of $\pistar$.
  \begin{align}
    \Dkl{\pistar}{\piref}
    &=\En_{x\sim\mu}\sbr*{
      \En_{y\sim\pistar(\cdot\mid{}x)}\sbr*{
        \log\frac1{\pt[0](x)}
      }
    }
    =\En_{x\sim\mu}\sbr*{\log\frac1{\pt[0](x)}}.
  \end{align}
  Finally, Jensen's inequality gives
  \begin{align}
    \En_{x\sim\mu}\sbr*{\log\frac1{\pt[0](x)}}
    \leq\log\En_{x\sim\mu}\sbr*{\frac1{\pt[0](x)}}
    =\log\Cresp.
  \end{align}
\end{proof}

\begin{lemma}[Closed form of exact mirror descent]
  \label{lem:exact-apmd-closed-form}
  \label{prop:exact-apmd-closed-form}
  For any current policy $\pit$ and $\eta>0$, the exact update in
  \cref{eq:exact-apmd-optimization} satisfies, for every prompt $x$ and response $y$,
  \begin{align}
    \piplust(y\mid{}x)
    =\frac{\pit(y\mid{}x)e^{\eta(r(x,y)-r(x,\pit))}}{\Zt(x)}.
    \label{eq:exact-apmd-update}
  \end{align}
  If $\pit[0]=\piref$ and $\pitp=\piplust$ at every iteration, then
  \begin{align}
    \pit(y\mid{}x)
    =\frac{\piref(y\mid{}x)e^{t\eta r(x,y)}}
    {\En_{y'\sim\piref(\cdot\mid{}x)}\sbr*{e^{t\eta r(x,y')}}},
    \qquad t\geq0.
    \label{eq:exact-apmd-closed-form}
  \end{align}
  In particular, $\piplust(\cdot\mid{}x)$ and $\pit(\cdot\mid{}x)$ have the same support,
  which remains equal to that of $\piref(\cdot\mid{}x)$ along the sequence.
\end{lemma}

\begin{proof}[\pfref{lem:exact-apmd-closed-form}]
  Fix a prompt $x$.
  A policy assigning positive probability outside the support of $\pit(\cdot\mid{}x)$
  has infinite KL divergence in \cref{eq:exact-apmd-optimization}.
  For any policy $\pi$ with finite KL divergence, we have
  \begin{align}
    &J_x(\pi)-\frac1\eta\Dkl{\pi(\cdot\mid{}x)}{\pit(\cdot\mid{}x)}
\\
    &\qquad=r(x,\pit)+\frac{\log\Zt(x)}{\eta}
    -\frac1\eta\sum_{y:\pit(y\mid{}x)>0}\pi(y\mid{}x)
    \log\frac{\pi(y\mid{}x)\Zt(x)}{\pit(y\mid{}x)e^{\eta(r(x,y)-r(x,\pit))}}
\\
    &\qquad\leq r(x,\pit)+\frac{\log\Zt(x)}{\eta}.
  \end{align}
  The sum is the KL divergence from $\pi(\cdot\mid{}x)$ to the distribution
  in \cref{eq:exact-apmd-update}, whose probabilities sum to one by the definition of $\Zt(x)$.
  It is nonnegative, with equality precisely at that distribution.
  This proves the closed form of $\piplust$.

  For the policy sequence, \cref{eq:exact-apmd-closed-form} holds at $t=0$.
  If it holds at iteration $t$, substituting into \cref{eq:exact-apmd-update}
  and canceling the common normalization factor gives
  \begin{align}
    \pitp(y\mid{}x)
    &=\frac{\piref(y\mid{}x)e^{t\eta r(x,y)}e^{\eta r(x,y)}}
    {\sum_{y'\in\cY}\piref(y'\mid{}x)e^{t\eta r(x,y')}e^{\eta r(x,y')}}
    \\
    &=\frac{\piref(y\mid{}x)e^{(t+1)\eta r(x,y)}}
    {\En_{y'\sim\piref(\cdot\mid{}x)}\sbr*{e^{(t+1)\eta r(x,y')}}}.
  \end{align}
  Induction proves the second claim.
  The support claims follow because the exponential factors are strictly positive.
\end{proof}

The following improvement bound is the contextual bandit analogue of
\citet[Lemma~17]{agarwal2020theory}.

\begin{lemma}[Improvement lower bound]
  \label{lem:exact-apmd-improvement}
  For the exact mirror descent iterates with $\eta>0$, at every iteration $t\geq0$,
  \begin{align}
    J(\pitp)-J(\pit)
    \geq\frac1\eta\En_{x\sim\mu}\sbr*{\log\Zt(x)}
    \geq0.
    \label{eq:exact-apmd-improvement}
  \end{align}
  For each prompt $x\in\cX$,
  \begin{align}
    J_x(\pitp)-J_x(\pit)
    \geq\frac{\log\Zt(x)}{\eta}
    \geq0.
    \label{eq:exact-apmd-monotone}
  \end{align}
\end{lemma}

\begin{proof}[\pfref{lem:exact-apmd-improvement}]
  Fix a prompt $x$.
  Jensen's inequality gives
  \begin{align}
    \log\Zt(x)
    =\log\En_{y\sim\pit(\cdot\mid{}x)}\sbr*{e^{\eta(r(x,y)-r(x,\pit))}}
    \geq\eta\En_{y\sim\pit(\cdot\mid{}x)}\sbr*{r(x,y)-r(x,\pit)}
    =0.
  \end{align}
  By \cref{lem:exact-apmd-closed-form}, for every response in the support of $\pit(\cdot\mid{}x)$,
  \begin{align}
    \log\frac{\pitp(y\mid{}x)}{\pit(y\mid{}x)}
    =\eta\rbr*{r(x,y)-r(x,\pit)}-\log\Zt(x).
    \label{eq:exact-apmd-log-ratio}
  \end{align}
  Taking the expectation under $y\sim\pitp(\cdot\mid{}x)$ gives
  \begin{align}
    \Dkl{\pitp(\cdot\mid{}x)}{\pit(\cdot\mid{}x)}
    =\eta\rbr*{J_x(\pitp)-J_x(\pit)}-\log\Zt(x).
    \label{eq:exact-apmd-improvement-identity}
  \end{align}
  Therefore,
  \begin{align}
    J_x(\pitp)-J_x(\pit)
    &=\frac1\eta\Dkl{\pitp(\cdot\mid{}x)}{\pit(\cdot\mid{}x)}
    +\frac{\log\Zt(x)}{\eta}
    \\
    &\geq\frac{\log\Zt(x)}{\eta}
    \geq0.
  \end{align}
  Averaging over prompts proves the bound on $J(\pitp)-J(\pit)$.
\end{proof}

\begin{lemma}[Response coverage improvement]
  \label{lem:exact-apmd-coverage}
  For the exact mirror descent iterates with $\eta>0$, at every iteration $t\geq0$,
  \begin{align}
    \ptp(x)\geq\pt(x),
    \qquad x\in\cX.
    \label{eq:exact-apmd-coverage}
  \end{align}
  Consequently,
  \begin{align}
    \Cresp[\pitp]\leq\Cresp[\pit].
  \end{align}
\end{lemma}

\begin{proof}[\pfref{lem:exact-apmd-coverage}]
  Fix a prompt $x$.
  Since $r(x,y)\leq r_x^\star$ for every response $y$,
  \begin{align}
    \Zt(x)
    =\En_{y\sim\pit(\cdot\mid{}x)}\sbr*{e^{\eta(r(x,y)-r(x,\pit))}}
    \leq e^{\eta(r_x^\star-r(x,\pit))}.
  \end{align}
  Summing \cref{eq:exact-apmd-update} over the optimal responses gives
  \begin{align}
    \ptp(x)
    &=\sum_{y\in\Ystarx}\frac{\pit(y\mid{}x)e^{\eta(r(x,y)-r(x,\pit))}}{\Zt(x)}
\\
    &=\frac{\pt(x)e^{\eta(r_x^\star-r(x,\pit))}}{\Zt(x)}
    \geq\pt(x).
    \label{eq:exact-apmd-coverage-update}
  \end{align}
  Taking reciprocals and averaging over prompts gives the response coverage bound.
\end{proof}

\begin{lemma}[One-step suboptimality]
  \label{lem:exact-apmd-one-step}
  For the exact mirror descent iterates with $\eta>0$, fix an iteration $t\geq0$
  and a prompt $x$ with $\pt(x)>0$.  Then
  \begin{align}
    \eta\rbr*{r_x^\star-J_x(\pitp)}
    &\leq\log\frac{\ptp(x)}{\pt(x)}.
    \label{eq:exact-apmd-one-step}
  \end{align}
\end{lemma}

\begin{proof}[\pfref{lem:exact-apmd-one-step}]
  Taking logarithms in \cref{eq:exact-apmd-coverage-update} gives
  \begin{align}
    \log\frac{\ptp(x)}{\pt(x)}
    =\eta\rbr*{r_x^\star-r(x,\pit)}-\log\Zt(x).
  \end{align}
  Subtracting \cref{eq:exact-apmd-improvement-identity} cancels the log partition function, yielding
  \begin{align}
    \eta\rbr*{r_x^\star-J_x(\pitp)}
    &=\log\frac{\ptp(x)}{\pt(x)}
    -\Dkl{\pitp(\cdot\mid{}x)}{\pit(\cdot\mid{}x)}
    \\
    &\leq\log\frac{\ptp(x)}{\pt(x)}.
  \end{align}
\end{proof}

\subsection{Softmax Linear Policies}\label{app:softmax-linear}

\subsubsection{Linear Reward and Mirror Descent Realizability}\label{app:linear-reward-realizability}

\begin{lemma}[Target realizability]
  \label{lem:softmax-target-realizability}
  Consider the softmax linear class $\Pi_\Theta$ with $\Theta=\bbR^d$.
  Fix $\pi_\theta\in\Pi_\Theta$ and $\eta>0$, and define
  \begin{align}
    \pi_\eta^\star(y\mid{}x)
    =\frac{\pi_\theta(y\mid{}x)e^{\eta r(x,y)}}
      {\sum_{y'}\pi_\theta(y'\mid{}x)e^{\eta r(x,y')}}.
  \end{align}
  Then $\pi_\eta^\star\in\Pi_\Theta$ if and only if there exist
  $\thetastar\in\bbR^d$ and $c:\cX\to\bbR$ such that
  \begin{align}
    r(x,y)=\phi(x,y)^\top\thetastar+c(x)
    \qquad\text{for all }(x,y)\in\cX\times\cY.
    \label{eq:reward-realizability-up-to-constant}
  \end{align}
  In this case,
  \begin{align}
    \pi_\eta^\star=\pi_{\theta+\eta\thetastar}.
    \label{eq:softmax-realizable-target}
  \end{align}
\end{lemma}

The inner loop of \framework approximates $\piplust$ within the policy class $\Pi_\Theta$.
The linear reward assumption in \cref{asm:softmax-linear} ensures that this target
is realizable at every iteration.
\Cref{lem:softmax-target-realizability} shows that target realizability is equivalent
to reward linearity up to a prompt-dependent additive constant,
which cancels when the target is normalized.

\begin{proof}[\pfref{lem:softmax-target-realizability}]
  Suppose $\pi_\eta^\star=\pi_{\theta'}$ for some $\theta'\in\bbR^d$.
  For every prompt $x$ and responses $y,y'$, the definition of $\pi_\eta^\star$
  and the softmax formula give
  \begin{align}
    \eta\rbr*{r(x,y)-r(x,y')}
    &=\log\frac{\pi_{\theta'}(y\mid{}x)}{\pi_{\theta'}(y'\mid{}x)}
      -\log\frac{\pi_\theta(y\mid{}x)}{\pi_\theta(y'\mid{}x)}
    \\
    &=\rbr*{\phi(x,y)-\phi(x,y')}^\top(\theta'-\theta).
  \end{align}
  Set $\thetastar=(\theta'-\theta)/\eta$.
  The displayed identity shows that $r(x,y)-\phi(x,y)^\top\thetastar$
  is independent of $y$ for each prompt $x$.
  Denoting this value by $c(x)$ gives \cref{eq:reward-realizability-up-to-constant}.

  Conversely, suppose \cref{eq:reward-realizability-up-to-constant} holds.
  The factor $e^{\eta c(x)}$ is common to all responses to $x$ and cancels
  in the normalization. Thus
  \begin{align}
    \pi_\eta^\star(y\mid{}x)
    &=\frac{\pi_\theta(y\mid{}x)e^{\eta r(x,y)}}
      {\sum_{y'}\pi_\theta(y'\mid{}x)e^{\eta r(x,y')}}
    \\
    &=\frac{\exp\rbr*{\phi(x,y)^\top(\theta+\eta\thetastar)}}
      {\sum_{y'}\exp\rbr*{\phi(x,y')^\top(\theta+\eta\thetastar)}}
    =\pi_{\theta+\eta\thetastar}(y\mid{}x),
  \end{align}
  proving \cref{eq:softmax-realizable-target} and membership in $\Pi_\Theta$.
\end{proof}

\subsubsection{Properties of Softmax Linear Policies}
We now prove various properties of the softmax linear parameterization in \cref{asm:softmax-linear}.
Define the feature diameter
\begin{align}
  \Dfeat
  \ldef{}
  \max_{x\in\cX,\,y,y'\in\cY}
  \nbr*{\phi(x,y)-\phi(x,y')}_2.
\end{align}
The following properties hold for each prompt $x\in\cX$.

\begin{lemma}[Softmax policy ratio bound]
  \label{lem:softmax-policy-ratio}
  For every $\theta,\theta'\in\bbR^d$ and $(x,y)\in\cX\times\cY$,
  \begin{align}
    \frac{\pi_{\theta'}(y\mid{}x)}{\pi_\theta(y\mid{}x)}
    \le
    \exp\rbr*{\Dfeat\nbr*{\theta'-\theta}_2}.
  \end{align}
\end{lemma}

\begin{proof}[\pfref{lem:softmax-policy-ratio}]
  The softmax formula gives
  \begin{align}
    \frac{\pi_{\theta'}(y\mid{}x)}{\pi_\theta(y\mid{}x)}
    &=
    \frac{\exp\rbr*{\phi(x,y)^\top(\theta'-\theta)}}{
      \En_{y'\sim\pi_\theta(\cdot\mid{}x)}
      \sbr*{\exp\rbr*{\phi(x,y')^\top(\theta'-\theta)}}
    }
    \\
    &=
    \frac{1}{
      \En_{y'\sim\pi_\theta(\cdot\mid{}x)}
      \sbr*{\exp\rbr*{(\phi(x,y')-\phi(x,y))^\top(\theta'-\theta)}}
    }
    \\
    &\le \exp\rbr*{\Dfeat\nbr*{\theta'-\theta}_2},
  \end{align}
  where the last inequality follows from
  $(\phi(x,y')-\phi(x,y))^\top(\theta'-\theta)
  \ge -\Dfeat\nbr*{\theta'-\theta}_2$
  for every $y'\in\cY$ by the Cauchy--Schwarz inequality.
\end{proof}

\begin{lemma}[Softmax KL bound]
  \label{lem:softmax-kl-bound}
  For every $\theta,\theta'\in\bbR^d$ and $x\in\cX$,
  \begin{align}
    \Dkl{\pi_\theta(\cdot\mid{}x)}{\pi_{\theta'}(\cdot\mid{}x)}
    \le\frac{\Dfeat^2}{8}\nbr*{\theta'-\theta}_2^2.
    \label{eq:softmax-kl-bound}
  \end{align}
\end{lemma}

\begin{proof}[\pfref{lem:softmax-kl-bound}]
  Fix $x\in\cX$. Under $y\sim\pi_\theta(\cdot\mid{}x)$, the random variable
  $\phi(x,y)^\top(\theta'-\theta)$ has range of length at most
  $\Dfeat\nbr*{\theta'-\theta}_2$ by the definition of $\Dfeat$.
  The softmax formula and Hoeffding's lemma therefore give
  \begin{align}
    \Dkl{\pi_\theta(\cdot\mid{}x)}{\pi_{\theta'}(\cdot\mid{}x)}
    &=\log\En_{y\sim\pi_\theta(\cdot\mid{}x)}
      \sbr*{e^{\phi(x,y)^\top(\theta'-\theta)}}
      -\En_{y\sim\pi_\theta(\cdot\mid{}x)}
      \sbr*{\phi(x,y)^\top(\theta'-\theta)}
    \\
    &\le\frac{\Dfeat^2}{8}\nbr*{\theta'-\theta}_2^2.
  \end{align}
\end{proof}

\begin{lemma}[Softmax score identity]
  \label{lem:softmax-score-identity}
  For every $\theta\in\bbR^d$ and $(x,y)\in\cX\times\cY$,
  \begin{align}
    \nabla\log\pi_\theta(y\mid{}x)
    =
    \phi(x,y)-\En_{y'\sim\pi_\theta(\cdot\mid{}x)}\sbr*{\phi(x,y')}.
  \end{align}
\end{lemma}

\begin{proof}[\pfref{lem:softmax-score-identity}]
  Differentiating the log policy gives
  \begin{align}
    \nabla\log\pi_\theta(y\mid{}x)
    &=
    \nabla\rbr*{
      \phi(x,y)^\top\theta
      -\log\sum_{y'\in\cY}\exp\rbr*{\phi(x,y')^\top\theta}
    }
    \\
    &=
    \phi(x,y)-\En_{y'\sim\pi_\theta(\cdot\mid{}x)}\sbr*{\phi(x,y')}.
  \end{align}
\end{proof}

\begin{lemma}[Softmax score bound]
  \label{lem:softmax-score-bound}
  For every $\theta\in\bbR^d$ and $(x,y)\in\cX\times\cY$,
  \begin{align}
    \nbr*{\nabla\log\pi_\theta(y\mid{}x)}_2
    \le \Dfeat.
  \end{align}
\end{lemma}

\begin{proof}[\pfref{lem:softmax-score-bound}]
  The score identity in \cref{lem:softmax-score-identity} gives
  \begin{align}
    \nbr*{\nabla\log\pi_\theta(y\mid{}x)}_2
    &=
    \nbr*{
      \En_{y'\sim\pi_\theta(\cdot\mid{}x)}\sbr*{\phi(x,y)-\phi(x,y')}
    }_2
    \\
    &\le
    \En_{y'\sim\pi_\theta(\cdot\mid{}x)}
    \sbr*{\nbr*{\phi(x,y)-\phi(x,y')}_2}
    \le \Dfeat.
  \end{align}
\end{proof}

\begin{lemma}[Softmax log-policy Hessian]
  \label{lem:softmax-log-hessian}
  For every $\theta\in\bbR^d$ and $(x,y)\in\cX\times\cY$,
  \begin{align}
    -\nabla^2\log\pi_\theta(y\mid{}x)
    =
    \operatorname{Cov}_{y'\sim\pi_\theta(\cdot\mid{}x)}\rbr*{\phi(x,y')}.
  \end{align}
\end{lemma}

\begin{proof}[\pfref{lem:softmax-log-hessian}]
  Differentiating the score identity in \cref{lem:softmax-score-identity} yields
  \begin{align}
    -\nabla^2\log\pi_\theta(y\mid{}x)
    &=
    \sum_{y'\in\cY}
    \pi_\theta(y'\mid{}x)\phi(x,y')
    \rbr*{\nabla\log\pi_\theta(y'\mid{}x)}^\top
    \\
    &=
    \En_{y'\sim\pi_\theta(\cdot\mid{}x)}\sbr*{\phi(x,y')\phi(x,y')^\top}
    \\
    &\quad-
    \En_{y'\sim\pi_\theta(\cdot\mid{}x)}\sbr*{\phi(x,y')}
    \En_{y'\sim\pi_\theta(\cdot\mid{}x)}\sbr*{\phi(x,y')}^\top
    \\
    &=
    \operatorname{Cov}_{y'\sim\pi_\theta(\cdot\mid{}x)}\rbr*{\phi(x,y')}.
  \end{align}
\end{proof}

\clearpage
\section{Coverage Collapse under RL with a Fixed Reference (Proof of \cref*{thm:lower-bound-main})}\label{app:proofs-motivation}
\label{app:motivation-overview}

This section provides the supplementary material for \cref{sec:lower-bound} and \cref{fig:coverage-collapse-leading}.

First, in \cref{app:figure-2-details} we provide the experimental details behind \cref{fig:coverage-collapse-leading,fig:reference-reanchoring-landscapes}
as well as the details for the visualization.
Then, in \cref{app:lower-bound-proof}, we provide the full statement of \cref{thm:lower-bound-main}
and the proof, including the construction and explicit constants for the result.

\begin{figure}[!htbp]
  \centering
  \includegraphics[width=0.92\linewidth]{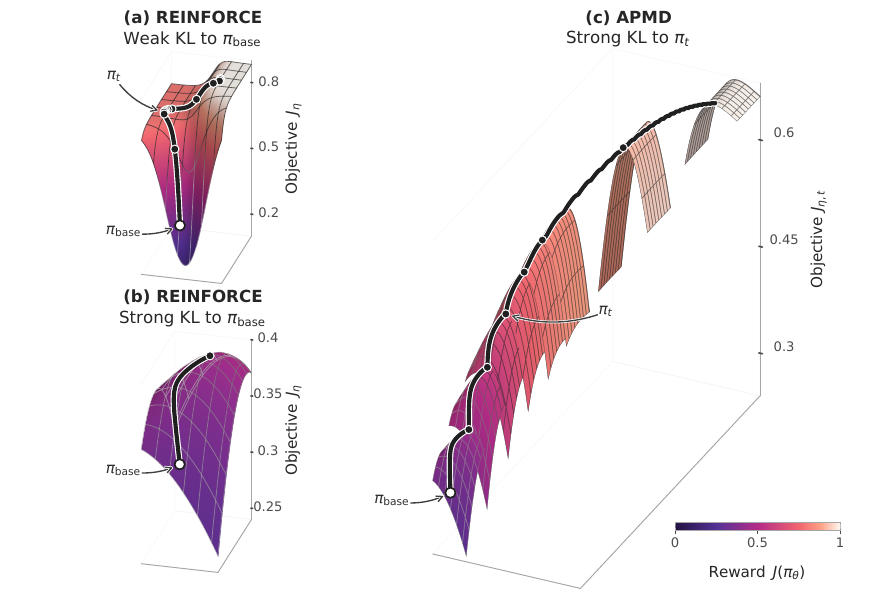}
  \input{sections/reference-reanchoring-landscapes-caption.tex}
  \label{fig:reference-reanchoring-landscapes}
\end{figure}

\subsection{Details for \cref{fig:coverage-collapse-leading,fig:reference-reanchoring-landscapes}}
\label{app:figure-2-details}

\para{Instance}
\Cref{fig:coverage-collapse-leading} uses a modification of the single-prompt bandit
constructed for \cref{thm:lower-bound-main} in \cref{app:lower-bound-proof}.
There are $m+2$ responses, with $m=100$ zero-reward responses.
The rewards and base model are
\begin{align}
  r(1)&=1, & r(2)&=0.8, & r(i)&=0 \quad(i\ge3),\\
  \piref(1)&=\frac1{16}, & \piref(2)&=\frac13, &
  \piref(i)&=\frac{1-1/16-1/3}{m} \quad(i\ge3).
\end{align}
We use tabular softmax policies
$\pi_\theta(i)=\exp(\theta_i)/\sum_{j=1}^{m+2}\exp(\theta_j)$,
where $\theta_i$ is the unnormalized score, or \emph{logit}, for response $i$.
Let $\thetaref$ denote the initial logit vector representing $\piref$.
We fix the target suboptimality at $\veps=0.01$.
Relative to the lower-bound construction, we increase the reward gap between
the optimal response and the closest alternative from $3\veps$ to $0.2$,
and reduce the base model's probability of the optimal response from $1/6$ to $1/16$.
These changes make coverage collapse and the reward plateau more visually apparent
than in the construction used for the worst-case analysis.

\para{Reading \cref{fig:coverage-collapse-leading}}
The horizontal axes in panels (b,c) are coordinates $(u,v)$ in a two-dimensional
parameter plane.
In the displayed view, $u$ increases from left to right, while $v$ increases
from the lower foreground to the upper background.
Increasing $u$ raises the optimal response's logit relative to that of response $2$.
Increasing $v$ raises both rewarding responses' logits relative to the zero-reward
logits, without changing the relative probability of responses $1$ and $2$.

Surface height gives the fixed-reference objective $\Jr$ (\cref{eq:fixed-ref-objective}),
while color shows expected reward in (b) and coverage $\Cresp[\pi_\theta]$ in (c).

\para{Parameter coordinates}
These coordinates represent the training trajectories exactly.
Both methods use exact policy gradients, as specified below.
The zero-reward logits start equal and receive equal updates by symmetry,
so they remain equal throughout training.
The gradient coordinates in \cref{eq:lower-bound-softmax-gradient} sum to zero,
so $\sum_{i=1}^{m+2}\theta_i$ remains fixed at its initial value.
The same identity holds with the current reference policy in place of $\piref$
for \framework{}, so both trajectories lie in the plane through
$\thetaref$ with orthonormal basis
\begin{align}
  b_1&=\frac{(1,-1,0,\ldots,0)}{\sqrt2}, &
  b_2&=\frac{(1,1,-2/m,\ldots,-2/m)}{\sqrt{2+4/m}}.
\end{align}
For a policy $\pi_\theta$ with logits $\theta$ in this plane, its coordinates and
reconstruction are
\begin{align}
  u&=b_1^\top(\theta-\thetaref), &
  v&=b_2^\top(\theta-\thetaref),\\
  \theta(u,v)&=\thetaref+u b_1+v b_2.
\end{align}
Thus, the initial base model has $(u,v)=(0,0)$.
Because the basis is orthonormal, the coordinates preserve Euclidean distances
and gradient steps within the plane.
Coverage is measured by
$\Cresp[\pi_{\theta(u,v)}]=1/\pi_{\theta(u,v)}(1)$, where smaller values indicate better coverage.

\para{Methods and tuning}
Both methods start from $\piref$ and perform $10{,}000$ gradient steps
with a constant learning rate $\alpha=1$.
\reinforce{} performs exact gradient ascent on
$\Jr(\pi_\theta)=J(\pi_\theta)-\eta^{-1}\Dkl{\pi_\theta}{\piref}$.
\framework{} instantiates \cref{alg:offline-pg} with exact population gradients
in place of the sampled gradient estimates.
At each outer iteration, it takes $K=100$ gradient ascent steps on
$\Jt(\pi_\theta)=J(\pi_\theta)-\eta^{-1}\Dkl{\pi_\theta}{\pit}$,
then uses $\pitp=\pi_t^K$ as the next reference policy.
It therefore performs $T=100$ outer iterations.
The horizontal axis in \cref{fig:coverage-collapse-leading}a counts individual gradient steps
$n=tK+k$, not reference resets.

For \reinforce{}, we choose $\eta\approx23.64513$ by solving
$1-J(\pistar_\eta)=0.9\veps$, where
$\pistar_\eta(i)\propto\piref(i)e^{\eta r(i)}$.
This is the strongest regularization for which the optimal regularized policy $\pistar_\eta$
has suboptimality at least $10\%$ below the target $\veps$.
For \framework{}, we fix $\eta=1/(\Bfeat\Bstar)=1/\sqrt{1+0.8^2}
\approx0.78087$, using standard basis vectors as features and the reward parameter
$\thetastar=r$.

To choose the learning rate, we compare $\alpha\in\{0.1,0.3,1\}$ in
runs with a fixed reference at both values of $\eta$.
We use $100{,}000$ steps for $\eta\approx23.64513$ and $1{,}000$ steps
for $\eta\approx0.78087$.
The selected learning rate $\alpha=1$ gives the fastest reward improvement
in both sweeps.

With $\alpha=1$, we test $K\in\{1,5,10,50,100\}$ for \framework{}
over $10{,}000$ steps.
The choice $K=10$ reaches reward $0.99$ fastest, in $520$ steps,
compared with $2{,}911$ for $K=100$.
We nevertheless use $K=100$ to show more optimization within each subproblem.
The common budget of $10{,}000$ steps extends beyond \reinforce{}'s first
attainment of reward $0.99$, at step $7{,}364$.

\para{Optimization landscapes}
\Cref{fig:reference-reanchoring-landscapes}c displays the optimization landscape of \framework in the same run as \cref{fig:coverage-collapse-leading}.
\Cref{fig:reference-reanchoring-landscapes}a matches \cref{fig:coverage-collapse-leading}b,
while \cref{fig:reference-reanchoring-landscapes}b is the strongly regularized version that matches the lowest manifold in \cref{fig:reference-reanchoring-landscapes}c.

\clearpage

\subsection{Proof of \cref{thm:lower-bound-main}}\label{app:lower-bound-proof}
\label{app:lower-bound-full-statement}

\para{Instance and notation}
Fix $\veps\in(0,\frac{1}{64}]$ and consider a single-prompt bandit with
response set $\cY=[m+2]$, where $m=\lceil1/\veps\rceil$ is the number of
zero-reward responses. The rewards are
\begin{align}
  r(1)=1,
  \qquad
  r(2)=1-3\veps,
  \qquad
  r(i)=0 \quad\text{for every } i\ge3.
  \label{eq:lower-bound-rewards}
\end{align}
The base model assigns twice as much probability to response $2$ as to the
optimal response $1$, and divides the remaining probability equally among
the zero-reward responses,
\begin{align}
  \piref(1)=\frac{1}{6},
  \qquad
  \piref(2)=\frac{1}{3},
  \qquad
  \piref(i)=\frac{1}{2m} \quad\text{for every } i\ge3.
  \label{eq:lower-bound-reference}
\end{align}
Thus $\Jstar=1$ and $\Cresp[\piref]=1/\piref(1)=6$.
We use the tabular softmax policy class
\begin{align}
  \pi_\theta(i)=\frac{\exp(\theta_i)}{\sum_{j=1}^{m+2}\exp(\theta_j)}.
\end{align}
Together with a one-hot feature vector for each response, this instance satisfies
\cref{asm:softmax-linear} with $\Rmax=1$.

For $\eta\in(0,\infty]$, write $\beta\ldef{}1/\eta$, with $1/\infty=0$.
The exact policy-gradient update in \cref{eq:fixed-ref-update} optimizes
the fixed-reference objective
\begin{align}
  \Jr(\pi)&\ldef{}J(\pi)-\beta\Dkl{\pi}{\piref},
  &\thetatp&=\thetat+\alpha\nabla\Jr(\pit).
\end{align}
We denote its maximizer over the policy simplex by
\begin{align}
  \pi_\eta^\star\ldef{}\argmax_{\pi}\Jr(\pi).
\end{align}

The following theorem is the full version of \cref{thm:lower-bound-main}.
It shows that the instance above has constant initial coverage, but becoming
and remaining $\veps$-optimal requires passing through a policy with exponentially worse coverage.

\begin{theorem}[Coverage collapse in \reinforce, full version of \protect\cref{thm:lower-bound-main}]
  \label{thm:lower-bound-full}
  Fix $\veps\in(0,\frac{1}{64}]$, and consider the single-prompt bandit in
  \cref{eq:lower-bound-rewards,eq:lower-bound-reference} with the tabular softmax policy class.
  Fix any $\eta>0$, initialize $\pit[0]=\piref$, and run
  \cref{eq:fixed-ref-update} with any fixed learning rate
  $\alpha\in(0,\frac{1}{5}]$.
  For every $T\ge1$, if $\Jstar-J(\pit)\le\veps$ for every $t\ge T$,
  then there exists an iteration $\rho$ with $0<\rho<T$ such that
  \begin{align}
    \frac{1}{7}e^{1/(4\sqrt{\veps})}
    <\Cresp[\pi_\rho]
    \le3e^{1/(4\sqrt{\veps})}
    \qquad\text{and}\qquad
    T
    \ge\frac{e^{1/(4\sqrt{\veps})}}{630\alpha\sqrt{\veps}}
    \ge\frac{\Cresp[\pi_\rho]}{1890\alpha\sqrt{\veps}}.
    \label{eq:lower-bound-full}
  \end{align}
  For this same $\rho$, the expected reward increases strictly throughout
  the coverage-collapse phase,
  \begin{align}
    J(\pitp)>J(\pit)
    \qquad\text{for every }0\le t<\rho.
    \label{eq:lower-bound-monotone-reward}
  \end{align}
  These conclusions also hold for $\eta=\infty$.  In this case,
  the lower bound on the number of gradient steps strengthens to
  \begin{align}
    T\ge\frac{2}{315\alpha}\exp\rbr*{\frac{1}{16\veps}}.
    \label{eq:lower-bound-unregularized}
  \end{align}

\end{theorem}

\Cref{thm:lower-bound-main} follows by suppressing the numerical constants.

We first show that eventual $\veps$-optimality requires
$\beta\le3\veps/\log4$ (\cref{lem:lower-bound-sustained-accuracy,eq:lower-bound-beta-range}).
By symmetry, the zero-reward responses have equal probabilities throughout training,
so the dynamics reduce to the probabilities of responses $1$, $2$, and the
total probability of the zero-reward responses.
We then analyze two phases.
\begin{itemize}
  \item During coverage collapse, response $2$ gains probability relative to
  response $1$ as probability moves away from the zero-reward responses.
  We show that the probability of response $1$ falls to order $e^{-1/(4\sqrt{\veps})}$,
  giving the coverage bounds in
  \cref{eq:lower-bound-coverage-collapse,eq:lower-bound-coverage-collapse-upper}.
  \item To reach an $\veps$-optimal policy, response $1$ must become at least as
  likely as response $2$ (\cref{eq:lower-bound-regret-from-ratio}).
  We bound how quickly $\pit(2)/\pit(1)$ can decrease from its exponentially
  large value after coverage collapse.
  Summing this bound gives the lower bound on the number of gradient steps in
  \cref{eq:lower-bound-recovery-time}.
\end{itemize}

\begin{lemma}[Sustained accuracy requires weak regularization]
  \label{lem:lower-bound-sustained-accuracy}
  Fix $\veps\in(0,1/64]$ and $m=\lceil1/\veps\rceil$.
  On $\cY=[m+2]$, let
  \begin{align}
    &r(1)=1,\qquad r(2)=1-3\veps,\qquad r(i)=0\quad(i\ge3),\\
    &\piref(1)=\frac16,\qquad \piref(2)=\frac13,\qquad
    \piref(i)=\frac{1}{2m}\quad(i\ge3).
  \end{align}
  For $\eta>0$, define $J(\pi)=\sum_i\pi(i)r(i)$ and
  $\Jr(\pi)=J(\pi)-\eta^{-1}\Dkl{\pi}{\piref}$.
  Starting from finite logits $\theta_0$ with $\pi_{\theta_0}=\piref$,
  run exact gradient ascent with a fixed $0<\alpha\le1/5$:
  \begin{align}
    \pi_t(i)=\frac{e^{\theta_{t,i}}}{\sum_j e^{\theta_{t,j}}},
    \qquad
    \theta_{t+1}=\theta_t+
    \alpha\left.\nabla\Jr(\pi_\theta)\right|_{\theta=\theta_t}.
  \end{align}
  If $1-J(\pi_t)\le\veps$ for every $t\ge T$ for some finite $T$, then
  \begin{align}
    1-J(\pi_\eta^\star)\le\veps,
    \qquad
    \eta\ge\frac{\log4}{3\veps},
    \qquad\text{where}\qquad
    \pi_\eta^\star(i)=
    \frac{\piref(i)e^{\eta r(i)}}{\sum_j\piref(j)e^{\eta r(j)}}.
  \end{align}
\end{lemma}

\begin{proof}[\pfref{lem:lower-bound-sustained-accuracy}]
  Define
  \begin{align}
    Q_t(i)&=r(i)-\eta^{-1}\log\frac{\pi_t(i)}{\piref(i)},\\
    g_{t,i}&=\pi_t(i)\rbr*{Q_t(i)-\sum_j\pi_t(j)Q_t(j)}.
  \end{align}
  Differentiating the objective gives $\theta_{t+1,i}=\theta_{t,i}+\alpha g_{t,i}$.
  In particular, $\sum_i g_{t,i}=0$, so $\sum_i\theta_{t,i}$ is constant.
  The zero-reward logits remain equal by symmetry of the initialization and update.

  We first rule out $\eta<1/5$ under the stated accuracy condition.
  For every $t\ge T$,
  \begin{align}
    \veps
    \ge 1-J(\pi_t)
    =3\veps\pi_t(2)+\sum_{i\ge3}\pi_t(i)
    \ge3\veps\rbr*{1-\pi_t(1)}.
  \end{align}
  Hence $\pi_t(1)\ge2/3$, $\pi_t(2)\le1/3$, and
  $\sum_{i\ge3}\pi_t(i)\le\veps$.
  Using symmetry for the zero-reward responses gives
  \begin{align}
    \frac{\pi_t(1)}{\piref(1)}&\ge4,
    &\frac{\pi_t(2)}{\piref(2)}&\le1,\\
    \frac{\pi_t(i)}{\piref(i)}&=2\sum_{j\ge3}\pi_t(j)\le2\veps\le1
    &&(i\ge3).
  \end{align}
  Therefore, for every $i\ne1$,
  \begin{align}
    Q_t(1)-Q_t(i)
    &\le1-\frac{\log4}{\eta}=-c,
    \qquad c\ldef{}\frac{\log4}{\eta}-1>0,\\
    g_{t,1}
    &\le-c\pi_t(1)\rbr*{1-\pi_t(1)}<0.
  \end{align}
  Thus $\theta_{t,1}$ decreases for $t\ge T$.
  Since $\pi_t(1)\ge2/3$, it remains the largest logit.
  Writing $S=\sum_i\theta_{0,i}$, conservation of the sum implies
  \begin{align}
    S-(m+1)\theta_{T,1}
    \le\theta_{t,i}\le\theta_{T,1}
    \qquad(t\ge T,\ i\in[m+2]).
  \end{align}
  All probabilities are consequently bounded away from zero uniformly in
  $t\ge T$.  The bound on $g_{t,1}$ then makes
  $\theta_{t+1,1}-\theta_{t,1}$ bounded above by a strictly negative constant,
  contradicting the displayed lower bound.

  It remains to consider $\eta\ge1/5$, for which $\alpha\le\eta$.
  We show directly that $\pi_t\to\pi_\eta^\star$.
  Set
  \begin{align}
    \theta_{\eta,i}^\star&=\log\piref(i)+\eta r(i),
    &s_{t,i}&=\theta_{t,i}-\theta_{\eta,i}^\star,\\
    \bar s_t&=\sum_i\pi_t(i)s_{t,i},
    &h&=\frac{\alpha}{\eta}\in(0,1].
  \end{align}
  Substitution into the gradient formula yields
  \begin{align}
    s_{t+1,i}
    =\rbr*{1-h\pi_t(i)}s_{t,i}+h\pi_t(i)\bar s_t.
  \end{align}
  Each new error is a convex combination of the old errors.
  If $m_t=\min_i s_{t,i}$ and $M_t=\max_i s_{t,i}$, then
  $m_0\le m_t\le M_t\le M_0$ and
  \begin{align}
    \pi_t(i)
    =\frac{\pi_\eta^\star(i)e^{s_{t,i}}}
    {\sum_j\pi_\eta^\star(j)e^{s_{t,j}}}
    \ge e^{m_0-M_0}\pi_\eta^\star(i)
    \ge\underline p,
    \qquad
    \underline p\ldef{}e^{m_0-M_0}\min_i\pi_\eta^\star(i)>0.
  \end{align}
  The common averaging term therefore gives
  \begin{align}
    m_{t+1}&\ge(1-h\underline p)m_t+h\underline p\bar s_t,\\
    M_{t+1}&\le(1-h\underline p)M_t+h\underline p\bar s_t,\\
    M_{t+1}-m_{t+1}&\le(1-h\underline p)(M_t-m_t).
  \end{align}
  It follows that $M_t-m_t\to0$, and the displayed expression for
  $\pi_t(i)$ gives $\pi_t\to\pi_\eta^\star$.
  By continuity of $J$, the accuracy condition implies
  $1-J(\pi_\eta^\star)\le\veps$.
  Finally, the explicit Gibbs weights give
  \begin{align}
    1-J(\pi_\eta^\star)
    =\frac{6\veps e^{-3\veps\eta}+3e^{-\eta}}
    {1+2e^{-3\veps\eta}+3e^{-\eta}}.
    \label{eq:lower-bound-regularization-bias}
  \end{align}
  Consequently,
  \begin{align}
    4\veps e^{-3\veps\eta}+3(1-\veps)e^{-\eta}\le\veps,
    \qquad
    e^{-3\veps\eta}\le\frac14,
  \end{align}
  which implies $\eta\ge\log4/(3\veps)$.
\end{proof}

\begin{proof}[\pfref{thm:lower-bound-full}]

  \para{Constraint on regularization}
  Fix any $T\ge1$ such that $\Jstar-J(\pi_t)\le\veps$ for every $t\ge T$.
  If $\eta<\infty$,
  the instance and update satisfy all the hypotheses of
  \cref{lem:lower-bound-sustained-accuracy}.  Hence
  \begin{align}
    0<\beta=\frac{1}{\eta}\le\frac{3\veps}{\log4}.
    \label{eq:lower-bound-beta-range}
  \end{align}
  For $\eta=\infty$,
  we have $\beta=0$, so the same upper bound holds automatically.
  In either case, $\pi_T$ is $\veps$-optimal.  The remaining argument uses
  only this fact and $0\le\beta\le3\veps/\log4$.

  \para{Exact dynamics of the probability ratios}
  By symmetry, all arms in $\{3,\ldots,m+2\}$ have the same logit and
  probability at every iteration.  It therefore suffices to track
  \begin{align}
    p_t \ldef{} \pi_t(1),
    \qquad
    q_t \ldef{} \pi_t(2),
    \qquad
    b_t \ldef{} \sum_{i=3}^{m+2}\pi_t(i),
  \end{align}
  together with the ratios
  \begin{align}
    u_t \ldef{} \frac{p_t}{b_t},
    \qquad
    x_t \ldef{} \frac{b_t}{q_t},
    \qquad
    y_t \ldef{} u_tx_t=\frac{p_t}{q_t},
    \qquad
    D_t \ldef{} 1+x_t+y_t.
    \label{eq:lower-bound-ratios}
  \end{align}
  In particular,
  \begin{align}
    p_t=\frac{y_t}{D_t},
    \qquad
    q_t=\frac{1}{D_t},
    \qquad
    b_t=\frac{x_t}{D_t},
  \end{align}
  and the initialization in \cref{eq:lower-bound-reference} gives
  \begin{align}
    u_0=\frac{1}{3},
    \qquad
    x_0=\frac{3}{2},
    \qquad
    y_0=\frac{1}{2}.
    \label{eq:lower-bound-ratio-initialization}
  \end{align}

  For an arm $i$, define
  \begin{align}
    Q_t(i)
    \ldef{}
    r(i)-\beta\log\frac{\pi_t(i)}{\piref(i)}.
  \end{align}
  The softmax gradient has coordinate
  \begin{align}
    \sbr*{\nabla \Jr(\pi_\theta)}_i\bigg|_{\theta=\theta_t}
    =
    \pi_t(i)\rbr*{Q_t(i)-\En_{j\sim\pi_t}[Q_t(j)]}.
    \label{eq:lower-bound-softmax-gradient}
  \end{align}
  Writing $Q_t(0)$ for the common value of $Q_t(i)$ on the arms $i\ge3$,
  the three relevant gaps are
  \begin{align}
    Q_t(1)-Q_t(2)
    &=3\veps-\beta\log(2y_t),
    \\
    Q_t(1)-Q_t(0)
    &=1-\beta\log(3u_t),
    \\
    Q_t(2)-Q_t(0)
    &=1-3\veps+\beta\log\frac{2x_t}{3}.
    \label{eq:lower-bound-Q-gaps}
  \end{align}
  Substituting these gaps into \cref{eq:lower-bound-softmax-gradient} yields
  \begin{align}
    u_{t+1}&=u_t\exp(\alpha A_t),
    &
    x_{t+1}&=x_t\exp(-\alpha B_t),
    &
    y_{t+1}&=y_t\exp(\alpha C_t),
    \label{eq:lower-bound-ratio-dynamics}
  \end{align}
  where $C_t=A_t-B_t$ and
  \begin{align}
    \frac{A_t}{u_t}
    ={}&
    \frac{x_t}{D_t^2}
    \sbr*{
      x_t+3\veps-\beta\log(2y_t)-\beta x_t\log(3u_t)
    }
\\
    &+
    \frac{x_t}{mD_t^2}
    \sbr*{
      x_t
      +\frac{1-3\veps+\beta\log(2x_t/3)}{u_t}
      -\beta x_t\log(3u_t)
    },
    \label{eq:lower-bound-At}
    \\
    B_t
    ={}&
    \frac{x_t}{D_t^2}
    \sbr*{
      1-3\veps(1+u_t)
      +\beta u_t\log(2y_t)
      +\beta\log\frac{2x_t}{3}
    }
\\
    &+
    \frac{x_t}{mD_t^2}
    \sbr*{
      y_t+1-3\veps
      -\beta y_t\log(3u_t)
      +\beta\log\frac{2x_t}{3}
    },
    \label{eq:lower-bound-Bt}
    \\
    C_t
    ={}&
    \frac{1}{D_t^2}
    \sbr*{
      x_t(y_t-1)+3\veps(x_t+2y_t)
      -\beta\rbr*{
        (x_t+2)y_t\log(2y_t)
        -(y_t-1)x_t\log\frac{2x_t}{3}
      }
    }.
    \label{eq:lower-bound-Ct}
  \end{align}

  The unregularized reward gap can also be written in terms of these ratios,
  \begin{align}
    \Jstar-J(\pi_t)
    =b_t+3\veps q_t
    =\frac{x_t+3\veps}{1+x_t+y_t}.
    \label{eq:lower-bound-regret}
  \end{align}
  Consequently, whenever $y_t<1$,
  \begin{align}
    \Jstar-J(\pi_t)
    >
    \frac{x_t+3\veps}{2+x_t}
    >\veps.
    \label{eq:lower-bound-regret-from-ratio}
  \end{align}

  \para{Coverage collapse}
  Define the log-ratio change and the threshold for the collapse phase by
  \begin{align}
    \ell_t \ldef{} \log\frac{x_0}{x_t},
    \qquad
    L_\veps \ldef{} \frac{1}{4\sqrt{\veps}}.
  \end{align}
  We first record the bounds that drive the initial phase.  As long as
  \begin{align}
    \frac{1}{3}\le u_t<4,
    \qquad
    y_t\le\frac{1}{2},
    \qquad
    0\le\ell_t\le L_\veps,
    \label{eq:lower-bound-collapse-region}
  \end{align}
  the quantities in \cref{eq:lower-bound-At,eq:lower-bound-Bt,eq:lower-bound-Ct}
  satisfy
  \begin{align}
    0&\le A_t\le\frac{1}{8},
    \\
    \frac{3}{4}\frac{x_t}{D_t^2}
    &\le B_t\le\frac{1}{5},
    \\
    0&\le\frac{A_t}{u_t}
    \le
    \frac{x_t}{D_t^2}\sbr*{x_t+\frac{15}{2}\veps(1+\ell_t)},
    \\
    C_t&<0.
    \label{eq:lower-bound-collapse-bounds}
  \end{align}
  We verify these bounds directly.  The identities $x_t=x_0e^{-\ell_t}$
  and $2y_t=3u_te^{-\ell_t}$ imply
  \begin{align}
    x_t\le 3y_t\le\frac{3}{2},
    \qquad
    0\le\log(3u_t)\le\ell_t,
    \qquad
    -\log(2y_t)=\ell_t-\log(3u_t),
    \label{eq:lower-bound-collapse-relations}
  \end{align}
  while $m^{-1}\le\veps$ and, by
  \cref{eq:lower-bound-beta-range},
  \begin{align}
    \beta\ell_t
    \le
    \frac{3\veps L_\veps}{\log4}
    \le
    \frac{3}{32\log4}
    <\frac{1}{14}.
  \end{align}
  Using these relations in \cref{eq:lower-bound-At} and dropping its negative
  terms gives
  \begin{align}
    \frac{A_t}{u_t}
    &\le
    \frac{x_t}{D_t^2}
    \sbr*{x_t+3\veps+\beta\ell_t+\veps x_t+3\veps}
    \\
    &\le
    \frac{x_t}{D_t^2}
    \sbr*{x_t+\frac{15}{2}\veps(1+\ell_t)}.
    \label{eq:lower-bound-At-upper}
  \end{align}
  Both brackets in \cref{eq:lower-bound-At} are nonnegative.  Indeed,
  $x_t\le1/(2u_t)$ and
  $\sup_{u>0}\log(3u)/u=3/\mathrm e$ give
  \begin{align}
    \beta x_t\log(3u_t)
    \le
    \frac{\beta}{2}\frac{\log(3u_t)}{u_t}
    \le
    \frac{3\beta}{2\mathrm e}
    <3\veps.
  \end{align}
  Hence the first bracket is at least
  $x_t+3\veps-3\beta/(2\mathrm e)>0$.  The second bracket is at least
  \begin{align}
    x_t
    +\frac{1-3\veps-\frac{3}{2}\beta\ell_t}{u_t}
    >0,
  \end{align}
  where we again used $x_t\le1/(2u_t)$ and
  $\log(3u_t)\le\ell_t$.  Thus $A_t\ge0$.

  For $B_t$, the $m^{-1}$-bracket in \cref{eq:lower-bound-Bt} obeys
  \begin{align}
    &y_t+1-3\veps-\beta y_t\log(3u_t)
    +\beta\log\frac{2x_t}{3}
\\
    &\qquad\ge
    1-3\veps-\frac{3}{2}\beta\ell_t
    >\frac{4}{5}.
  \end{align}
  Substituting \cref{eq:lower-bound-collapse-relations} into the other bracket
  in \cref{eq:lower-bound-Bt} gives
  \begin{align}
    \frac{D_t^2}{x_t}B_t
    \ge
    1-3\veps(1+u_t)
    +\beta u_t\log(3u_t)
    -\beta(u_t+1)L_\veps.
    \label{eq:lower-bound-Bt-lower-pre}
  \end{align}
  To lower-bound this expression uniformly, let $z=\sqrt{\veps}\le1/8$,
  $a=\log(3u_t)$, and $\kappa=3/\log4$. Since $a\le L_\veps$, the coefficient of
  $\beta$ is negative, so the right-hand side is minimized at
  $\beta=\kappa\veps$. The resulting loss from $1$ is
  \begin{align}
    3z^2(1+u_t)
    -\kappa z^2u_ta
    +\frac{\kappa z}{4}(1+u_t).
  \end{align}
  This loss is increasing in $a$ because $a\le L_\veps$ and its derivative has
  the sign of
  \begin{align}
    \frac{\kappa}{4}
    +z\sbr*{3-\kappa(a+1)}
    \ge
    \frac{\kappa}{4}
    +z\sbr*{3-\kappa(L_\veps+1)}
    =z\kappa(\log4-1)>0.
  \end{align}
  If $L_\veps\le\log12$, its maximum is therefore attained at $a=L_\veps$, where it is
  \begin{align}
    \frac{3+e^{L_\veps}}{16L_\veps^2}
    +\frac{\kappa}{16L_\veps}
    \le
    \frac{3+e^2}{64}
    +\frac{\kappa}{32}
    <\frac{1}{4}.
  \end{align}
  Here both terms are decreasing on $L_\veps\in[2,\log12]$.  If
  $L_\veps\ge\log12$, the maximum is attained at $a=\log12$ and is increasing in
  $z\le1/(4\log12)$, so the same bound follows by evaluating at
  $L_\veps=\log12$.  Thus the right-hand side of
  \cref{eq:lower-bound-Bt-lower-pre} is at least $3/4$.

  Conversely, the nonpositive logarithmic terms in
  \cref{eq:lower-bound-Bt} and the relation $x_t\le3y_t$ imply
  \begin{align}
    B_t
    \le
    \frac{x_t}{D_t^2}\rbr*{1+\frac{1+y_t}{m}}
    \le
    \frac{3}{16}\rbr*{1+\frac{3\veps}{2}}
    <\frac{1}{5}.
  \end{align}
  Here we used
  $x_t/D_t^2\le x_t/(1+4x_t/3)^2\le3/16$.

  Multiplying \cref{eq:lower-bound-At} by $u_t$ and again dropping negative
  terms gives the envelope
  \begin{align}
    A_t
    \le
    \frac{x_ty_t+3\veps y_t-\beta y_t\log(2y_t)
    +\veps x_t(1+y_t)}{D_t^2}.
  \end{align}
  On the region in \cref{eq:lower-bound-collapse-region}, these terms are at
  most $1/12$, $3\veps/2$, $3\veps/(2\mathrm e\log4)$, and $\veps/4$,
  respectively.  Thus $A_t<1/8$.
  It remains to verify the sign of $C_t$.  Let
  \begin{align}
    a_t\ldef{}\log(3u_t),
    \qquad
    \delta_t\ldef{}\ell_t-a_t=-\log(2y_t).
  \end{align}
  Dividing \cref{eq:lower-bound-Ct} by $x_t/D_t^2$ and using
  \cref{eq:lower-bound-collapse-relations} gives
  \begin{align}
    \frac{D_t^2}{x_t}C_t
    =
    y_t-1
    +3\veps(1+2u_t)
    +\beta\sbr*{
      (1+2u_t)\ell_t-(y_t+2u_t)a_t
    }.
    \label{eq:lower-bound-Ct-collapse-form}
  \end{align}
  Dropping the term $-\beta y_ta_t$ bounds the right-hand side by
  \begin{align}
    h(\ell_t,\delta_t)
    \ldef{}
    \frac{1}{2}\exp(-\delta_t)-1
    +3\veps\sbr*{
      1+\frac{2}{3}\exp(\ell_t-\delta_t)
    }
    +\beta\sbr*{
      \ell_t
      +\frac{2}{3}\exp(\ell_t-\delta_t)\delta_t
    }.
  \end{align}
  For fixed $\ell_t$, this function is strictly decreasing in $\delta_t$:
  when $\delta_t\le1$, this follows from $\beta<3\veps$, and when
  $\delta_t\ge1$, every term in its derivative is nonpositive.
  The constraints $u_t<4$ and $y_t\le1/2$ give
  $\delta_t\ge\max\{0,\ell_t-\log12\}$, so it remains to check the two
  boundary cases.  If $\ell_t\le\log12$, then
  \begin{align}
    h(\ell_t,0)
    =
    -\frac{1}{2}
    +3\veps\sbr*{1+\frac{2}{3}\exp(\ell_t)}
    +\beta\ell_t
    <-\frac{1}{8}.
    \label{eq:lower-bound-Ct-collapse-case-one}
  \end{align}
  Indeed, if $L_\veps\le\log12$, the positive part is at most
  \begin{align}
    \frac{3+2e^{L_\veps}}{16L_\veps^2}
    +\frac{\kappa}{16L_\veps}
    \le
    \frac{3+2e^2}{64}
    +\frac{\kappa}{32}
    <\frac{3}{8}.
  \end{align}
  The derivative of the displayed upper bound changes sign at most once,
  from negative to positive, on $L_\veps\in[2,\log12]$; its maximum is therefore
  attained at an endpoint, and the value at $L_\veps=2$ is larger.  If
  $L_\veps>\log12$, the same bound follows by evaluating at $L_\veps=\log12$.

  If $\ell_t\ge\log12$, then
  \begin{align}
    h(\ell_t,\ell_t-\log12)
    =
    6\exp(-\ell_t)-1
    +27\veps
    +\beta\sbr*{9\ell_t-8\log12}.
    \label{eq:lower-bound-Ct-collapse-case-two}
  \end{align}
  This is convex in $\ell_t$, so its maximum on $[\log12,L_\veps]$ occurs at an
  endpoint.  At $\ell_t=\log12$, it is smaller than $-1/8$ by the preceding
  bound.  At $\ell_t=L_\veps$, substituting $\veps=1/(16L_\veps^2)$ and
  $\beta\le\kappa/(16L_\veps^2)$ shows that the expression is decreasing for
  $L_\veps\ge\log12$; hence it is again at most its value at $L_\veps=\log12$.

  Thus $C_t<0$, completing the proof of
  \cref{eq:lower-bound-collapse-bounds}.

  Let
  \begin{align}
    \rho
    \ldef{}
    \min\cbr*{t\ge0:u_t\ge4\ \text{or}\ \ell_t>L_\veps}.
    \label{eq:lower-bound-rho}
  \end{align}
  The initialization in \cref{eq:lower-bound-ratio-initialization} and
  \cref{eq:lower-bound-collapse-bounds} imply inductively that, for every
  $t<\rho$, $u_t$ is nondecreasing, $x_t$ is nonincreasing, and
  $y_t\le1/2$.
  If this hitting time is infinite, then
  these properties hold for all $t$.  By
  \cref{eq:lower-bound-regret-from-ratio}, no iterate is $\veps$-optimal.
  We may therefore suppose that $\rho<\infty$.

  We next show that the exit occurs through $\ell_\rho>L_\veps$.  Suppose instead
  that $u_\rho\ge4$ and $\ell_\rho\le L_\veps$.  Define
  \begin{align}
    s_t\ldef{}\frac{1}{u_t},
    \qquad
    \Delta_t\ldef{}\ell_{t+1}-\ell_t=\alpha B_t.
  \end{align}
  For every $t<\rho$, \cref{eq:lower-bound-collapse-bounds} and
  $\alpha\le1/5$ give $0\le\Delta_t\le1/25$.  Hence
  \begin{align}
    s_{t+1}-s_t
    &=
    \frac{1}{u_t}\rbr*{e^{-\alpha A_t}-1}
\\
    &\ge
    -\alpha\frac{A_t}{u_t}
\\
    &\ge
    -\frac{4}{3}\Delta_t
    \sbr*{x_t+\frac{15}{2}\veps(1+\ell_t)}.
    \label{eq:lower-bound-s-recursion}
  \end{align}
  Also, $1-e^{-z}\ge(24/25)z$ for $z\in[0,1/25]$ implies
  \begin{align}
    \sum_{t=0}^{\rho-1}x_t\Delta_t
    \le
    \frac{25}{24}\sum_{t=0}^{\rho-1}(x_t-x_{t+1})
    \le\frac{25}{16}.
    \label{eq:lower-bound-x-sum}
  \end{align}
  Since $\ell_t$ is nondecreasing, a left Riemann sum gives
  \begin{align}
    \sum_{t=0}^{\rho-1}(1+\ell_t)\Delta_t
    \le
    \ell_\rho+\frac{1}{2}\ell_\rho^2
    \le
    L_\veps+\frac{1}{2}L_\veps^2.
    \label{eq:lower-bound-ell-sum}
  \end{align}
  Summing \cref{eq:lower-bound-s-recursion} and using
  \cref{eq:lower-bound-x-sum,eq:lower-bound-ell-sum} yields
  \begin{align}
    s_\rho
    &\ge
    3-\frac{25}{12}
    -10\veps\rbr*{L_\veps+\frac{1}{2}L_\veps^2}
\\
    &\ge
    3-\frac{25}{12}-\frac{5}{8}
    =\frac{7}{24}.
  \end{align}
  This contradicts $u_\rho\ge4$, which would imply $s_\rho\le1/4$.
  Therefore $\ell_\rho>L_\veps$.

  Since $u_{\rho-1}<4$, $A_{\rho-1}\le1/8$, and $\alpha\le1/5$,
  \begin{align}
    u_\rho<4e^{1/40}.
  \end{align}
  Together with $x_\rho=x_0e^{-\ell_\rho}$, this gives
  \begin{align}
    y_\rho=u_\rho x_\rho
    <6\exp(1/40-L_\veps)
    <7\exp(-L_\veps).
    \label{eq:lower-bound-y-rho}
  \end{align}
  In particular, $p_\rho=y_\rho/D_\rho<y_\rho$, and hence
  \begin{align}
    \Cresp[\pi_\rho]
    =\frac{1}{p_\rho}
    >\frac{1}{7}\exp\rbr*{\frac{1}{4\sqrt{\veps}}}.
    \label{eq:lower-bound-coverage-collapse}
  \end{align}
  Conversely, since
  $\ell_{\rho-1}\le L_\veps$ and
  $\ell_\rho-\ell_{\rho-1}=\alpha B_{\rho-1}\le1/25$, while
  $u_\rho\ge u_0=1/3$, we have
  \begin{align}
    \Cresp[\pi_\rho]
    &=1+\frac{1}{u_\rho}+\frac{1}{y_\rho}
    =1+\frac{1}{u_\rho}
    +\frac{2\exp(\ell_\rho)}{3u_\rho}
\\
    &\le
    4+2\exp\rbr*{L_\veps+\frac{1}{25}}
    \le
    3\exp(L_\veps).
    \label{eq:lower-bound-coverage-collapse-upper}
  \end{align}
  The last inequality uses $L_\veps\ge2$.

  We also verify \cref{eq:lower-bound-monotone-reward}.
  For every $t<\rho$, \cref{eq:lower-bound-ratio-dynamics,eq:lower-bound-collapse-bounds}
  give $u_{t+1}\ge u_t$ and $x_{t+1}<x_t$.
  In the coordinates of \cref{eq:lower-bound-ratios}, the expected reward is
  \begin{align}
    J(\pit)=\frac{u_tx_t+1-3\veps}{1+(1+u_t)x_t}.
  \end{align}
  This expression is increasing in $u_t$. Hence
  \begin{align}
    J(\pitp)-J(\pit)
    &\ge
    \frac{u_tx_{t+1}+1-3\veps}{1+(1+u_t)x_{t+1}}
    -\frac{u_tx_t+1-3\veps}{1+(1+u_t)x_t}
\\
    &=
    \frac{(1-3\veps-3\veps u_t)(x_t-x_{t+1})}
    {\rbr*{1+(1+u_t)x_{t+1}}\rbr*{1+(1+u_t)x_t}}
    >0.
  \end{align}
  The last inequality follows from $x_{t+1}<x_t$ and
  $1-3\veps-3\veps u_t>1-15\veps\ge49/64$,
  since $u_t<4$ and $\veps\le1/64$.

  \para{Gradient steps required to reach an $\veps$-optimal policy}
  We first establish two bounds that hold whenever $u_t\ge1/3$ and $y_t<1$.
  Rewriting \cref{eq:lower-bound-At} using $x_t=y_t/u_t$ gives
  \begin{align}
    A_t
    =
    \frac{x_t}{D_t^2}R(u_t,y_t),
  \end{align}
  where
  \begin{align}
    R(u,y)
    \ldef{}
    3\veps u-\beta u\log(2y)
    +\rbr*{1+\frac{1}{m}}y\rbr*{1-\beta\log(3u)}
    +\frac{1-3\veps+\beta\log(2y/(3u))}{m}.
    \label{eq:lower-bound-R}
  \end{align}
  Write $a=\log(3u)$ and $s=\log(1/y)$.  Grouping the terms proportional
  to $s$ rewrites this expression as
  \begin{align}
    R(u,y)
    ={}&
    u\rbr*{3\veps-\beta\log2}
    +\beta\rbr*{u-\frac{1}{m}}s
\\
    &+
    \rbr*{1+\frac{1}{m}}y(1-\beta a)
    +\frac{1-3\veps+\beta\log2-\beta a}{m}.
  \end{align}
  By \cref{eq:lower-bound-beta-range},
  \begin{align}
    3\veps-\beta\log2
    \ge\frac{3}{2}\veps.
    \label{eq:lower-bound-recovery-margin}
  \end{align}
  Every term in the expression
  for $R(u,y)$ is therefore nonnegative for $u\in[1/3,4]$ and $y<1$,
  and the first term is positive.  Here we also used
  $\beta\log12<1$ and $m^{-1}\le\veps$.
  For $u\ge4$, direct differentiation gives
  \begin{align}
    \partial_uR(u,y)
    &\ge
    \frac{3}{2}\veps
    -\frac{\beta}{4}\rbr*{1+\frac{2}{m}}
    >0.
  \end{align}
  Since $R(4,y)>0$ by the preceding case, this proves
  \begin{align}
    A_t>0
    \qquad\text{whenever}\qquad
    u_t\ge\frac{1}{3}\ \text{and}\ y_t<1.
    \label{eq:lower-bound-global-At}
  \end{align}

  Under the same conditions, $x_t=y_t/u_t\le3y_t$.  Applying this relation
  to \cref{eq:lower-bound-Ct} yields
  \begin{align}
    C_t
    \le
    18\veps y_t\rbr*{1+\log\frac{1}{y_t}}.
    \label{eq:lower-bound-global-Ct}
  \end{align}
  To see this, the reward term contributes at most $15\veps y_t$.
  The term containing $-\log(2y_t)$ is nonpositive when $y_t\ge1/2$;
  when $y_t<1/2$, it contributes at most
  $8\veps y_t\log(1/y_t)$.  Finally, the remaining logarithmic term is
  positive only when $x_t<3/2$.  Writing $x_t=c_ty_t$ with $c_t\le3$ gives
  \begin{align}
    x_t\log\frac{3}{2x_t}
    &=
    c_ty_t\log\frac{1}{y_t}
    +y_tc_t\log\frac{3}{2c_t}
    \\
    &\le
    3y_t\log\frac{1}{y_t}+y_t.
  \end{align}
  By \cref{eq:lower-bound-beta-range}, its contribution is at most
  $3\veps y_t+7\veps y_t\log(1/y_t)$.
  Combining the three bounds proves
  \cref{eq:lower-bound-global-Ct}.

  Let
  \begin{align}
    \tau_*
    \ldef{}
    \min\cbr*{t\ge\rho:y_t\ge1}.
  \end{align}
  If $\tau_*=\infty$, no iterate after $\rho$ is $\veps$-optimal, so the
  theorem follows.  Otherwise, \cref{eq:lower-bound-global-At} and the fact
  that $u_\rho\ge u_0=1/3$ show that $u_t\ge1/3$ for every
  $t\in[\rho,\tau_*]$.  For $t<\tau_*$, the inequality
  $e^{-z}\ge1-z$ and \cref{eq:lower-bound-global-Ct} therefore give the
  reciprocal recursion
  \begin{align}
    \frac{1}{y_{t+1}}
    &=
    \frac{1}{y_t}e^{-\alpha C_t}
\\
    &\ge
    \frac{1}{y_t}\rbr*{1-\alpha C_t}
\\
    &\ge
    \frac{1}{y_t}
    -18\alpha\veps\rbr*{1+\log\frac{1}{y_t}}.
    \label{eq:lower-bound-reciprocal-recursion}
  \end{align}

  Choose $t_{\max}\in[\rho,\tau_*)$ maximizing $1/y_t$.  Summing
  \cref{eq:lower-bound-reciprocal-recursion} from $t_{\max}$ to
  $\tau_*-1$, and using $y_{\tau_*}^{-1}\le1$ and
  $y_t^{-1}\le y_{t_{\max}}^{-1}$ for $t<\tau_*$, gives
  \begin{align}
    \tau_*-\rho
    \ge
    \frac{y_{t_{\max}}^{-1}-1}
    {18\alpha\veps\rbr*{1+\log(y_{t_{\max}}^{-1})}}.
    \label{eq:lower-bound-recovery-pre}
  \end{align}
  The function $f(z)=(z-1)/(1+\log z)$ is increasing on $z>1$ because
  \begin{align}
    f'(z)=\frac{\log z+1/z}{(1+\log z)^2}>0.
  \end{align}
  By \cref{eq:lower-bound-y-rho},
  $y_{t_{\max}}^{-1}\ge y_\rho^{-1}>e^{L_\veps}/7$.  Since $L_\veps\ge2$,
  \begin{align}
    f\rbr*{\frac{e^{L_\veps}}{7}}
    &=
    e^{L_\veps}\frac{1/7-e^{-L_\veps}}{1+L_\veps-\log7}
    \ge
    \frac{e^{L_\veps}}{140L_\veps}.
  \end{align}
  Here we used $1/7-e^{-L_\veps}\ge1/7-e^{-2}>1/140$ and
  $1+L_\veps-\log7\le L_\veps$.
  Substituting this into \cref{eq:lower-bound-recovery-pre} and using
  $\veps L_\veps=\sqrt{\veps}/4$ yields
  \begin{align}
    \tau_*-\rho
    \ge
    \frac{1}{630\alpha\sqrt{\veps}}
    \exp\rbr*{\frac{1}{4\sqrt{\veps}}}.
    \label{eq:lower-bound-recovery-time}
  \end{align}

  By \cref{eq:lower-bound-regret-from-ratio}, every $\veps$-optimal iterate
  must occur at or after $\tau_*$.  Hence any $T$ satisfying
  $\Jstar-J(\pi_T)\le\veps$ obeys
  \begin{align}
    T
    \ge \tau_*
    \ge \tau_* - \rho
    \ge
    \frac{1}{630\alpha\sqrt{\veps}}
    \exp\rbr*{\frac{1}{4\sqrt{\veps}}}
    \ge
    \frac{\Cresp[\pi_\rho]}{1890\alpha\sqrt{\veps}}.
  \end{align}
  Moreover, \cref{eq:lower-bound-coverage-collapse,eq:lower-bound-coverage-collapse-upper}
  hold at the same index $\rho<\tau_*\le T$, proving that
  \begin{align}
    \Cresp[\pi_\rho]
    =
    \exp\rbr*{\Theta\rbr*{\frac{1}{\sqrt{\veps}}}}.
  \end{align}
  Since $u_0=1/3<4$ and $\ell_0=0$,
  the hitting time satisfies $\rho\ge1$, so $0<\rho<T$.
  Together with the preceding display, this establishes
  \cref{eq:lower-bound-full}, and \cref{thm:lower-bound-main} follows by
  suppressing the constants.

  \para{Stronger bound without regularization}
  When $\beta=0$, the term growing with $\ell_t$ in
  \cref{eq:lower-bound-At-upper} disappears, allowing a longer collapse phase.
  For this final argument, set $L_\veps=1/(16\veps)$ and redefine $\rho$
  using this threshold. Let $\tau_*=\min\cbr*{t\ge\rho:y_t\ge1}$.
  On the region in \cref{eq:lower-bound-collapse-region},
  \cref{eq:lower-bound-At} gives
  \begin{align}
    0\le\frac{A_t}{u_t}
    &\le\frac{x_t}{D_t^2}
    \sbr*{x_t+3\veps+\veps x_t+3\veps}
    \le\frac{x_t}{D_t^2}\sbr*{x_t+\frac{15}{2}\veps}.
  \end{align}
  The other bounds in \cref{eq:lower-bound-collapse-bounds} remain valid;
  \cref{eq:lower-bound-Bt} gives
  $B_t\ge(1-15\veps)x_t/D_t^2\ge(49/64)x_t/D_t^2$,
  and \cref{eq:lower-bound-Ct} gives
  $C_t\le(-1/2+27\veps)x_t/D_t^2<0$.
  The upper bounds on $A_t$ and $B_t$ above do not depend on the threshold when $\beta=0$.
  The same induction therefore keeps the iterates in the collapse region
  for $t<\rho$.  Moreover, $\rho<\infty$, since otherwise
  \cref{eq:lower-bound-regret-from-ratio} would rule out the
  $\veps$-optimality of $\pi_T$.
  Thus, if $u_\rho\ge4$ and $\ell_\rho\le L_\veps$, repeating the arguments for
  \cref{eq:lower-bound-s-recursion,eq:lower-bound-x-sum} with the sharper bound on $A_t/u_t$
  and using $\sum_{t<\rho}\Delta_t=\ell_\rho$ yields
  \begin{align}
    s_\rho
    &\ge3-\frac{4}{3}\sum_{t<\rho}x_t\Delta_t
    -10\veps\sum_{t<\rho}\Delta_t
    \\
    &\ge3-\frac{25}{12}-10\veps L_\veps
    =\frac{7}{24}>\frac{1}{4}.
  \end{align}
  Hence the exit again occurs through $\ell_\rho>L_\veps$, and
  the argument for \cref{eq:lower-bound-y-rho} gives $y_\rho<7e^{-L_\veps}$.
  Since $L_\veps\ge4$, repeating the argument for
  \cref{eq:lower-bound-recovery-pre} and using the monotonicity of $f$ and
  $f(e^{L_\veps}/7)\ge e^{L_\veps}/(140L_\veps)$ gives
  \begin{align}
    T\ge\tau_*-\rho
    \ge\frac{e^{L_\veps}}{2520\alpha\veps L_\veps}
    =\frac{2}{315\alpha}\exp\rbr*{\frac{1}{16\veps}},
  \end{align}
  proving \cref{eq:lower-bound-unregularized}.

\end{proof}

\clearpage
\part{\frameworklong}\label{part:appendix-framework}

\section{The KL Divergence Criterion (Proof of \cref*{thm:kl-regret-main})}\label{app:kl}
\label{app:kl-shared-statements}
This section proves \cref{thm:kl-regret-main},
which establishes the KL divergence criterion
sufficient for convergence and coverage preservation in \framework.
In \cref{app:kl-shared-supporting} we first develop supporting lemmas,
then prove \cref{thm:kl-regret-main} in \cref{app:kl-shared-global}.

\para{Notation}
Throughout this section, fix $\eta>0$, integer $T\geq1$,
and a given sequence of policies $\cbr*{\pit}_{t=0}^T$ with $\pit[0]=\piref$.
Recall that for all $0 \leq t < T$, the exact \framework target is given by
\begin{align}
  \piplust(y\mid x)
  &\ldef{}\pit(y\mid x)e^{\eta(r(x,y)-\btstar(x))},
  \quad\text{where}\quad
  \btstar(x)
  \ldef{}\frac1\eta\log\En_{y\sim\pit(\cdot\mid x)}\sbr*{e^{\eta r(x,y)}}.
  \label{eq:kl-shared-update}
\end{align}
We also introduce notation for a fixed prompt $x$.
Denote the expected reward under $\pi$ given $x$ to be
$J_x(\pi)\ldef{}\En_{y\sim\pi(\cdot\mid x)}\sbr*{r(x,y)}$,
and the maximum reward at a prompt $x$ to be 
$r_x^\star\ldef{}\max_{y\in\cY}r(x,y)$.
In addition, we denote $\pt(x)$ to be probability mass of $\pit$ over optimal responses,
\begin{align}
  \pt(x)&\ldef{}\pit(\Ystarx\mid x),
  \quad\text{where}\quad
  \Ystarx&\ldef{}\cbr*{y\in\cY:r(x,y)=r_x^\star},
  \label{eq:kl-shared-reward-notation}
\end{align}
where $\Ystarx$ is the set of optimal responses at $x$.
Lastly, for shorthand we use the following to denote the KL divergence error of $\pitp$,
\begin{align}
  \mathsf{KL}_t(x)
  &\ldef{}\Dkl{\pitp(\cdot\mid x)}{\piplust(\cdot\mid x)},
  &\mathsf{KL}_t
  &\ldef{}\En_{x\sim\mu}\sbr*{\mathsf{KL}_t(x)},
\end{align}
and the average error over $T$ iterations to be,
\begin{align}
  \ol{\mathsf{KL}}(x)
  &\ldef{}\frac1T\sum_{t=0}^{T-1}\mathsf{KL}_t(x),
  &\ol{\mathsf{KL}}
  &\ldef{}\frac1T\sum_{t=0}^{T-1}\mathsf{KL}_t.
  \label{eq:kl-shared-errors}
\end{align}

\subsection{Supporting Lemmas}
\label{app:kl-shared-supporting}

We first develop supporting estimates for approximate updates.

Later, \cref{app:general-apmd} reuses these lemmas to extend \cref{thm:kl-regret-main}
to general target rewards and comparator policies,
while \cref{app:mse-accumulation} adapts their arguments to
mean-squared approximation error.
They are organized as follows.

\begin{itemize}

  \item (Approximate improvement)
  \Cref{lem:kl-shared-improvement} shows that the KL divergence
  $\Dkl{\pitp}{\piplust}$ controls how much $\pitp$ can decrease
  in reward relative to $\pit$.
  It is the approximate version of \cref{lem:exact-apmd-improvement}.

  \item (Coverage approximately preserved)
  \cref{lem:kl-shared-coverage-accumulation} bounds how much
  the probabilities of upper reward quantiles
  degrade from $\pit[0]$ to $\pit$, in terms of the accumulated KL divergence error.

 \item (Error accumulation) \cref{lem:kl-shared-accumulation}
  quantifies how approximation errors perturb sequential exponential
  reweighting under the exact \framework updates.
  The perturbation is measured through
  a given function of interest $f$ and bounded in terms of the
  cumulative KL error.

  \item (Technical lemmas)
  \cref{lem:kl-shared-normalizer} and \cref{lem:kl-shared-hellinger}.
\end{itemize}

\begin{lemma}[Approximate improvement]
  \label{lem:kl-shared-improvement}
  Fix $x \in \cX$.
  For every $0\leq t<T$,
  \begin{align}
    J_x(\pit)&\leq\btstar(x)
    \leq J_x(\pitp)+\frac{\mathsf{KL}_t(x)}\eta.
    \label{eq:kl-shared-improvement}
  \end{align}
\end{lemma}

\begin{proof}[\pfref{lem:kl-shared-improvement}]
  We obtain the first inequality using Jensen's inequality,
  \begin{align}
    \eta J_x(\pit)
    \leq\log\En_{y\sim\pit(\cdot\mid x)}\sbr*{e^{\eta r(x,y)}}
    =\eta\btstar(x).
  \end{align}
  For the second inequality,
  expanding the KL divergence gives

  \begin{align}
    \mathsf{KL}_t(x)
    &= \En_{y\sim\pitp(\cdot\mid x)}\sbr*{\log\frac{\pitp(y\mid x)}{\pit(y\mid x)} - \eta r(x,y) + \eta\btstar(x)}
    \\
    &= \Dkl{\pitp(\cdot\mid x)}{\pit(\cdot\mid x)} - \eta J_x(\pitp) + \eta\btstar(x)
    \\
    &\ge -\eta J_x(\pitp) + \eta\btstar(x),
  \end{align}
  by nonnegativity of KL divergence. Rearranging gives the second inequality.

\end{proof}

\begin{lemma}[Coverage from accumulated KL error]
  \label{lem:kl-shared-coverage-accumulation}
  Fix $x\in\cX$ and $a\in[0,\Rmax]$, and let
  $\cY_a(x)=\cbr*{y\in\cY:r(x,y)\geq a}$.
  Then, for every $0\leq t\leq T$,
  \begin{align}
    \sqrt{\pit(\cY_a(x)\mid x)}
    \geq\sqrt{\piref(\cY_a(x)\mid x)}
      -\sum_{s=0}^{t-1}\sqrt{\mathsf{KL}_s(x)}.
    \label{eq:kl-shared-coverage-accumulation-root}
  \end{align}
\end{lemma}

\begin{proof}[\pfref{lem:kl-shared-coverage-accumulation}]

  For every $0\leq t<T$, \cref{lem:kl-shared-hellinger} gives
  \begin{align}
    \sum_y\rbr*{\sqrt{\piplust(y\mid x)}-\sqrt{\pitp(y\mid x)}}^2
    \leq\mathsf{KL}_t(x).
  \end{align}
  Applying the reverse triangle inequality to the
  vectors $(\sqrt{\pitp(y\mid x)})_{y\in\cY_a(x)}$ and
  $(\sqrt{\piplust(y\mid x)})_{y\in\cY_a(x)}$ then gives
  \begin{align}
    \sqrt{\pitp(\cY_a(x)\mid x)}
    &\geq\sqrt{\piplust(\cY_a(x)\mid x)}
      -\rbr*{\sum_{y\in\cY_a(x)}
        \rbr*{\sqrt{\piplust(y\mid x)}-\sqrt{\pitp(y\mid x)}}^2}^{1/2}
    \\
    &\geq\sqrt{\piplust(\cY_a(x)\mid x)}-\sqrt{\mathsf{KL}_t(x)}.
    \label{eq:kl-shared-coverage-from-accumulation-step}
  \end{align}

  The exact update also satisfies
  \begin{align}
    &e^{\eta\btstar(x)}
      \rbr*{\piplust(\cY_a(x)\mid x)-\pit(\cY_a(x)\mid x)}
\\
    &\qquad=\sum_{y\in\cY_a(x)}\sum_{y'\notin\cY_a(x)}
      \pit(y\mid x)\pit(y'\mid x)
      \rbr*{e^{\eta r(x,y)}-e^{\eta r(x,y')}}
    \geq0.
    \label{eq:kl-shared-upper-set-tilt}
  \end{align}
  Combining with \cref{eq:kl-shared-coverage-from-accumulation-step} gives
  \begin{align}
    \sqrt{\pitp(\cY_a(x)\mid x)}
    \geq\sqrt{\pit(\cY_a(x)\mid x)}-\sqrt{\mathsf{KL}_t(x)}.
  \end{align}

  For any $0\leq t\leq T$, summing over $0\leq s<t$ gives
  \begin{align}
    \sqrt{\pit(\cY_a(x)\mid x)}
    \geq\sqrt{\piref(\cY_a(x)\mid x)}
      -\sum_{s=0}^{t-1}\sqrt{\mathsf{KL}_s(x)}.
    \label{eq:kl-shared-coverage-root}
  \end{align}

\end{proof}

The intuition for the following lemma is that under exact updates,
$\frac{\pit(y\mid x)}{\piref(y\mid x)}
= e^{\eta\sum_{s=0}^{t-1}(r(x,y)-\btstar[s](x))}$.
This result characterizes how this relationship is perturbed by cumulative approximation errors,
as measured in expectation over $\piref$ for a given function of interest $f(x,y)$.
The square-root form is better suited to accounting for approximation
errors measured in KL divergence,
which accumulate additively when updates are unrolled.

\begin{lemma}[Accumulation of KL error]
  \label{lem:kl-shared-accumulation}
  Let $f:\cX\times\cY\to[0,\infty)$ be bounded.
  For every $1\leq t\leq T$,
  \begin{align}
    \En_{\piref}\sbr*{f(x,y)\sqrt{\frac{\pit(y\mid x)}{\piref(y\mid x)}}}
    &\geq\En_{\piref}\sbr*{
      f(x,y)e^{\frac\eta2\sum_{s=0}^{t-1}(r(x,y)-\btstar[s](x))}}
    \\
    &\quad-\sum_{s=1}^t\sqrt{\mathsf{KL}_{s-1}}
      \rbr*{\En_{\piref}\sbr*{
        f(x,y)^2e^{\eta\sum_{u=s}^{t-1}(r(x,y)-\btstar[u](x))}}}^{1/2}.
    \label{eq:kl-shared-accumulation-exact}
  \end{align}
\end{lemma}

\begin{proof}[\pfref{lem:kl-shared-accumulation}]
  By \cref{lem:kl-shared-hellinger}, for every $x$ and $0\leq t<T$,
  \begin{align}
    \sum_y\rbr*{\sqrt{\piplust(y\mid x)}-\sqrt{\pitp(y\mid x)}}^2
    \leq\mathsf{KL}_t(x).
    \label{eq:kl-shared-update-hellinger}
  \end{align}
  At each $(x,y)$, the definition of the exact update $\piplust$ gives
  \begin{align}
    \sqrt{\pitp(y\mid x)}
    =\sqrt{\pit(y\mid x)}e^{\frac\eta2(r(x,y)-\btstar(x))}
      -\rbr*{\sqrt{\piplust(y\mid x)}-\sqrt{\pitp(y\mid x)}}.
  \end{align}
  For any $1\leq t\leq T$, unrolling over $0\leq s<t$ gives
  \begin{align}
    \sqrt{\pit(y\mid x)}
    &=\sqrt{\piref(y\mid x)}
      e^{\frac\eta2\sum_{s=0}^{t-1}(r(x,y)-\btstar[s](x))}
    \\
    &\quad-\sum_{s=1}^t
      \rbr*{\sqrt{\piplust[s-1](y\mid x)}-\sqrt{\pit[s](y\mid x)}}
      e^{\frac\eta2\sum_{u=s}^{t-1}(r(x,y)-\btstar[u](x))}.
    \label{eq:kl-shared-unrolling}
  \end{align}
  Fix $x$, multiply by $\sqrt{\piref(y\mid x)}f(x,y)$,
  and sum over responses. For each term in the sum on the RHS,
  the Cauchy--Schwarz inequality and \cref{eq:kl-shared-update-hellinger} give
  \begin{align}
    &\sum_y\sqrt{\piref(y\mid x)}f(x,y)
      \rbr*{\sqrt{\piplust[s-1](y\mid x)}-\sqrt{\pit[s](y\mid x)}}
      e^{\frac\eta2\sum_{u=s}^{t-1}(r(x,y)-\btstar[u](x))}
    \\
    &\qquad\leq
      \rbr*{\sum_y\rbr*{\sqrt{\piplust[s-1](y\mid x)}-\sqrt{\pit[s](y\mid x)}}^2}^{1/2}
      \rbr*{\En_{y\sim\piref(\cdot\mid x)}\sbr*{
        f(x,y)^2e^{\eta\sum_{u=s}^{t-1}(r(x,y)-\btstar[u](x))}}}^{1/2}
    \\
    &\qquad\leq\sqrt{\mathsf{KL}_{s-1}(x)}
      \rbr*{\En_{y\sim\piref(\cdot\mid x)}\sbr*{
        f(x,y)^2e^{\eta\sum_{u=s}^{t-1}(r(x,y)-\btstar[u](x))}}}^{1/2}.
  \end{align}
  Substituting into \cref{eq:kl-shared-unrolling} yields
  \begin{align}
    &\En_{y\sim\piref(\cdot\mid x)}\sbr*{
      f(x,y)\sqrt{\frac{\pit(y\mid x)}{\piref(y\mid x)}}}
    \\
    &\quad\geq\En_{y\sim\piref(\cdot\mid x)}\sbr*{
      f(x,y)e^{\frac\eta2\sum_{s=0}^{t-1}(r(x,y)-\btstar[s](x))}}
    \\
    &\qquad-\sum_{s=1}^t\sqrt{\mathsf{KL}_{s-1}(x)}
      \rbr*{\En_{y\sim\piref(\cdot\mid x)}\sbr*{
        f(x,y)^2e^{\eta\sum_{u=s}^{t-1}(r(x,y)-\btstar[u](x))}}}^{1/2}.
    \label{eq:kl-shared-accumulation-prompt}
  \end{align}
  Taking the expectation over $x\sim\mu$ and applying the Cauchy--Schwarz
  inequality to each error term proves \cref{eq:kl-shared-accumulation-exact}.
\end{proof}

\begin{lemma}
  \label{lem:kl-shared-normalizer}
  Fix $x \in \cX$. For every $0\leq t<T$,
  \begin{align}
    \btstar(x)&\leq J_x(\piT)+\frac{T\ol{\mathsf{KL}}(x)}\eta.
    \label{eq:kl-shared-normalizer}
  \end{align}
\end{lemma}

\begin{proof}[\pfref{lem:kl-shared-normalizer}]

  Fix $0\leq t<T$. For every $t<s<T$,
  \cref{lem:kl-shared-improvement} gives

  \begin{align}
    J_x(\pit[s])-J_x(\pit[s+1])
    \leq\frac{\mathsf{KL}_s(x)}\eta.
  \end{align}
  Summing these inequalities over $t<s<T$, we obtain
  \begin{align}
    J_x(\pitp)-J_x(\piT)
    &=
    \sum_{s=t+1}^{T-1}
    \rbr*{J_x(\pit[s])-J_x(\pit[s+1])}
    \leq \frac1\eta \sum_{s=t+1}^{T-1}\mathsf{KL}_s(x).
  \end{align}
  Then, applying the second inequality in \cref{lem:kl-shared-improvement} at iteration $t$
  and combining with the preceding inequality, we have

  \begin{align}
    \btstar(x)
    &\le J_x(\pitp) + \frac{\mathsf{KL}_t(x)}\eta
    \le J_x(\piT) + \frac1\eta \sum_{s=t}^{T-1}\mathsf{KL}_s(x).
  \end{align}

  The result follows from
  $\sum_{s=t}^{T-1}\mathsf{KL}_s(x) \le \sum_{s=0}^{T-1} \mathsf{KL}_s(x) = T \ol{\mathsf{KL}}(x)$.
\end{proof}

\begin{lemma}[KL--Hellinger lower bound]
  \label{lem:kl-shared-hellinger}
  For any probability distributions $p,q \in \Delta(\cY)$,
  \begin{align}
    \Dkl{p}{q}\geq\sum_{y\in\cY}\rbr*{\sqrt{p(y)}-\sqrt{q(y)}}^2.
    \label{eq:kl-shared-hellinger}
  \end{align}
\end{lemma}

\begin{proof}[\pfref{lem:kl-shared-hellinger}]
  The inequality $-\log u\geq2(1-\sqrt{u})$ gives
  \begin{align}
    \Dkl{p}{q}
    &\geq2\sum_{y:p(y)>0}p(y)\rbr*{1-\sqrt{\frac{q(y)}{p(y)}}}
    \\
    &=2-2\sum_y\sqrt{p(y)q(y)}
    =\sum_y\rbr*{\sqrt{p(y)}-\sqrt{q(y)}}^2.
  \end{align}
  If $q(y)=0<p(y)$ for some $y$, the KL divergence is infinite so the claim in the lemma is immediate.
\end{proof}

\subsection{Proof of \cref{thm:kl-regret-main}}
\label{app:kl-shared-global}

We first state \cref{thm:kl-shared-regret}, which extends the exact policy mirror descent analysis in \cref{thm:exact-apmd-rate} to approximate updates,
where approximation error is quantified by the KL divergence $\Dkl{\pitp}{\piplust}$.
The proof of \cref{thm:kl-regret-main} directly follows below;
substituting the prescribed inner KL tolerance into \cref{thm:kl-shared-regret} yields the main theorem.
Finally, we prove \cref{thm:kl-shared-regret}.

\begin{theorem}[Global convergence under \framework{}]
  \label{thm:kl-shared-regret}
  The final policy $\piT$ satisfies
  \begin{align}
    \Jstar-J(\piT)
    &\leq\frac{\En_{x\sim\mu}\sbr*{\log\rbr*{1/\pt[0](x)}}}{\eta T}
      +\frac1\eta\sum_{t=0}^{T-1}\mathsf{KL}_t
    \\
    &\quad+\frac{2\sqrt e}{\eta}
      \sqrt{\Cresp\sum_{t=0}^{T-1}\frac{\mathsf{KL}_t}{t+1}}.
    \label{eq:kl-shared-regret}
  \end{align}
  Moreover, for any $c\in(0,1)$, with probability at least
  $1-\frac{\sqrt{\Cresp}}{1-\sqrt c}
    \sum_{t=0}^{T-1}\sqrt{\mathsf{KL}_t}$ over $x\sim\mu$,
  \begin{align}
    \pit(\Ystarx\mid x)\geq c\piref(\Ystarx\mid x),
    \qquad\forall 0\leq t\leq T.
    \label{eq:kl-shared-regret-coverage}
  \end{align}
\end{theorem}

\begin{proof}[\pfref{thm:kl-regret-main}]
  We first consider the coverage claim in \cref{thm:kl-regret-main}.
  We apply \cref{thm:kl-shared-regret} with $c=1/2$
  and the prescribed inner KL tolerance,
  \begin{align}
    \mathsf{KL}_t\leq\ein
    =\frac1{64\Cresp T^2\log(eT)},
    \qquad 0\leq t<T.
    \label{eq:kl-shared-tolerance}
  \end{align}
  Since $\sum_{t=0}^{T-1}\sqrt{\mathsf{KL}_t}\leq T\sqrt{\ein}$,
  the theorem gives
  \begin{align}
    \Pr_{x\sim\mu}\rbr*{
      \pit(\Ystarx\mid x)\geq\tfrac12\piref(\Ystarx\mid x),
      \ \forall 0\leq t\leq T}
    &\geq1-(2+\sqrt2)T\sqrt{\Cresp\ein}
    \\
    &=1-\frac{2+\sqrt2}{8\sqrt{\log(eT)}}\geq1-\frac{1}{2\sqrt{\log(eT)}},
  \end{align}
  which is the claimed coverage guarantee.
  To prove the convergence guarantee in \cref{thm:kl-regret-main},
  we again apply \cref{thm:kl-shared-regret} with \cref{eq:kl-shared-tolerance}.
  Since $\En_{x\sim\mu}\sbr*{\log(1/\pt[0](x))}\leq\log\Cresp$
  by Jensen's inequality and $\sum_{t=1}^T1/t\leq\log(eT)$, we obtain
  \begin{align}
    \Jstar-J(\piT)
    &\leq\frac{\log\Cresp}{\eta T}
      +\frac{T\ein}{\eta}
      +\frac{2\sqrt e}{\eta}\sqrt{\Cresp\ein\log(eT)}
    \\
    &\leq\frac{\log\Cresp+1/64+\sqrt e/4}{\eta T}
    \leq\frac{\log(2\Cresp)}{\eta T}.
  \end{align}
  This proves the convergence claim in \cref{eq:framework-convergence}.
\end{proof}

Next, we prove \cref{thm:kl-shared-regret},
the approximate analogue of
the mirror descent convergence guarantee in \cref{thm:exact-apmd-rate}.
We use \cref{lem:kl-shared-prompt-reward}, proved below, for the
suboptimality bound and \cref{lem:kl-shared-coverage-accumulation}
for the coverage guarantee.

\begin{proof}[\pfref{thm:kl-shared-regret}]
  We first prove the convergence bound by applying
  \cref{lem:kl-shared-prompt-reward} with $t=T$ and averaging over prompts.
  Since $\Jstar-J(\piT)=\En_{x\sim\mu}\sbr*{r_x^\star-J_x(\piT)}$, this gives
  \begin{align}
    \Jstar-J(\piT)
    &\leq\frac{\En_{x\sim\mu}\sbr*{\log\rbr*{1/\pt[0](x)}}}{\eta T}
      +\frac1\eta\sum_{t=0}^{T-1}\mathsf{KL}_t
    \\
    &\quad+\frac{2\sqrt e}{\eta}
      \En_{x\sim\mu}\sbr*{\sqrt{\frac1{\pt[0](x)}
        \sum_{t=0}^{T-1}\frac{\mathsf{KL}_t(x)}{t+1}}}.
  \end{align}
  To further bound the last term,
  the Cauchy--Schwarz inequality gives
  \begin{align}
    \En_{x\sim\mu}\sbr*{\sqrt{\frac1{\pt[0](x)}
      \sum_{t=0}^{T-1}\frac{\mathsf{KL}_t(x)}{t+1}}}
    &\leq\sqrt{\En_{x\sim\mu}\sbr*{\frac1{\pt[0](x)}}
      \En_{x\sim\mu}\sbr*{\sum_{t=0}^{T-1}
        \frac{\mathsf{KL}_t(x)}{t+1}}}
    \\
    &=\sqrt{\Cresp\sum_{t=0}^{T-1}\frac{\mathsf{KL}_t}{t+1}}.
  \end{align}
  Combining the above two panels gives \cref{eq:kl-shared-regret}.

  Next, to prove the coverage guarantee, fix $c\in(0,1)$ and apply
  \cref{lem:kl-shared-coverage-accumulation} with $a=r_x^\star$.
  For every prompt $x$ satisfying
  $\sum_{s=0}^{T-1}\sqrt{\mathsf{KL}_s(x)/\pt[0](x)}\leq1-\sqrt c$,
  the lemma gives
  \begin{align}
    \sqrt{\pt(x)}
    \geq\sqrt{\pt[0](x)}-\sum_{s=0}^{t-1}\sqrt{\mathsf{KL}_s(x)}
    \geq\sqrt{c\pt[0](x)},
    \qquad\forall 0\leq t\leq T.
  \end{align}
  Thus, $\pit(\Ystarx\mid x)\geq c\piref(\Ystarx\mid x)$
  for all $0\leq t\leq T$ on every prompt satisfying the above condition.
  It remains to bound the probability that this condition does not hold, or
  $\sum_{s=0}^{T-1}\sqrt{\mathsf{KL}_s(x)/\pt[0](x)} > 1-\sqrt c$.
  Markov's inequality and the Cauchy--Schwarz inequality give
  \begin{align}
    \Pr_{x\sim\mu}\rbr*{
      \sum_{s=0}^{T-1}\sqrt{\frac{\mathsf{KL}_s(x)}{\pt[0](x)}}
      >1-\sqrt c}
    &\leq\frac1{1-\sqrt c}\sum_{s=0}^{T-1}
      \En_{x\sim\mu}\sbr*{
        \sqrt{\frac{\mathsf{KL}_s(x)}{\pt[0](x)}}}
    \\
    &\leq\frac{\sqrt{\Cresp}}{1-\sqrt c}
      \sum_{s=0}^{T-1}\sqrt{\mathsf{KL}_s}.
  \end{align}
  This proves the stated coverage guarantee.
\end{proof}

We now prove the key lemma used in the proof above,
an approximate version of the per-prompt suboptimality bound
in \cref{thm:exact-apmd-rate}.
The intuition is as follows.
Under exact updates, the inequality
$\pt(x)\geq\pt[0](x)e^{\eta t(r_x^\star-J_x(\pit))}$
yields the mirror descent rate by rearranging.
For approximate \framework updates, we use \cref{lem:kl-shared-hellinger}
to bound the change in the probability of $\Ystarx$ at each iteration.

\begin{lemma}
  \label{lem:kl-shared-prompt-reward}
  For every prompt $x$ and every $1\leq t\leq T$,
  \begin{align}
    r_x^\star-J_x(\pit)
    &\leq\frac{\log\rbr*{1/\pt[0](x)}}{\eta t}
      +\frac1\eta\sum_{s=0}^{t-1}\mathsf{KL}_s(x)
    +\frac{2\sqrt e}{\eta}
      \sqrt{\frac1{\pt[0](x)}
        \sum_{s=0}^{t-1}\frac{\mathsf{KL}_s(x)}{s+1}}.
    \label{eq:kl-shared-prompt-regret}
  \end{align}
\end{lemma}

\begin{proof}[\pfref{lem:kl-shared-prompt-reward}]
  Fix $x\in\cX$ and $1\leq t\leq T$.
  We first bound $\pt[s+1](x)$ in terms of $\pt[s](x)$ and
  $\mathsf{KL}_s(x)$.
  Since $r(x,y)=r_x^\star$ on $\Ystarx$, the exact update satisfies
  \begin{align}
    \piplust[s](\Ystarx\mid x)=\pt[s](x)e^{\eta(r_x^\star-\btstar[s](x))}.
  \end{align}
  For $0\leq s<t$, apply the reverse triangle inequality to the vectors
  $(\sqrt{\pit[s+1](y\mid x)})_{y\in\Ystarx}$ and
  $(\sqrt{\piplust[s](y\mid x)})_{y\in\Ystarx}$.
  Then using \cref{lem:kl-shared-hellinger}, we obtain
  \begin{align}
    \sqrt{\pt[s+1](x)}
    &\geq\sqrt{\piplust[s](\Ystarx\mid x)}
      -\rbr*{\sum_{y\in\Ystarx}
        \rbr*{\sqrt{\pit[s+1](y\mid x)}
          -\sqrt{\piplust[s](y\mid x)}}^2}^{1/2}
    \\
    &\geq\sqrt{\pt[s](x)}
      e^{\frac\eta2(r_x^\star-\btstar[s](x))}
      -\sqrt{\mathsf{KL}_s(x)}.
  \end{align}
  Rearranging this inequality
  and unrolling from iterations $t$ to $0$ gives
  \begin{align}
    \sqrt{\pt[0](x)}
    &\leq\sqrt{\pt(x)}
      e^{-\frac\eta2\sum_{u=0}^{t-1}(r_x^\star-\btstar[u](x))}
    +\sum_{s=1}^t\sqrt{\mathsf{KL}_{s-1}(x)}
      e^{-\frac\eta2\sum_{u=0}^{s-1}(r_x^\star-\btstar[u](x))}.
  \end{align}
  Next, we translate the exponents in the above bound to $r_x^\star-J_x(\pit)$.
  Applying \cref{lem:kl-shared-normalizer} gives, for every $0 \le s < t$,
  $\btstar[s](x)\leq J_x(\pit)+\eta^{-1}\sum_{u=0}^{t-1}\mathsf{KL}_u(x)$, so
  \begin{align}
    \sqrt{\pt[0](x)}
    &\leq\sqrt{\pt(x)}
      e^{-\frac{\eta t}{2}(r_x^\star-J_x(\pit))
        +\frac t2\sum_{u=0}^{t-1}\mathsf{KL}_u(x)}
    \\
    &\quad+\sum_{s=1}^t\sqrt{\mathsf{KL}_{s-1}(x)}
      e^{-\frac{\eta s}{2}(r_x^\star-J_x(\pit))
        +\frac s2\sum_{u=0}^{t-1}\mathsf{KL}_u(x)}.
    \label{eq:kl-shared-discounted-mass}
  \end{align}

  Finally, to prove the lemma claim from \cref{eq:kl-shared-discounted-mass}, define
  \begin{align}
    z_t(x)&\ldef{}\eta\rbr*{r_x^\star-J_x(\pit)}
      -\sum_{s=0}^{t-1}\mathsf{KL}_s(x)
      -\frac{\log\rbr*{1/\pt[0](x)}}t,
    \\
    A_t(x)&\ldef{}\sqrt{\frac1{\pt[0](x)}
      \sum_{s=1}^t\frac{\mathsf{KL}_{s-1}(x)}s}.
  \end{align}
  By the definition of $z_t(x)$, it suffices to prove
  \begin{align}
    z_t(x)\leq2A_t(x),
    \label{eq:kl-shared-excess-target}
  \end{align}
  which we return to in \cref{eq:kl-shared-excess-panel-final}.
  We will assume going forwards that $z_t(x) > 0$,
  otherwise \cref{eq:kl-shared-excess-target} is immediate.

  Dividing the bound in \cref{eq:kl-shared-discounted-mass} by
  $\sqrt{\pt[0](x)}$ and substituting the definition of $z_t(x)$ gives
  \begin{align}
    1
    &\leq\sqrt{\pt(x)}e^{-tz_t(x)/2}
      +\frac1{\sqrt{\pt[0](x)}}\sum_{s=1}^t
        \sqrt{\mathsf{KL}_{s-1}(x)}\,\pt[0](x)^{s/(2t)}e^{-sz_t(x)/2}.
  \end{align}
  Using $\pt[0](x), \pt(x) \le 1$ and rearranging,
  we obtain
  \begin{align}
    1 -  e^{-tz_t(x)/2}
    &\leq
      \frac1{\sqrt{\pt[0](x)}}\sum_{s=1}^t
        \sqrt{\mathsf{KL}_{s-1}(x)}\,e^{-sz_t(x)/2}.
    \\
    &\le
    \frac1{\sqrt{\pt[0](x)}}
      \rbr*{\sum_{s=1}^t\frac{\mathsf{KL}_{s-1}(x)}s}^{1/2}
      \rbr*{\sum_{s=1}^t s e^{-sz_t(x)}}^{1/2}
    \\
    &=A_t(x)\rbr*{\sum_{s=1}^t s e^{-sz_t(x)}}^{1/2},
    \label{eq:kl-shared-excess-mass}
  \end{align}
  where we have used the Cauchy-Schwarz inequality in the second line,
  followed by the definition of $A_t(x)$.
  Next, to bound the sum on the RHS,
  \begin{align}
    \sum_{s=1}^t s e^{-sz_t(x)}
    &=
    \sum_{s=1}^t \rbr*{\sum_{j=1}^s e^{-s z_t(x) / 2}} e^{-sz_t(x) / 2}
    \\
    &\le
    \sum_{s=1}^t \rbr*{\sum_{j=1}^s e^{-j z_t(x) / 2}} e^{-sz_t(x) / 2}
    \\
    &\le
    \sum_{s=1}^t \rbr*{\sum_{j=1}^t e^{-j z_t(x) / 2}} e^{-sz_t(x) / 2}
    =
    \rbr*{\sum_{s=1}^t e^{-sz_t(x) / 2}}^2.
  \end{align}
  Substituting the above panel back into \cref{eq:kl-shared-excess-mass},
  we obtain
  \begin{align}
    1-e^{-tz_t(x)/2}
    &\le
    A_t(x)
    \rbr*{\sum_{s=1}^t e^{-sz_t(x) / 2}}
    =
    A_t(x)\frac{1-e^{-tz_t(x)/2}}{e^{z_t(x)/2}-1}
    .
  \end{align}
  Finally, rearranging the above gives \cref{eq:kl-shared-excess-target},
  \begin{align}
    z_t(x)\leq2\log\rbr*{1+A_t(x)}\leq2A_t(x).
  \end{align}
  This proves the lemma claim since,
  using the definition of $z_t(x)$,
  \begin{align}
    \eta\rbr*{r_x^\star-J_x(\pit)}
    &=z_t(x)+\sum_{s=0}^{t-1}\mathsf{KL}_s(x)
      +\frac{\log\rbr*{1/\pt[0](x)}}t
    \\
    &\leq2A_t(x)+\sum_{s=0}^{t-1}\mathsf{KL}_s(x)
      +\frac{\log\rbr*{1/\pt[0](x)}}t
    \\
    &=\frac{\log\rbr*{1/\pt[0](x)}}t
      +\sum_{s=0}^{t-1}\mathsf{KL}_s(x)
      +2\sqrt{\frac1{\pt[0](x)}
        \sum_{s=0}^{t-1}\frac{\mathsf{KL}_s(x)}{s+1}},
      \label{eq:kl-shared-excess-panel-final}
  \end{align}
  where we have used $z_t(x) \le 2A_t(x)$ in the second line
  and the definition of $A_t(x)$ in the last equality.
  Dividing both sides by $\eta$ yields \cref{eq:kl-shared-prompt-regret}.
\end{proof}

\clearpage
\section{The Mean-Squared Error Criterion (Proof of \cref*{thm:mse-regret-main})}\label{app:mse-accumulation}
This section proves \cref{thm:mse-regret-main},
which establishes the mean-squared error criterion
sufficient for convergence and coverage preservation in \framework.
We first develop supporting lemmas in
\cref{app:mse-accum-supporting}, then prove \cref{thm:mse-regret-main} in
\cref{app:mse-accum-global}.

\para{Notation}
Throughout this section, fix $\eta>0$ and $T\geq1$,
and a given sequence of policies
$\cbr*{\pit}_{t=0}^T$ with $\pit[0]=\piref$ and $\pit > 0$.
We retain the notation of \cref{app:kl-shared-statements},
including for a fixed prompt $x$
the notation $J_x(\pi)$, $r_x^\star$, $\Ystarx$, $\pt(x)$, and $\cY_a(x)$.
For $0\leq t<T$, define the centered residual of $\pitp$ to be
\begin{align}
  s_t(x,y)
  &\ldef{}\frac1\eta\log\frac{\pitp(y\mid x)}{\pit(y\mid x)}
  -\En_{y'\sim\pit(\cdot\mid x)}\sbr*{
    \frac1\eta\log\frac{\pitp(y'\mid x)}{\pit(y'\mid x)}}
  -\rbr*{r(x,y)-J_x(\pit)},
  \label{eq:mse-accum-residual}
\end{align}
and denote $S \ldef{} \max_{0\leq t<T}\nbr*{s_t}_\infty$ as in \cref{thm:mse-regret-main}.

Under \cref{asm:softmax-linear}, the definition of $s_t$ gives
\begin{align}
  s_t(x,y)
  =\rbr*{\phi(x,y)-\En_{y'\sim\pit(\cdot\mid{}x)}\sbr*{\phi(x,y')}}^\top
    \rbr*{\frac{\thetatp-\thetat}{\eta}-\thetastar}.
  \label{eq:mse-accum-linear-residual}
\end{align}

We denote the mean-squared error conditional on $x$ by $\mset(x)$,
\begin{align}
  \mathsf{MSE}_t(x)
  &\ldef{}\En_{y\sim\pit(\cdot\mid x)}\sbr*{s_t(x,y)^2},
  &\text{where}\quad\mset
  &\ldef{}\En_{x\sim\mu}\sbr*{\mathsf{MSE}_t(x)}.
\end{align}

Throughout this section, we will call upon the following rearrangement of $\pitp$ for convenience,
and use the shifted normalization constant $\wt \bt$. Define this to be
\begin{align}
  \widetilde b_t(x)
  &\ldef{}\frac1\eta\log\En_{y\sim\pit(\cdot\mid x)}\sbr*{
    e^{\eta(r(x,y)+s_t(x,y))}},
\end{align}
then rearranging the definition of $s_t$ gives
\begin{align}
  \pitp(y\mid x)
  &=\pit(y\mid x)e^{\eta(r(x,y)+s_t(x,y)-\widetilde b_t(x))}.
  \label{eq:mse-accum-update}
\end{align}

\subsection{Supporting Lemmas}
\label{app:mse-accum-supporting}

We first state and prove the supporting lemmas used in \cref{app:mse-accum-global},
adapting the arguments of \cref{app:kl-shared-supporting}
to the mean-squared error.
They are organized as follows.
\begin{itemize}
  \item (Approximate improvement)
  \Cref{lem:mse-accum-improvement} bounds how much performance can drop from $\pit$ to any $\pit[s]$ by the cumulative mean-squared errors.

  \item (Coverage approximately preserved)
  \Cref{lem:mse-accum-coverage} bounds how much the probabilities of
  upper reward quantiles degrade from $\pit[0]$ to $\pit$,
  in terms of the accumulated mean-squared error.
\end{itemize}

\begin{lemma}[Exponential moment of the centered residual]
  \label{lem:mse-accum-residual-moment}
  Fix $x\in\cX$. For every $0\leq t<T$,
  \begin{align}
    \log\En_{y\sim\pit(\cdot\mid x)}\sbr*{e^{\eta s_t(x,y)}}
    \leq\frac{\eta^2e^{\eta S}}2\mathsf{MSE}_t(x).
    \label{eq:mse-accum-residual-mgf}
  \end{align}
\end{lemma}

\begin{proof}[\pfref{lem:mse-accum-residual-moment}]
  Fix $0\leq t<T$.
  Since $\abr*{s_t(x,y)}\leq S$, Taylor's theorem gives
  \begin{align}
    e^{\eta s_t(x,y)}\leq1+\eta s_t(x,y)
    +\frac{\eta^2e^{\eta S}}2s_t(x,y)^2.
  \end{align}
  We therefore obtain
  \begin{align}
    \log\En_{y\sim\pit(\cdot\mid x)}\sbr*{e^{\eta s_t(x,y)}}
    &\leq\log\rbr*{1+\frac{\eta^2e^{\eta S}}2
      \En_{y\sim\pit(\cdot\mid x)}\sbr*{s_t(x,y)^2}}
    \\
    &\leq\frac{\eta^2e^{\eta S}}2\mathsf{MSE}_t(x),
  \end{align}
  where the first inequality uses the fact that $\En_{y' \sim \pit(\cdot \mid{} x)}\sbr*{s_t(x,y')} = 0$,
  and the second uses $\log(1+u)\leq u$.
\end{proof}

\begin{lemma}[Approximate improvement]
  \label{lem:mse-accum-improvement}
  Fix $x\in\cX$. For every $0\leq t<T$,
  \begin{align}
    J_x(\pit)\leq\widetilde b_t(x)
    \leq J_x(\pitp)+\frac{\eta e^{\eta S}}2\mathsf{MSE}_t(x).
    \label{eq:mse-accum-improvement-1}
  \end{align}
  More generally, for every $0\leq s<t\leq T$,
  \begin{align}
    J_x(\pit[s])\leq\widetilde b_s(x)
    \leq J_x(\pit)+\frac{\eta e^{\eta S}}2
      \sum_{u=s}^{t-1}\mathsf{MSE}_u(x).
    \label{eq:mse-accum-improvement}
  \end{align}
\end{lemma}

\begin{proof}[\pfref{lem:mse-accum-improvement}]
  Fix any $0\leq t<T$.
  We first prove the lower bounds in both \cref{eq:mse-accum-improvement-1} and \cref{eq:mse-accum-improvement}.
  Jensen's inequality and the definition of $s_t(x,y)$ give
  \begin{align}
    \eta\widetilde b_t(x)
    \geq\eta\En_{y\sim\pit(\cdot\mid x)}\sbr*{r(x,y)+s_t(x,y)}
    =\eta J_x(\pit).
      \label{eq:mse-approx-improve-1}
  \end{align}

  We next prove the upper bound in \cref{eq:mse-accum-improvement-1}.
  Recall that
  \begin{align}
    \pitp(y\mid{}x)
    = \pit(y\mid{}x)e^{\eta(r(x,y)+s_t(x,y)-\widetilde b_t(x))}.
  \end{align}
  Taking logarithms and rearranging, we obtain
  \begin{align}
    \eta\widetilde b_t(x)
    =\eta r(x,y)+\eta s_t(x,y)
      -\log\frac{\pitp(y\mid x)}{\pit(y\mid x)}.
  \end{align}
  Taking expectation under $\pitp(\cdot\mid x)$ gives
  \begin{align}
    \eta\widetilde b_t(x)
    &=\eta J_x(\pitp)
      +\eta\En_{y\sim\pitp(\cdot\mid x)}\sbr*{s_t(x,y)}
      -\Dkl{\pitp(\cdot\mid x)}{\pit(\cdot\mid x)}
    \\
    &=\eta J_x(\pitp)
      +\En_{y\sim\pitp(\cdot\mid x)}\sbr*{
        \log\rbr*{e^{\eta s_t(x,y)}
          \frac{\pit(y\mid x)}{\pitp(y\mid x)}}}
    \\
    &\leq\eta J_x(\pitp)
      +\log\En_{y\sim\pit(\cdot\mid x)}\sbr*{e^{\eta s_t(x,y)}}
    \\
    &\leq\eta J_x(\pitp)
      +\frac{\eta^2e^{\eta S}}2\mathsf{MSE}_t(x).
      \label{eq:mse-approx-improve-2}
  \end{align}
  The first inequality uses Jensen's inequality, and the second uses
  \cref{lem:mse-accum-residual-moment}.
  Thus, combining \cref{eq:mse-approx-improve-1} and \cref{eq:mse-approx-improve-2}, we have shown that
  \begin{align}
    J_x(\pit) \le \wt{b}_t(x)
    \le
    J_x(\pitp) + \frac{\eta e^{\eta S}}{2} \mset(x),
      \label{eq:mse-approx-improve-3}
  \end{align}
  which proves the claim in \cref{eq:mse-accum-improvement-1}.
  Finally, we prove the upper bound for general $0\leq s<t\leq T$ in \cref{eq:mse-accum-improvement}.
  We can write
  \begin{align}
    \widetilde b_s(x)-J_x(\pit)
    &=\widetilde b_s(x)-J_x(\pit[s+1])
      +\sum_{u=s+1}^{t-1}\rbr*{J_x(\pit[u])-J_x(\pit[u+1])}
    \\
    &\leq\frac{\eta e^{\eta S}}2\mathsf{MSE}_s(x)
      +\frac{\eta e^{\eta S}}2\sum_{u=s+1}^{t-1}\mathsf{MSE}_u(x)
    \\
    &=\frac{\eta e^{\eta S}}2\sum_{u=s}^{t-1}\mathsf{MSE}_u(x),
  \end{align}
  where we have used \cref{eq:mse-approx-improve-3} for each $s \le u \le t-1$ in the second line.
  Rearranging proves the upper bound in \cref{eq:mse-accum-improvement}.
\end{proof}

\begin{lemma}[Coverage preserved approximately]
  \label{lem:mse-accum-coverage}
  Fix $x\in\cX$ and let
  $\cY_a(x)=\cbr*{y\in\cY:r(x,y)\geq a}$ for $a\in[0,\Rmax]$.
  Then, for every $0\leq t\leq T$,
  \begin{align}
    \sqrt{\pit(\cY_a(x)\mid x)}
    &\geq e^{-\frac{\eta^2e^{\eta S}}4\sum_{s=0}^{t-1}\mathsf{MSE}_s(x)}
      \sqrt{\piref(\cY_a(x)\mid x)}
      -\frac\eta2\sum_{s=0}^{t-1}\sqrt{\mathsf{MSE}_s(x)}.
    \label{eq:mse-accum-coverage-root}
  \end{align}
\end{lemma}

\begin{proof}[\pfref{lem:mse-accum-coverage}]
  For $0\leq t<T$, define the conditional distribution
  \begin{align}
    q_t(y\mid x)
    \ldef{}\frac{\pit(y\mid x)e^{\eta s_t(x,y)}}{
      \En_{y'\sim\pit(\cdot\mid x)}[e^{\eta s_t(x,y')}] }.
  \end{align}
  Note that $\pitp(y\mid{}x) = \frac{q_t(y\mid{}x) e^{\eta r(x,y)}}{\En_{y' \sim q_t(\cdot \mid{} x)}\sbr*{e^{\eta r(x,y')}}}$ from \cref{eq:mse-accum-update}.
  Therefore, the same calculation as in \cref{eq:kl-shared-upper-set-tilt} gives
  \begin{align}
    \pitp(\cY_a(x) \mid{} x) \ge q_t(\cY_a(x) \mid{} x).
    \label{eq:pitp-qt-ordering}
  \end{align}

  Next, since $1 \le e^u -u \le e^u + \abr*{u}$,
  \begin{align}
    \sqrt{\pit(y\mid{}x)}
    &\le
    \sqrt{\pit(y\mid{}x)}e^{\eta s_t(x,y) / 2} + \frac{\eta}{2} \sqrt{\pit(y\mid{}x)} \abr*{s_t(x,y)}
    \\
    &=
  \sqrt{q_t(y\mid x)} \rbr*{\En_{y'\sim\pit(\cdot\mid x)} \sbr*{e^{\eta s_t(x,y')}}}^{1/2}
  + \frac{\eta}{2} \sqrt{\pit(y\mid{}x)} \abr*{s_t(x,y)},
  \end{align}
  where the second line uses the definition of $q_t$.

Then applying the above inequality to each $y \in \cY_a(x)$,
\begin{align}
  \pit(\cY_a(x)\mid x)
  &=\sum_{y\in\cY_a(x)}\pit(y\mid x)
  \\
  &\leq q_t(\cY_a(x)\mid x)
    \En_{y'\sim\pit(\cdot\mid x)}
      \sbr*{e^{\eta s_t(x,y')}}
  + \frac{\eta^2}4 \sum_{y\in\cY_a(x)}\pit(y\mid x)s_t(x,y)^2
  \\
  &\quad+\eta
    \rbr*{\En_{y'\sim\pit(\cdot\mid x)}
      \sbr*{e^{\eta s_t(x,y')}}}^{1/2}
    \sum_{y\in\cY_a(x)}
      \sqrt{q_t(y\mid x)\pit(y\mid x)}\abr*{s_t(x,y)}
  \\
  &\leq q_t(\cY_a(x)\mid x)
    \En_{y'\sim\pit(\cdot\mid x)}
      \sbr*{e^{\eta s_t(x,y')}}
  +\frac{\eta^2}4 \sum_{y\in\cY_a(x)}\pit(y\mid x)s_t(x,y)^2
  \\
  &\quad+\eta\sqrt{q_t(\cY_a(x)\mid x)}
    \rbr*{\En_{y'\sim\pit(\cdot\mid x)}
      \sbr*{e^{\eta s_t(x,y')}}}^{1/2}
    \rbr*{\sum_{y\in\cY_a(x)}
      \pit(y\mid x)s_t(x,y)^2}^{1/2}
  \\
  &\leq q_t(\cY_a(x)\mid x)
    \En_{y'\sim\pit(\cdot\mid x)}
      \sbr*{e^{\eta s_t(x,y')}}
    +\frac{\eta^2}4\mathsf{MSE}_t(x)
  \\
  &\quad+\eta\sqrt{q_t(\cY_a(x)\mid x)}
    \rbr*{\En_{y'\sim\pit(\cdot\mid x)}
      \sbr*{e^{\eta s_t(x,y')}}}^{1/2}
    \sqrt{\mathsf{MSE}_t(x)}
  \\
  &=\rbr*{
    \sqrt{q_t(\cY_a(x)\mid x)}
    \rbr*{\En_{y'\sim\pit(\cdot\mid x)}
      \sbr*{e^{\eta s_t(x,y')}}}^{1/2}
    +\frac\eta2\sqrt{\mathsf{MSE}_t(x)}
  }^2.
\end{align}

Therefore,
\begin{align}
  \sqrt{\pit(\cY_a(x)\mid x)}
  &\leq\sqrt{q_t(\cY_a(x)\mid x)}
    \rbr*{\En_{y\sim\pit(\cdot\mid x)}
      \sbr*{e^{\eta s_t(x,y)}}}^{1/2}
    +\frac\eta2\sqrt{\mathsf{MSE}_t(x)}.
\end{align}

For the RHS,
\cref{eq:pitp-qt-ordering} gives $q_t(\cY_a(x)\mid x)\leq\pitp(\cY_a(x)\mid x)$,
and \cref{lem:mse-accum-residual-moment} gives
\begin{align}
  \log\En_{y\sim\pit(\cdot\mid x)}\sbr*{e^{\eta s_t(x,y)}}
  \leq\frac{\eta^2e^{\eta S}}2\mathsf{MSE}_t(x).
\end{align}

Therefore,
  \begin{align}
    \sqrt{\pit(\cY_a(x)\mid x)}
    \leq e^{\frac{\eta^2e^{\eta S}}4\mathsf{MSE}_t(x)}
      \sqrt{\pitp(\cY_a(x)\mid x)}
      +\frac\eta2\sqrt{\mathsf{MSE}_t(x)}.
    \label{eq:mse-accum-coverage-step}
  \end{align}
  For $1\leq t\leq T$, applying \cref{eq:mse-accum-coverage-step} at iteration $t-1$
  and unrolling for $0\leq s<t$ gives
  \begin{align}
    \sqrt{\pit(\cY_a(x) \mid{} x)}
    &\ge
  e^{\frac{-\eta^2e^{\eta S}}4\mathsf{MSE}_{t-1}(x)} \sqrt{\pit[t-1](\cY_a(x) \mid{} x)}
  - \frac{\eta}{2} e^{\frac{-\eta^2e^{\eta S}}4\mathsf{MSE}_{t-1}(x)} \sqrt{\mset[t-1](x)}
  \\
    &\geq e^{-\frac{\eta^2e^{\eta S}}4\sum_{s=0}^{t-1}\mathsf{MSE}_s(x)}
      \sqrt{\piref(\cY_a(x)\mid x)}
    \\
    &\quad-\frac\eta2\sum_{s=0}^{t-1}
      e^{-\frac{\eta^2e^{\eta S}}4\sum_{u=s}^{t-1}\mathsf{MSE}_u(x)}
      \sqrt{\mathsf{MSE}_s(x)}
    \\
    &\geq e^{-\frac{\eta^2e^{\eta S}}4\sum_{s=0}^{t-1}\mathsf{MSE}_s(x)}
      \sqrt{\piref(\cY_a(x)\mid x)}
      -\frac\eta2\sum_{s=0}^{t-1}\sqrt{\mathsf{MSE}_s(x)}.
    \label{eq:mse-accum-unroll}
  \end{align}
  The case $t=0$ holds with equality.
\end{proof}

\subsection{Proof of \cref{thm:mse-regret-main}}
\label{app:mse-accum-global}

We first state \cref{thm:mse-accum-regret}, which extends the exact policy mirror descent
analysis in \cref{thm:exact-apmd-rate} to approximate updates using mean-squared error.
The proof of \cref{thm:mse-regret-main} directly follows;
following that, we prove \cref{thm:mse-accum-regret}.

\begin{theorem}[Global convergence under \framework{}]
  \label{thm:mse-accum-regret}
  The final policy $\piT$ satisfies
  \begin{align}
    \Jstar-J(\piT)
    &\leq\frac{\En_{x\sim\mu}\sbr*{\log\rbr*{1/\pt[0](x)}}}{\eta T}
      +\frac{\eta e^{\eta S}}2\sum_{t=0}^{T-1}\mathsf{MSE}_t
    +\sqrt{\Cresp\sum_{t=0}^{T-1}\frac{\mathsf{MSE}_t}{t+1}}.
    \label{eq:mse-accum-regret}
  \end{align}
  Moreover, for any $c\in(0,1)$, with probability at least
  \begin{align}
    1-\frac{\eta^2e^{\eta S}}{4(1-\sqrt c)}
      \sum_{t=0}^{T-1}\mathsf{MSE}_t
      -\frac{\eta\sqrt{\Cresp}}{2(1-\sqrt c)}
        \sum_{t=0}^{T-1}\sqrt{\mathsf{MSE}_t}
  \end{align}
  over $x\sim\mu$,
  \begin{align}
    \pit(\Ystarx\mid x)\geq c\piref(\Ystarx\mid x),
    \qquad\forall 0\leq t\leq T.
    \label{eq:mse-accum-regret-coverage}
  \end{align}
\end{theorem}

\begin{proof}[\pfref{thm:mse-regret-main}]
  We first prove the coverage claim by applying \cref{thm:mse-accum-regret}
  with $c=1/2$ and the prescribed MSE tolerance in
  \cref{eq:framework-mse-tolerance},
  \begin{align}
    \mathsf{MSE}_t
    \leq\frac1{4\eta^2T^2\rbr*{\Cresp\log(eT)+e^{\eta S}}},
    \qquad 0\leq t<T.
    \label{eq:mse-accum-main-tolerance}
  \end{align}
  Substituting into the theorem gives
  \begin{align}
    &\Pr_{x\sim\mu}\rbr*{
      \exists\,0\leq t\leq T:\;
      \pit(\Ystarx\mid x)<\tfrac12\piref(\Ystarx\mid x)}
    \\
    &\quad\leq(2+\sqrt2)\rbr*{
      \frac{e^{\eta S}}{16T\rbr*{\Cresp\log(eT)+e^{\eta S}}}
      +\frac14\sqrt{\frac{\Cresp}{\Cresp\log(eT)+e^{\eta S}}}}
    \\
    &\quad\leq\frac{2+\sqrt2}{8\sqrt{\log(eT)}}\rbr*{
      \frac{e^{\eta S}}{\Cresp\log(eT)+e^{\eta S}}
      +1+\frac{\Cresp\log(eT)}{\Cresp\log(eT)+e^{\eta S}}}
    \\
    &\quad=\frac{2+\sqrt2}{4\sqrt{\log(eT)}}\leq\frac{7}{8\sqrt{\log(eT)}}.
  \end{align}
  The second inequality uses $\sqrt u\leq(1+u)/2$ for $u\geq0$.
  This proves the coverage claim.

  To prove convergence, we again substitute \cref{eq:mse-accum-main-tolerance} into \cref{thm:mse-accum-regret}.
  Since $\En_{x\sim\mu}\sbr*{\log(1/\pt[0](x))}\leq\log\Cresp$
  by Jensen's inequality, we obtain
  \begin{align}
    \Jstar-J(\piT)
    &\leq\frac{\log\Cresp}{\eta T}
      +\frac{e^{\eta S}}
        {8\eta T\rbr*{\Cresp\log(eT)+e^{\eta S}}}
      +\frac1{2\eta T}
      \sqrt{\frac{\Cresp\sum_{t=1}^T1/t}{\Cresp\log(eT)+e^{\eta S}}}
    \\
    &\leq\frac{\log\Cresp}{\eta T}
      +\frac{e^{\eta S}}
        {8\eta T\rbr*{\Cresp\log(eT)+e^{\eta S}}}
      +\frac1{2\eta T}
      \sqrt{\frac{\Cresp\log(eT)}{\Cresp\log(eT)+e^{\eta S}}}
    \\
    &\leq\frac{\log\Cresp+1/8+1/2}{\eta T}
      \leq\frac{\log(2\Cresp)}{\eta T}.
  \end{align}
  The second inequality uses $\sum_{t=1}^T1/t\leq\log(eT)$.
  This proves the convergence claim.
\end{proof}

Next, we prove \cref{thm:mse-accum-regret},
which extends the exact mirror descent convergence guarantee
in \cref{thm:exact-apmd-rate} to handle approximation error measured in $\mset$.
We use \cref{lem:mse-accum-prompt-reward}, proved below, for the
suboptimality bound and \cref{lem:mse-accum-coverage} for the coverage guarantee.

\begin{proof}[\pfref{thm:mse-accum-regret}]
  We first prove the convergence bound by applying
  \cref{lem:mse-accum-prompt-reward} with $t=T$ and averaging over prompts.
  Since $\Jstar-J(\piT)=\En_{x\sim\mu}\sbr*{r_x^\star-J_x(\piT)}$, this gives
  \begin{align}
    \Jstar-J(\piT)
    &\leq\frac{\En_{x\sim\mu}\sbr*{\log\rbr*{1/\pt[0](x)}}}{\eta T}
      +\frac{\eta e^{\eta S}}2\sum_{t=0}^{T-1}\mathsf{MSE}_t
    +\En_{x\sim\mu}\sbr*{
      \sqrt{\frac1{\pt[0](x)}
        \sum_{t=0}^{T-1}\frac{\mathsf{MSE}_t(x)}{t+1}}}
    \\
    &\leq\frac{\En_{x\sim\mu}\sbr*{\log\rbr*{1/\pt[0](x)}}}{\eta T}
      +\frac{\eta e^{\eta S}}2\sum_{t=0}^{T-1}\mathsf{MSE}_t
    + \sqrt{\Cresp\sum_{t=0}^{T-1}\frac{\mathsf{MSE}_t}{t+1}},
  \end{align}
  using Cauchy--Schwarz in the second line,
  which gives \cref{eq:mse-accum-regret}.

  Next, to prove the coverage guarantee, fix $c\in(0,1)$ and apply
  \cref{lem:mse-accum-coverage} with $a=r_x^\star$.
  For every $0\leq t\leq T$, we obtain
  \begin{align}
    \sqrt{\frac{\pt(x)}{\pt[0](x)}}
    &\geq e^{-\frac{\eta^2e^{\eta S}}4
        \sum_{s=0}^{t-1}\mathsf{MSE}_s(x)}
      -\frac\eta2\sum_{s=0}^{t-1}
        \sqrt{\frac{\mathsf{MSE}_s(x)}{\pt[0](x)}}
    \\
    &\geq1-\frac{\eta^2e^{\eta S}}4
        \sum_{s=0}^{T-1}\mathsf{MSE}_s(x)
      -\frac\eta2\sum_{s=0}^{T-1}
        \sqrt{\frac{\mathsf{MSE}_s(x)}{\pt[0](x)}}.
  \end{align}
  The second inequality uses $e^{-u}\geq1-u$ and $t\leq T$.
  Thus, every prompt satisfying
  \begin{align}
    \frac{\eta^2e^{\eta S}}4\sum_{s=0}^{T-1}\mathsf{MSE}_s(x)
      +\frac\eta2\sum_{s=0}^{T-1}
        \sqrt{\frac{\mathsf{MSE}_s(x)}{\pt[0](x)}}
    \leq1-\sqrt c
    \label{eq:mse-accum-coverage-small-error}
  \end{align}
  has
  \begin{align}
    \sqrt{\pt(x)/\pt[0](x)}\geq\sqrt c,
  \end{align}
  and hence
  $\pit(\Ystarx\mid x)\geq c\piref(\Ystarx\mid x)$,
  for all $0\leq t\leq T$.
  It remains to bound the probability that
  \cref{eq:mse-accum-coverage-small-error} does not hold.
  We have
  \begin{align}
    &\Pr_{x\sim\mu}\rbr*{
      \frac{\eta^2e^{\eta S}}4\sum_{s=0}^{T-1}\mathsf{MSE}_s(x)
      +\frac\eta2\sum_{s=0}^{T-1}
        \sqrt{\frac{\mathsf{MSE}_s(x)}{\pt[0](x)}}>1-\sqrt c}
    \\
    &\quad\leq\frac1{1-\sqrt c}\rbr*{
      \frac{\eta^2e^{\eta S}}4\sum_{s=0}^{T-1}\mathsf{MSE}_s
      +\frac\eta2\sum_{s=0}^{T-1}\En_{x\sim\mu}\sbr*{
        \sqrt{\frac{\mathsf{MSE}_s(x)}{\pt[0](x)}}}}
    \\
    &\quad\leq\frac1{1-\sqrt c}\rbr*{
      \frac{\eta^2e^{\eta S}}4\sum_{s=0}^{T-1}\mathsf{MSE}_s
      +\frac{\eta\sqrt{\Cresp}}2
        \sum_{s=0}^{T-1}\sqrt{\mathsf{MSE}_s}}.
  \end{align}
  The first inequality uses Markov's inequality, and the second uses
  Cauchy--Schwarz and the definition of $\Cresp$.
  This proves the coverage guarantee in \cref{eq:mse-accum-regret-coverage}.
\end{proof}

The remaining per-prompt bound uses the residual form of the update on $\Ystarx$.

\begin{lemma}
  \label{lem:mse-accum-prompt-reward}
  For every prompt $x$ and every $1\leq t\leq T$,
  \begin{align}
    r_x^\star-J_x(\pit)
    &\leq\frac{\log\rbr*{1/\pt[0](x)}}{\eta t}
      +\frac{\eta e^{\eta S}}2\sum_{s=0}^{t-1}\mathsf{MSE}_s(x)
      +\sqrt{\frac1{\pt[0](x)}
        \sum_{s=0}^{t-1}\frac{\mathsf{MSE}_s(x)}{s+1}}.
    \label{eq:mse-accum-prompt-regret}
  \end{align}
\end{lemma}

\begin{proof}[\pfref{lem:mse-accum-prompt-reward}]
  Fix $x\in\cX$ and $1\leq t\leq T$.
  We bound the excess over the first two terms in the claimed inequality.
  Define
  \begin{align}
    z_t(x)\ldef{}\eta\rbr*{r_x^\star-J_x(\pit)}
      -\frac{\eta^2e^{\eta S}}2\sum_{s=0}^{t-1}\mathsf{MSE}_s(x)
      -\frac{\log\rbr*{1/\pt[0](x)}}t.
  \end{align}
  The desired bound is equivalent to
  \begin{align}
    z_t(x)\leq\eta\sqrt{\frac1{\pt[0](x)}
      \sum_{s=0}^{t-1}\frac{\mathsf{MSE}_s(x)}{s+1}},
    \label{eq:mse-accum-excess-target}
  \end{align}
  and it suffices to prove that \cref{eq:mse-accum-excess-target} holds.
  We may assume $z_t(x)>0$, since otherwise it holds immediately.

  First, observe that by \cref{lem:mse-accum-improvement}, for every $0\leq s<t$,
  \begin{align}
    \eta\rbr*{r_x^\star-\widetilde b_s(x)}
    &\geq\eta\rbr*{r_x^\star-J_x(\pit)}
      -\frac{\eta^2e^{\eta S}}2\sum_{u=s}^{t-1}\mathsf{MSE}_u(x)
    \geq z_t(x)+\frac{\log\rbr*{1/\pt[0](x)}}t.
    \label{eq:mse-accum-normalizer-gap}
  \end{align}
  Using \cref{eq:mse-accum-normalizer-gap}, we lower bound
  $\sqrt{\pt[s+1](x)}$ in terms of $z_t(x)$.
  Since $r(x,y)=r_x^\star$ on $\Ystarx$,
  the update for $\pitp[s]$ in \cref{eq:mse-accum-update} gives
  \begin{align}
    \sqrt{\pt[s+1](x)}
    &=\sqrt{\pt[s](x)}e^{\frac\eta2(r_x^\star-\widetilde b_s(x))}
      \rbr*{\sum_{y\in\Ystarx}
        \frac{\pit[s](y\mid x)}{\pt[s](x)}e^{\eta s_s(x,y)}}^{1/2}
    \\
    &\geq\pt[0](x)^{-1/(2t)}\sqrt{\pt[s](x)}e^{z_t(x)/2}
      \rbr*{\sum_{y\in\Ystarx}
        \frac{\pit[s](y\mid x)}{\pt[s](x)}e^{\eta s_s(x,y)}}^{1/2},
  \end{align}
  where in the inequality we use \cref{eq:mse-accum-normalizer-gap}.
  We next bound the contribution of $s_s(x,y)$ by $\mathsf{MSE}_s(x)$.
  Using Jensen's inequality on the sum in the parentheses,
  \begin{align}
    \sqrt{\pt[s+1](x)}
    &\geq\pt[0](x)^{-1/(2t)}\sqrt{\pt[s](x)}
      \exp\rbr*{\frac{z_t(x)}2+
        \frac\eta{2\pt[s](x)}\sum_{y\in\Ystarx}\pit[s](y\mid x)s_s(x,y)}
    \\
    &\geq\pt[0](x)^{-1/(2t)}\rbr*{
      \rbr*{1+\frac{z_t(x)}2}\sqrt{\pt[s](x)}
      +\frac\eta{2\sqrt{\pt[s](x)}}
        \sum_{y\in\Ystarx}\pit[s](y\mid x)s_s(x,y)}
    \\
    &\geq\pt[0](x)^{-1/(2t)}\rbr*{
      \rbr*{1+\frac{z_t(x)}2}\sqrt{\pt[s](x)}
      -\frac\eta2\sqrt{\mathsf{MSE}_s(x)}},
    \label{eq:mse-accum-optimal-mass-step}
  \end{align}
  where the second inequality above uses
  $e^u\geq1+u$, and the third uses Cauchy--Schwarz,
  \begin{align}
    \sum_{y\in\Ystarx}\pit[s](y\mid x)s_s(x,y)
    &\geq-\sqrt{\sum_{y\in\Ystarx}\pit[s](y\mid x)}
      \sqrt{\sum_{y\in\Ystarx}\pit[s](y\mid x)s_s(x,y)^2}
    \\
    &\geq-\sqrt{\pt[s](x)\mathsf{MSE}_s(x)}.
  \end{align}

  We now use \cref{eq:mse-accum-optimal-mass-step} to prove
  \cref{eq:mse-accum-excess-target}.
  Rearranging the bound on $\sqrt{\pt[s+1](x)}$ gives
  \begin{align}
    \sqrt{\pt[s](x)}
    \leq\frac{\pt[0](x)^{1/(2t)}}{1+z_t(x)/2}\sqrt{\pt[s+1](x)}
      +\frac{\eta\sqrt{\mathsf{MSE}_s(x)}}{2\rbr*{1+z_t(x)/2}}.
  \end{align}
  Unrolling this inequality through iterations $t-1,\ldots,0$ and dividing by
  $\sqrt{\pt[0](x)}$, we obtain
  \begin{align}
    1
    \leq\frac{\sqrt{\pt(x)}}{\rbr*{1+z_t(x)/2}^t}
      +\frac\eta{2\sqrt{\pt[0](x)}}\sum_{s=1}^t
        \frac{\pt[0](x)^{(s-1)/(2t)}\sqrt{\mathsf{MSE}_{s-1}(x)}}
          {\rbr*{1+z_t(x)/2}^s}.
  \end{align}
  Using $\pt[0](x),\pt(x)\leq1$ and rearranging,
  \begin{align}
    1-\rbr*{1+z_t(x)/2}^{-t}
    &\leq\frac\eta{2\sqrt{\pt[0](x)}}\sum_{s=1}^t
      \frac{\sqrt{\mathsf{MSE}_{s-1}(x)}}{\rbr*{1+z_t(x)/2}^s}
    \\
    &\leq\frac\eta2\sqrt{\frac1{\pt[0](x)}
      \sum_{s=1}^t\frac{\mathsf{MSE}_{s-1}(x)}s}
      \rbr*{\sum_{s=1}^t\frac{s}{\rbr*{1+z_t(x)/2}^{2s}}}^{1/2},
    \label{eq:mse-accum-weighted-mass}
  \end{align}
  where the second inequality uses Cauchy--Schwarz.
  To bound the sum in the rightmost term of \cref{eq:mse-accum-weighted-mass},
  \begin{align}
    \sum_{s=1}^t\frac{s}{\rbr*{1+z_t(x)/2}^{2s}}
    &\leq\sum_{s=1}^t\sum_{j=1}^s
      \frac1{\rbr*{1+z_t(x)/2}^{s+j}}
    \\
    &\leq\rbr*{\sum_{s=1}^t\frac1{\rbr*{1+z_t(x)/2}^s}}^2
      =\frac4{z_t(x)^2}\rbr*{1-\rbr*{1+z_t(x)/2}^{-t}}^2.
  \end{align}
  Substituting this bound into \cref{eq:mse-accum-weighted-mass} gives
  \begin{align}
    1-\rbr*{1+z_t(x)/2}^{-t}
    &\leq\frac\eta{z_t(x)}\sqrt{\frac1{\pt[0](x)}
      \sum_{s=1}^t\frac{\mathsf{MSE}_{s-1}(x)}s}
      \rbr*{1-\rbr*{1+z_t(x)/2}^{-t}}.
  \end{align}
  Since $z_t(x)>0$, $1-\rbr*{1+z_t(x)/2}^{-t} > 0$.
  Multiplying both sides by $z_t(x) / (1 - (1 + z_t(x) / 2)^{-t})$ gives
  \begin{align}
    z_t(x)
    &\leq\eta\sqrt{\frac1{\pt[0](x)}
      \sum_{s=1}^t\frac{\mathsf{MSE}_{s-1}(x)}s}
    =\eta\sqrt{\frac1{\pt[0](x)}
      \sum_{s=0}^{t-1}\frac{\mathsf{MSE}_s(x)}{s+1}}.
  \end{align}
  This proves \cref{eq:mse-accum-excess-target} and the claimed lemma statement.
\end{proof}

\section{Additional Result: KL Divergence Criterion for General Comparators}\label{app:general-apmd}
In this section, we extend \cref{thm:kl-regret-main}
to general comparator policies and performance targets.

\textbf{(\cref{app:general-kl})}
In \cref{thm:general-target-comparator}, we establish sufficient conditions in terms of $T$ and the inner KL tolerance $\Dkl{\pitp}{\piplust} \le \ein$ for $J(\pi) - J(\piT) \le \veps$. 
For a given $\pi$ the tolerance $\ein$ scales with the coverage of the base model over $\pi$,
and the guarantee holds for every comparator $\pi$.
The conditions in \cref{eq:general-target-reward-conditions} also guarantee $J - J(\piT) \le \veps$, where $J \in [0, \Jstar]$ is a target performance.

We omit analogous results for the mean-squared error criterion
(\cref{thm:mse-regret-main}), as well as end-to-end analyses
for algorithms such as \ppolitenc, which follow straightforwardly
from the results here.

\textbf{(\cref{app:general-kl-comparator-policy-proofs})}
Here we prove the results in \cref{app:general-kl}.
The proofs in this section build on
the supporting lemmas in \cref{app:kl-shared-supporting}.

\para{Notation}
Fix an $\eta>0$ and $T\geq1$,
as well as a sequence of policies $\cbr*{\pit}_{t=0}^T$
with $\pit[0]=\piref$ with $\pit > 0$ for all $t$.
We retain the setup and notation of \cref{app:kl},
including the per-prompt rewards $J_x(\pi)$
and approximation errors $\mathsf{KL}_t(x)$ and $\ol{\mathsf{KL}}(x)$.
The inner KL tolerance $\ein\geq0$ is a free parameter in this section.

\subsection{Extension of \cref{thm:kl-regret-main} to General Comparator Policies}
\label{app:general-kl}

In this section, we write $\pifin = \piT$
and consider performance targets of the form $J(\pi) - J(\piT)$,
where $\pi$ is a given comparator policy.
Our convergence guarantees will scale with $\Cpibase{\pi}$,
the response coverage of $\piref$ over $\pi$, defined below.

\begin{definition}[Coverage coefficient]\label{def:general-coverage}
  Fix a comparator policy $\pi$.
  The coverage of $\pi$ by another policy $\pi'$ is measured by
  \begin{align}
    \Cpi{\pi}{\pi'}
    \ldef{}\En_{x\sim\mu}\sbr*{
      \sum_{y\in\cY}\frac{\pi(y\mid x)^2}{\pi'(y\mid x)}
    }
    =\En_\pi\sbr*{\frac{\pi(y\mid x)}{\pi'(y\mid x)}}.
    \label{eq:general-target-c2}
  \end{align}
  For the base model, we use $\Cpibase{\pi}\equiv\Cpi{\pi}{\piref}$.
  For the policy iterate $\pit$ we use $\Cpit{\pi}\equiv\Cpi{\pi}{\pit}$.
\end{definition}

The comparator policy $\pi$ need not be optimal or belong to the learner's policy class.
An example is the distribution induced by \bonlong on the base model with $N$ responses per prompt, denoted $\pibon$,
for which $\Cpibase{\pibon} \le N$ \citep[Lemma~F.1]{huang2025bestofn}.
Another example is the optimal policy induced by conditioning the base model on optimal responses, given by
$\piref^\star \ldef{}\frac{\piref(y\mid x)\one{y\in\Ystarx}} {\piref(\Ystarx\mid x)}$.
This comparator policy has $J(\piref^\star) = \Jstar$; moreover,
\begin{align}
  \Cpibase{\piref^\star}
  &= \underbrace{\Cresp}_{\text{\cref{def:response-coverage}}}.
  \label{eq:coverage-equivalence-1}
\end{align}
Therefore, the performance target of the main body (\cref{eq:question-response})
is a special case of the formulation here with comparator policy $\pi = \pirefstar$.

We now extend the convergence guarantee and KL criterion of \cref{thm:kl-regret-main}
to hold for any comparator policy $\pi$, where the inner tolerance $\ein$ now depends on $\Cpibase{\pi}$,
the coverage of the base model over the comparator.

\begin{theorem}[Convergence relative to a comparator]
  \label{thm:general-target-comparator}
  Fix $\eta>0$ and a sequence of policies $\cbr*{\pit}_{t=0}^{T}$
  with $\pit[0]=\piref$ and $\pit>0$ for all $t$.
  Recall that $\mathsf{KL}_t = \Dkl{\pitp}{\piplust}$.
  Then it holds for every comparator policy $\pi$ that
  \begin{align}
    J(\pi)-J(\piT)
    &\leq
    \frac{\Dkl{\pi}{\piref}}{\eta T}
    +\frac{1}{\eta}\sum_{t=0}^{T-1}\mathsf{KL}_t
    +\frac{4}{\eta}\sqrt{
      \Cpibase{\pi}\sum_{t=0}^{T-1}\frac{\mathsf{KL}_t}{t+1}
    },
    \label{eq:general-target-comparator}
  \end{align}
  where $\Cpibase{\pi}$ is given in \cref{def:general-coverage}.
  In particular, for any target suboptimality $\veps\in(0,\Rmax]$,
  we have $J(\pi)-J(\piT)\leq\veps$ whenever
  \begin{align}
    T
    &\geq\left\lceil
      \frac{\log\rbr*{2\Cpibase{\pi}}}{\eta\veps}
    \right\rceil,
    &
    \Dkl{\pitp}{\piplust}
    &\leq\ein\ldef{}
    \frac{1}{64\Cpibase{\pi}T^2\log(eT)},
    \qquad \forall 0\leq t<T.
    \label{eq:general-target-error-condition}
  \end{align}
\end{theorem}

\cref{thm:general-target-comparator} holds simultaneously for all comparator policies $\pi$.
Taking $\pi=\pirefstar$ (\cref{eq:coverage-equivalence-1}) recovers the convergence guarantee of \cref{thm:kl-regret-main}.
More generally, smaller KL errors $\Dkl{\pitp}{\piplust}$
allow the final policy to compete with comparators having larger coverage coefficients.

In particular, to achieve $J-J(\piT)\le\veps$ for a target performance
$J\in[0,\Jstar]$, it suffices to have an inner KL tolerance that scales inversely
with the best response coverage (lowest $\Cpibase{\pi}$) of any policy with $J(\pi) \ge J$,
  \begin{align}
    T
    &=
    \left\lceil \frac{\inf_{\pi : J(\pi) \ge J}{} \log(2 \Cpibase{\pi})}{\eta\veps} \right\rceil
    ;
    \qquad
    \Dkl{\pitp}{\piplust}
    \le
    \frac{1}{ 64 \rbr*{\inf_{\pi : J(\pi) \ge J}{} \Cpibase{\pi}} T^2 \log(eT)},
    \quad \forall 0 \le t < T
    .
    \label{eq:general-target-reward-conditions}
  \end{align}
  Lastly, under \cref{asm:softmax-linear},
  combining \cref{thm:general-target-comparator} with the inner loop convergence result for \ppolitenc (\cref{thm:noclip-inner})
  yields end-to-end sample complexity and gradient step complexity guarantees.
  Choosing $\eta=1/(\Bfeat\Bstar)$,
  for any comparator policy $\pi$,
  \ppolitenc with other parameters set according to $\Cpibase{\pi}$
  returns a final iterate satisfying $J(\pi)-J(\pifin)\le\veps$
  with probability at least $1-\delta$,
  using gradient steps and sampled responses, respectively,
  \begin{align}
    \Nstep(\veps,\delta; \pi)
    &=\wt O\rbr*{\frac{\Cpibase{\pi}\Bfeat^3\Bstar^3}{\veps^3}},
    \qquad
    \Nsample(\veps,\delta; \pi)
    =\wt O\rbr*{
      \frac{\Cpibase{\pi}\Bfeat^3\Bstar^3}{\veps^3}
      \sbr*{d+\log\frac1\delta}
    }.
    \label{eq:general-target-noclip-complexity}
  \end{align}

\subsection{Proofs for \cref{app:general-kl}}
\label{app:general-kl-comparator-policy-proofs}

We first prove \cref{thm:general-target-comparator}, then the reward-target guarantee
under \cref{eq:general-target-reward-conditions}.

\begin{proof}[\pfref{thm:general-target-comparator}]
    We will use the following reward threshold from \cref{lem:general-target-excess}.
  Define
  \begin{align}
    \tau_\pi(x,y)
    =J_x(\piT)+\frac1\eta\sum_{t=0}^{T-1}\mathsf{KL}_t(x)
    +\frac{\log\rbr*{\pi(y\mid x)/\piref(y\mid x)}-1}{\eta T},
  \end{align}
  where $\pi(y\mid{}x) > 0$, and $\tau_\pi(x,y) = 0$ otherwise.
  Using $r(x,y)\leq\tau_\pi(x,y)+[r(x,y)-\tau_\pi(x,y)]_+$, we obtain
  \begin{align}
    J(\pi)-J(\piT)
    &\leq\En_\pi\sbr*{\tau_\pi(x,y)-J_x(\piT)}
      +\En_\pi\sbr*{[r(x,y)-\tau_\pi(x,y)]_+}
    \\
    &=\frac{\Dkl{\pi}{\piref}-1}{\eta T}
      +\frac1\eta\sum_{t=0}^{T-1}\mathsf{KL}_t
      +\En_\pi\sbr*{[r(x,y)-\tau_\pi(x,y)]_+}.
  \end{align}
  \cref{lem:general-target-excess} bounds the last term as
  \begin{align}
    \En_\pi\sbr*{[r(x,y)-\tau_\pi(x,y)]_+}
    \leq\frac1{\eta T}
    +\frac2\eta\sqrt{
      e\Cpibase{\pi}\sum_{t=0}^{T-1}\frac{\mathsf{KL}_t}{t+1}
    },
  \end{align}
  and substituting gives
  \begin{align}
    J(\pi)-J(\piT)
    &\leq\frac{\Dkl{\pi}{\piref}}{\eta T}
    +\frac1\eta\sum_{t=0}^{T-1}\mathsf{KL}_t
    +\frac2\eta\sqrt{
      e\Cpibase{\pi}\sum_{t=0}^{T-1}\frac{\mathsf{KL}_t}{t+1}
    }.
    \label{eq:general-target-comparator-positive}
  \end{align}
  This proves the first theorem claim in \cref{eq:general-target-comparator},
  which uses $2\sqrt e\leq4$.

  Next, for the second claim in the theorem,
  Jensen's inequality gives
  \begin{align}
    \Dkl{\pi}{\piref}
    =\En_\pi\sbr*{\log\frac{\pi(y\mid x)}{\piref(y\mid x)}}
    \leq\log\En_\pi\sbr*{\frac{\pi(y\mid x)}{\piref(y\mid x)}}
    =\log\Cpibase{\pi}.
    \label{eq:general-target-kl-coverage}
  \end{align}
  Under the KL tolerance $\Dkl{\pitp}{\piplust} \le \ein$ from \cref{eq:general-target-error-condition},
  \begin{align}
    \sum_{t=0}^{T-1}\mathsf{KL}_t&\leq T\ein,
    &\sum_{t=0}^{T-1}\frac{\mathsf{KL}_t}{t+1}
    &\leq\ein\sum_{t=1}^T\frac1t\leq\ein\log(eT).
  \end{align}
  Substituting these bounds into \cref{eq:general-target-comparator}
    and the bounds on $T$ and $\ein$ from \cref{eq:general-target-error-condition} gives
  \begin{align}
    J(\pi)-J(\piT)
    &\leq\frac{\log\Cpibase{\pi}}{\eta T}
    +\frac{T\ein}{\eta}
    +\frac4\eta\sqrt{\Cpibase{\pi}\ein\log(eT)}
    \\
    &=\frac1{\eta T}\rbr*{
      \log\Cpibase{\pi}
      +\frac1{64\Cpibase{\pi}\log(eT)}+\frac12
    }
    \\
    &\leq\frac{\log\rbr*{2\Cpibase{\pi}}}{\eta T}
    \leq\veps.
  \end{align}
  This proves the second theorem claim.
\end{proof}

\begin{proof}[Proof of \cref{eq:general-target-reward-conditions}]
  Fix a target performance $J\in[0,\Jstar]$ and write
  $C\ldef{}\inf_{\pi:\,J(\pi)\ge J}\Cpibase{\pi}$.
  The choices in \cref{eq:general-target-reward-conditions} give
  $T=\lceil\log(2C)/(\eta\veps)\rceil$ and
  $\mathsf{KL}_t\le\ein\ldef{}1/(64CT^2\log(eT))$ for every $0\le t<T$.
  For any comparator policy $\pi$ with $J(\pi)\ge J$,
  applying \cref{eq:general-target-comparator} with the KL bound
  \cref{eq:general-target-kl-coverage} gives
  \begin{align}
    J-J(\piT)
    &\le J(\pi)-J(\piT)
    \\
    &\le\frac{\log\Cpibase{\pi}}{\eta T}
      +\frac{T\ein}{\eta}
      +\frac4\eta\sqrt{\Cpibase{\pi}\ein\log(eT)},
  \end{align}
  where we used $\mathsf{KL}_t\le\ein$ and
  $\sum_{t=1}^T1/t\le\log(eT)$.
  The right-hand side is continuous and increasing in $\Cpibase{\pi}$,
  so taking the infimum over these comparators and substituting $\ein$ yields
  \begin{align}
    J-J(\piT)
    &\le\frac{\log C}{\eta T}
      +\frac{T\ein}{\eta}
      +\frac4\eta\sqrt{C\ein\log(eT)}
    \\
    &=\frac1{\eta T}\rbr*{\log C+\frac1{64C\log(eT)}+\frac12}
    \le\frac{\log(2C)}{\eta T}
    \le\veps.
  \end{align}
  The last inequality uses the choice of $T$ in
  \cref{eq:general-target-reward-conditions}.
\end{proof}

We now state and prove the supporting lemmas used in the proof of \cref{thm:general-target-comparator}.

\begin{lemma}
  \label{lem:general-target-excess}
  For any comparator policy $\pi$, define
  the reward threshold $\tau_\pi : \cX \times \cY \rightarrow (-\infty, \infty)$ to be
  \begin{align}
    \tau_\pi(x,y)
    \ldef{}
    J_x(\piT)+\frac1\eta\sum_{t=0}^{T-1}\mathsf{KL}_t(x)
    +\frac{\log\rbr*{\pi(y\mid x)/\piref(y\mid x)}-1}{\eta T},
    \label{eq:general-target-cutoff}
  \end{align}
  where $\pi(y\mid{}x)>0$, and $\tau_\pi(x,y)=0$ otherwise.
  Then
  \begin{align}
    \En_\pi\sbr*{[r(x,y)-\tau_\pi(x,y)]_+}
    \leq\frac1{\eta T}
    +\frac2\eta\sqrt{
      e\Cpibase{\pi}\sum_{t=0}^{T-1}\frac{\mathsf{KL}_t}{t+1}
    }.
    \label{eq:general-target-excess-c2}
  \end{align}
\end{lemma}

\begin{proof}[\pfref{lem:general-target-excess}]
  To bound \cref{eq:general-target-excess-c2}, define
  \begin{align}
    f(x,y)
    \ldef{}\frac{\pi(y\mid x)}{\piref(y\mid x)}
    [r(x,y)-\tau_\pi(x,y)]_+.
    \label{eq:general-target-excess-function}
  \end{align}
  Where $\pi(y\mid{}x)>0$, the definition of $\tau_\pi$ gives
  \begin{align}
    0\leq f(x,y)
    &\leq\frac{\pi(y\mid x)}{\eta T\piref(y\mid x)}
      \sbr*{1+\eta T\Rmax-\log\frac{\pi(y\mid x)}{\piref(y\mid x)}}_+
    \leq\frac{e^{\eta T\Rmax}}{\eta T},
  \end{align}
  where the last inequality uses
  $\sup_{u>0} \cbr*{u[1+\eta T\Rmax-\log u]_+}=e^{\eta T\Rmax}$.
  Since $f(x,y)=0$ otherwise, $f$ is nonnegative and bounded, with
  \begin{align}
    \En_{\piref}\sbr*{f(x,y)}
    =\En_\pi\sbr*{[r(x,y)-\tau_\pi(x,y)]_+}<\infty.
  \end{align}

  Then,
  applying \cref{lem:kl-shared-accumulation} at $t=T$ to the bounded nonnegative function
  $ f(x,y)e^{\frac\eta2\sum_{s=0}^{T-1}(\btstar[s](x)-r(x,y))}$,
  we obtain
  \begin{align}
    \En_{\piref}\sbr*{f(x,y)}
    &\leq\En_{\piref}\sbr*{
      f(x,y)e^{\frac\eta2\sum_{s=0}^{T-1}(\btstar[s](x)-r(x,y))}
      \sqrt{\frac{\piT(y\mid x)}{\piref(y\mid x)}}}
    \\
    &\quad+\sum_{t=1}^T\sqrt{\mathsf{KL}_{t-1}}
      \rbr*{\En_{\piref}\sbr*{
        f(x,y)^2e^{\eta\sum_{s=0}^{t-1}(\btstar[s](x)-r(x,y))}}}^{1/2}
    \\
    &\leq
    \rbr*{\En_{\piref}\sbr*{
      f(x,y)^2e^{-\eta T(r(x,y)-J_x(\piT))+T\sum_{s=0}^{T-1}\mathsf{KL}_s(x)}
    }}^{1/2}
    \\
    &\quad+\sum_{t=1}^T\sqrt{\mathsf{KL}_{t-1}}
    \rbr*{\En_{\piref}\sbr*{
      f(x,y)^2e^{-\eta t(r(x,y)-J_x(\piT))+t\sum_{s=0}^{T-1}\mathsf{KL}_s(x)}
    }}^{1/2}.
    \label{eq:general-target-accumulation}
  \end{align}
    The second inequality uses Cauchy--Schwarz,
  and the following inequality from \cref{lem:kl-shared-normalizer} to bound the exponents,
  which states for every $1\leq t\leq T$ that,
  \begin{align}
    \eta\sum_{s=0}^{t-1}(\btstar[s](x)-r(x,y))
    \leq-\eta t(r(x,y)-J_x(\piT))
    +t\sum_{s=0}^{T-1}\mathsf{KL}_s(x).
  \end{align}

  We now bound the expectations on the RHS of \cref{eq:general-target-accumulation}.
  Where $\pi(y\mid x)>0$, the definition of $\tau_\pi$ gives
  \begin{align}
    &-\eta t(r(x,y)-J_x(\piT))+t\sum_{s=0}^{T-1}\mathsf{KL}_s(x)
    =-\eta t(r(x,y)-\tau_\pi(x,y))
    +\frac tT\rbr*{1-\log\frac{\pi(y\mid x)}{\piref(y\mid x)}}.
  \end{align}
  Since $f(x,y)=0$ otherwise, substituting this identity and the definition of $f$ gives,
  for every $1\leq t\leq T$,
  \begin{align}
    &\En_{\piref}\sbr*{
      f(x,y)^2e^{-\eta t(r(x,y)-J_x(\piT))+t\sum_{s=0}^{T-1}\mathsf{KL}_s(x)}
    }
    \\
    &\qquad=
    e^{t/T}\En_\pi\sbr*{
      \rbr*{\frac{\pi(y\mid x)}{\piref(y\mid x)}}^{1-t/T}
      [r(x,y)-\tau_\pi(x,y)]_+^2
      e^{-\eta t(r(x,y)-\tau_\pi(x,y))}
    }.
    \label{eq:general-target-excess-moment}
  \end{align}

  For the first term on the RHS of \cref{eq:general-target-accumulation},
  taking $t=T$ in \cref{eq:general-target-excess-moment} gives
  \begin{align}
    \En_{\piref}\sbr*{
      f(x,y)^2e^{-\eta T(r(x,y)-J_x(\piT))+T\sum_{s=0}^{T-1}\mathsf{KL}_s(x)}
    }
    &=e\En_\pi\sbr*{
      [r(x,y)-\tau_\pi(x,y)]_+^2e^{-\eta T(r(x,y)-\tau_\pi(x,y))}
    }
    \\
    &\qquad\leq\frac{\En_\pi\sbr*{[r(x,y)-\tau_\pi(x,y)]_+}}{\eta T},
    \label{eq:general-target-terminal}
  \end{align}
  where the last inequality uses $\sup_{z\geq0}ze^{-\eta Tz}=1/(e\eta T)$.

  Next, for the second term of \cref{eq:general-target-accumulation},
  the Cauchy--Schwarz inequality gives
  \begin{align}
    &\left(
      \sum_{t=1}^T\sqrt{\mathsf{KL}_{t-1}}
      \rbr*{\En_{\piref}\sbr*{
        f(x,y)^2e^{-\eta t(r(x,y)-J_x(\piT))+t\sum_{s=0}^{T-1}\mathsf{KL}_s(x)}
      }}^{1/2}
    \right)^2
    \\
    &\qquad\leq
    \rbr*{\sum_{t=1}^T\frac{\mathsf{KL}_{t-1}}t}
    \sum_{t=1}^Tt\En_{\piref}\sbr*{
      f(x,y)^2e^{-\eta t(r(x,y)-J_x(\piT))+t\sum_{s=0}^{T-1}\mathsf{KL}_s(x)}
    }.
    \label{eq:general-target-error-cs}
  \end{align}
  To bound the second sum on the RHS of \cref{eq:general-target-error-cs},
  we use the following two inequalities.
  For $u>0$ and $s\in[0,1]$,
  \begin{align}
    e^s u^{2-s}
    =(u^2)^{1-s}(eu)^s
    \leq\max\{u^2,eu\}
    \leq\frac e2(1+u^2),
    \label{eq:general-kl-inequality-1}
  \end{align}
  which also holds for $u = 0$.
  Further, for $z>0$,
  \begin{align}
    z^2\sum_{t=1}^\infty te^{-tz}
    =\frac{z^2e^{-z}}{(1-e^{-z})^2}
    =\rbr*{\frac{z}{e^{z/2}-e^{-z/2}}}^2
    \leq1,
    \label{eq:general-kl-inequality-2}
  \end{align}
  since $e^{z/2}-e^{-z/2}\geq z$.
  Rewriting \cref{eq:general-target-excess-moment} as an expectation under $\piref$
  and applying these inequalities gives
  \begin{align}
    &\sum_{t=1}^Tt\En_{\piref}\sbr*{
      f(x,y)^2e^{-\eta t(r(x,y)-J_x(\piT))+t\sum_{s=0}^{T-1}\mathsf{KL}_s(x)}
    }
    \\
    &\qquad=\En_{\piref}\sbr*{
      [r(x,y)-\tau_\pi(x,y)]_+^2
      \sum_{t=1}^Tt e^{t/T}
      \rbr*{\frac{\pi(y\mid x)}{\piref(y\mid x)}}^{2-t/T}
      e^{-\eta t(r(x,y)-\tau_\pi(x,y))}
    }
    \\
    &\qquad\leq\frac e2\En_{\piref}\sbr*{
      \rbr*{1+\rbr*{\frac{\pi(y\mid x)}{\piref(y\mid x)}}^2}
      [r(x,y)-\tau_\pi(x,y)]_+^2
      \sum_{t=1}^Tt e^{-\eta t(r(x,y)-\tau_\pi(x,y))}
    }
    \\
    &\qquad\leq\frac{e}{2\eta^2}
      \En_{\piref}\sbr*{1+\rbr*{\frac{\pi(y\mid x)}{\piref(y\mid x)}}^2}
    =\frac{e(1+\Cpibase{\pi})}{2\eta^2}
    \leq\frac{e\Cpibase{\pi}}{\eta^2}.
    \label{eq:general-target-weighted-moment}
  \end{align}
  The first inequality above uses $u=\pi(y\mid x)/\piref(y\mid x)$ and $s=t/T$ in \cref{eq:general-kl-inequality-1}.
  The second uses $z=\eta(r(x,y)-\tau_\pi(x,y))$ where $r(x,y)>\tau_\pi(x,y)$ in \cref{eq:general-kl-inequality-2}.

  Substituting \cref{eq:general-target-weighted-moment}
  into \cref{eq:general-target-error-cs}, we obtain
  \begin{align}
    &\left(
      \sum_{t=1}^T\sqrt{\mathsf{KL}_{t-1}}
      \rbr*{\En_{\piref}\sbr*{
        f(x,y)^2e^{-\eta t(r(x,y)-J_x(\piT))+t\sum_{s=0}^{T-1}\mathsf{KL}_s(x)}
      }}^{1/2}
    \right)^2
    \\
    &\qquad\leq\frac{e\Cpibase{\pi}}{\eta^2}
      \sum_{t=1}^T\frac{\mathsf{KL}_{t-1}}t.
    \label{eq:general-target-error-sum}
  \end{align}

  Substituting \cref{eq:general-target-terminal,eq:general-target-error-sum}
  into \cref{eq:general-target-accumulation} gives
  \begin{align}
    \En_\pi\sbr*{[r(x,y)-\tau_\pi(x,y)]_+}
    &\leq\sqrt{\frac{\En_\pi\sbr*{[r(x,y)-\tau_\pi(x,y)]_+}}{\eta T}}
    +\frac1\eta\sqrt{
      e\Cpibase{\pi}\sum_{t=0}^{T-1}\frac{\mathsf{KL}_t}{t+1}
    }
    \\
    &\leq\frac12\En_\pi\sbr*{[r(x,y)-\tau_\pi(x,y)]_+}
    +\frac1{2\eta T}
    +\frac1\eta\sqrt{
      e\Cpibase{\pi}\sum_{t=0}^{T-1}\frac{\mathsf{KL}_t}{t+1}
    }.
  \end{align}
  Rearranging proves \cref{eq:general-target-excess-c2}.
\end{proof}

\begin{lemma}[Coverage of the optimal reward]
  \label{lem:general-target-optimal-coverage}
  The policy $\pirefstar$ defined in \cref{app:general-kl} satisfies
  $J(\pirefstar)=\Jstar$ and
  \begin{align}
    \Cpibase{\pirefstar}
    =\inf_{\pi:\,J(\pi)\ge\Jstar}\Cpibase{\pi}
    =\En_{x\sim\mu}\sbr*{\frac1{\piref(\Ystarx\mid x)}}
    =\Cresp.
  \end{align}
\end{lemma}

\begin{proof}[\pfref{lem:general-target-optimal-coverage}]
  By definition, $\pirefstar$ is supported on $\Ystarx$, so
  $J(\pirefstar)=\Jstar$. Substituting its probabilities into
  \cref{eq:general-target-c2} gives
  \begin{align}
    \Cpibase{\pirefstar}
    &=\En_{x\sim\mu}\sbr*{
      \sum_{y\in\Ystarx}\frac{\piref(y\mid x)}{\piref(\Ystarx\mid x)^2}}
    =\En_{x\sim\mu}\sbr*{\frac1{\piref(\Ystarx\mid x)}}
    =\Cresp.
  \end{align}
  Thus the infimum over policies with reward at least $\Jstar$ is at most $\Cresp$.
  Conversely, any such policy $\pi$ has $J(\pi)=\Jstar$ and
  $\pi(\Ystarx\mid x)=1$ for $\mu$-almost every prompt.
  Cauchy--Schwarz therefore gives
  \begin{align}
    \sum_{y\in\Ystarx}\frac{\pi(y\mid x)^2}{\piref(y\mid x)}
    \ge\frac{\rbr*{\sum_{y\in\Ystarx}\pi(y\mid x)}^2}
      {\piref(\Ystarx\mid x)}
    =\frac1{\piref(\Ystarx\mid x)}.
  \end{align}
  Averaging over $x\sim\mu$ yields $\Cpibase{\pi}\ge\Cresp$.
  Taking the infimum over these policies proves the reverse inequality.
\end{proof}

\clearpage
\part{\Cref*{sec:pg}: PPO-Style Algorithms}\label{part:appendix-ppo}

\section{\ppolitenc: Global Convergence (Proof of \cref*{cor:pg-main})}\label{app:noclip}

This section proves \cref{cor:pg-main}, which establishes that \ppolitenc is \covpres
under \cref{asm:softmax-linear}.
We first show that the inner loop of \cref{alg:offline-pg}, which reuses a fixed batch,
approximately maximizes the regularized population objective.
The proof adapts the population policy gradient argument in \cref{app:population-pg}.
We control the distance to a parameter realizing the exact mirror descent update,
then use this control to bound the objective gap of the final inner iterate.
The KL divergence criterion in \cref{thm:kl-shared-regret} converts this inner guarantee
into global convergence and coverage preservation.
The section is organized as follows.
\begin{itemize}
  \item \cref{app:noclip-supporting} develops the supporting lemmas for the population and empirical objectives.
  \item \cref{app:noclip-inner} proves convergence of the inner optimization loop of \cref{alg:offline-pg} on a fixed batch.
  \item \cref{app:noclip-global} states the full version of \cref{cor:pg-main}
  and proves global convergence using the KL divergence criterion in \cref{thm:kl-shared-regret}.
\end{itemize}

\para{Notation}
Throughout this section, suppose that \cref{asm:softmax-linear} holds, and define the feature diameter
\begin{align}
  \Dfeat\ldef{}\max_{x,y,y'}\nbr*{\phi(x,y)-\phi(x,y')}_2\le2\Bfeat.
  \label{eq:noclip-size-notation}
\end{align}
For an outer iteration $t$ and regularization parameter $\eta>0$, the exact mirror descent update is
\begin{align}
  \piplust(y\mid x)&=\pit(y\mid x)e^{\eta(r(x,y)-\btstar(x))},
  &\btstar(x)&\ldef{}\frac1\eta\log
  \En_{y\sim\pit(\cdot\mid x)}\sbr*{e^{\eta r(x,y)}}.
  \label{eq:noclip-exact-update}
\end{align}
For a candidate parameter $\theta$, define the importance weight and regularized reward,
\begin{align}
  w_t^{\pi_\theta}(x,y)&\ldef{}\frac{\pi_\theta(y\mid{}x)}{\pit(y\mid{}x)},
  &\ell_\theta(x,y)&\ldef{}r(x,y)-\frac1\eta\log w_t^{\pi_\theta}(x,y).
  \label{eq:noclip-weight-reward}
\end{align}
We also write
\begin{align}
  \widetilde\phi_\theta(x,y)
  &\ldef{}\phi(x,y)-\En_{y'\sim\pi_\theta(\cdot\mid{}x)}\sbr*{\phi(x,y')},
  &\Sigma_\theta
  &\ldef{}\En_{x\sim\mu,\,y\sim\pi_\theta(\cdot\mid{}x)}
  \sbr*{\widetilde\phi_\theta(x,y)\widetilde\phi_\theta(x,y)^\top}.
  \label{eq:noclip-score-covariance}
\end{align}
By \cref{lem:softmax-score-identity}, $\widetilde\phi_\theta=\nabla\log\pi_\theta$.

We define the empirical version of the objective in \cref{eq:pg-objective} by
\begin{align}
  \widehat{J}_{\eta,t}(\pi_\theta)
  \ldef{}\frac1N\sum_{i=1}^N
  \sbr*{w_t^{\pi_\theta}(x_i,y_i)r_i
  -\frac1\eta\rbr*{w_t^{\pi_\theta}(x_i,y_i)\log w_t^{\pi_\theta}(x_i,y_i)-w_t^{\pi_\theta}(x_i,y_i)+1}},
  \label{eq:noclip-empirical-objective}
\end{align}
where $\cD_t=\cbr*{(x_i,y_i,r_i)}_{i=1}^N$ is the batch drawn in \cref{alg:offline-pg}.
Although \cref{alg:offline-pg} does not explicitly use this objective,
the gradient of \cref{eq:noclip-empirical-objective}
is exactly the gradient estimate used in \cref{alg:offline-pg}.
We use $\wh{J}_{\eta, t}$ in the proof to prove that the inner optimization loop converges relative to $\piplust$ on the empirical objective,
which we then transfer to performance on the population objective $\Jtr$.

\subsection{Supporting Lemmas}
\label{app:noclip-supporting}

The supporting lemmas are organized as follows.
\begin{itemize}
  \item \cref{lem:noclip-objective-kl} identifies the population objective gap with the KL divergence to $\piplust$.
  \item \cref{lem:noclip-target-realizability} establishes realizability of $\piplust$ and defines its target parameter $\thetatarget$.

  \item \cref{lem:noclip-empirical-gradient}, \cref{lem:noclip-target-gradient} and \cref{lem:noclip-gradient-self-bound}
  establish properties of the gradient and gradient norm.

  \item \cref{lem:noclip-population-geometry} establishes the local concavity of the population RL objective.

  \item \cref{lem:noclip-empirical-smoothness} bounds the smoothness of the empirical objective on a parameter ball.

  \item \cref{lem:noclip-gradient-concentration} and \cref{lem:noclip-gap-concentration}
  are concentration results for the empirical objective and functions of the empirical gradient.
\end{itemize}
Throughout this subsection, we fix an outer iteration $t$ and condition on the
history before its batch is drawn, so $\pit$ is fixed.
All probabilistic statements are over the batch $\cD_t$ drawn in \cref{alg:offline-pg}.

\begin{lemma}
  \label{lem:noclip-empirical-gradient}
  Let $\widehat{J}_{\eta,t}$ be the empirical objective in
  \cref{eq:noclip-empirical-objective}. For every $\theta\in\bbR^d$,
  \begin{align}
    \nabla\widehat{J}_{\eta,t}(\pi_\theta)
    =\frac1N\sum_{i=1}^N w_t^{\pi_\theta}(x_i,y_i)\ell_\theta(x_i,y_i)
      \nabla\log\pi_\theta(y_i\mid x_i).
    \label{eq:noclip-empirical-gradient}
  \end{align}
  In particular, the estimate in \cref{eq:alg-offline-pg-gradient} equals
  $\nabla\widehat{J}_{\eta,t}(\pitk)$ at each inner iterate of \cref{alg:offline-pg}.
\end{lemma}

\begin{proof}[\pfref{lem:noclip-empirical-gradient}]
  For a sample $(x,y)$, let $F_\theta(x,y)$ denote the expression inside the
  brackets in \cref{eq:noclip-empirical-objective}.
  Since $\pit$ is fixed, $\nabla w_t^{\pi_\theta}=w_t^{\pi_\theta}\nabla\log\pi_\theta$.
  Differentiating gives
  \begin{align}
    \nabla F_\theta(x,y)
    &=\sbr*{r(x,y)-\frac1\eta\rbr*{\log w_t^{\pi_\theta}(x,y)+1-1}}
      \nabla w_t^{\pi_\theta}(x,y)
    \\
    &=w_t^{\pi_\theta}(x,y)\ell_\theta(x,y)\widetilde\phi_\theta(x,y),
    \label{eq:noclip-single-sample-gradient}
  \end{align}
  where $\widetilde\phi_\theta=\nabla\log\pi_\theta$ by
  \cref{lem:softmax-score-identity}. Averaging over $\cD_t$ proves
  \cref{eq:noclip-empirical-gradient}; substituting $\theta=\thetatk$ gives
  the estimate in \cref{eq:alg-offline-pg-gradient}.
\end{proof}

\begin{lemma}
  \label{lem:noclip-objective-kl}
  For any $\eta>0$, $\pit>0$, and policy $\pi$,
  \begin{align}
    \Jtr(\piplust)-\Jtr(\pi)=\frac1\eta\Dkl{\pi}{\piplust}.
    \label{eq:noclip-objective-kl}
  \end{align}
\end{lemma}

\begin{proof}[\pfref{lem:noclip-objective-kl}]
  The closed form of $\piplust$ in \cref{lem:exact-apmd-closed-form} gives
  $\log(\piplust(y\mid{}x)/\pit(y\mid{}x))=\eta(r(x,y)-\btstar(x))$.
  Hence
  \begin{align}
    \Dkl{\pi}{\piplust}
    &=\En_\pi\sbr*{\log\frac{\pi(y\mid{}x)}{\pit(y\mid{}x)}
      -\eta r(x,y)+\eta \btstar(x)}
    \\
    &=-\eta\Jtr(\pi)+\eta\En_{x\sim\mu}\sbr*{\btstar(x)}.
  \end{align}
  Substituting $\pi=\piplust$ gives
  $\Jtr(\piplust)=\En_{x\sim\mu}[\btstar(x)]$, proving the identity.
\end{proof}

\begin{lemma}
  \label{lem:noclip-target-realizability}
  Suppose that \cref{asm:softmax-linear} holds,
  and let $\Pi_\Theta$ be the softmax linear policy class.
  For every outer iteration $t$
  and $\eta>0$, define the target parameter
  \begin{align}
    \thetatarget\ldef{}\thetat+\eta\thetastar.
    \label{eq:noclip-target-parameter}
  \end{align}
  Then $\piplust=\pi_{\thetatarget}\in\Pi_\Theta$ is realizable.
\end{lemma}

\begin{proof}[\pfref{lem:noclip-target-realizability}]
  Linear realizability and the softmax formula give
  \begin{align}
    \pi_{\thetatarget}(y\mid x)
    &=\frac{\exp\rbr*{\phi(x,y)^\top\thetat+\eta r(x,y)}}
      {\sum_{y'}\exp\rbr*{\phi(x,y')^\top\thetat+\eta r(x,y')}}
    =\piplust(y\mid x).
  \end{align}
\end{proof}

\begin{lemma}
  \label{lem:noclip-target-gradient}
  For every $\eta>0$ and $\theta\in\bbR^d$, with $\thetatarget$ defined in
  \cref{lem:noclip-target-realizability},
  \begin{align}
    \nabla\Jtr(\pi_\theta)
    &=\frac1\eta\Sigma_\theta(\thetatarget-\theta).
    \label{eq:noclip-population-gradient}
  \end{align}
  Therefore,
  \begin{align}
    \inner*{\nabla\Jtr(\pi_\theta),\thetatarget-\theta}
    &=\frac1\eta\nbr*{\thetatarget-\theta}_{\Sigma_\theta}^2.
    \label{eq:noclip-gradient-alignment}
  \end{align}
\end{lemma}

\begin{proof}[\pfref{lem:noclip-target-gradient}]
  We separate the part of the regularized reward that varies with the response
  from the part that depends only on the prompt.
  \begin{align}
    \ell_\theta(x,y)
    &=\frac1\eta\phi(x,y)^\top(\thetatarget-\theta)
    +\frac1\eta\log\En_{y'\sim\pit(\cdot\mid x)}
      \sbr*{e^{\phi(x,y')^\top(\theta-\thetat)}}.
  \end{align}
  Since the score has conditional mean zero under $\pi_\theta$, the second term
  makes no contribution to the population gradient. Thus
  \begin{align}
    \nabla\Jtr(\pi_\theta)
    &=\En_{\pi_\theta}\sbr*{\widetilde\phi_\theta(x,y)\ell_\theta(x,y)}
    \\
    &=\frac1\eta\En_{\pi_\theta}
      \sbr*{\widetilde\phi_\theta(x,y)\widetilde\phi_\theta(x,y)^\top}
      (\thetatarget-\theta)
    =\frac1\eta\Sigma_\theta(\thetatarget-\theta).
  \end{align}
  Taking the inner product with $\thetatarget-\theta$ proves the alignment identity.
\end{proof}

\begin{lemma}
  \label{lem:noclip-gradient-self-bound}
  For every $\eta>0$ and $\theta\in\bbR^d$,
  \begin{align}
    \nbr*{\nabla\Jtr(\pi_\theta)}_2^2
    &\le\frac{\Dfeat^2}{\eta}
      \inner*{\nabla\Jtr(\pi_\theta),\thetatarget-\theta}.
    \label{eq:noclip-gradient-self-bound}
  \end{align}
\end{lemma}

\begin{proof}[\pfref{lem:noclip-gradient-self-bound}]
  By \cref{lem:softmax-score-bound}, $0\preceq\Sigma_\theta\preceq\Dfeat^2I$.
  Combining this with \cref{lem:noclip-target-gradient} gives
  \begin{align}
    \nbr*{\nabla\Jtr(\pi_\theta)}_2^2
    &=\frac1{\eta^2}(\thetatarget-\theta)^\top\Sigma_\theta^2(\thetatarget-\theta)
    \\
    &\le\frac{\Dfeat^2}{\eta^2}\nbr*{\thetatarget-\theta}_{\Sigma_\theta}^2
    =\frac{\Dfeat^2}{\eta}\inner*{\nabla\Jtr(\pi_\theta),\thetatarget-\theta}.
  \end{align}
  This bound controls the squared gradient term in the recursion for the squared
  distance to $\thetatarget$ in \cref{lem:noclip-distance-control}.
\end{proof}

\begin{lemma}[Adapted from \protect\cref{lem:pg-local-convex}]
  \label{lem:noclip-population-geometry}
  For every $\eta>0$ and $\theta\in\bbR^d$,
  \begin{align}
    \Jtr(\piplust)-\Jtr(\pi_\theta)
    \le\frac12\nbr*{\frac{\piplust}{\pi_\theta}}_\infty
    \inner*{\nabla\Jtr(\pi_\theta),\thetatarget-\theta}.
  \end{align}

  For a parameter radius $R>0$, define $W_R\ldef{}e^{\Dfeat R}$.
  In particular, if $\nbr*{\theta-\thetatarget}_2\le R$, then
  \begin{align}
    \Jtr(\piplust)-\Jtr(\pi_\theta)
    \le\frac{W_R}{2}
    \inner*{\nabla\Jtr(\pi_\theta),\thetatarget-\theta}.
    \label{eq:noclip-local-gap}
  \end{align}

\end{lemma}

\begin{proof}[\pfref{lem:noclip-population-geometry}]

  By \cref{eq:noclip-objective-kl}, it suffices to bound $\Dkl{\pi_\theta}{\piplust}$.
  We express this divergence as an integral of feature variances along the
  segment from $\theta$ to $\thetatarget$, then compare them with the variance
  under $\pi_\theta$.
  For a fixed prompt $x$, let
  $F_x(s)\ldef{}\log\En_{y\sim\pi_\theta(\cdot\mid{}x)}
  [e^{s\phi(x,y)^\top(\thetatarget-\theta)}]$.
  Differentiating twice and applying Taylor's theorem gives
  \begin{align}
    \Dkl{\pi_\theta(\cdot\mid{}x)}{\piplust(\cdot\mid{}x)}
    &=F_x(1)-F_x(0)-F_x'(0)
    \\
    &=\int_0^1(1-s)
    \Var_{y\sim\pi_{\theta+s(\thetatarget-\theta)}(\cdot\mid{}x)}
    \rbr*{\phi(x,y)^\top(\thetatarget-\theta)}\,ds.
  \end{align}
  To compare these variances with the variance under $\pi_\theta$, write
  $m_x\ldef{}\sup_y\phi(x,y)^\top(\thetatarget-\theta)$.
  The softmax formula gives, for $0\le s\le1$,
  \begin{align}
    \nbr*{\frac{\pi_{\theta+s(\thetatarget-\theta)}(\cdot\mid{}x)}
      {\pi_\theta(\cdot\mid{}x)}}_\infty
    &=\frac{1}{\En_{y\sim\pi_\theta(\cdot\mid{}x)}
      [e^{s(\phi(x,y)^\top(\thetatarget-\theta)-m_x)}]}
    \\
    &\le\frac{1}{\En_{y\sim\pi_\theta(\cdot\mid{}x)}
      [e^{\phi(x,y)^\top(\thetatarget-\theta)-m_x}]}
    =\nbr*{\frac{\piplust(\cdot\mid{}x)}{\pi_\theta(\cdot\mid{}x)}}_\infty,
  \end{align}
  where the inequality uses $\phi(x,y)^\top(\thetatarget-\theta)-m_x\le0$.
  A variance is no larger than the second moment about any fixed center.
  Centering at the mean under $\pi_\theta$ and changing measure therefore gives
  \begin{align}
    &\Var_{y\sim\pi_{\theta+s(\thetatarget-\theta)}(\cdot\mid{}x)}
      \rbr*{\phi(x,y)^\top(\thetatarget-\theta)}
    \\
    &\qquad\le\nbr*{\frac{\piplust(\cdot\mid{}x)}{\pi_\theta(\cdot\mid{}x)}}_\infty
    \En_{y\sim\pi_\theta(\cdot\mid{}x)}
    \sbr*{\rbr*{\widetilde\phi_\theta(x,y)^\top(\thetatarget-\theta)}^2}.
  \end{align}
  Integrating over $s$ and averaging over $x$, we conclude that
  \begin{align}
    \Jtr(\piplust)-\Jtr(\pi_\theta)
    &=\frac1\eta\Dkl{\pi_\theta}{\piplust}
    \\
    &\le\frac1{2\eta}\nbr*{\frac{\piplust}{\pi_\theta}}_\infty
    \nbr*{\thetatarget-\theta}_{\Sigma_\theta}^2
    \\
    &=\frac12\nbr*{\frac{\piplust}{\pi_\theta}}_\infty
      \inner*{\nabla\Jtr(\pi_\theta),\thetatarget-\theta},
  \end{align}
  using \cref{lem:noclip-objective-kl,eq:noclip-gradient-alignment} in the first and last steps.
  If $\nbr*{\theta-\thetatarget}_2\le R$, then
  $\nbr*{\piplust/\pi_\theta}_\infty\le W_R$ by \cref{lem:softmax-policy-ratio},
  proving \cref{eq:noclip-local-gap}.
\end{proof}

\begin{lemma}
  \label{lem:noclip-empirical-smoothness}
  Let $\widehat{J}_{\eta,t}$ be defined in \cref{eq:noclip-empirical-objective}.
  Fix $\eta>0$ and a parameter radius $R>0$, and define the parameter ball
  \begin{align}
    \mathcal B_R\ldef{}\cbr*{\theta\in\bbR^d:\nbr*{\theta-\thetatarget}_2\le R}.
  \end{align}
  For every $\theta\in\mathcal B_R$,
  \begin{align}
    w_t^{\pi_\theta}(x,y) = \frac{\pi_\theta(y\mid{}x)}{\pit(y\mid{}x)}
    \le W_{\eta,R}\ldef{}W_Re^{\eta\Bfeat\Bstar},
    \qquad \forall (x,y),
    \label{eq:noclip-weight-bound}
  \end{align}
  where $W_R=e^{\Dfeat R}$ is defined in \cref{lem:noclip-population-geometry}.

  For every batch, $\widehat{J}_{\eta,t}$ is $\widehat L_{\eta,R}$-smooth on $\mathcal B_R$.
  In particular, for any $\theta,\theta'\in\mathcal B_R$,

  \begin{align}
    \widehat{J}_{\eta,t}(\pi_{\theta'})
    \ge\widehat{J}_{\eta,t}(\pi_\theta)
    +\inner*{\nabla\widehat{J}_{\eta,t}(\pi_\theta),\theta'-\theta}
    -\frac{\widehat L_{\eta,R}}2\nbr*{\theta'-\theta}_2^2,
    \label{eq:noclip-empirical-smoothness}
  \end{align}
  where
  \begin{align}
    \widehat L_{\eta,R}&\ldef{}\frac{\Dfeat^2}{\eta}W_{\eta,R}
      \rbr*{1+\frac54\log W_{\eta,R}}.
    \label{eq:noclip-local-bounds}
  \end{align}
\end{lemma}

\begin{proof}[\pfref{lem:noclip-empirical-smoothness}]

  Fix $\theta\in\mathcal B_R$.
  Since $0\le r(x,y)\le\Bfeat\Bstar$, we have $0\le \btstar(x)\le\Bfeat\Bstar$.
  The identity $\piplust/\pit=e^{\eta(r-\btstar)}$ and
  \cref{lem:softmax-policy-ratio} then give
  \begin{align}
    w_t^{\pi_\theta}(x,y)
    &=
    \frac{\piplust(y\mid{}x)}{\pit(y\mid{}x)}
    \frac{\pi_\theta(y\mid{}x)}{\piplust(y\mid{}x)}
    =\frac{\pi_\theta(y\mid{}x)}{\piplust(y\mid{}x)}
    e^{\eta(r(x,y)-\btstar(x))}
    \le e^{\Dfeat R+\eta\Bfeat\Bstar}=W_{\eta,R},
    \\
    \abs*{\ell_\theta(x,y)}
    &=\abs*{\btstar(x)+\frac1\eta\log\frac{\piplust(y\mid{}x)}{\pi_\theta(y\mid{}x)}}
    \le\Bfeat\Bstar+\frac{\Dfeat R}{\eta}=\frac{\log W_{\eta,R}}{\eta}.
  \end{align}
  Let $F_\theta(x,y)$ denote the expression inside the brackets in
  \cref{eq:noclip-empirical-objective}. Its gradient is given by
  \cref{eq:noclip-single-sample-gradient}. Since
  $\nabla w_t^{\pi_\theta}=w_t^{\pi_\theta}\widetilde\phi_\theta$ and
  $\nabla\ell_\theta=-\widetilde\phi_\theta/\eta$, differentiating once more gives
  \begin{align}
    \nabla^2F_\theta(x,y)
    &=w_t^{\pi_\theta}(x,y)\sbr*{
      \ell_\theta(x,y)\nabla^2\log\pi_\theta(y\mid{}x)
      +\rbr*{\ell_\theta(x,y)-\frac1\eta}
      \widetilde\phi_\theta(x,y)\widetilde\phi_\theta(x,y)^\top}.
  \end{align}
  By \cref{lem:softmax-log-hessian}, for every unit vector $v$,
  \begin{align}
    -v^\top\nabla^2\log\pi_\theta(y\mid{}x)v
    =\Var_{y'\sim\pi_\theta(\cdot\mid{}x)}\rbr*{v^\top\phi(x,y')}
    \le\frac{\Dfeat^2}{4}.
  \end{align}
  The last inequality follows by bounding variance by the squared deviation from
  the midpoint of an interval of width at most $\Dfeat$.
  Together with $\nbr*{\widetilde\phi_\theta(x,y)}_2\le\Dfeat$, this yields
  \begin{align}
    \nbr*{\nabla^2F_\theta(x,y)}_{\mathrm{op}}
    &\le w_t^{\pi_\theta}(x,y)\sbr*{\frac{\Dfeat^2}{4}\abs*{\ell_\theta(x,y)}
      +\Dfeat^2\rbr*{\abs*{\ell_\theta(x,y)}+\frac1\eta}}
    \\
    &\le\frac{\Dfeat^2}{\eta}W_{\eta,R}
      \rbr*{1+\frac54\log W_{\eta,R}}=\widehat L_{\eta,R}.
  \end{align}
  Averaging over the batch gives the same bound on the Hessian of
  $\widehat{J}_{\eta,t}$ throughout $\mathcal B_R$.
  For $\theta,\theta'\in\mathcal B_R$, the segment joining them lies in $\mathcal B_R$
  by convexity, so Taylor's theorem proves \cref{eq:noclip-empirical-smoothness}.
\end{proof}

\begin{lemma}
  \label{lem:noclip-gradient-concentration}
  Let $\widehat{J}_{\eta,t}$ be defined in \cref{eq:noclip-empirical-objective}.
  Fix $\eta,R>0$, $N\ge1$, and $\delta\in(0,1)$, and let
  $W_{\eta,R}$ and $\widehat L_{\eta,R}$ be as in \cref{eq:noclip-weight-bound,eq:noclip-local-bounds}.
  Define
  \begin{align}
    \veps_{\mathsf{dir}}
    &\ldef{}\frac{W_{\eta,R}\log W_{\eta,R}}{\eta N}
    \rbr*{2\log W_{\eta,R}+\frac23\Dfeat R}
    \rbr*{d\log(1+2N)+\log\frac4\delta}
    \\
    &\qquad+\frac{9R}{4N}\rbr*{R\widehat L_{\eta,R}
      +\frac{\Dfeat W_{\eta,R}\log W_{\eta,R}}{\eta}},
    \\
    \veps_{\mathsf{grad}}
    &\ldef{}\frac{2\Dfeat W_{\eta,R}\log W_{\eta,R}}{\eta}
    \sqrt{\frac2N\rbr*{d\log\rbr*{5(1+2N)}+\log\frac4\delta}}
    +\frac{2R\widehat L_{\eta,R}}{N}.
    \label{eq:noclip-gradient-concentration-errors}
  \end{align}
  With probability at least $1-\delta$, simultaneously for every $\theta$ with
  $\nbr*{\theta-\thetatarget}_2\le R$,
  \begin{align}
    \abs*{\inner*{\nabla\widehat{J}_{\eta,t}(\pi_\theta)
      -\nabla\Jtr(\pi_\theta),\thetatarget-\theta}}
    \le\frac14\inner*{\nabla\Jtr(\pi_\theta),\thetatarget-\theta}
      &+\veps_{\mathsf{dir}},
    \label{eq:noclip-directional-concentration}
    \\
    \nbr*{\nabla\widehat{J}_{\eta,t}(\pi_\theta)-\nabla\Jtr(\pi_\theta)}_2
    &\le\veps_{\mathsf{grad}}.
    \label{eq:noclip-gradient-concentration}
  \end{align}

\end{lemma}

\begin{proof}[\pfref{lem:noclip-gradient-concentration}]
  We first control the gradient error in the direction $\thetatarget-\theta$, then
  bound its Euclidean norm on the same parameter ball.
  By \cref{eq:noclip-single-sample-gradient}, the contribution of one sample to the empirical gradient is
  \begin{align}
    Z_\theta(x,y)\ldef{}w_t^{\pi_\theta}(x,y)\ell_\theta(x,y)\widetilde\phi_\theta(x,y).
  \end{align}
  Its expectation under $\pit$ is $\nabla\Jtr(\pi_\theta)$, while its
  average over $\cD_t$ is $\nabla\widehat{J}_{\eta,t}(\pi_\theta)$.
  The bounds in \cref{lem:noclip-empirical-smoothness} and its proof give, throughout the ball,
  \begin{align}
    \nbr*{Z_\theta(x,y)}_2&\le\frac{\Dfeat W_{\eta,R}\log W_{\eta,R}}{\eta},
    \label{eq:noclip-gradient-envelope-proof}
    \\
    \nbr*{Z_\theta(x,y)-Z_{\theta'}(x,y)}_2
    &\le\widehat L_{\eta,R}\nbr*{\theta-\theta'}_2.
    \label{eq:noclip-gradient-lipschitz-proof}
  \end{align}
  Changing measure to $\pi_\theta$ bounds the second moment of this directional
  contribution in terms of the population gradient's alignment with $\thetatarget-\theta$.
  \begin{align}
    \En_{\pit}\sbr*{\inner*{Z_\theta(x,y),\thetatarget-\theta}^2}
    &=\En_{\pi_\theta}\sbr*{w_t^{\pi_\theta}(x,y)\ell_\theta(x,y)^2
      \inner*{\widetilde\phi_\theta(x,y),\thetatarget-\theta}^2}
    \\
    &\le\frac{W_{\eta,R}(\log W_{\eta,R})^2}{\eta^2}
      \nbr*{\thetatarget-\theta}_{\Sigma_\theta}^2
    \\
    &=\frac{W_{\eta,R}(\log W_{\eta,R})^2}{\eta}
      \inner*{\nabla\Jtr(\pi_\theta),\thetatarget-\theta},
    \label{eq:noclip-directional-second-moment}
  \end{align}
  where the last equality is \cref{eq:noclip-gradient-alignment}.
  Also, $\abs*{\inner*{Z_\theta(x,y),\thetatarget-\theta}}
  \le R\Dfeat W_{\eta,R}\log W_{\eta,R}/\eta$.
  Thus, for a fixed $\theta$ and $u>0$, Bernstein's inequality
  \citep[Section~2.8]{vershynin2018high} and $\sqrt{ab}\le a/4+b$ give, with probability at least $1-2e^{-u}$,
  \begin{align}
    &\abs*{\inner*{\nabla\widehat{J}_{\eta,t}(\pi_\theta)
      -\nabla\Jtr(\pi_\theta),\thetatarget-\theta}}
    \\
    &\quad\le\sqrt{\frac{2W_{\eta,R}(\log W_{\eta,R})^2u}{\eta N}
      \inner*{\nabla\Jtr(\pi_\theta),\thetatarget-\theta}}
      +\frac{2R\Dfeat W_{\eta,R}\log W_{\eta,R}}{3\eta N}u
    \\
    &\quad\le\frac14\inner*{\nabla\Jtr(\pi_\theta),\thetatarget-\theta}
      +\frac{W_{\eta,R}\log W_{\eta,R}}{\eta N}
        \rbr*{2\log W_{\eta,R}+\frac23R\Dfeat}u.
    \label{eq:noclip-fixed-directional-bernstein}
  \end{align}
  Take a Euclidean $R/N$-net $\mathcal N_R$ of the ball, with
  $\abr*{\mathcal N_R}\le(1+2N)^d$ \citep[Section~4.2]{vershynin2018high}.
  Applying \cref{eq:noclip-fixed-directional-bernstein} on this net with
  $u=d\log(1+2N)+\log(4/\delta)$ gives an event of probability at least $1-\delta/2$.
  To extend the bound to the whole ball, \cref{eq:noclip-gradient-envelope-proof,eq:noclip-gradient-lipschitz-proof} imply
  \begin{align}
    &\abs*{\inner*{Z_\theta(x,y),\thetatarget-\theta}
      -\inner*{Z_{\theta'}(x,y),\thetatarget-\theta'}}
    \\
    &\quad\le\nbr*{Z_\theta(x,y)-Z_{\theta'}(x,y)}_2\nbr*{\thetatarget-\theta}_2
      +\nbr*{Z_{\theta'}(x,y)}_2\nbr*{\theta-\theta'}_2
    \\
    &\quad\le\rbr*{R\widehat L_{\eta,R}
      +\frac{\Dfeat W_{\eta,R}\log W_{\eta,R}}{\eta}}
      \nbr*{\theta-\theta'}_2.
    \label{eq:noclip-directional-lipschitz-proof}
  \end{align}
  For any $\theta$ in the ball, choose $\bar\theta\in\mathcal N_R$ within distance $R/N$.
  Replacing $\bar\theta$ by $\theta$ in the empirical and population directional
  terms on the left of \cref{eq:noclip-fixed-directional-bernstein}, and in the
  population alignment on the right, adds at most
  \begin{align}
    \rbr*{2+\frac14}\rbr*{R\widehat L_{\eta,R}
      +\frac{\Dfeat W_{\eta,R}\log W_{\eta,R}}{\eta}}\frac RN.
  \end{align}
  This is the second term in $\veps_{\mathsf{dir}}$, proving
  \cref{eq:noclip-directional-concentration}.

  For the Euclidean norm, let $\mathcal V$ be a $1/2$-net of the unit sphere
  with $\abr*{\mathcal V}\le5^d$.
  By \cref{eq:noclip-gradient-envelope-proof} and Hoeffding's inequality
  \citep[Section~2.2]{vershynin2018high}, for fixed
  $(\bar\theta,v)\in\mathcal N_R\times\mathcal V$,
  \begin{align}
    \abs*{\inner*{\nabla\widehat{J}_{\eta,t}(\pi_{\bar\theta})
      -\nabla\Jtr(\pi_{\bar\theta}),v}}
    \le\frac{\Dfeat W_{\eta,R}\log W_{\eta,R}}{\eta}\sqrt{\frac{2u}{N}}
  \end{align}
  with probability at least $1-2e^{-u}$.
  Setting $u=d\log(5(1+2N))+\log(4/\delta)$ and taking a union bound gives
  an event of probability at least $1-\delta/2$ on which this holds for every
  $(\bar\theta,v)\in\mathcal N_R\times\mathcal V$.
  Approximating an arbitrary unit vector by $v\in\mathcal V$ loses at most
  one half of the gradient-error norm, so
  \begin{align}
    \nbr*{\nabla\widehat{J}_{\eta,t}(\pi_{\bar\theta})
      -\nabla\Jtr(\pi_{\bar\theta})}_2
    \le\frac{2\Dfeat W_{\eta,R}\log W_{\eta,R}}{\eta}\sqrt{\frac{2u}{N}},
    \qquad \bar\theta\in\mathcal N_R.
  \end{align}
  Finally, \cref{eq:noclip-gradient-lipschitz-proof} bounds the change in each
  of the empirical and population gradients by $R\widehat L_{\eta,R}/N$
  when passing from $\bar\theta$ to $\theta$.
  This proves \cref{eq:noclip-gradient-concentration}; intersecting the two
  events establishes the simultaneous guarantee.
\end{proof}

\begin{lemma}
  \label{lem:noclip-gap-concentration}
  Let $\widehat{J}_{\eta,t}$ be defined in \cref{eq:noclip-empirical-objective}.
  Fix $\eta,R>0$, $N\ge1$, and $\delta\in(0,1)$, and define
  \begin{align}
    \veps_{\mathsf{gap}}\ldef{}\frac{W_{\eta,R}}{\eta N}\Bigg[
      &\rbr*{8\eta^2(\Bfeat\Bstar)^2+4\Dfeat R+\frac23\log W_{\eta,R}}
      \rbr*{d\log(1+2N)+\log\frac2\delta}
      \\
      &+\frac94\Dfeat R\log W_{\eta,R}\Bigg],
    \label{eq:noclip-gap-concentration-error}
  \end{align}
  where $W_{\eta,R}$ is defined in \cref{eq:noclip-weight-bound}.
  With probability at least $1-\delta$, simultaneously for every $\theta$ with
  $\nbr*{\theta-\thetatarget}_2\le R$,
  \begin{align}
    &\abs*{\rbr*{\widehat{J}_{\eta,t}(\piplust)-\widehat{J}_{\eta,t}(\pi_\theta)}
      -\rbr*{\Jtr(\piplust)-\Jtr(\pi_\theta)}}
    \le\frac14\rbr*{\Jtr(\piplust)-\Jtr(\pi_\theta)}+\veps_{\mathsf{gap}}.
    \label{eq:noclip-gap-concentration}
  \end{align}
\end{lemma}

\begin{proof}[\pfref{lem:noclip-gap-concentration}]

  We bound the second moment of each sample's contribution to the empirical
  objective gap by a multiple of the population objective gap. Bernstein's
  inequality and a parameter net then give a uniform concentration bound,
  as in \cref{lem:noclip-gradient-concentration}.
  Let $Y_\theta(x,y)$ be the difference between the contributions to
  \cref{eq:noclip-empirical-objective} at $\thetatarget$ and $\theta$, and write
  $v_\theta(x,y)\ldef{}\pi_\theta(y\mid{}x)/\piplust(y\mid{}x)
  =w_t^{\pi_\theta}(x,y)/w_t^{\piplust}(x,y)$.
  Substituting $\log w_t^{\piplust}=\eta(r-\btstar)$ gives
  \begin{align}
    Y_\theta(x,y)
    =w_t^{\piplust}(x,y)\Bigg[
      &\btstar(x)\rbr*{1-v_\theta(x,y)}
      \\
      &+\frac1\eta\rbr*{v_\theta(x,y)\log v_\theta(x,y)-v_\theta(x,y)+1}\Bigg].
    \label{eq:noclip-gap-summand}
  \end{align}
  Its expectation under $\pit$ is $\Jtr(\piplust)-\Jtr(\pi_\theta)$, while its
  average over $\cD_t$ is $\widehat{J}_{\eta,t}(\piplust)-\widehat{J}_{\eta,t}(\pi_\theta)$.
  The term containing $\btstar(x)$ has conditional mean zero under $\pit$,
  since $\En_{y\sim\piplust(\cdot\mid{}x)}[v_\theta(x,y)]=1$,
  but it must still be included in the second moment.
  On the parameter ball,
  \begin{align}
    w_t^{\piplust}(x,y)&\le e^{\eta\Bfeat\Bstar},
    &W_R^{-1}\le v_\theta(x,y)&\le W_R,
    &0\le \btstar(x)&\le\Bfeat\Bstar.
  \end{align}
  For any $v\in[W_R^{-1},W_R]$,
  \begin{align}
    |1-v|&\le W_R,
    &0\le v\log v-v+1&\le W_R\log W_R,
    &(v-1)^2&\le2W_R(v\log v-v+1).
    \label{eq:noclip-scalar-kl-bounds}
  \end{align}
  The last inequality follows because $v\log v-v+1$ and its first derivative
  vanish at $v=1$, while its second derivative is $1/v\ge1/W_R$ on this interval.
  These bounds and \cref{eq:noclip-gap-summand} imply
  \begin{align}
    |Y_\theta(x,y)|
    \le e^{\eta\Bfeat\Bstar}W_R
      \rbr*{\Bfeat\Bstar+\frac{\log W_R}{\eta}}
    =\frac{W_{\eta,R}\log W_{\eta,R}}{\eta}.
  \end{align}
  Moreover, using $(a+b)^2\le2a^2+2b^2$ and changing measure to $\piplust$ gives
  \begin{align}
    \En_{\pit}[Y_\theta^2]
    &\le2e^{\eta\Bfeat\Bstar}(\Bfeat\Bstar)^2
      \En_{\piplust}\sbr*{(1-v_\theta)^2}
    \\
    &\quad+\frac{2e^{\eta\Bfeat\Bstar}}{\eta^2}
        \En_{\piplust}\sbr*{(v_\theta\log v_\theta-v_\theta+1)^2}
    \\
    &\le W_{\eta,R}\rbr*{4(\Bfeat\Bstar)^2+\frac{2\log W_R}{\eta^2}}
      \En_{\piplust}\sbr*{v_\theta\log v_\theta-v_\theta+1}
    \\
    &=W_{\eta,R}\rbr*{4\eta(\Bfeat\Bstar)^2+\frac{2\Dfeat R}{\eta}}
      \rbr*{\Jtr(\piplust)-\Jtr(\pi_\theta)}.
    \label{eq:noclip-gap-second-moment}
  \end{align}
  The second step uses \cref{eq:noclip-scalar-kl-bounds}; the last step uses
  \begin{align}
    \En_{\piplust}\sbr*{v_\theta\log v_\theta-v_\theta+1}
    &=\En_{\pi_\theta}\sbr*{\log\frac{\pi_\theta}{\piplust}}
    =\Dkl{\pi_\theta}{\piplust}
    =\eta\rbr*{\Jtr(\piplust)-\Jtr(\pi_\theta)},
  \end{align}
  where the first equality again uses that both policies have total mass one,
  and the last equality is \cref{lem:noclip-objective-kl}.

  Take a Euclidean $R/N$-net $\mathcal N_R$ with $\abr*{\mathcal N_R}\le(1+2N)^d$,
  and set $u=d\log(1+2N)+\log(2/\delta)$.
  Bernstein's inequality \citep[Section~2.8]{vershynin2018high},
  \cref{eq:noclip-gap-second-moment}, and a union bound imply that, with
  probability at least $1-\delta$, simultaneously for every $\bar\theta\in\mathcal N_R$,
  \begin{align}
    &\abs*{\rbr*{\widehat{J}_{\eta,t}(\piplust)-\widehat{J}_{\eta,t}(\pi_{\bar\theta})}
      -\rbr*{\Jtr(\piplust)-\Jtr(\pi_{\bar\theta})}}
    \\
    &\quad\le\sqrt{\frac{2W_{\eta,R}u}{N}
      \rbr*{4\eta(\Bfeat\Bstar)^2+\frac{2\Dfeat R}{\eta}}
      \rbr*{\Jtr(\piplust)-\Jtr(\pi_{\bar\theta})}}
      +\frac{2W_{\eta,R}\log W_{\eta,R}}{3\eta N}u
    \\
    &\quad\le\frac14\rbr*{\Jtr(\piplust)-\Jtr(\pi_{\bar\theta})}
      +\frac{W_{\eta,R}}{\eta N}
        \rbr*{8\eta^2(\Bfeat\Bstar)^2+4\Dfeat R+\frac23\log W_{\eta,R}}u,
  \end{align}
  where the last step is $\sqrt{ab}\le a/4+b$.
  To extend to the whole ball, \cref{eq:noclip-single-sample-gradient} gives
  \begin{align}
    \nabla Y_\theta(x,y)
    &=-w_t^{\pi_\theta}(x,y)\ell_\theta(x,y)\widetilde\phi_\theta(x,y),
    &\nbr*{\nabla Y_\theta(x,y)}_2
    &\le\frac{\Dfeat W_{\eta,R}\log W_{\eta,R}}{\eta}.
  \end{align}
  For each $\theta$ in the ball, choose $\bar\theta\in\mathcal N_R$ within distance $R/N$.
  Replacing $\bar\theta$ by $\theta$ in the empirical and population gaps on
  the left and the population gap on the right adds at most
  \begin{align}
    \rbr*{2+\frac14}\frac{\Dfeat W_{\eta,R}\log W_{\eta,R}}{\eta}\frac RN
    =\frac{9\Dfeat R W_{\eta,R}\log W_{\eta,R}}{4\eta N}.
  \end{align}
  This is the remaining term in $\veps_{\mathsf{gap}}$, proving the claim.
\end{proof}

\subsection{Inner Loop Convergence}
\label{app:noclip-inner}

We now prove convergence of the inner loop of \cref{alg:offline-pg}, which reuses
one batch for all $K$ gradient steps. The proof extends the population policy
gradient argument in \cref{app:population-pg} by controlling the parameter drift
caused by sampling error. A sufficiently large batch keeps the iterates near
the target parameter, where the empirical objective is smooth and its gradient
and objective gap concentrate uniformly.
Throughout the subsection, we condition on the history before the batch $\cD_t$ is drawn.

\begin{theorem}[Inner loop convergence of \ppolitenc]
  \label{thm:noclip-inner}
  Fix an outer iteration $t$, integers $K,N\ge1$, and $\delta\in(0,1)$.
  Run the inner loop of \cref{alg:offline-pg},
  initialized at $\thetatk[0]=\thetat$, with
  \begin{align}
    \eta=\frac1{\Bfeat\Bstar},
    \qquad\alpha=\frac1{60e^5\Bfeat^3\Bstar}.
    \label{eq:noclip-inner-parameters}
  \end{align}
  If
  \begin{align}
    N\ge2K\log\rbr*{\frac{8[5(1+2N)]^d}{\delta}},
    \label{eq:noclip-inner-sample-condition}
  \end{align}
  then, with probability at least $1-\delta$, for every $0\le k\le K$ and every $(x,y)$,

  \begin{align}
    \nbr*{\thetatk-\thetatarget}_2&<R=2\eta\Bstar,
    \\
    \frac{\pitk(y\mid{}x)}{\pit(y\mid{}x)}&\le e^{\Dfeat R+\eta\Bfeat\Bstar},
    &\frac{\piplust(y\mid{}x)}{\pitk(y\mid{}x)}&\le e^{\Dfeat R},
    \label{eq:noclip-inner-policy-ratios}
  \end{align}
  and the final policy satisfies
  \begin{align}
    \Jtr(\piplust)-\Jtr(\pitk[K])
    \le150e^9\Bfeat\Bstar\sbr*{\frac1K+
      \frac1N\log\rbr*{\frac{8[5(1+2N)]^d}{\delta}}}.
    \label{eq:noclip-inner-rate}
  \end{align}

\end{theorem}

\begin{proof}[\pfref{thm:noclip-inner}]

  Set $R=2\eta\Bstar=2/\Bfeat$. Recall the policy ratio bounds
  \begin{align}
    W_R&=e^{\Dfeat R},
    &W_{\eta,R}&=e^{\Dfeat R+\eta\Bfeat\Bstar}
  \end{align}
  from \cref{lem:noclip-population-geometry,eq:noclip-weight-bound}.
  On the ball of radius $R$ around $\thetatarget$, these bound
  $\piplust/\pi_\theta$ and $\pi_\theta/\pit$, respectively.
  The empirical objective is smooth on this ball with constant
  $\widehat L_{\eta,R}$ defined in \cref{eq:noclip-local-bounds}.
  Since $\Dfeat R\le4$ and $\eta\Bfeat\Bstar=1$,
  \begin{align}
    W_R&\le e^4,
    &W_{\eta,R}&\le e^5,
    &\widehat L_{\eta,R}&\le29e^5\Bfeat^3\Bstar<\frac1\alpha.
    \label{eq:noclip-inner-specialized-bounds}
  \end{align}
  Also, $\alpha\Dfeat^2/\eta\le1/(15e^5)<3/4$, which verifies the learning rate
  condition in \cref{lem:noclip-distance-control}.

  Apply \cref{lem:noclip-gradient-concentration,lem:noclip-gap-concentration}
  with failure probability $\delta/2$ each. By a union bound, with probability
  at least $1-\delta$, the directional and Euclidean norm bounds on the gradient
  error in \cref{eq:noclip-directional-concentration,eq:noclip-gradient-concentration}
  and the objective-gap bound in \cref{eq:noclip-gap-concentration} hold
  simultaneously for every $\theta$ in the ball.
  Fix a batch for which these inequalities hold, and let
  $\veps_{\mathsf{dir}}$, $\veps_{\mathsf{grad}}$, and $\veps_{\mathsf{gap}}$
  denote the errors in
  \cref{eq:noclip-gradient-concentration-errors,eq:noclip-gap-concentration-error}
  evaluated with failure probability $\delta/2$.
  Abbreviate the common logarithmic factor by
  \begin{align}
    u\ldef{}\log\rbr*{\frac{8[5(1+2N)]^d}{\delta}}.
  \end{align}
  Then the sample condition is $N\ge2Ku$, and $u\ge\log120>9/2$.
  Substituting \cref{eq:noclip-inner-specialized-bounds} into the error bounds gives
  \begin{align}
    \veps_{\mathsf{dir}}
    &\le\frac{e^5\Bfeat\Bstar}{N}\rbr*{\frac{190}{3}u+306}
    \le\frac{394e^5\Bfeat\Bstar}{3N}u,
    \\
    \veps_{\mathsf{grad}}
    &\le e^5\Bfeat^2\Bstar\rbr*{20\sqrt{\frac{2u}{N}}+\frac{116}{N}}
    \le33e^5\Bfeat^2\Bstar\sqrt{\frac{2u}{N}},
    \\
    \veps_{\mathsf{gap}}
    &\le\frac{e^5\Bfeat\Bstar}{N}\rbr*{\frac{82}{3}u+45}
    \le\frac{112e^5\Bfeat\Bstar}{3N}u.
    \label{eq:noclip-inner-error-bounds}
  \end{align}
  The last bounds for $\veps_{\mathsf{dir}}$ and $\veps_{\mathsf{gap}}$
  use $u>9/2$. For $\veps_{\mathsf{grad}}$, the condition $N\ge2Ku\ge2u$ gives
  $116/\sqrt{2Nu}\le58/u<13$, which bounds its $N^{-1}$ term by a multiple of
  $\sqrt{2u/N}$.

  We first show that all iterates remain in the ball of radius $R$ around $\thetatarget$.
  The accumulated error in \cref{eq:noclip-localization-recursion} satisfies
  \begin{align}
    K\rbr*{2\alpha\veps_{\mathsf{dir}}+2\alpha^2\veps_{\mathsf{grad}}^2}
    &\le\frac{5029}{900}\frac{Ku}{\Bfeat^2N}
    \le\frac{5029}{1800\Bfeat^2}<\frac3{\Bfeat^2}.
    \label{eq:noclip-inner-localization-error}
  \end{align}
  The initialization satisfies
  $\nbr*{\thetatk[0]-\thetatarget}_2^2=\eta^2\Bstar^2=1/\Bfeat^2$.
  Whenever the current iterate lies in the ball, \cref{lem:noclip-distance-control}
  bounds the increase in squared distance by
  $2\alpha\veps_{\mathsf{dir}}+2\alpha^2\veps_{\mathsf{grad}}^2$.
  Induction therefore gives
  \begin{align}
    \nbr*{\thetatk-\thetatarget}_2^2
    &\le\frac1{\Bfeat^2}
      +k\rbr*{2\alpha\veps_{\mathsf{dir}}+2\alpha^2\veps_{\mathsf{grad}}^2}
    <\frac4{\Bfeat^2}=R^2,
    \qquad 0\le k\le K.
    \label{eq:noclip-inner-localization}
  \end{align}
  Indeed, \cref{eq:noclip-inner-localization-error} ensures that the next iterate
  stays strictly inside the ball whenever the recursion is applied.
  Applying \cref{lem:noclip-empirical-smoothness,lem:softmax-policy-ratio} on this ball yields
  \begin{align}
    \frac{\pitk(y\mid{}x)}{\pit(y\mid{}x)}
    &\le W_{\eta,R},
    &\frac{\piplust(y\mid{}x)}{\pitk(y\mid{}x)}
    &\le W_R.
  \end{align}
  Substituting the definitions of $W_R$ and $W_{\eta,R}$ proves
  \cref{eq:noclip-inner-policy-ratios}.
  Both endpoints of every inner update now lie in the ball, and
  $\alpha\widehat L_{\eta,R}<1$ by \cref{eq:noclip-inner-specialized-bounds}.
  Together with the uniform concentration bounds fixed above, these conditions
  verify the hypotheses of \cref{lem:noclip-progress} for every inner update.
  Its proof also gives the empirical ascent bound \cref{eq:noclip-empirical-ascent},
  so the empirical objective is nondecreasing.

  We next bound the final objective gap. Sum \cref{eq:noclip-progress} over
  $0\le k<K$ and telescope the squared distances and empirical objective increments.
  Using $W_R\le e^4$ and the initial distance gives
  \begin{align}
    \sum_{k=0}^{K-1}\rbr*{\widehat{J}_{\eta,t}(\piplust)-\widehat{J}_{\eta,t}(\pitk)}
    &\le\frac{5e^4\eta^2\Bstar^2}{12\alpha}
      +\frac{5e^4}{6}\rbr*{\widehat{J}_{\eta,t}(\pitk[K])-\widehat{J}_{\eta,t}(\pitk[0])}
    \\
    &\quad+K\rbr*{\frac{5e^4}{6}\veps_{\mathsf{dir}}+\veps_{\mathsf{gap}}}.
  \end{align}
  Each empirical gap in the sum is at least the final empirical gap, because the empirical objective is nondecreasing.
  Moving that gap to the left gives
  \begin{align}
    &\rbr*{K+\frac{5e^4}{6}}
      \rbr*{\widehat{J}_{\eta,t}(\piplust)-\widehat{J}_{\eta,t}(\pitk[K])}
    \\
    &\quad\le\frac{5e^4\eta^2\Bstar^2}{12\alpha}
      +\frac{5e^4}{6}\rbr*{\widehat{J}_{\eta,t}(\piplust)-\widehat{J}_{\eta,t}(\pitk[0])}
      +K\rbr*{\frac{5e^4}{6}\veps_{\mathsf{dir}}+\veps_{\mathsf{gap}}}.
    \label{eq:noclip-inner-final-empirical-gap}
  \end{align}
  The initial population gap is at most $\Bfeat\Bstar$, because
  $\Jtr(\piplust)\le J(\piplust)\le\Bfeat\Bstar$ and $\Jtr(\pit)=J(\pit)\ge0$.
  By \cref{lem:noclip-gap-concentration},
  \begin{align}
    \widehat{J}_{\eta,t}(\piplust)-\widehat{J}_{\eta,t}(\pitk[0])
    &\le\frac54\Bfeat\Bstar+\veps_{\mathsf{gap}},
    \\
    \Jtr(\piplust)-\Jtr(\pitk[K])
    &\le\frac43\rbr*{\widehat{J}_{\eta,t}(\piplust)-\widehat{J}_{\eta,t}(\pitk[K])}
      +\frac43\veps_{\mathsf{gap}}.
  \end{align}
  Substitute the initial empirical gap bound into \cref{eq:noclip-inner-final-empirical-gap}
  and divide by $K+5e^4/6$. The two coefficients of $\veps_{\mathsf{gap}}$ sum to one, giving
  \begin{align}
    &\widehat{J}_{\eta,t}(\piplust)-\widehat{J}_{\eta,t}(\pitk[K])
    \\
    &\quad\le
      \frac{5e^4\eta^2\Bstar^2/(12\alpha)+25e^4\Bfeat\Bstar/24}{K+5e^4/6}
      +\frac{5e^4K}{6K+5e^4}\veps_{\mathsf{dir}}+\veps_{\mathsf{gap}}.
  \end{align}
  Converting to the population gap and using $K+5e^4/6\ge K$ yields
  \begin{align}
    \Jtr(\piplust)-\Jtr(\pitk[K])
    &\le\rbr*{\frac{100e^9}{3}+\frac{25e^4}{18}}\frac{\Bfeat\Bstar}{K}
      +\frac{10e^4}{9}\veps_{\mathsf{dir}}+\frac83\veps_{\mathsf{gap}}.
    \label{eq:noclip-inner-rate-errors}
  \end{align}
  Substituting \cref{eq:noclip-inner-error-bounds} and the definition of $u$
  proves \cref{eq:noclip-inner-rate}, since
  \begin{align}
    \frac{100e^9}{3}+\frac{25e^4}{18}&<150e^9,
    &\frac{3940e^9}{27}+\frac{896e^5}{9}&<150e^9.
  \end{align}
\end{proof}

\begin{lemma}[Control of the parameter distance, adapted from \protect\cref{lem:population-pg-progress}]
  \label{lem:noclip-distance-control}
  Let $\widehat{J}_{\eta,t}$ be defined in \cref{eq:noclip-empirical-objective}.
  Fix $\eta,R>0$, $N\ge1$, $\delta\in(0,1)$, and $\alpha>0$ with
  $4\alpha\Dfeat^2\le3\eta$.
  Let $\veps_{\mathsf{dir}}$ and $\veps_{\mathsf{grad}}$ be defined in
  \cref{eq:noclip-gradient-concentration-errors} with failure probability $\delta/2$.
  With probability at least $1-\delta/2$, the directional and norm bounds on the
  gradient error in \cref{lem:noclip-gradient-concentration} hold throughout
  the ball of radius $R$ around $\thetatarget$.
  On this event, for every $\theta$ in the ball, the empirical gradient update
  $\theta^+=\theta+\alpha\nabla\widehat{J}_{\eta,t}(\pi_\theta)$ satisfies
  \begin{align}
    \nbr*{\theta^+-\thetatarget}_2^2
    \le\nbr*{\theta-\thetatarget}_2^2
      +2\alpha\veps_{\mathsf{dir}}+2\alpha^2\veps_{\mathsf{grad}}^2.
    \label{eq:noclip-localization-recursion}
  \end{align}
\end{lemma}

\begin{proof}[\pfref{lem:noclip-distance-control}]
  We bound the increase in squared distance caused by replacing the population gradient
  with its empirical estimate. Expanding the update gives
  \begin{align}
    \nbr*{\theta^+-\thetatarget}_2^2
    &=\nbr*{\theta-\thetatarget}_2^2
      -2\alpha\inner*{\nabla\widehat{J}_{\eta,t}(\pi_\theta),\thetatarget-\theta}
      +\alpha^2\nbr*{\nabla\widehat{J}_{\eta,t}(\pi_\theta)}_2^2.
    \label{eq:noclip-localization-expansion}
  \end{align}
  On the concentration event, the directional and norm bounds give
  \begin{align}
    -2\alpha\inner*{\nabla\widehat{J}_{\eta,t}(\pi_\theta),\thetatarget-\theta}
    &\le-\frac{3\alpha}{2}\inner*{\nabla\Jtr(\pi_\theta),\thetatarget-\theta}
      +2\alpha\veps_{\mathsf{dir}},
    \label{eq:noclip-localization-direction}
    \\
    \nbr*{\nabla\widehat{J}_{\eta,t}(\pi_\theta)}_2^2
    &\le2\nbr*{\nabla\Jtr(\pi_\theta)}_2^2+2\veps_{\mathsf{grad}}^2
    \\
    &\le\frac{2\Dfeat^2}{\eta}
      \inner*{\nabla\Jtr(\pi_\theta),\thetatarget-\theta}
      +2\veps_{\mathsf{grad}}^2,
    \label{eq:noclip-empirical-gradient-bound}
  \end{align}
  where the last inequality uses \cref{eq:noclip-gradient-self-bound}.
  Substituting into the distance expansion yields
  \begin{align}
    \nbr*{\theta^+-\thetatarget}_2^2
    &\le\nbr*{\theta-\thetatarget}_2^2
      -\alpha\rbr*{\frac32-\frac{2\alpha\Dfeat^2}{\eta}}
      \inner*{\nabla\Jtr(\pi_\theta),\thetatarget-\theta}
    \\
    &\quad+2\alpha\veps_{\mathsf{dir}}+2\alpha^2\veps_{\mathsf{grad}}^2.
  \end{align}
  The alignment is nonnegative by \cref{eq:noclip-gradient-alignment}, and
  $4\alpha\Dfeat^2\le3\eta$ makes its coefficient nonpositive.
  Dropping this term proves the claim.
\end{proof}

\begin{lemma}[One-step objective progress]
  \label{lem:noclip-progress}
  Let $\widehat{J}_{\eta,t}$ be defined in \cref{eq:noclip-empirical-objective}.
  Fix $\eta,R>0$, $N\ge1$, $\delta\in(0,1)$, and $\alpha>0$ with
  $\alpha\widehat L_{\eta,R}\le1$.
  Let $\veps_{\mathsf{dir}}$ and $\veps_{\mathsf{gap}}$ be defined in
  \cref{eq:noclip-gradient-concentration-errors,eq:noclip-gap-concentration-error}
  with failure probability $\delta/2$.
  With probability at least $1-\delta$, the directional gradient error and
  objective-gap error satisfy \cref{eq:noclip-directional-concentration,eq:noclip-gap-concentration}
  throughout the ball of radius $R$ around $\thetatarget$.
  On this event, every empirical gradient update
  $\theta^+=\theta+\alpha\nabla\widehat{J}_{\eta,t}(\pi_\theta)$
  with $\nbr*{\theta-\thetatarget}_2\le R$ and $\nbr*{\theta^+-\thetatarget}_2\le R$ satisfies
  \begin{align}
    \widehat{J}_{\eta,t}(\piplust)-\widehat{J}_{\eta,t}(\pi_\theta)
    &\le\frac{5W_R}{12\alpha}
      \rbr*{\nbr*{\theta-\thetatarget}_2^2-\nbr*{\theta^+-\thetatarget}_2^2}
    \\
    &\quad+\frac{5W_R}{6}
      \rbr*{\widehat{J}_{\eta,t}(\pi_{\theta^+})-\widehat{J}_{\eta,t}(\pi_\theta)}
      +\frac{5W_R}{6}\veps_{\mathsf{dir}}+\veps_{\mathsf{gap}}.
    \label{eq:noclip-progress}
  \end{align}
\end{lemma}

\begin{proof}[\pfref{lem:noclip-progress}]
  Apply \cref{lem:noclip-gradient-concentration,lem:noclip-gap-concentration}
  with failure probability $\delta/2$ each. By a union bound, with probability
  at least $1-\delta$, the directional gradient error bound
  \cref{eq:noclip-directional-concentration} and the objective gap error bound
  \cref{eq:noclip-gap-concentration} hold simultaneously for every $\theta$ in
  the ball of radius $R$ around $\thetatarget$.
  Fix a batch for which these bounds hold and a parameter $\theta$ in this ball.

  We first bound the empirical objective gap by the population objective gap.
  Taking the upper bound on the difference between these gaps in
  \cref{eq:noclip-gap-concentration} gives
  \begin{align}
    \widehat{J}_{\eta,t}(\piplust)-\widehat{J}_{\eta,t}(\pi_\theta)
    &\le\rbr*{\Jtr(\piplust)-\Jtr(\pi_\theta)}
      +\frac14\rbr*{\Jtr(\piplust)-\Jtr(\pi_\theta)}+\veps_{\mathsf{gap}}
    \\
    &=\frac54\rbr*{\Jtr(\piplust)-\Jtr(\pi_\theta)}+\veps_{\mathsf{gap}}.
  \end{align}
  Since $\nbr*{\theta-\thetatarget}_2\le R$, the local concavity bound
  \cref{eq:noclip-local-gap} gives
  \begin{align}
    \Jtr(\piplust)-\Jtr(\pi_\theta)
    &\le\frac{W_R}{2}
      \inner*{\nabla\Jtr(\pi_\theta),\thetatarget-\theta}.
  \end{align}
  Substituting this bound for the population objective gap yields
  \begin{align}
    \widehat{J}_{\eta,t}(\piplust)-\widehat{J}_{\eta,t}(\pi_\theta)
    &\le\frac{5W_R}{8}
      \inner*{\nabla\Jtr(\pi_\theta),\thetatarget-\theta}+\veps_{\mathsf{gap}}.
    \label{eq:noclip-gap-population-alignment}
  \end{align}

  We next replace the population gradient in this bound by the empirical gradient.
  The lower bound on the directional gradient error in
  \cref{eq:noclip-directional-concentration} implies
  \begin{align}
    \inner*{\nabla\widehat{J}_{\eta,t}(\pi_\theta),\thetatarget-\theta}
    &\ge\inner*{\nabla\Jtr(\pi_\theta),\thetatarget-\theta}
      -\frac14\inner*{\nabla\Jtr(\pi_\theta),\thetatarget-\theta}
      -\veps_{\mathsf{dir}}
    \\
    &=\frac34\inner*{\nabla\Jtr(\pi_\theta),\thetatarget-\theta}
      -\veps_{\mathsf{dir}}.
  \end{align}
  Moving the error term to the left and dividing by $3/4$ gives
  \begin{align}
    \inner*{\nabla\Jtr(\pi_\theta),\thetatarget-\theta}
    &\le\frac43\rbr*{
      \inner*{\nabla\widehat{J}_{\eta,t}(\pi_\theta),\thetatarget-\theta}
      +\veps_{\mathsf{dir}}}.
  \end{align}
  Substituting into \cref{eq:noclip-gap-population-alignment} and distributing
  the factor $\frac{5W_R}{8}\cdot\frac43=\frac{5W_R}{6}$, we obtain
  \begin{align}
    \widehat{J}_{\eta,t}(\piplust)-\widehat{J}_{\eta,t}(\pi_\theta)
    &\le\frac{5W_R}{8}\cdot\frac43\rbr*{
      \inner*{\nabla\widehat{J}_{\eta,t}(\pi_\theta),\thetatarget-\theta}
      +\veps_{\mathsf{dir}}}+\veps_{\mathsf{gap}}
    \\
    &=\frac{5W_R}{6}
      \inner*{\nabla\widehat{J}_{\eta,t}(\pi_\theta),\thetatarget-\theta}
      +\frac{5W_R}{6}\veps_{\mathsf{dir}}+\veps_{\mathsf{gap}}.
    \label{eq:noclip-empirical-convexity}
  \end{align}
  Both parameters lie in the ball of radius $R$ around $\thetatarget$.
  Smoothness of the empirical objective (\cref{lem:noclip-empirical-smoothness})
  and $\alpha\widehat L_{\eta,R}\le1$ therefore imply
  \begin{align}
    \widehat{J}_{\eta,t}(\pi_{\theta^+})-\widehat{J}_{\eta,t}(\pi_\theta)
    &\ge\alpha\rbr*{1-\frac{\alpha\widehat L_{\eta,R}}2}
      \nbr*{\nabla\widehat{J}_{\eta,t}(\pi_\theta)}_2^2
    \ge\frac\alpha2\nbr*{\nabla\widehat{J}_{\eta,t}(\pi_\theta)}_2^2.
    \label{eq:noclip-empirical-ascent}
  \end{align}
  Using this bound for the squared gradient term in the update identity gives
  \begin{align}
    \inner*{\nabla\widehat{J}_{\eta,t}(\pi_\theta),\thetatarget-\theta}
    &=\frac{\nbr*{\theta-\thetatarget}_2^2-\nbr*{\theta^+-\thetatarget}_2^2}{2\alpha}
      +\frac\alpha2\nbr*{\nabla\widehat{J}_{\eta,t}(\pi_\theta)}_2^2
    \\
    &\le\frac{\nbr*{\theta-\thetatarget}_2^2-\nbr*{\theta^+-\thetatarget}_2^2}{2\alpha}
      +\widehat{J}_{\eta,t}(\pi_{\theta^+})-\widehat{J}_{\eta,t}(\pi_\theta).
  \end{align}
  Substituting into \cref{eq:noclip-empirical-convexity} proves the claim.
\end{proof}

\subsection{Proof of \cref{cor:pg-main}}
\label{app:noclip-global}

We now state the full version of \cref{cor:pg-main}. Its proof combines the inner
loop guarantee with the KL divergence criterion in \cref{thm:kl-shared-regret},
which bounds suboptimality and coverage loss in terms of the error in each
approximate mirror descent update.
By \cref{lem:noclip-objective-kl}, an inner objective gap of at most $\ein/\eta$
gives the required KL tolerance $\ein$.
Under \cref{asm:softmax-linear}, $0\le r(x,y)=\phi(x,y)^\top\thetastar\le\Bfeat\Bstar$,
so we take the reward bound to satisfy $\Rmax\le\Bfeat\Bstar$.

\begin{corollary}[\ppolitenc is \covpres]
  \label{cor:noclip-regret}
  Suppose that \cref{asm:softmax-linear} holds and $\Cresp<\infty$.
  Fix $\veps\in(0,\Rmax]$ and $\delta\in(0,1)$.
  Run \cref{alg:offline-pg} from $\pit[0]=\piref=\pi_{\theta_0}$ with
  \begin{align}
    \eta&=\frac1{\Bfeat\Bstar},
    &T&=\left\lceil\frac{\Bfeat\Bstar\log(2\Cresp)}{\veps}\right\rceil,
    &\alpha&=\frac1{60e^5\Bfeat^3\Bstar},
    \\
    \ein&=\frac1{64\Cresp T^2\log(eT)},
    \label{eq:noclip-outer-choices}
    \\
    K&=\left\lceil\frac{225e^9}{\ein}\right\rceil,
    &N&=\left\lceil4K\rbr*{d\log(80Kd)+\log\frac{8T}{\delta}}\right\rceil.
    \label{eq:noclip-inner-choices}
  \end{align}
  Here $\ein$ is the tolerance on $\Dkl{\pitp}{\piplust}$ in \cref{eq:kl-shared-tolerance}.
  With probability at least $1-\delta$, the returned policy satisfies
  \begin{align}
    \Jstar-J(\pifin)\le\veps.
    \label{eq:noclip-global-conclusion}
  \end{align}
  On the same event, the policy sequence satisfies
  \begin{align}
    \Pr_{x\sim\mu}\rbr*{
      \pit(\Ystarx\mid x)\geq\tfrac12\piref(\Ystarx\mid x)
      \text{ for all }0\leq t\leq T}
    \geq1-\frac{2+\sqrt2}{8\sqrt{\log(eT)}}.
    \label{eq:noclip-coverage-guarantee}
  \end{align}
  The total numbers of gradient steps and sampled responses satisfy
  \begin{align}
    \Nstep(\veps,\delta)
    &=\wt O\rbr*{\frac{\Cresp(\Bfeat\Bstar)^3}{\veps^3}},
    &\Nsample(\veps,\delta)
    &=\wt O\rbr*{\frac{\Cresp(\Bfeat\Bstar)^3
      \rbr*{d+\log(1/\delta)}}{\veps^3}}.
    \label{eq:noclip-total-complexity}
  \end{align}
\end{corollary}

\begin{proof}[\pfref{cor:noclip-regret}]
  We first verify that every inner loop reaches the KL tolerance required by
  \cref{thm:kl-shared-regret}. Condition on the history before outer iteration $t$ and apply \cref{thm:noclip-inner}
  to its fresh batch with failure probability $\delta/T$.
  To verify the sample condition, the choice of $N$ in \cref{eq:noclip-inner-choices} gives
  \begin{align}
    4K\rbr*{d\log(80Kd)+\log\frac{8T}{\delta}}
    \le N\le5K\rbr*{d\log(80Kd)+\log\frac{8T}{\delta}}.
  \end{align}
  Consequently,
  \begin{align}
    \log\rbr*{\frac{8T[5(1+2N)]^d}{\delta}}
    &\le\log\frac{8T}{\delta}
      +d\log\rbr*{75K\rbr*{d\log(80Kd)+\log\frac{8T}{\delta}}}
    \\
    &=\log\frac{8T}{\delta}+d\log(80Kd)
      +d\log\rbr*{\frac{15}{16d}\rbr*{d\log(80Kd)+\log\frac{8T}{\delta}}}
    \\
    &\le2\rbr*{d\log(80Kd)+\log\frac{8T}{\delta}},
    \label{eq:noclip-corollary-log-bound}
  \end{align}
  where the last inequality uses $\log z\le z$ for $z>0$.
  This verifies \cref{eq:noclip-inner-sample-condition} and bounds the logarithmic term in
  \cref{eq:noclip-inner-rate} by $N/(2K)$.
  Therefore, with conditional probability at least $1-\delta/T$,
  \begin{align}
    \Jtr(\piplust)-\Jtr(\pitp)
    &\le150e^9\Bfeat\Bstar\rbr*{\frac1K+\frac1{2K}}
    =\frac{225e^9\Bfeat\Bstar}{K}\le\frac{\ein}{\eta}.
    \label{eq:noclip-inner-target}
  \end{align}
  Taking expectations over the history and a union bound over outer iterations shows
  that this holds simultaneously for all $0\le t<T$ with probability at least $1-\delta$.

  We next use the KL error bound to prove reward convergence and coverage preservation.
  On this event, \cref{lem:noclip-objective-kl} gives
  \begin{align}
    \Dkl{\pitp}{\piplust}
    =\eta\rbr*{\Jtr(\piplust)-\Jtr(\pitp)}
    \le\ein=\frac1{64\Cresp T^2\log(eT)},
    \qquad 0\le t<T.
  \end{align}
  The policies are nonzero everywhere under the softmax parameterization and $\pit[0]=\piref$,
  so \cref{thm:kl-shared-regret} applies.
  Using Jensen's inequality and
  $\sum_{t=0}^{T-1}(t+1)^{-1}\leq\log(eT)$, we obtain
  \begin{align}
    \Jstar-J(\pifin)
    &\leq\frac{\log\Cresp}{\eta T}
      +\frac{T\ein}{\eta}
      +\frac{2\sqrt e}{\eta}\sqrt{\Cresp\ein\log(eT)}
    \\
    &\leq\frac{\log\Cresp+1/64+\sqrt e/4}{\eta T}
    \leq\frac{\log(2\Cresp)}{\eta T}
    \leq\veps.
  \end{align}
  The second line substitutes $\ein$ and the choice of $T$.

  For coverage, apply \cref{eq:kl-shared-regret-coverage} with $c=1/2$
  on the same event. Since each KL error is at most $\ein$,
  \begin{align}
    &\Pr_{x\sim\mu}\rbr*{
      \pit(\Ystarx\mid x)\geq\tfrac12\piref(\Ystarx\mid x)
      \text{ for all }0\leq t\leq T}
    \\
    &\qquad\geq1-(2+\sqrt2)T\sqrt{\Cresp\ein}
      =1-\frac{2+\sqrt2}{8\sqrt{\log(eT)}}.
  \end{align}
  This proves \cref{eq:noclip-coverage-guarantee}.

  Finally, each outer iteration uses $K$ gradient steps and $N$ sampled responses,
  so the total counts are $TK$ and $TN$, respectively.
  Since $\veps\le\Rmax\le\Bfeat\Bstar$ and $\Cresp\ge1$,
  the choices of $T$ and $\ein$ in \cref{eq:noclip-outer-choices} give
  \begin{align}
    \frac{\Bfeat\Bstar\log(2\Cresp)}{\veps}
    \le T\le\frac{3\Bfeat\Bstar\log(2\Cresp)}{\veps},
    \qquad
    \frac1{\ein}
    =64\Cresp T^2\log(eT)
    =\wt O\rbr*{\frac{\Cresp(\Bfeat\Bstar)^2}{\veps^2}}.
  \end{align}
  Substituting these bounds into the definitions of $K$ and $N$ in \cref{eq:noclip-inner-choices} gives
  \begin{align}
    K&=\wt O\rbr*{\frac{\Cresp(\Bfeat\Bstar)^2}{\veps^2}},
    &N&=\wt O\rbr*{K\rbr*{d+\log(T/\delta)}}.
  \end{align}
  Multiplying by $T$ proves \cref{eq:noclip-total-complexity}.
\end{proof}

\Cref{cor:pg-main} follows from \cref{cor:noclip-regret} upon writing
$B=\Bfeat\Bstar$ and choosing $K,N$ as in \cref{eq:noclip-inner-choices}.
The explicit learning rate is $\alpha=1/(60e^5\Bfeat^2B)=\Theta(1/(\Bfeat^2B))$,
and \cref{eq:noclip-total-complexity} gives the stated gradient step and sample bounds.

\section{\clippg: Global Convergence (Proof of \cref*{cor:clippg-main})}\label{app:clip}
This section proves \cref{cor:clippg-main} for \clippg (\cref{alg:clip-pg}).
The section is organized as follows.
\begin{itemize}
  \item \cref{app:clip-supporting} develops the supporting lemmas for the clipped objective.
  \item \cref{app:clip-global} states the full corollary and proves global convergence and coverage.
  \item \cref{app:clip-gradient} gives the gradient of the clipped objective.
  \item \cref{app:clip-softmax} proves inner and global convergence for softmax linear policies
  with an estimated KL penalty and sufficiently wide clipping thresholds.
  \item \cref{app:clip-comparison} compares the objectives of \ppolitenc and \clippg with those of \ppo and \grpo.
\end{itemize}

We assume throughout this section that the policy class is finite with cardinality $\abr*{\Pi_\Theta}$.
Extensions to continuous policy classes require uniform concentration bounds
for the chosen class.

\para{Notation and objectives}
Throughout this section, we use signed rewards,
\begin{align}
  -\Rmax\le r(x,y)\le\Rmax,
  \qquad \forall (x,y)\in\cX\times\cY.
  \label{eq:clip-reward-bound}
\end{align}
For an outer iteration $t$, a regularization parameter $\eta>0$, and $\pit>0$, write
\begin{align}
  w_t^\pi(x,y)&\ldef{}\frac{\pi(y\mid{}x)}{\pit(y\mid{}x)},
  &\Jtr(\pi)&\ldef{}J(\pi)-\frac1\eta\Dkl{\pi}{\pit}.
  \label{eq:clip-weight-regularized-objective}
\end{align}
For $0\le\cminus<1$ and $\cplus\ge0$, use the clipping thresholds from \cref{alg:clip-pg},
\begin{align}
  [z]_{1-\cminus}^{1+\cplus}
  \ldef{}\min\cbr*{1+\cplus,\max\cbr*{1-\cminus,z}}.
\end{align}
Recall the population objective in \cref{eq:clip-pg-objective},
\begin{align}
  \Jtclip(\pi)
  \ldef{}\En_{\pit}\sbr*{
    \min\cbr*{
      w_t^\pi(x,y)r(x,y),
      [w_t^\pi(x,y)]_{1-\cminus}^{1+\cplus}r(x,y)
    }
  }
  -\frac1\eta\Dkl{\pi}{\pit}.
  \label{eq:clip-population-objective}
\end{align}

\subsection{Supporting Lemmas}
\label{app:clip-supporting}

The supporting lemmas are as follows.
\begin{itemize}
  \item \cref{lem:clip-gap-control} derives clipping thresholds that guarantee $\piplust$ maximizes $\Jtclip$,
  which then implies that maximizing $\Jtclip$ controls $\Dkl{\pi}{\piplust}$.
  \item \cref{lem:clip-concentration} proves the concentration inequalities for $\hJtclip$.
  \item \cref{lem:clip-inner} then bounds suboptimality on $\Jtclip$ by the statistical error
  and the empirical $\vepsopt$ (Line~\ref{line:clip-pg-erm}).
\end{itemize}
Throughout the rest of this subsection, fix $\eta$, $t$, and a given $\pit$.

\begin{lemma}
  \label{lem:clip-gap-control}
  If
  \begin{align}
    1-\cminus\le e^{-2\eta\Rmax},
    \qquad
    1+\cplus\ge e^{2\eta\Rmax},
    \label{eq:clip-unclipped-target}
  \end{align}
  then it holds that
  $1 - \cminus \le \frac{\piplust(y\mid{}x)}{\pit(y\mid{}x)} \le 1 + \cplus$ for all $(x,y)$,
  that is, $\piplust$ is not clipped,
  and for every $\pi \in \Pi_\Theta$
  \begin{align}
    \Jtclip(\piplust)-\Jtclip(\pi)
    \ge \Jtr(\piplust)-\Jtr(\pi)
    =\frac1\eta\Dkl{\pi}{\piplust}.
    \label{eq:clip-gap-controls-kl}
  \end{align}
\end{lemma}

\begin{proof}[\pfref{lem:clip-gap-control}]
  Recall that
  $
    \piplust(y\mid{}x)
    =\frac{\pit(y\mid{}x)e^{\eta r(x,y)}}
    {\En_{y'\sim\pit(\cdot\mid{}x)}\sbr*{e^{\eta r(x,y')}}}
  $.
  Since $-\Rmax\le r(x,y)\le\Rmax$, we have
  \begin{align}
    e^{-\eta\Rmax}
    &\le\En_{y\sim\pit(\cdot\mid{}x)}\sbr*{e^{\eta r(x,y)}}
    \le e^{\eta\Rmax},
    \\
    e^{-2\eta\Rmax}
    &\le w_t^{\piplust}(x,y)
    =\frac{e^{\eta r(x,y)}}
    {\En_{y'\sim\pit(\cdot\mid{}x)}\sbr*{e^{\eta r(x,y')}}}
    \le e^{2\eta\Rmax}.
    \label{eq:clip-target-ratio}
  \end{align}
  Thus clipping leaves the objective value for $\piplust$ unchanged.
  For any other policy, taking the minimum in \cref{eq:clip-population-objective}
  can only decrease its objective value, so
  \begin{align}
    \Jtclip(\piplust) - \Jtr(\piplust) = 0,
    \qquad
    \Jtclip(\pi) - \Jtr(\pi) \le 0.
  \end{align}
  Combining these inequalities proves the first part of \cref{eq:clip-gap-controls-kl},
  and the second follows from the identity established for the unclipped objective in \cref{lem:noclip-objective-kl}.
\end{proof}

\begin{lemma}
  \label{lem:clip-concentration}
  Suppose that $1-\cminus\le e^{-2\eta\Rmax}$ and $1+\cplus\ge e^{2\eta\Rmax}$.
  For any $\delta\in(0,1)$, with probability at least $1-\delta$,
  simultaneously for every $\pi\in\Pi_\Theta$,
  \begin{align}
    \Jtclip(\piplust)-\Jtclip(\pi)
    \le 4\rbr*{\hJtclip(\piplust)-\hJtclip(\pi)}
    +\vepsstat(\delta),
    \label{eq:clip-concentration}
  \end{align}
  where
  \begin{align}
    \vepsstat(\delta)
    \ldef{}
    &\rbr*{
      16\eta\Rmax\max\cbr*{1+\cplus,e^{1+3\eta\Rmax}}
      +\frac83\rbr*{1+\cplus+e^{1+3\eta\Rmax}}
    }
\\
    &\qquad{}\times\frac{\Rmax\log\!\rbr*{|\Pi_\Theta|/\delta}}{N}.
    \label{eq:clip-statistical-error}
  \end{align}
\end{lemma}

\begin{proof}[\pfref{lem:clip-concentration}]
  We will bound the difference between the population and empirical objective gaps
  uniformly over $\Pi_\Theta$.
  Fix a policy $\pi \in \Pi_\Theta$.
  Write the population and empirical objective gaps as
  \begin{align}
    \Delta_\pi&\ldef{}\Jtclip(\piplust)-\Jtclip(\pi),
    &\wh{\Delta}_\pi&\ldef{}\hJtclip(\piplust)-\hJtclip(\pi).
  \end{align}
  We write $\En_t=\En_{\pit}$, and let $\wh{\En}_t$ denote the empirical average over $\cD_t$.
  Since both $\Jtclip(\pi)$ and $\hJtclip(\pi)$ use the population KL divergence $-\eta^{-1}\Dkl{\pi}{\pit}$,
  this term cancels when we subtract the two gaps above.
  Moreover, the clipping thresholds ensure that $\piplust$ is not clipped
  (\cref{lem:clip-gap-control}). We therefore have
  \begin{align}
    \Delta_\pi-\wh{\Delta}_\pi
    =(\En_t-\wh{\En}_t)\sbr*{Z_\pi(x,y)},
    \label{eq:clip-gap-deviation}
  \end{align}
  where
  \begin{align}
    Z_\pi(x,y)
    \ldef{}&w_t^{\piplust}(x,y)r(x,y)
      -\min\cbr*{w_t^\pi(x,y)r(x,y),[w_t^\pi(x,y)]_{1-\cminus}^{1+\cplus}r(x,y)}.
  \end{align}

  For this proof, define
  \begin{align}
    r_+(x,y)&\ldef{}\max\cbr*{r(x,y),0},
    &r_-(x,y)&\ldef{}\max\cbr*{-r(x,y),0}.
  \end{align}
  We expand the clipped reward according to the sign of the reward,
  rewriting it in terms of $r_+(x,y)$ and $r_-(x,y)$,
  \begin{align}
    \min\cbr*{w_t^\pi(x,y)r(x,y),[w_t^\pi(x,y)]_{1-\cminus}^{1+\cplus}r(x,y)}
    &=
    r_+(x,y)\min\cbr*{w_t^\pi(x,y),1+\cplus}
    \\
    &\quad-r_-(x,y)\max\cbr*{w_t^\pi(x,y),1-\cminus}.
    \label{eq:clip-signed-reward}
  \end{align}
  Substituting into the definition of $Z_\pi(x,y)$ gives
  \begin{align}
    Z_\pi(x,y)
    ={}&r_+(x,y)w_t^{\piplust}(x,y)
    \rbr*{1-\min\cbr*{\frac{\pi(y\mid{}x)}{\piplust(y\mid{}x)},\frac{1+\cplus}{w_t^{\piplust}(x,y)}}}
    \\
    &+r_-(x,y)w_t^{\piplust}(x,y)
    \rbr*{\max\cbr*{\frac{\pi(y\mid{}x)}{\piplust(y\mid{}x)},\frac{1-\cminus}{w_t^{\piplust}(x,y)}}-1}.
    \label{eq:clip-reward-excess}
  \end{align}

  In \cref{eq:clip-reward-excess}, the second term containing $r_+(x,y)$ is bounded,
  whereas the first term containing $r_-(x,y)$ can grow with $\frac{\pi(y\mid{}x)}{\piplust(y\mid{}x)}$.
  To apply Bernstein's inequality, we first replace $Z_\pi(x,y)$ by a bounded random variable.
  Letting $M_\eta \ldef{} e^{1 + 2\eta \Rmax}$, we define this to be
  \begin{align}
    \widetilde Z_\pi(x,y)
    \ldef{}Z_\pi(x,y)\one{
      r(x,y)\ge0\ \text{or}\ \frac{\pi(y\mid{}x)}{\piplust(y\mid{}x)}\le M_\eta
    }.
    \label{eq:clip-truncated-excess}
  \end{align}

  We next bound the second moment of $\widetilde Z_\pi(x,y)$ in terms of $\Delta_\pi$,
  which will control the variance in Bernstein's inequality,
  and the following steps culminate in \cref{eq:clip-excess-second-moment}.
  Squaring the two terms with disjoint supports in \cref{eq:clip-reward-excess}
  and applying the truncation gives
  \begin{align}
    \widetilde Z_\pi(x,y)^2
    ={}&
    r_+(x,y)^2
    \rbr*{w_t^{\piplust}(x,y)-\min\cbr*{w_t^\pi(x,y),1+\cplus}}^2
    \\
    &+
    r_-(x,y)^2
    \rbr*{\max\cbr*{w_t^\pi(x,y),1-\cminus}-w_t^{\piplust}(x,y)}^2
    \one{\frac{\pi(y\mid{}x)}{\piplust(y\mid{}x)}\le M_\eta}.
  \end{align}
  We will bound the above in terms of the KL divergence,
  for which we first establish useful bounds for $\wt Z_\pi$,
  which will use the following inequality.
  Define $h(u)\ldef{}u\log u-u+1$, with $h(0)=1$ by continuity.
  Integrating $h''(u)=1/u$ twice from $1$, where $h(1)=h'(1)=0$, gives
  \begin{align}
    h(u)
    \ldef{} u\log u - u + 1
    \ge
    \frac{(u-1)^2}{2\max\cbr*{u,1}},
    \qquad u\ge0.
    \label{eq:clip-scalar-kl}
  \end{align}

  Next, we handle two cases separately based on the sign of the rewards.
  Suppose $r(x,y)\ge0$. Then \cref{eq:clip-truncated-excess} gives
  $\widetilde Z_\pi(x,y)=Z_\pi(x,y)$.
  To apply \cref{eq:clip-scalar-kl}, take
  $u=\min\cbr*{w_t^\pi(x,y),1+\cplus}/w_t^{\piplust}(x,y)$.
  By \cref{lem:clip-gap-control}, $w_t^{\piplust}(x,y)\le1+\cplus$, so
  $\max\cbr*{1,u}\le(1+\cplus)/w_t^{\piplust}(x,y)$.
  If $w_t^\pi(x,y)\le1+\cplus$, then $u=\pi(y\mid{}x)/\piplust(y\mid{}x)$;
  otherwise, $1\le u\le\pi(y\mid{}x)/\piplust(y\mid{}x)$.
  Since $h$ is increasing on $[1,\infty)$, in either case
  $h(u)\le h\!\rbr*{\pi(y\mid{}x)/\piplust(y\mid{}x)}$.
  Substituting this choice of $u$ into \cref{eq:clip-scalar-kl} therefore gives
  \begin{align}
    &\rbr*{w_t^{\piplust}(x,y)-\min\cbr*{w_t^\pi(x,y),1+\cplus}}^2
    =\rbr*{w_t^{\piplust}(x,y)}^2(1-u)^2
    \\
    &\qquad\le
    2\rbr*{w_t^{\piplust}(x,y)}^2\max\cbr*{1,u}h(u)
    \\
    &\qquad\le
    2(1+\cplus)w_t^{\piplust}(x,y)
    h\!\rbr*{\frac{\pi(y\mid{}x)}{\piplust(y\mid{}x)}}.
    \label{eq:clip-upper-scalar-bound}
  \end{align}
  Multiplying both sides by $r(x,y)^2 = r_+(x,y)^2$ and using $\abr*{r(x,y)} \le \Rmax$ on the RHS gives
  \begin{align}
    \widetilde Z_\pi(x,y)^2
    \le
    2\Rmax^2(1+\cplus)w_t^{\piplust}(x,y)
    h\!\rbr*{\frac{\pi(y\mid{}x)}{\piplust(y\mid{}x)}}.
    \label{eq:positive-reward-result}
  \end{align}

  Now suppose $r(x,y)<0$.
  If $\pi(y\mid{}x)/\piplust(y\mid{}x)>M_\eta$ then
  $\widetilde Z_\pi(x,y)=0$ by \cref{eq:clip-truncated-excess}.
  Therefore, we assume that both $r(x,y) < 0$ and $\pi(y\mid{}x)/\piplust(y\mid{}x)\le M_\eta$
  for the next set of derivations.

  Apply \cref{eq:clip-scalar-kl} with
  $u=\max\cbr*{w_t^\pi(x,y),1-\cminus}/w_t^{\piplust}(x,y)$.
  Since $1-\cminus\le w_t^{\piplust}(x,y)$ by \cref{lem:clip-gap-control},
  we have $\max\cbr*{1,u}\le M_\eta$.
  If $w_t^\pi(x,y)\ge1-\cminus$, then $u=\pi(y\mid{}x)/\piplust(y\mid{}x)$;
  otherwise, $\pi(y\mid{}x)/\piplust(y\mid{}x)\le u\le1$.
  Since $h$ is decreasing on $[0,1]$, in either case
  $h(u)\le h\!\rbr*{\pi(y\mid{}x)/\piplust(y\mid{}x)}$.
  Thus,
  \begin{align}
    &\rbr*{\max\cbr*{w_t^\pi(x,y),1-\cminus}-w_t^{\piplust}(x,y)}^2
    =\rbr*{w_t^{\piplust}(x,y)}^2(u-1)^2
    \\
    &\qquad\le
    2\rbr*{w_t^{\piplust}(x,y)}^2\max\cbr*{1,u}h(u)
    \\
    &\qquad\le
    2M_\eta\rbr*{w_t^{\piplust}(x,y)}^2
    h\!\rbr*{\frac{\pi(y\mid{}x)}{\piplust(y\mid{}x)}}.
    \label{eq:clip-lower-scalar-bound}
  \end{align}
  The last line above uses $\pi(y\mid{}x)/\piplust(y\mid{}x)\le M_\eta$,
  as, in the second line, the lower clipping threshold alone does not bound $u$ from above.
  For negative rewards, $e^{\eta r(x,y)}\le1$ and
  $r(x,y')\ge-\Rmax$ for every $y'$, so
  \begin{align}
    w_t^{\piplust}(x,y)
    =\frac{e^{\eta r(x,y)}}
    {\En_{y'\sim\pit(\cdot\mid{}x)}\sbr*{e^{\eta r(x,y')}}}
    \le e^{\eta\Rmax}.
    \label{eq:clip-negative-reward-ratio}
  \end{align}
  Multiplying \cref{eq:clip-lower-scalar-bound} by $r(x,y)^2$
  and using \cref{eq:clip-negative-reward-ratio} gives the following bound
  when $\pi(y\mid{}x)/\piplust(y\mid{}x)\le M_\eta$.
  \begin{align}
    \widetilde Z_\pi(x,y)^2
    \le
    2\Rmax^2M_\eta e^{\eta\Rmax}w_t^{\piplust}(x,y)
    h\!\rbr*{\frac{\pi(y\mid{}x)}{\piplust(y\mid{}x)}}.
    \label{eq:negative-reward-result}
  \end{align}
  Since $\widetilde Z_\pi(x,y) = 0$ when $\pi(y\mid{}x)/\piplust(y\mid{}x)>M_\eta$,
  it therefore holds for all $x,y$ such that $r(x,y) < 0$.

  We now bound the second moment of $\wt Z_\pi$ in terms of the KL divergence.
  Combining \cref{eq:positive-reward-result} and \cref{eq:negative-reward-result}
  and averaging under $\pit$ gives
  \begin{align}
    \En_t\sbr*{\widetilde Z_\pi(x,y)^2}
    &\le
    2\Rmax^2\max\cbr*{1+\cplus,M_\eta e^{\eta\Rmax}}
    \En_{\piplust}\sbr*{h\!\rbr*{\frac{\pi(y\mid{}x)}{\piplust(y\mid{}x)}}}.
  \end{align}
  For the expectation on the right,
  since $\pit(y\mid{}x)w_t^{\piplust}(x,y)=\piplust(y\mid{}x)$,
  first we have the identity that
  \begin{align}
    \En_{\piplust}\sbr*{h\!\rbr*{\frac{\pi(y\mid{}x)}{\piplust(y\mid{}x)}}}
    &=
    \En_{\piplust}\sbr*{
      \frac{\pi(y\mid{}x)}{\piplust(y\mid{}x)}
      \log\frac{\pi(y\mid{}x)}{\piplust(y\mid{}x)}
      -\frac{\pi(y\mid{}x)}{\piplust(y\mid{}x)}+1}
    \\
    &=\Dkl{\pi}{\piplust}.
    \label{eq:clip-h-kl-identity}
  \end{align}
  Combining the above two panels we obtain
  \begin{align}
    \En_t\sbr*{\widetilde Z_\pi(x,y)^2}
    &\le
    2\Rmax^2\max\cbr*{1+\cplus,e^{1+3\eta\Rmax}}
    \Dkl{\pi}{\piplust}
    \\
    &\le
    2\eta\Rmax^2\max\cbr*{1+\cplus,e^{1+3\eta\Rmax}}\Delta_\pi.
    \label{eq:clip-excess-second-moment}
  \end{align}
  The last inequality uses \cref{lem:clip-gap-control}
  and gives our bound on the second moment.
  
  In addition to this, for Bernsteins' inequality we need a bound on the range of the random variables.
  For $r(x,y)\ge0$, use
  $0\le\min\cbr*{w_t^\pi(x,y),1+\cplus}\le1+\cplus$ and
  $w_t^{\piplust}(x,y)\le e^{2\eta\Rmax}$ from \cref{lem:clip-gap-control}.
  For $r(x,y)<0$ with $\pi(y\mid{}x)/\piplust(y\mid{}x)\le M_\eta$, use
  $0\le\max\cbr*{w_t^\pi(x,y),1-\cminus}\le M_\eta w_t^{\piplust}(x,y)$
  and \cref{eq:clip-negative-reward-ratio}.
  When $r(x,y)<0$ and $\pi(y\mid{}x)/\piplust(y\mid{}x)>M_\eta$,
  we have $\widetilde Z_\pi(x,y)=0$. Together, these bounds give
  \begin{align}
    -(1+\cplus)\Rmax
    &\le\widetilde Z_\pi(x,y)\le e^{2\eta\Rmax}\Rmax,
    &&r(x,y)\ge0,
    \\
    -e^{\eta\Rmax}\Rmax
    &\le\widetilde Z_\pi(x,y)\le M_\eta e^{\eta\Rmax}\Rmax,
    &&r(x,y)<0.
  \end{align}
  Using $1+\cplus\ge e^{2\eta\Rmax}$ and $M_\eta=e^{1+2\eta\Rmax}$, we then have
  \begin{align}
    -(1+\cplus)\Rmax
    \le\widetilde Z_\pi(x,y)
    \le e^{1+3\eta\Rmax}\Rmax
    ,
  \end{align}
  and
  \begin{align}
    \abr*{\widetilde Z_\pi(x,y)-\En_t[\widetilde Z_\pi(x,y)]}
    \le\Rmax\rbr*{1+\cplus+e^{1+3\eta\Rmax}}.
  \end{align}
  For a fixed $\pi$, Bernstein's inequality \citep[Section~2.8]{vershynin2018high}
  therefore gives, with probability at least $1-\delta/|\Pi_\Theta|$,
  \begin{align}
    (\En_t-\wh{\En}_t)\sbr*{\widetilde Z_\pi(x,y)}
    \le{}&
    \sqrt{\frac{2\Var_{\pit}[\widetilde Z_\pi(x,y)]
      \log\!\rbr*{|\Pi_\Theta|/\delta}}{N}}
    +
    \frac{2\Rmax\rbr*{1+\cplus+e^{1+3\eta\Rmax}}
      \log\!\rbr*{|\Pi_\Theta|/\delta}}{3N}.
      \label{eq:bernstein-1}
  \end{align}
  The variance is bounded by the second moment,
  and using \cref{eq:clip-excess-second-moment} gives
  \begin{align}
    \Var_{\pit}[\wt Z_\pi(x,y)]
    &\le
    \En_t[\widetilde Z_\pi(x,y)^2]
    \\
    &\le
    2\eta\Rmax^2\max\cbr*{1+\cplus,e^{1+3\eta\Rmax}}\Delta_\pi.
  \end{align}
  Substituting this bound into \cref{eq:bernstein-1} and taking a union bound
  over $\Pi_\Theta$ gives, with probability at least $1-\delta$,
  simultaneously for every $\pi\in\Pi_\Theta$,
  \begin{align}
    (\En_t-\wh{\En}_t)\sbr*{\widetilde Z_\pi(x,y)}
    \le{}&
    2\sqrt{
      \frac{\eta\Rmax^2
      \max\cbr*{1+\cplus,e^{1+3\eta\Rmax}}
      \log\!\rbr*{|\Pi_\Theta|/\delta}}{N}\Delta_\pi
    }
    \\
    &+
    \frac{2\Rmax\rbr*{1+\cplus+e^{1+3\eta\Rmax}}
    \log\!\rbr*{|\Pi_\Theta|/\delta}}{3N}.
    \label{eq:clip-truncated-concentration}
  \end{align}

  Finally, we translate this bound into one on the objective gap using the
  relation in \cref{eq:clip-gap-deviation}.
  To do so, we will show that $Z_\pi(x,y)-\widetilde Z_\pi(x,y)$ is nonnegative
  and has expectation at most $\Delta_\pi/2$ under $\pit$.
  Whenever $\pi(y\mid{}x)/\piplust(y\mid{}x)>M_\eta>1$,
  \cref{lem:clip-gap-control} gives
  $w_t^\pi(x,y)>w_t^{\piplust}(x,y)\ge1-\cminus$.
  Using \cref{eq:clip-reward-excess,eq:clip-truncated-excess}, we therefore have
  \begin{align}
    Z_\pi(x,y)-\widetilde Z_\pi(x,y)
    &=
    r_-(x,y)\rbr*{w_t^\pi(x,y)-w_t^{\piplust}(x,y)}
    \one{\frac{\pi(y\mid{}x)}{\piplust(y\mid{}x)}>M_\eta}
    \ge0.
    \label{eq:clip-nonnegative-tail}
  \end{align}
  To bound the expectation of this difference, recall that $h(u)\ge0$ and
  $h(u)/u=\log u-1+1/u$ is increasing for $u>1$. It follows that
  \begin{align}
    u\one{u>M_\eta}
    \le\frac{h(u)}{\log M_\eta-1+1/M_\eta},
    \qquad u\ge0.
  \end{align}
  Since $\pit(y\mid{}x)w_t^{\piplust}(x,y)=\piplust(y\mid{}x)$,
  taking expectations in \cref{eq:clip-nonnegative-tail} gives
  \begin{align}
    \En_t\sbr*{Z_\pi(x,y)-\widetilde Z_\pi(x,y)}
    &=
    \En_{\piplust}\sbr*{
      r_-(x,y)\rbr*{\frac{\pi(y\mid{}x)}{\piplust(y\mid{}x)}-1}
      \one{\frac{\pi(y\mid{}x)}{\piplust(y\mid{}x)}>M_\eta}}
    \\
    &\le
    \Rmax\En_{\piplust}\sbr*{
      \frac{\pi(y\mid{}x)}{\piplust(y\mid{}x)}
      \one{\frac{\pi(y\mid{}x)}{\piplust(y\mid{}x)}>M_\eta}}
    \\
    &\le
    \frac{\Rmax\Dkl{\pi}{\piplust}}
    {\log M_\eta-1+1/M_\eta}
    \\
    &\le
    \frac{\eta\Rmax}{2\eta\Rmax+e^{-1-2\eta\Rmax}}\Delta_\pi
    \le\frac12\Delta_\pi.
    \label{eq:clip-tail-bound}
  \end{align}
  The first inequality uses $0\le r_-(x,y)\le\Rmax$.
  The second applies the scalar inequality above with
  $u=\pi(y\mid{}x)/\piplust(y\mid{}x)$ and then uses \cref{eq:clip-h-kl-identity}.
  The last line uses \cref{lem:clip-gap-control} and
  $\log M_\eta-1+1/M_\eta=2\eta\Rmax+e^{-1-2\eta\Rmax}$.

  Under the event that \cref{eq:clip-truncated-concentration} holds for every
  $\pi\in\Pi_\Theta$,
  by \cref{eq:clip-nonnegative-tail},
  $\wh{\En}_t[Z_\pi(x,y)-\widetilde Z_\pi(x,y)]\ge0$.
  Combining this with \cref{eq:clip-truncated-concentration,eq:clip-tail-bound}
  in \cref{eq:clip-gap-deviation} gives
  \begin{align}
    \Delta_\pi-\wh{\Delta}_\pi
    &=
    (\En_t-\wh{\En}_t)[\widetilde Z_\pi(x,y)]
    +(\En_t-\wh{\En}_t)[Z_\pi(x,y)-\widetilde Z_\pi(x,y)]
    \\
    &\le
    (\En_t-\wh{\En}_t)[\widetilde Z_\pi(x,y)]
    +\En_t[Z_\pi(x,y)-\widetilde Z_\pi(x,y)]
    \label{eq:clip-deviation-truncation}
    \\
    &\le
    \frac12\Delta_\pi
    +2\sqrt{
      \frac{\eta\Rmax^2
      \max\cbr*{1+\cplus,e^{1+3\eta\Rmax}}
      \log\!\rbr*{|\Pi_\Theta|/\delta}}{N}\Delta_\pi
    }
    \\
    &\quad+
    \frac{2\Rmax\rbr*{1+\cplus+e^{1+3\eta\Rmax}}
    \log\!\rbr*{|\Pi_\Theta|/\delta}}{3N}
    \\
    &\le
    \frac34\Delta_\pi
    +\frac{4\eta\Rmax^2
      \max\cbr*{1+\cplus,e^{1+3\eta\Rmax}}
      \log\!\rbr*{|\Pi_\Theta|/\delta}}{N}
    \\
    &\quad+
    \frac{2\Rmax\rbr*{1+\cplus+e^{1+3\eta\Rmax}}
      \log\!\rbr*{|\Pi_\Theta|/\delta}}{3N}
    =\frac34\Delta_\pi+\frac14\vepsstat(\delta).
  \end{align}
  The last inequality uses $2\sqrt{ab}\le a/4+4b$ with
  $a=\Delta_\pi$ and
  $b=\eta\Rmax^2\max\cbr*{1+\cplus,e^{1+3\eta\Rmax}}
  \log\!\rbr*{|\Pi_\Theta|/\delta}/N$.
  The final equality uses the definition of $\vepsstat(\delta)$ in
  \cref{eq:clip-statistical-error}.
  Rearranging gives $\Delta_\pi\le4\wh{\Delta}_\pi+\vepsstat(\delta)$,
  proving \cref{eq:clip-concentration}.
\end{proof}

\begin{lemma}
  \label{lem:clip-inner}
  Suppose that $1-\cminus\le e^{-2\eta\Rmax}$ and $1+\cplus\ge e^{2\eta\Rmax}$.
  If $\pitp\in\Pi_\Theta$ satisfies
  \begin{align}
    \hJtclip(\pitp)\ge\sup_{\pi\in\Pi_\Theta}\hJtclip(\pi)-\vepsopt,
  \end{align}
  then, for any $\delta\in(0,1)$, with probability at least $1-\delta$,
  \begin{align}
    \Jtr(\piplust)-\Jtr(\pitp)
    =\frac1\eta\Dkl{\pitp}{\piplust}
    \le4\vepsopt+\vepsstat(\delta),
    \label{eq:clip-inner-guarantee}
  \end{align}
  where $\vepsstat(\delta)$ is defined in \cref{eq:clip-statistical-error}.
\end{lemma}

\begin{proof}[\pfref{lem:clip-inner}]
  Since $\piplust\in\Pi_\Theta$, Line~\ref{line:clip-pg-erm} gives
  \begin{align}
    \hJtclip(\piplust)-\hJtclip(\pitp)\le\vepsopt.
  \end{align}
  Since the concentration inequality in \cref{lem:clip-concentration} applies to every policy in $\Pi_\Theta$, including $\pitp$,
  combining it with \cref{lem:clip-gap-control} and the above panel give
  \begin{align}
    \frac1\eta\Dkl{\pitp}{\piplust}
    &\le\Jtclip(\piplust)-\Jtclip(\pitp)
    \\
    &\le4\rbr*{\hJtclip(\piplust)-\hJtclip(\pitp)}+\vepsstat(\delta)
    \le4\vepsopt+\vepsstat(\delta).
  \end{align}
\end{proof}

\subsection{Proof of \cref{cor:clippg-main}}
\label{app:clip-global}

We now give the full version of \cref{cor:clippg-main}.
The result depends on the complexity of the finite policy classes considered in this section,
given by $\log|\Pi_\Theta|$.
We first use this result to prove \cref{cor:clippg-main} then prove \cref{cor:clip-pg}.

\begin{corollary}[\clippg is \covpres]
  \label{cor:clip-global}
  \label{cor:clip-pg}
  Suppose that $r(x,y)\in[-\Rmax,\Rmax]$ and $\Pi_\Theta$ is finite.
  Assume that every policy in $\Pi_\Theta$ has full support,
  $\Dkl{\pi}{\pi'}<\infty$ for all $\pi,\pi'\in\Pi_\Theta$,
  and $\piplust\in\Pi_\Theta$ at every outer iteration.
  Fix $\eta>0$, $\veps\in(0,2\Rmax]$, and $\delta\in(0,1)$.
  Run \cref{alg:clip-pg} with $\pit[0]=\piref\in\Pi_\Theta$ and
  \begin{align}
    \cminus&=1-e^{-2\eta\Rmax},
    &\cplus&=e^{2\eta\Rmax}-1,
    &T&=\left\lceil\frac{\log(2\Cresp)}{\eta\veps}\right\rceil.
    \label{eq:clip-outer-parameters}
  \end{align}
  Set

  \begin{align}
    \ein&=\frac1{64\Cresp T^2\log(eT)},
    &\vepsopt&\le\frac{\ein}{8\eta},
    \label{eq:clip-inner-tolerance}
    \\
    N&=\left\lceil
      \frac{16\eta\Rmax}{3\ein}e^{2\eta\Rmax}
      \sbr*{1+(1+6\eta\Rmax)e^{1+\eta\Rmax}}
      \log\frac{|\Pi_\Theta|T}{\delta}
    \right\rceil.
    \label{eq:clip-sample-parameters}
  \end{align}

  Here $\ein$ is the right-hand side of \cref{eq:kl-shared-tolerance}.
  With probability at least $1-\delta$ over the batches, the returned policy satisfies
  \begin{align}
    \Jstar-J(\pifin)\le\veps,
    \label{eq:clip-final-guarantee}
  \end{align}
  and the policy sequence satisfies

  \begin{align}
    \Pr_{x\sim\mu}\rbr*{
      \pit(\Ystarx\mid{}x)\ge\frac12\piref(\Ystarx\mid{}x)
      \text{ for all }0\le t\le T
    }
    \ge1-\frac{2+\sqrt2}{8\sqrt{\log(eT)}}.
    \label{eq:clip-coverage-guarantee}
  \end{align}

  In particular, with $\eta=1/\Rmax$, the number of sampled responses is

  \begin{align}
    \Nsample(\veps,\delta)=TN
    =\wt O\rbr*{
      \frac{\Cresp\Rmax^3}{\veps^3}
      \log\frac{2|\Pi_\Theta|}{\delta}
    }.
    \label{eq:clip-sample-complexity}
  \end{align}

\end{corollary}

\begin{proof}[\pfref{cor:clippg-main}]
 \Cref{cor:clippg-main} follows from \cref{cor:clip-global} by setting
  $\eta=1/\Rmax$ and choosing $N$ as in \cref{eq:clip-sample-parameters}.
  The KL tolerance in \cref{eq:clip-inner-tolerance} is sufficient for \cref{eq:framework-kl-tolerance} (\cref{thm:kl-regret-main}),
  and the same inner optimization requirement $\vepsopt\le\ein/(8\eta)$ applies.
  The sample complexity bound is then given by \cref{eq:clip-sample-complexity}.
\end{proof}

\begin{proof}[\pfref{cor:clip-global}]
  We first show that the KL errors are at most $\ein$ simultaneously over all
  outer iterations with probability at least $1-\delta$.
  We will then apply \cref{thm:kl-shared-regret} to obtain convergence and coverage.

  The thresholds in \cref{eq:clip-outer-parameters} satisfy the conditions of \cref{lem:clip-inner}.
  Substituting $1+\cplus=e^{2\eta\Rmax}$ into \cref{eq:clip-statistical-error} gives

  \begin{align}
    \vepsstat\rbr*{\frac{\delta}{T}}
    &=
    \frac{8\Rmax}{3N}e^{2\eta\Rmax}
    \sbr*{1+(1+6\eta\Rmax)e^{1+\eta\Rmax}}
    \log\frac{|\Pi_\Theta|T}{\delta}
    \\
    &\le\frac{\ein}{2\eta},
    \label{eq:clip-final-statistical-error}
  \end{align}

  where the inequality uses the choice of $N$ in \cref{eq:clip-sample-parameters}.
  Fix an outer iteration $t$ and condition on the preceding batches
  $\cD_0,\ldots,\cD_{t-1}$, which fixes $\pit$ before $\cD_t$ is drawn.
  Applying \cref{lem:clip-inner} with failure probability $\delta/T$ gives
  \begin{align}
    \Pr\rbr*{
      \Dkl{\pitp}{\piplust}
      \le\eta\rbr*{4\vepsopt+\vepsstat(\delta/T)}
      \,\middle|\,\cD_0,\ldots,\cD_{t-1}
    }\ge1-\frac{\delta}{T}.
  \end{align}
  Since $4\vepsopt\le\ein/(2\eta)$ by \cref{eq:clip-inner-tolerance},
  the statistical and optimization errors together give a KL error of at most $\ein$.
  Averaging over the preceding batches and taking a union bound over $0\le t<T$,
  we obtain, with probability at least $1-\delta$,

  \begin{align}
    \Dkl{\pitp}{\piplust}
    &\le\eta\rbr*{4\vepsopt+\vepsstat(\delta/T)}
    \le\ein
    =\frac1{64\Cresp T^2\log(eT)},
    \qquad 0\le t<T.
    \label{eq:clip-final-kl-error}
  \end{align}

  Fix any policy sequence satisfying \cref{eq:clip-final-kl-error} for the remainder
  of the proof.

  To obtain the reward guarantee, apply \cref{thm:kl-shared-regret} to
  $r(x,y)+\Rmax\in[0,2\Rmax]$.
  Adding this constant preserves the exact targets, KL errors, reward gaps,
  and sets of optimal responses, so the theorem applies to the same policy sequence.
  Jensen's inequality gives
  $\En_{x\sim\mu}\sbr*{\log(1/\piref(\Ystarx\mid{}x))}\le\log\Cresp$.
  Combining this with \cref{eq:clip-final-kl-error} and
  $\sum_{t=0}^{T-1}(t+1)^{-1}\le\log(eT)$ in \cref{thm:kl-shared-regret} yields
  \begin{align}
    \Jstar-J(\pifin)
    &\leq\frac{\log\Cresp}{\eta T}
      +\frac{T\ein}{\eta}
      +\frac{2\sqrt e}{\eta}\sqrt{\Cresp\ein\log(eT)}
    \\
    &\leq\frac{\log\Cresp+1/64+\sqrt e/4}{\eta T}
    \leq\frac{\log(2\Cresp)}{\eta T}
    \leq\veps.
    \label{eq:clip-final-regret-calculation}
  \end{align}
  The second line substitutes $\ein$ and uses $\Cresp\ge1$ and $\log(eT)\ge1$.
  The remaining inequalities use $1/64+\sqrt e/4\le\log2$ and the choice of $T$,
  proving \cref{eq:clip-final-guarantee}.

  The coverage guarantee in \cref{thm:kl-shared-regret} also applies to this
  policy sequence. Taking $c=1/2$ and using the bound $\ein$ on each KL error gives
  \begin{align}
    &\Pr_{x\sim\mu}\rbr*{
      \pit(\Ystarx\mid x)\geq\tfrac12\piref(\Ystarx\mid x)
      \text{ for all }0\leq t\leq T}
    \\
    &\qquad\geq1-(2+\sqrt2)T\sqrt{\Cresp\ein}
      =1-\frac{2+\sqrt2}{8\sqrt{\log(eT)}}.
  \end{align}
  This proves \cref{eq:clip-coverage-guarantee}.

  Finally, each of the $T$ outer iterations draws $N$ responses.
  Substituting $\ein$ from \cref{eq:clip-inner-tolerance} into the choice of $N$
  in \cref{eq:clip-sample-parameters}, and accounting for the ceiling, gives
  \begin{align}
    TN
    &\le T+
    \frac{1024}{3}\eta\Rmax\Cresp e^{2\eta\Rmax}
    \sbr*{1+(1+6\eta\Rmax)e^{1+\eta\Rmax}}
    T^3\log(eT)\log\frac{|\Pi_\Theta|T}{\delta}.
    \label{eq:clip-response-count}
  \end{align}
  In particular, taking $\eta=1/\Rmax$ and using $\veps\le2\Rmax$ yields
  \begin{align}
    T&=O\rbr*{\frac{\Rmax\log(2\Cresp)}{\veps}},
    \\
    TN
    &=O\rbr*{
      \frac{\Cresp\Rmax^3\log^3(2\Cresp)}{\veps^3}
      \log(eT)\log\frac{2|\Pi_\Theta|T}{\delta}
    }.
  \end{align}
  Suppressing logarithmic factors gives \cref{eq:clip-sample-complexity}.

\end{proof}

\subsection{Additional Result: Gradient Ascent for Softmax Linear Policies}
\label{app:clip-inner}

\Cref{cor:clip-global} assumes that each inner loop meets the optimization tolerance
in \cref{eq:clip-pg-erm}, but leaves the optimization procedure abstract.
We now study gradient ascent as a concrete way to implement the inner loop.
\Cref{app:clip-gradient} defines the inner loop for differentiable policies,
using the same batch to estimate the reward and KL gradients.
\Cref{app:clip-softmax} then gives an end-to-end guarantee for these updates
under softmax linear policies and sufficiently wide clipping thresholds.

\subsubsection{Gradient Ascent on the Clipped Objective}
\label{app:clip-gradient}
\label{app:addl-clip-pg}

Fix an outer iteration $t$ and a differentiable policy class $\Pi_\Theta=\cbr*{\pi_\theta}$,
with $\pit=\pi_{\thetat}$.
For the discussion below, assume that the clipped reward term is differentiable
for almost every $(x,y)$ under $\pit$ and that we can differentiate under the expectation.

\para{Gradient computation}
The population objective in \cref{eq:clip-population-objective} has gradient
\begin{align}
  &\nabla\Jtclip(\pi_\theta)
  \\
  &\quad=\En_{\pit}\sbr*{
    w_t^{\pi_\theta}(x,y)r(x,y)\nabla\log\pi_\theta(y\mid{}x)\,
    \one{r(x,y)\ge0,\ w_t^{\pi_\theta}(x,y)<1+\cplus}
  }
  \\
  &\qquad+\En_{\pit}\sbr*{
    w_t^{\pi_\theta}(x,y)r(x,y)\nabla\log\pi_\theta(y\mid{}x)\,
    \one{r(x,y)<0,\ w_t^{\pi_\theta}(x,y)>1-\cminus}
  }
  \\
  &\qquad-\frac1\eta\nabla\Dkl{\pi_\theta}{\pit}.
  \label{eq:clip-gradient}
\end{align}
Clipping sets a response's contribution to the reward gradient to zero when its policy ratio
exceeds $1+\cplus$ for a positive reward or falls below $1-\cminus$ for a negative reward.
It leaves the KL gradient unchanged.

For the inner loop, we estimate both the gradient using the batch
$\cD_t$ drawn as in \cref{alg:clip-pg},
where the gradient estimate is
\begin{align}
  \nabla\hJtclip[t,\mathrm{est}](\pi_\theta)
  &=\frac1N\sum_{i=1}^N
    w_t^{\pi_\theta}(x_i,y_i)r_i\nabla\log\pi_\theta(y_i\mid{}x_i)\,
    \one{r_i\ge0,\ w_t^{\pi_\theta}(x_i,y_i)<1+\cplus}
  \\
  &\qquad+\frac1N\sum_{i=1}^N
    w_t^{\pi_\theta}(x_i,y_i)r_i\nabla\log\pi_\theta(y_i\mid{}x_i)\,
    \one{r_i<0,\ w_t^{\pi_\theta}(x_i,y_i)>1-\cminus}
  \\
  &\qquad-\frac1{\eta N}\sum_{i=1}^N
    w_t^{\pi_\theta}(x_i,y_i)\log w_t^{\pi_\theta}(x_i,y_i)
    \nabla\log\pi_\theta(y_i\mid{}x_i).
  \label{eq:clip-empirical-gradient}
\end{align}
Similar to \ppolitenc, this is the gradient of the empirical version of the objective given by
\begin{align}
  \hJtclip[t,\mathrm{est}](\pi)
  \ldef{}&\frac1N\sum_{i=1}^N\biggl[
  \min\cbr*{
    w_t^\pi(x_i,y_i)r_i,
    [w_t^\pi(x_i,y_i)]_{1-\cminus}^{1+\cplus}r_i
  }
  \\
  &\qquad-\frac1\eta
  \sbr*{w_t^\pi(x_i,y_i)\log w_t^\pi(x_i,y_i)-w_t^\pi(x_i,y_i)+1}\biggr].
  \label{eq:clip-estimated-kl-objective}
\end{align}

\para{Inner optimization loop}
The inner loop initializes at $\thetat$ and reuses $\cD_t$ for $K$ steps
with the gradient in \cref{eq:clip-empirical-gradient},
\begin{align}
  \thetatk[0]&=\thetat,
  &\thetatkp&=\thetatk+\alpha\nabla\hJtclip[t,\mathrm{est}](\pitk),
  &\pitk&=\pi_{\thetatk},
  \qquad 0\le k<K,
  \label{eq:clip-estimated-kl-update}
\end{align}
and returns $\thetatp=\thetatk[K]$ and $\pitp=\pitk[K]$.

\subsubsection{Convergence for Softmax Linear Policies}
\label{app:clip-softmax}

To bound the sample and gradient step complexity for the full algorithm,
we adopt the setting of \cref{cor:pg-main}
with softmax linear policies (\cref{asm:softmax-linear}) and nonnegative rewards $0\le r(x,y)\le\Rmax$.
Note that here the policy class is not finite.

The analysis of \ppolitenc already shows that, with high probability, KL regularization
keeps its inner iterates close to the target $\piplust$,
and bounds the policy ratios as absolute constants throughout training.
Thus, with sufficiently side clipping thresholds of constant magnitude the ratios will remain unclipped,
and we can largely reuse the analysis of \ppolitenc in \cref{app:noclip},
as the clipping in \ppolite leaves every gradient update unchanged.
The following corollary therefore recovers the complexity guarantees of \ppolitenc.

\begin{corollary}
  \label{cor:clip-softmax}
  Suppose that \cref{asm:softmax-linear} holds and $0\le r(x,y)\le\Rmax$.
  Fix an outer iteration $t$ and run \cref{eq:clip-estimated-kl-update}
  with $\eta,\alpha,K,N,\delta$ satisfying the conditions of \cref{thm:noclip-inner}.
  If
  \begin{align}
    1-\cminus\le e^{-5},
    \qquad
    1+\cplus\ge e^5,
    \label{eq:clip-softmax-thresholds}
  \end{align}
  then, with probability at least $1-\delta$, clipping is inactive at every inner iterate and
  \begin{align}
    \Jtr(\piplust)-\Jtr(\pitk[K])
    \le190e^9\Bfeat\Bstar\sbr*{\frac1K+
      \frac1N\log\rbr*{\frac{8[5(1+2N)]^d}{\delta}}}.
    \label{eq:clip-softmax-inner-rate}
  \end{align}

  Consequently, under the assumptions and parameter choices of \cref{cor:noclip-regret},
  replacing its inner updates by \cref{eq:clip-estimated-kl-update} with
  \cref{eq:clip-softmax-thresholds} gives, with probability at least $1-\delta$,
  \begin{align}
    \Jstar-J(\pifin)\le\veps.
    \label{eq:clip-softmax-global-guarantee}
  \end{align}
  On the same event, the policy sequence satisfies
  \begin{align}
    \Pr_{x\sim\mu}\rbr*{
      \pit(\Ystarx\mid x)\geq\tfrac12\piref(\Ystarx\mid x)
      \text{ for all }0\leq t\leq T}
    \geq1-\frac{2+\sqrt2}{8\sqrt{\log(eT)}}.
    \label{eq:clip-softmax-coverage-guarantee}
  \end{align}
  The numbers of gradient steps and sampled responses satisfy the bounds in
  \cref{eq:noclip-total-complexity}.

\end{corollary}

\para{Discussion: Comparison to \cref{cor:clippg-main}}
\Cref{cor:clip-softmax} bounds both the samples and gradient steps needed for convergence,
without assuming that the inner loop satisfies \cref{eq:clip-pg-erm}.
Its guarantee relies on clipping remaining inactive, so it does not establish
an additional benefit from clipping in this regime.
\Cref{cor:clip-global} allows clipping to be active and gives a guarantee for any
inner procedure that meets the empirical optimization tolerance, but does not
bound the number of gradient steps needed to do so.

\begin{proof}[\pfref{cor:clip-softmax}]
  We show that the clipped updates coincide with those of \ppolitenc,
  so its convergence guarantees apply.
  Fix an outer iteration $t$ and condition on the history before drawing $\cD_t$.

  Let $\cbr*{\wt \thetatk}_{k=0}^K$ denote the iterates of \cref{alg:offline-pg} using the same batch and initialization at $\thetat$,
  with all other parameters the same.
  By \cref{lem:noclip-target-realizability}, the parameter
  $\thetatarget=\thetat+\eta\thetastar$ realizes $\piplust$.
  \Cref{thm:noclip-inner} gives, with probability at least $1-\delta$,
  \begin{align}
    \nbr*{\widetilde{\thetatk}-\thetatarget}_2<2\eta\Bstar,
    \qquad 0\le k\le K.
    \label{eq:clip-softmax-localization}
  \end{align}
  Going forward we condition on the event that \cref{eq:clip-softmax-localization} holds.
  Under this event, we next show that clipping remains inactive for all policy iterates
  by comparing each iterate with $\piplust$, which obeys the following policy ratio bound.
  Since $0\le r(x,y)\le\Bfeat\Bstar$, \cref{eq:noclip-exact-update} gives, for every $(x,y)$,
  \begin{align}
    e^{-\eta\Bfeat\Bstar}
    \le\frac{\piplust(y\mid{}x)}{\pit(y\mid{}x)}
    \le e^{\eta\Bfeat\Bstar}.
  \end{align}
  Applying \cref{lem:softmax-policy-ratio} in both directions, with feature diameter
  at most $2\Bfeat$, now gives for $\wt \thetatk$,
  \begin{align}
    \abs*{\log\frac{\pi_{\widetilde{\thetatk}}(y\mid{}x)}{\pit(y\mid{}x)}}
    &\le
    \abs*{\log\frac{\pi_{\widetilde{\thetatk}}(y\mid{}x)}{\piplust(y\mid{}x)}}
    +\abs*{\log\frac{\piplust(y\mid{}x)}{\pit(y\mid{}x)}}
    \\
    &\le2\Bfeat\nbr*{\widetilde{\thetatk}-\thetatarget}_2
      +\eta\Bfeat\Bstar
    <5\eta\Bfeat\Bstar=5.
    \label{eq:clip-softmax-ratio-bound}
  \end{align}
  The clipped updates therefore produce the same inner iterates as \ppolitenc.
  By \cref{eq:clip-softmax-ratio-bound,eq:clip-softmax-thresholds}, clipping leaves
  the gradient unchanged at every $\widetilde{\thetatk}$.
  Thus, starting from the common initialization $\thetat$, induction gives
  $\thetatk=\widetilde{\thetatk}$ for all $0\le k\le K$ for the inner optimization loop.
  Thus, \ppolite also inherits the global optimality guarantee of \ppolitenc.

\end{proof}

\subsection{Comparison with \ppo and \grpo}
\label{app:clip-comparison}
\label{app:ppo-grpo-comparison}

Like \ppo and \grpo, both \ppolitenc and \clippg reuse a batch sampled from $\pit$
throughout the inner policy updates.
The objectives below use our contextual bandit notation and probabilities of complete responses.
They omit token averaging and normalization by the group standard deviation.

\para{\ppo}
For a fixed value estimate $\wh V_t(x)$, the contextual bandit advantage estimate is
$\wh A_t(x,y)\ldef{}r(x,y)-\wh V_t(x)$.
The clipped surrogate of \citet{schulman2017ppo} then takes the form
\begin{align}
  \Jt^{\mathsf{PPO}}(\pi)
  =\En_{\pit}\sbr*{
    \min\cbr*{
      w_t^\pi(x,y)\wh A_t(x,y),
      [w_t^\pi(x,y)]_{1-\cminus}^{1+\cplus}\wh A_t(x,y)
    }
  }.
  \label{eq:clip-ppo-objective}
\end{align}
The usual symmetric clipping interval corresponds to $\cminus=\cplus$.
This surrogate clips the importance weights applied to the advantages;
an explicit KL penalty is a separate design choice.

\para{\grpo}
In place of a value estimate, \grpo uses the mean reward of a group of responses as its baseline
\citep{shao2024deepseekmath}.
For $y^1,\ldots,y^G$ drawn independently from $\pit(\cdot\mid{}x)$, define
\begin{align}
  \bar r_t(x)&\ldef{}\frac1G\sum_{j=1}^G r(x,y^j),
  &\wh A_t(x,y^j)&\ldef{}r(x,y^j)-\bar r_t(x).
\end{align}
To display the choice of reference explicitly, we write a surrogate for complete responses
with a population KL penalty to $\piref$,
\begin{align}
  \Jt^{\mathsf{GRPO}}(\pi)
  ={}&
  \En_{\substack{x\sim\mu,\\y^1,\ldots,y^G\overset{\mathrm{iid}}{\sim}\pit(\cdot\mid{}x)}}
  \sbr*{\frac1G\sum_{j=1}^G
    \min\cbr*{
      w_t^\pi(x,y^j)\wh A_t(x,y^j),
      [w_t^\pi(x,y^j)]_{1-\cminus}^{1+\cplus}\wh A_t(x,y^j)
    }
  }
\\
  &-\frac1\eta\Dkl{\pi}{\piref}.
  \label{eq:clip-grpo-objective}
\end{align}
The original \grpo algorithm normalizes the centered rewards by their group standard deviation
and uses token-level ratios and a sampled KL penalty.
Our simplified objective in \cref{eq:clip-grpo-objective} makes the roles of the baseline
and regularization reference explicit for comparison with \clippg.

\para{Regularization, clipping, and baselines}
\framework resets the regularization reference to $\pit$ at each outer iteration,
and the inner optimization tolerance ensures that $\pitp$ approximates the resulting target $\piplust$.
For \clippg, clipping and the population KL penalty together control the statistical error
without requiring bounded candidate policy ratios (\cref{lem:clip-concentration}).
Clipping modifies the objective; it does not constrain those ratios to lie in the clipping interval.

For a fixed prompt-dependent baseline $\wh V_t(x)$, the unclipped objective satisfies
\begin{align}
  \En_{\pit}\sbr*{w_t^\pi(x,y)\rbr*{r(x,y)-\wh V_t(x)}}
  =J(\pi)-\En_{x\sim\mu}\sbr*{\wh V_t(x)}.
\end{align}
The baseline term is independent of $\pi$, so subtracting it leaves the population gradient unchanged.
With clipping, subtracting a baseline can change the sign of an advantage
and thus which clipping threshold applies, altering the objective and potentially its maximizer.
The result proved here therefore concerns the objective using observed rewards in \cref{eq:clip-population-objective};
extending it to estimated advantages or advantages normalized within each group requires accounting for those changes.

\clearpage
\part{\Cref*{sec:regression}: Squared-Loss Algorithms}\label{part:appendix-squared-loss}

\section{Overview of Kimi-Style Algorithms}\label{app:kimi-shared}
This section characterizes Kimi-style objectives with general baselines and
establishes supporting results for their analysis.
\begin{itemize}
  \item 
  \Cref{app:kimi-general} describes the Kimi-style objective for general baseline functions,
  which is biased relative to the \framework target $\piplust$.
  We then bound the error from this bias and also characterize the objective curvature for any baseline.

  \item 
  \Cref{app:kimi-baseline-discussion} gives an overview of the mean and normalizer baselines,
  and characterizes their individual objective biases and curvature.
  We then summarize our theoretical results in \cref{tab:kimi-baseline-comparison} and compare the sample and gradient step complexities of these two baseline choices.

  \item In \cref{app:kimi-local-supporting-lemmas} we state and prove supporting lemmas for general baselines,
  used for the above characterizations and in our baseline-specific convergence proofs.
  These results can be used to extend our results to other baseline choices.
\end{itemize}

\subsection{General Kimi-Style Objectives}\label{app:kimi-general}
Fix an iteration $t$, a current policy $\pit>0$, and $\eta>0$.
Given a baseline function $\bt:\cX\rightarrow\bbR$ that is independent of the candidate policy $\pi$,
recall that the Kimi objective (\cref{eq:kimi-objective-intro}) is
\begin{align}
  \Ltkimib(\pi)
  =
  \En_{x\sim\mu,\,y\sim\pit(\cdot\mid{}x)}
  \sbr*{
    \rbr*{
      \frac{1}{\eta}\log\frac{\pi(y\mid{}x)}{\pit(y\mid{}x)}
      -r(x,y)+\bt(x)
    }^2
  }.
  \label{eq:kimi-objective-general}
  \label{eq:general-kimi-population-loss}
\end{align}

The normalization function and the exact \framework update are
\begin{align}
  \btstar(x)
  &\ldef{}\frac1\eta\log\En_{y\sim\pit(\cdot\mid{}x)}\sbr*{e^{\eta r(x,y)}},
  \label{eq:kimi-normalizer}
  \\
  \piplust(y\mid{}x)
  &=\pit(y\mid{}x)e^{\eta\rbr*{r(x,y)-\btstar(x)}}.
  \label{eq:kimi-exact-update}
\end{align}

\para{Objective bias}
Kimi-style objectives are generally biased.
In particular, by \cref{lem:kimi-local-objectives}, the loss at the exact \framework update $\piplust$ is
\begin{align}
  \Ltkimib(\piplust)
  =\En_{x\sim\mu}\sbr*{\rbr*{\btstar(x)-\bt(x)}^2}.
\end{align}
Thus $\piplust$ minimizes the objective with zero loss when $\bt=\btstar$,
while a different baseline need not preserve this minimizer.
We show (\cref{lem:kimi-local-objectives}) that every policy $\pi>0$ satisfies
\begin{align}
  \mset(\pi)
  \le
  \Ltkimib(\pi)-\Ltkimib(\piplust)
  +
  \En_{x\sim\mu}
  \sbr*{
    \rbr*{\btstar(x)-\bt(x)}^2
    -
    \rbr*{r(x,\pit)-\bt(x)}_+^2
  }.
  \label{eq:general-kimi-mse-bias}
\end{align}

The correction in this upper bound is independent of the candidate policy.
Consequently, small excess loss guarantees small MSE up to an error controlled
by the baseline. This bound need not be tight for a particular baseline;
\cref{app:kimi-baseline-discussion} gives the comparisons used for the mean and
normalizer baselines.

\para{Curvature}
Under the softmax linear parameterization in \cref{asm:softmax-linear},
\cref{lem:kimi-local-curvature} shows that $\Ltkimib(\pi_\theta)$ is globally convex
whenever $\bt(x)\le r(x,\pit)$ for every $x$.
More generally, fix a radius $R>0$ and suppose that
$\bt(x)-r(x,\pit)\le a$ for every $x$ and some $a\ge0$.
The same lemma gives convexity on the ball $\nbr*{\theta-\thetat}_2\le R$ whenever
$\eta a e^{\Dfeat R}\le1$, where $\Dfeat$ is the feature diameter in
\cref{eq:kimi-feature-diameter}.
Thus the comparison with $\btstar$ controls the objective bias, while the
comparison with $r(x,\pit)$ controls these curvature guarantees.

\subsection{Mean and Normalizer Baselines: Bias, Curvature, and Complexity}

\label{app:kimi-baseline-discussion}

We consider two choices of baseline.
The comparisons below use \cref{asm:softmax-linear} and the notation of
\cref{app:kimi-local-supporting-lemmas}.

\para{Mean baseline (\cref{app:kimi})}
The method \kimipg (\cref{alg:kimi-main}) uses the mean baseline
$\bt(x)=r(x,\pit)$.
For this choice, we write $\Lkimi=\Ltkimib$, so the population objective is
\begin{align}
  \Lkimi(\pi)
  =\En_{x\sim\mu,\,y\sim\pit(\cdot\mid{}x)}
  \sbr*{\rbr*{\frac1\eta\log\frac{\pi(y\mid{}x)}{\pit(y\mid{}x)}-r(x,y)+r(x,\pit)}^2}.
  \label{eq:mean-kimi-population-loss}
\end{align}
The algorithm estimates $r(x,\pit)$ by averaging the rewards of $G$ additional responses
sampled independently of the response used to evaluate the loss.

\begin{itemize}[leftmargin=1.5em]
  \item \emph{Bias.}
  Increasing $G$ does not change the expected baseline $r(x,\pit)$ or its
  difference from $\btstar(x)$.
  The sampled mean is unbiased, and its variance cancels from the expected excess
  loss by \cref{lem:kimi-local-expected-loss}.
  The remaining baseline error satisfies
  $0\le\btstar(x)-r(x,\pit)\le\eta\Dstar^2/8$
  by \cref{eq:mean-kimi-normalization-gap}.
  The excess-loss comparison in \cref{lem:mean-kimi-geometric-loss-comparison} gives
  \begin{align}
    \mset(\pi)
    \le\frac43\rbr*{\Lkimi(\pi)-\Lkimi(\piplust)}
      +\frac{\eta^4\Dstar^6e^{\eta\Dstar}}{48}.
  \end{align}
  The inner convergence theorem requires
  $\eta^4\Dstar^6\le\einsq\le\Dstar^2$
  (\cref{eq:mean-kimi-geometric-inner-accuracy}),
  which implies $\eta\Dstar\le1$ and controls this bias by a constant multiple of $\einsq$.

  \item \emph{Curvature.}
  The population objective $\Lkimi(\pi_\theta)$ is globally convex by
  \cref{lem:mean-kimi-global-curvature}.
  By \cref{lem:kimi-local-regularity} with $A=\Dstar$, it is also
  $2\Dfeat^2/\eta^2$-smooth on the ball $\nbr*{\theta-\thetat}_2\le4\eta\Bstar$
  when $\eta\Dstar\le1/3$.
\end{itemize}

\para{Normalizer baseline (\cref{app:kimi-norm})}
The method \kiminorm (\cref{alg:kimi-norm}) approximates the exact baseline
$\btstar$ in \cref{eq:kimi-normalizer}.
We write $\Lkiminorm=\Ltkimib$ for this exact baseline.
The algorithm uses $G$ independent responses
$y_1,\ldots,y_G\sim\pit(\cdot\mid{}x)$ and sets
\begin{align}
  \hbt(x)
  \ldef{}\frac1\eta\log\rbr*{\frac1G\sum_{j=1}^G e^{\eta r(x,y_j)}}.
  \label{eq:kimi-baselines}
\end{align}
These responses are sampled independently of the response used to evaluate the loss.
The auxiliary population objective $\tLkiminorm$ uses
$\tbt(x)=\En\sbr*{\hbt(x)\mid x}$ as its baseline
(\cref{eq:kimi-common-auxiliary-loss,app:kimi-norm-shared-lemmas}).
By \cref{lem:kimi-local-expected-loss},
it differs from $\En\sbr*{\hLkiminorm}$ only by the baseline variance and hence
has the same gradients and Hessians. These generally differ from those of
$\Lkiminorm$, whose baseline is the exact $\btstar$.

\begin{itemize}[leftmargin=1.5em]
  \item \emph{Bias.}
  Increasing $G$ reduces the estimation bias.
  By \cref{lem:kimi-norm-estimation-bias},
  \begin{align}
    0\le\btstar(x)-\En\sbr*{\hbt(x)\mid x}
    \le\frac{\rbr*{e^{\eta\Dstar}-1}^2}{8\eta G}.
  \end{align}
  The decomposition in \cref{eq:kimi-local-objective-decomposition} gives
  $\mset(\pi)\le\tLkiminorm(\pi)$, and the bias bound above yields
  \begin{align}
    \tLkiminorm(\piplust)
    \le\frac{\rbr*{e^{\eta\Dstar}-1}^4}{64\eta^2G^2}
  \end{align}
  as in \cref{lem:kimi-norm-comparator-loss}.

  \item \emph{Curvature.}
  The expected baseline satisfies $\tbt(x)-r(x,\pit)\le\eta\Dstar^2/8$
  by \cref{lem:kimi-norm-estimation-bias,lem:mean-kimi-baseline-gap}.
  Applying \cref{lem:kimi-local-curvature} with $a=\eta\Dstar^2/8$
  gives convexity of $\tLkiminorm$ on the ball
  $\nbr*{\theta-\thetat}_2\le4\eta\Bstar$ when $\eta\Dstar\le1/3$.
  On this ball, \cref{lem:kimi-local-regularity} with $A=\Dstar$
  also gives $2\Dfeat^2/\eta^2$-smoothness.
\end{itemize}

\para{Comparison of sample and gradient step complexities (\cref{tab:kimi-baseline-comparison})}
The two baselines lead to different ways of controlling approximation error.
The mean baseline reduces bias by decreasing $\eta$, while the normalizer
baseline does so by increasing the number of additional responses $G$.
In particular, \cref{cor:kimi-norm-regret} keeps $\eta=1/(3\Rmax+4\Dstar)$ fixed
and chooses $G$ according to \cref{eq:kimi-regret-sample-parameters} to meet the
regression accuracy $\einsq$ in \cref{eq:kimi-regret-inner-parameters}.
This gives $T=\wt\Theta\rbr*{(\Rmax+\Dstar)/\veps}$ outer iterations.

For fixed problem parameters as $\veps\to0$, the bounds for the normalizer baseline
save a factor of $1/\sqrt{\veps}$ in outer iterations and gradient steps,
while the bound on sampled responses is larger by a factor of $1/\sqrt{\veps}$, up to logarithmic factors.
\Cref{tab:kimi-baseline-comparison} gives the parameter choices and complexities
from \cref{cor:mean-kimi-geometric-regret,cor:kimi-norm-regret}.
Since the additional responses are used only to form the baseline,
they can be collected in parallel and are not needed for gradient computation.

\begin{table}[H]
  \centering
  \begin{tabular}{lcc}
    \toprule
    & \makecell{\kimipg (\cref{cor:mean-kimi-geometric-regret})}
    & \kiminorm (\cref{cor:kimi-norm-regret}) \\
    \midrule
    Regularization $\eta$
    & $\wt O\rbr*{\frac{\sqrt{\veps}}{\Cresp^{1/4}\Dstar^{3/2}}}$
    & $1/(3\Rmax+4\Dstar)$ \\
    Inner learning rate $\alpha$
    & \makecell{$\frac{\eta^2}{\Dfeat^2(2+\eta\Dstar)}$}
    & \makecell{$\eta^2/(8\Dfeat^2)$} \\
    Additional responses per prompt $G$
    & $1$
    & $\wt\Theta\rbr*{\frac{\sqrt{\Cresp}\Dstar^2}{(\Rmax+\Dstar)\veps}}$ \\
    Outer iterations $T$
    & $\wt O\rbr*{\frac{\Cresp^{1/4}\Dstar^{3/2}}{\veps^{3/2}}}$
    & $\wt\Theta\rbr*{(\Rmax+\Dstar)/\veps}$ \\
    Total responses
    & $\wt O\rbr*{\frac{\Cresp^{5/4}\Dstar^{7/2}\rbr*{d+\log(1/\delta)}}{\veps^{7/2}}}$
    & $\wt O\rbr*{\frac{\Cresp^{3/2}\Dstar^4\rbr*{d+\log(1/\delta)}}{\veps^4}}$ \\
    Gradient steps $KT$
    & $\wt O\rbr*{\frac{\Cresp^{5/4}\Dstar^{7/2}}{\veps^{7/2}}}$
    & $\wt O\rbr*{\frac{\Cresp\Dstar^2(\Rmax+\Dstar)}{\veps^3}}$ \\
    \bottomrule
  \end{tabular}
  \caption{Parameter and complexity bounds as $\veps\to0$ with problem parameters fixed,
  from \cref{cor:mean-kimi-geometric-regret,cor:kimi-norm-regret}.
  Total responses are $2NT$ for \kimipg and $N(G+1)T$ for \kiminorm.
}
  \label{tab:kimi-baseline-comparison}

\end{table}

\subsection{Supporting Lemmas for a General Baseline}
\label{app:kimi-local-supporting-lemmas}

\para{Notation}
Fix an outer iteration $t$, a current policy $\pit>0$, and $\eta>0$.
Let $r:\cX\times\cY\to[0,\Rmax]$ and let $\bt:\cX\to\bbR$ be the
baseline being estimated. For $G\ge1$, form $\hbt(x)$ from independent
responses $y_1',\ldots,y_G'\sim\pit(\cdot\mid{}x)$.
Both $\bt$ and the law of $\hbt(x)$ are independent of the candidate policy.
The conditional mean and variance of the estimate are
\begin{align}
  \tbt(x)
  &\ldef{}\En_{y_1',\ldots,y_G'\sim\pit(\cdot\mid{}x)}
    \sbr*{\hbt(x)},
  \\
  \rbr*{\sigmatG(x)}^2
  &\ldef{}\Var_{y_1',\ldots,y_G'\sim\pit(\cdot\mid{}x)}
    \sbr*{\hbt(x)}.
  \label{eq:kimi-common-baseline-moments}
\end{align}
We leave the additional response samples implicit in $\hbt$.
The population objective with baseline $\bt$ is
\begin{align}
  \Ltkimib(\pi)
  \ldef{}\En_{x\sim\mu,\,y\sim\pit(\cdot\mid{}x)}\sbr*{\rbr*{
    \frac1\eta\log\frac{\pi(y\mid{}x)}{\pit(y\mid{}x)}
    -r(x,y)+\bt(x)
  }^2}.
  \label{eq:kimi-common-population-loss}
\end{align}
We abbreviate this joint expectation by $\En_{\pit}$.
Draw $N$ independent triples $(x_i,y_i,\hbt(x_i))$, where
$x_i\sim\mu$, $y_i\sim\pit(\cdot\mid{}x_i)$, and each $\hbt(x_i)$ is
formed from $G$ additional responses independently of $y_i$ given $x_i$.
Define the empirical and auxiliary population objectives by
\begin{align}
  \hLtkimib(\pi)
  &\ldef{}\frac1N\sum_{i=1}^N\rbr*{
    \frac1\eta\log\frac{\pi(y_i\mid{}x_i)}{\pit(y_i\mid{}x_i)}
    -r(x_i,y_i)+\hbt(x_i)
  }^2,
  \label{eq:kimi-common-empirical-loss}
  \\
  \tLtkimib(\pi)
  &\ldef{}\En_{\pit}\sbr*{\rbr*{
    \frac1\eta\log\frac{\pi(y\mid{}x)}{\pit(y\mid{}x)}
    -r(x,y)+\tbt(x)
  }^2}.
  \label{eq:kimi-common-auxiliary-loss}
\end{align}
The superscript $b$ identifies the baseline being estimated in all three
objectives, while the auxiliary objective uses $\tbt$.

Recall from \cref{asm:softmax-linear} that
$\Bfeat\ldef{}\max_{x,y}\nbr*{\phi(x,y)}_2$ and $\Bstar\ldef{}\nbr*{\thetastar}_2$.
We also use the feature diameter and the resulting bound on the reward range,
\begin{align}
  \Dfeat&\ldef{}\max_{x,y,y'}\nbr*{\phi(x,y)-\phi(x,y')}_2,
  &\Dstar&\ldef{}\Dfeat\Bstar.
  \label{eq:kimi-feature-diameter}
\end{align}
Thus $\Dfeat\le2\Bfeat$ and $\Dstar\le2\Bfeat\Bstar$.
Under the linear reward assumption in \cref{asm:softmax-linear},
\begin{align}
  \abs*{r(x,y)-r(x,y')}\le\Dstar,
  \qquad\forall x\in\cX,\ y,y'\in\cY.
  \label{eq:kimi-reward-range}
\end{align}
We retain $\Rmax$ and $\Dstar$ separately in the appendix bounds.
The main-body bounds use $B\ldef{}\Bfeat\Bstar$.
Under linear rewards, we can take $\Rmax\le B$, and $\Dstar\le2B$.
Throughout the Kimi analyses, $\einsq$ denotes a local regression accuracy,
while $\ein$ retains its meaning as the KL tolerance in \cref{eq:framework-kl-tolerance}.

For a positive policy $\pi$ write its centered logits and residual as
\begin{align}
  l_{t,\eta}^\pi(x,y)
  &\ldef{}\frac1\eta\log\frac{\pi(y\mid{}x)}{\pit(y\mid{}x)}
  -\En_{y'\sim\pit(\cdot\mid{}x)}\sbr*{\frac1\eta\log\frac{\pi(y'\mid{}x)}{\pit(y'\mid{}x)}},
  \\
  s_t^\pi(x,y)
  &\ldef{}l_{t,\eta}^\pi(x,y)-\rbr*{r(x,y)-r(x,\pit)}.
  \label{eq:kimi-residual}
\end{align}
The conditional and average mean-squared errors are
\begin{align}
  \mset(\pi;x)&\ldef{}\En_{y\sim\pit(\cdot\mid{}x)}\sbr*{s_t^\pi(x,y)^2},
  &\mset(\pi)&\ldef{}\En_{x\sim\mu}\sbr*{\mset(\pi;x)}.
  \label{eq:kimi-mse}
\end{align}

For the feature map in \cref{asm:softmax-linear}, write
\begin{align}
  \Sigmat\ldef{}\En_{x\sim\mu}\sbr*{
    \operatorname{Cov}_{y\sim\pit(\cdot\mid{}x)}\sbr*{\phi(x,y)}}.
  \label{eq:kimi-local-feature-covariance}
\end{align}
Under \cref{asm:softmax-linear}, the centered residual $s_t^{\pi_\theta}$ has
the representation in \cref{eq:mse-accum-linear-residual} with $\thetatp$ replaced by $\theta$.

\begin{lemma}
  \label{lem:kimi-local-objectives}
  For any baseline $\bt$, every policy $\pi>0$ satisfies
  \begin{align}
    \mset(\pi)\le\Ltkimib(\pi)
    &=\mset(\pi)
      +\En_{x\sim\mu}\sbr*{\rbr*{
        r(x,\pit)-\bt(x)
        +\frac1\eta\Dkl{\pit(\cdot\mid{}x)}{\pi(\cdot\mid{}x)}
      }^2},
    \label{eq:kimi-local-objective-decomposition}
    \\
    \Ltkimib(\piplust)
    &=\En_{x\sim\mu}\sbr*{\rbr*{\btstar(x)-\bt(x)}^2}.
    \label{eq:kimi-local-comparator-loss}
  \end{align}
  The same identities hold for $\tLtkimib$ with $\bt$ replaced by $\tbt$.
\end{lemma}

\begin{proof}[\pfref{lem:kimi-local-objectives}]
  The definition of $s_t^\pi$ in \cref{eq:kimi-residual} gives
  \begin{align}
    \frac1\eta\log\frac{\pi(y\mid{}x)}{\pit(y\mid{}x)}-r(x,y)+\bt(x)
    =s_t^\pi(x,y)
      -\rbr*{r(x,\pit)-\bt(x)
        +\frac1\eta\Dkl{\pit(\cdot\mid{}x)}{\pi(\cdot\mid{}x)}}.
  \end{align}
  Substituting into the definition of $\Ltkimib$ (\cref{eq:kimi-common-population-loss}) gives
  \begin{align}
    \Ltkimib(\pi)
    &=\En_{\pit}\sbr*{s_t^\pi(x,y)^2}
      -2\En_{x\sim\mu}\sbr*{\rbr*{
        r(x,\pit)-\bt(x)
        +\frac1\eta\Dkl{\pit(\cdot\mid{}x)}{\pi(\cdot\mid{}x)}}
        \En_{y\sim\pit(\cdot\mid{}x)}[s_t^\pi(x,y)]}
    \\
    &\quad+\En_{x\sim\mu}\sbr*{\rbr*{
      r(x,\pit)-\bt(x)
      +\frac1\eta\Dkl{\pit(\cdot\mid{}x)}{\pi(\cdot\mid{}x)}}^2}
    \\
    &=\mset(\pi)+\En_{x\sim\mu}\sbr*{\rbr*{
      r(x,\pit)-\bt(x)
      +\frac1\eta\Dkl{\pit(\cdot\mid{}x)}{\pi(\cdot\mid{}x)}}^2}
      \ge\mset(\pi).
  \end{align}
  The second equality uses
  $\En_{y\sim\pit(\cdot\mid{}x)}[s_t^\pi(x,y)]=0$.
  This proves \cref{eq:kimi-local-objective-decomposition}.
  For $\piplust$, the exact update in \cref{eq:kimi-exact-update} gives
  \begin{align}
    \frac1\eta\log\frac{\piplust(y\mid{}x)}{\pit(y\mid{}x)}-r(x,y)+\bt(x)
    =\bt(x)-\btstar(x).
  \end{align}
  Substituting this into $\Ltkimib$ proves \cref{eq:kimi-local-comparator-loss}.
  Finally, the result for $\tLtkimib$ follows by substituting $\bt = \tbt$.
\end{proof}

\begin{lemma}
  \label{lem:kimi-local-expected-loss}
  For any $G\ge1$, let $\hLtkimib$ (\cref{eq:kimi-common-empirical-loss})
  be computed from $N$ independent triples $(x_i,y_i,\hbt(x_i))$, with
  $\hbt(x_i)$ formed from responses $y'_{i,1},\ldots,y'_{i,G}$.
  For every policy $\pi>0$ fixed independently of this batch,
  \begin{align}
    \En_{\substack{x_i\sim\mu,\ 1\le i\le N\\
      y_i,y'_{i,1},\ldots,y'_{i,G}\sim\pit(\cdot\mid{}x_i)}}
      \sbr*{\hLtkimib(\pi)}
    =\tLtkimib(\pi)
      +\En_{x\sim\mu}\sbr*{\rbr*{\sigmatG(x)}^2}.
    \label{eq:kimi-local-expected-loss}
  \end{align}
\end{lemma}

\begin{proof}[\pfref{lem:kimi-local-expected-loss}]
  Since $\hbt(x_i)$ is independent of $y_i$ conditional on $x_i$,
  its conditional mean and variance given $(x_i,y_i)$ are
  $\tbt(x_i)$ and $(\sigmatG(x_i))^2$
  (\cref{eq:kimi-common-baseline-moments}).
  Thus
  \begin{align}
    &\En_{\substack{x_i\sim\mu\\
      y_i,y'_{i,1},\ldots,y'_{i,G}\sim\pit(\cdot\mid{}x_i)}}
      \sbr*{\rbr*{
        \frac1\eta\log\frac{\pi(y_i\mid{}x_i)}{\pit(y_i\mid{}x_i)}
        -r(x_i,y_i)+\hbt(x_i)}^2}
    \\
    &\quad=\En_{x_i\sim\mu,\,y_i\sim\pit(\cdot\mid{}x_i)}\sbr*{
      \rbr*{\frac1\eta\log\frac{\pi(y_i\mid{}x_i)}{\pit(y_i\mid{}x_i)}
        -r(x_i,y_i)+\tbt(x_i)}^2}
      +\En_{x_i\sim\mu}\sbr*{\rbr*{\sigmatG(x_i)}^2}.
  \end{align}
  Averaging over $i$ proves the result by the definitions of
  $\hLtkimib$ (\cref{eq:kimi-common-empirical-loss}) and
  $\tLtkimib$ (\cref{eq:kimi-common-auxiliary-loss}).
\end{proof}

\begin{lemma}
  \label{lem:kimi-local-policy-ratio-second-moment}
  For the softmax linear policy class in \cref{asm:softmax-linear}, fix a parameter radius $R>0$.
  Every $\theta$ with $\nbr*{\theta-\thetat}_2\le R$ satisfies
  \begin{align}
    \En_{y\sim\pit(\cdot\mid{}x)}\sbr*{
      \rbr*{\frac{\pi_\theta(y\mid{}x)}{\pit(y\mid{}x)}-1}^2}
    \le\frac{\Dfeat^2R^2e^{\Dfeat R}}4,
    \qquad
    \forall x \in \cX.
    \label{eq:kimi-local-policy-ratio-second-moment}
  \end{align}
\end{lemma}

\begin{proof}[\pfref{lem:kimi-local-policy-ratio-second-moment}]
  Fix $x\in\cX$ and $\theta$ with $\nbr*{\theta-\thetat}_2\le R$.
  Under the softmax linear parameterization,
  \begin{align}
    \frac{\pi_\theta(y\mid{}x)/\pit(y\mid{}x)}
      {\pi_\theta(y'\mid{}x)/\pit(y'\mid{}x)}
    =e^{\inner*{\phi(x,y)-\phi(x,y'),\theta-\thetat}}
    \le e^{\Dfeat R},
    \qquad y,y'\in\cY.
  \end{align}
  Let $m=\inf_y\pi_\theta(y\mid{}x)/\pit(y\mid{}x)>0$.
  The preceding inequality gives
  $m\le\pi_\theta(y\mid{}x)/\pit(y\mid{}x)\le e^{\Dfeat R}m$, hence
  \begin{align}
    \rbr*{\frac{\pi_\theta(y\mid{}x)}{\pit(y\mid{}x)}}^2
    \le (1+e^{\Dfeat R})m\frac{\pi_\theta(y\mid{}x)}{\pit(y\mid{}x)}
      -e^{\Dfeat R}m^2.
  \end{align}
  Since $\En_{y\sim\pit(\cdot\mid{}x)}[\pi_\theta(y\mid{}x)/\pit(y\mid{}x)]=1$,
  \begin{align}
    \En_{y\sim\pit(\cdot\mid{}x)}\sbr*{
      \rbr*{\frac{\pi_\theta(y\mid{}x)}{\pit(y\mid{}x)}-1}^2}
    &=\En_{y\sim\pit(\cdot\mid{}x)}\sbr*{
      \rbr*{\frac{\pi_\theta(y\mid{}x)}{\pit(y\mid{}x)}}^2}-1
    \\
    &\le(1+e^{\Dfeat R})m-e^{\Dfeat R}m^2-1
      \le\frac{(e^{\Dfeat R}-1)^2}{4e^{\Dfeat R}}
      \le\frac{\Dfeat^2R^2e^{\Dfeat R}}4.
  \end{align}
  The last inequality uses $e^a-1\le ae^a$ for $a\ge0$.
\end{proof}

\begin{lemma}
  \label{lem:kimi-local-curvature}
  Suppose that $\pi_\theta$ has the softmax linear parameterization in
  \cref{asm:softmax-linear}.
  Fix $\eta,R>0$ and $a\ge0$ such that $\bt(x)-r(x,\pit)\le a$ for every $x$.
  Every $\theta$ with $\nbr*{\theta-\thetat}_2\le R$ satisfies
  \begin{align}
    \nabla^2\Ltkimib(\pi_\theta)
    \succeq\frac2{\eta^2}\rbr*{1-\eta a e^{\Dfeat R}}\Sigmat.
    \label{eq:kimi-local-curvature}
  \end{align}
  In particular, if $\bt(x)\le r(x,\pit)$ for every $x$, then
  $\Ltkimib(\pi_\theta)$ is globally convex and
  \begin{align}
    \nabla^2\Ltkimib(\pi_\theta)
    \succeq\frac2{\eta^2}\Sigmat,
    \qquad \theta\in\bbR^d.
    \label{eq:kimi-global-curvature}
  \end{align}
  The same statements hold for $\tLtkimib$ with $\bt$ replaced by $\tbt$.
\end{lemma}

\begin{proof}[\pfref{lem:kimi-local-curvature}]
  Fix $\theta$ with $\nbr*{\theta-\thetat}_2\le R$.
  We differentiate the decomposition of $\Ltkimib$ in
  \cref{eq:kimi-local-objective-decomposition} of \cref{lem:kimi-local-objectives}.
  Write
  \begin{align}
    q_\theta(x)\ldef{}r(x,\pit)-\bt(x)
      +\frac1\eta\Dkl{\pit(\cdot\mid{}x)}{\pi_\theta(\cdot\mid{}x)}.
  \end{align}
  Then $\Ltkimib(\pi_\theta)=\mset(\pi_\theta)+\En_{x\sim\mu}[q_\theta(x)^2]$.

  Using $\nabla\log\pi_\theta(y\mid{}x)
    =\phi(x,y)-\En_{y'\sim\pi_\theta(\cdot\mid{}x)}[\phi(x,y')]$
  (\cref{lem:softmax-score-identity}) in the definition of $s_t^{\pi_\theta}$
  (\cref{eq:kimi-residual}) gives
  \begin{align}
    \nabla s_t^{\pi_\theta}(x,y)
    &=\frac1\eta\rbr*{\nabla\log\pi_\theta(y\mid{}x)
      -\En_{y'\sim\pit(\cdot\mid{}x)}[\nabla\log\pi_\theta(y'\mid{}x)]}
    \\
    &=\frac1\eta\rbr*{\phi(x,y)
      -\En_{y'\sim\pit(\cdot\mid{}x)}[\phi(x,y')]}.
  \end{align}
  This gradient is independent of $\theta$, so
  \begin{align}
    \nabla^2\mset(\pi_\theta)
    &=2\En_{x\sim\mu,\,y\sim\pit(\cdot\mid{}x)}\sbr*{
      \nabla s_t^{\pi_\theta}(x,y)
      \nabla s_t^{\pi_\theta}(x,y)^\top}
      =\frac2{\eta^2}\Sigmat.
  \end{align}
  Also, $-\nabla^2\log\pi_\theta(y\mid{}x)
    =\operatorname{Cov}_{y'\sim\pi_\theta(\cdot\mid{}x)}[\phi(x,y')]$
  (\cref{lem:softmax-log-hessian}), so
  $\nabla^2q_\theta(x)
    =\eta^{-1}\operatorname{Cov}_{y\sim\pi_\theta(\cdot\mid{}x)}[\phi(x,y)]$.
  Hence
  \begin{align}
    \nabla^2\Ltkimib(\pi_\theta)
    &=\frac2{\eta^2}\Sigmat
      +2\En_{x\sim\mu}\sbr*{
        \nabla q_\theta(x)\nabla q_\theta(x)^\top
        +\frac{q_\theta(x)}\eta
          \operatorname{Cov}_{y\sim\pi_\theta(\cdot\mid{}x)}[\phi(x,y)]}
    \\
    &\succeq\frac2{\eta^2}\Sigmat
      -\frac{2a}\eta\En_{x\sim\mu}\sbr*{
        \operatorname{Cov}_{y\sim\pi_\theta(\cdot\mid{}x)}[\phi(x,y)]}.
    \label{eq:proof:kimi-baseline-curvature-decomposition}
  \end{align}
  The inequality uses
  $q_\theta(x)\ge r(x,\pit)-\bt(x)\ge-a$ and
  $\nabla q_\theta(x)\nabla q_\theta(x)^\top\succeq0$.

  We now bound the second term in
  \cref{eq:proof:kimi-baseline-curvature-decomposition} using
  $\pi_\theta(y\mid{}x)\le e^{\Dfeat R}\pit(y\mid{}x)$
  (\cref{lem:softmax-policy-ratio}). For every $h\in\bbR^d$,
  \begin{align}
    \Var_{y\sim\pi_\theta(\cdot\mid{}x)}[h^\top\phi(x,y)]
    &\le\En_{y\sim\pi_\theta(\cdot\mid{}x)}\sbr*{\rbr*{
      h^\top\phi(x,y)-\En_{y'\sim\pit(\cdot\mid{}x)}[h^\top\phi(x,y')]}^2}
    \\
    &\le e^{\Dfeat R}\Var_{y\sim\pit(\cdot\mid{}x)}[h^\top\phi(x,y)].
  \end{align}
  The first inequality uses $\Var(Z)\le\En[(Z-c)^2]$ for any constant $c$.
  Thus
  $\En_{x\sim\mu}[\operatorname{Cov}_{y\sim\pi_\theta(\cdot\mid{}x)}[\phi(x,y)]]
    \preceq e^{\Dfeat R}\Sigmat$.
  Substituting into \cref{eq:proof:kimi-baseline-curvature-decomposition} gives
  \begin{align}
    \nabla^2\Ltkimib(\pi_\theta)
    \succeq\frac2{\eta^2}\Sigmat
      -\frac{2a e^{\Dfeat R}}\eta\Sigmat
    =\frac2{\eta^2}\rbr*{1-\eta a e^{\Dfeat R}}\Sigmat.
  \end{align}
  If $\bt(x)\le r(x,\pit)$ for every $x$, take $a=0$.
  The bound is then independent of $R$, so it holds for every $\theta$,
  and $\Sigmat\succeq0$ gives global convexity.
  Substituting $\tbt$ for $\bt$ proves the statements for $\tLtkimib$.
\end{proof}

\begin{lemma}
  \label{lem:kimi-local-regularity}
  Suppose that $\pi_\theta$ has the softmax linear parameterization in
  \cref{asm:softmax-linear}.
  Fix $\eta,R>0$ and $A\ge0$ such that
  $\abs*{\bt(x)-r(x,y)}\le A$ for all $x,y$.
  Every $\theta$ with $\nbr*{\theta-\thetat}_2\le R$ satisfies
  \begin{align}
    \nbr*{\nabla\Ltkimib(\pi_\theta)}_2
    &\le\frac{2\Dfeat}{\eta^2}\rbr*{\Dfeat R+\eta A},
    \label{eq:kimi-local-gradient-bound}
    \\
    \nbr*{\nabla^2\Ltkimib(\pi_\theta)}_{\op}
    &\le\frac{\Dfeat^2}{\eta^2}
      \max\cbr*{2,1+\frac{\Dfeat R+\eta A}{2}}.
    \label{eq:kimi-local-smoothness}
  \end{align}
  The same statement holds for $\hLtkimib$ with $\bt$ replaced by $\hbt$,
  and for $\tLtkimib$ with $\bt$ replaced by $\tbt$.
\end{lemma}

\begin{proof}[\pfref{lem:kimi-local-regularity}]
  Fix $\theta$ with $\nbr*{\theta-\thetat}_2\le R$ and $x,y$, and write
  \begin{align}
    q_\theta(x,y)
    \ldef{}\frac1\eta\log\frac{\pi_\theta(y\mid{}x)}{\pit(y\mid{}x)}
      -r(x,y)+\bt(x).
  \end{align}
  We use $\abs{\log(\pi_\theta(y\mid{}x)/\pit(y\mid{}x))}\le\Dfeat R$
  (\cref{lem:softmax-policy-ratio}) and
  $\nbr{\nabla\log\pi_\theta(y\mid{}x)}_2\le\Dfeat$
  (\cref{lem:softmax-score-bound}) to obtain
  \begin{align}
    \abs{q_\theta(x,y)}
    &\le\frac1\eta\abs*{\log\frac{\pi_\theta(y\mid{}x)}{\pit(y\mid{}x)}}
      +\abs{\bt(x)-r(x,y)}\le A+\frac{\Dfeat R}{\eta},
    \label{eq:proof:kimi-local-residual-bounds}
    \\
    \nbr{\nabla q_\theta(x,y)}_2
    &=\frac1\eta\nbr{\nabla\log\pi_\theta(y\mid{}x)}_2
      \le\frac{\Dfeat}{\eta}.
    \label{eq:proof:kimi-local-residual-gradient}
  \end{align}
  Consequently,
  \begin{align}
    \nbr*{\nabla\rbr*{q_\theta(x,y)^2}}_2
    =2\abs{q_\theta(x,y)}\nbr{\nabla q_\theta(x,y)}_2
    \le\frac{2\Dfeat}{\eta^2}\rbr*{\Dfeat R+\eta A}.
  \end{align}

  To bound the Hessian, fix $\nbr{h}_2=1$.
  Using $-\nabla^2\log\pi_\theta(y\mid{}x)
    =\operatorname{Cov}_{y'\sim\pi_\theta(\cdot\mid{}x)}[\phi(x,y')]$
  (\cref{lem:softmax-log-hessian}), differentiation gives
  \begin{align}
    h^\top\nabla^2\rbr*{q_\theta(x,y)^2}h
    =\frac2{\eta^2}\sbr*{
      \inner*{\nabla\log\pi_\theta(y\mid{}x),h}^2
      -\eta q_\theta(x,y)
        \Var_{y'\sim\pi_\theta(\cdot\mid{}x)}[h^\top\phi(x,y')]}.
    \label{eq:proof:kimi-local-sample-hessian}
  \end{align}
  To control these two terms together, let $c$ be the midpoint of the range
  of $h^\top\phi(x,\cdot)$, so
  \begin{align}
    c\ldef{}\frac12\rbr*{
      \inf_{y'}h^\top\phi(x,y')+\sup_{y'}h^\top\phi(x,y')},
    \qquad
    \abs{h^\top\phi(x,y')-c}\le\frac{\Dfeat}{2},
    \quad y'\in\cY.
  \end{align}
  It follows that
  \begin{align}
    \Var_{y'\sim\pi_\theta(\cdot\mid{}x)}[h^\top\phi(x,y')]
    &=\En_{y'\sim\pi_\theta(\cdot\mid{}x)}\sbr*{(h^\top\phi(x,y')-c)^2}
      -\rbr*{\En_{y'\sim\pi_\theta(\cdot\mid{}x)}[h^\top\phi(x,y')]-c}^2
    \\
    &\le\frac{\Dfeat^2}{4}
      -\rbr*{\En_{y'\sim\pi_\theta(\cdot\mid{}x)}[h^\top\phi(x,y')]-c}^2
      \le\frac{\Dfeat^2}{4}.
    \label{eq:proof:kimi-local-projection-variance}
  \end{align}
  Since $\inner*{\nabla\log\pi_\theta(y\mid{}x),h}
    =h^\top\phi(x,y)-\En_{y'\sim\pi_\theta(\cdot\mid{}x)}[h^\top\phi(x,y')]$
  (\cref{lem:softmax-score-identity}), we also obtain
  \begin{align}
    &\inner*{\nabla\log\pi_\theta(y\mid{}x),h}^2
      +2\Var_{y'\sim\pi_\theta(\cdot\mid{}x)}[h^\top\phi(x,y')]
    \\
    &\quad\le\rbr*{\frac{\Dfeat}{2}
      +\abs*{\En_{y'\sim\pi_\theta(\cdot\mid{}x)}[h^\top\phi(x,y')]-c}}^2
      +\frac{\Dfeat^2}{2}
      -2\rbr*{\En_{y'\sim\pi_\theta(\cdot\mid{}x)}[h^\top\phi(x,y')]-c}^2
    \\
    &\quad=\Dfeat^2-\rbr*{\frac{\Dfeat}{2}
      -\abs*{\En_{y'\sim\pi_\theta(\cdot\mid{}x)}[h^\top\phi(x,y')]-c}}^2
      \le\Dfeat^2.
    \label{eq:proof:kimi-local-score-variance}
  \end{align}
  Returning to \cref{eq:proof:kimi-local-sample-hessian}, we obtain
  \begin{align}
    \abs*{h^\top\nabla^2\rbr*{q_\theta(x,y)^2}h}
    &\le\frac2{\eta^2}\rbr*{
      \inner*{\nabla\log\pi_\theta(y\mid{}x),h}^2
      +\eta\abs{q_\theta(x,y)}
        \Var_{y'\sim\pi_\theta(\cdot\mid{}x)}[h^\top\phi(x,y')]}
    \\
    &\le\frac2{\eta^2}\rbr*{\Dfeat^2
      +\rbr*{\eta\abs{q_\theta(x,y)}-2}_+
        \Var_{y'\sim\pi_\theta(\cdot\mid{}x)}[h^\top\phi(x,y')]}
    \\
    &\le\frac{\Dfeat^2}{\eta^2}
      \max\cbr*{2,1+\frac{\Dfeat R+\eta A}{2}}.
  \end{align}
  The second inequality uses \cref{eq:proof:kimi-local-score-variance}.
  The last uses the variance bound \cref{eq:proof:kimi-local-projection-variance}
  and $\abs{q_\theta(x,y)}\le A+\Dfeat R/\eta$ from
  \cref{eq:proof:kimi-local-residual-bounds}.
  Taking the supremum over $\nbr{h}_2=1$ gives
  \begin{align}
    \nbr*{\nabla^2\rbr*{q_\theta(x,y)^2}}_{\op}
    =\sup_{\nbr{h}_2=1}\abs*{h^\top\nabla^2\rbr*{q_\theta(x,y)^2}h}
    \le\frac{\Dfeat^2}{\eta^2}
      \max\cbr*{2,1+\frac{\Dfeat R+\eta A}{2}}.
  \end{align}

  The definition of $\Ltkimib$ (\cref{eq:kimi-common-population-loss}) now gives
  \begin{align}
    \nbr*{\nabla\Ltkimib(\pi_\theta)}_2
    &\le\En_{x\sim\mu,\,y\sim\pit(\cdot\mid{}x)}\sbr*{
      \nbr*{\nabla\rbr*{q_\theta(x,y)^2}}_2}
      \le\frac{2\Dfeat}{\eta^2}\rbr*{\Dfeat R+\eta A},
    \\
    \nbr*{\nabla^2\Ltkimib(\pi_\theta)}_{\op}
    &\le\En_{x\sim\mu,\,y\sim\pit(\cdot\mid{}x)}\sbr*{
      \nbr*{\nabla^2\rbr*{q_\theta(x,y)^2}}_{\op}}
      \le\frac{\Dfeat^2}{\eta^2}
        \max\cbr*{2,1+\frac{\Dfeat R+\eta A}{2}}.
  \end{align}
  This proves \cref{eq:kimi-local-gradient-bound,eq:kimi-local-smoothness}.
  For $\hLtkimib$ (\cref{eq:kimi-common-empirical-loss}), replace $\bt$ by
  $\hbt$ in $q_\theta$. The same bounds hold at each sampled $(x_i,y_i)$, so
  \begin{align}
    \nbr*{\nabla\hLtkimib(\pi_\theta)}_2
    &\le\frac1N\sum_{i=1}^N\nbr*{\nabla\rbr*{q_\theta(x_i,y_i)^2}}_2
      \le\frac{2\Dfeat}{\eta^2}\rbr*{\Dfeat R+\eta A},
    \\
    \nbr*{\nabla^2\hLtkimib(\pi_\theta)}_{\op}
    &\le\frac1N\sum_{i=1}^N\nbr*{\nabla^2\rbr*{q_\theta(x_i,y_i)^2}}_{\op}
      \le\frac{\Dfeat^2}{\eta^2}
        \max\cbr*{2,1+\frac{\Dfeat R+\eta A}{2}}.
  \end{align}
  The bounds for $\tLtkimib$ (\cref{eq:kimi-common-auxiliary-loss}) follow by
  substituting $\bt=\tbt$ in the population calculation.
  If $\abs{\hbt(x)-r(x,y)}\le A$, then
  \begin{align}
    \abs{\tbt(x)-r(x,y)}
    \le\En_{y_1',\ldots,y_G'\sim\pit(\cdot\mid{}x)}[\abs{\hbt(x)-r(x,y)}]
    \le A.
  \end{align}
\end{proof}

\begin{lemma}
  \label{lem:kimi-local-concentration}
  Suppose that \cref{asm:softmax-linear} holds.
  Let $A\ge0$ satisfy $\abs*{\hbt(x)-r(x,y)}\le A$ for all $x,y$.
  Fix $\eta,R>0$, an integer $N\ge1$, and $\delta\in(0,1)$, with $R\ge\eta\Bstar$.
  Define
  \begin{align}
    \zeta_N(\delta;R,A)
    \ldef{}\frac{160e^{\Dfeat R}}{N}
      \rbr*{A+\frac{\Dfeat R}{\eta}}^2
      \rbr*{d\log(1+2N)+\log\frac4\delta}.
    \label{eq:kimi-local-statistical-error}
  \end{align}
  With probability at least $1-\delta$, simultaneously for every
  $\theta$ with $\nbr*{\theta-\thetat}_2\le R$,
  \begin{align}
    \abs*{\inner*{
      \nabla\hLtkimib(\pi_\theta)
      -\nabla\tLtkimib(\pi_\theta),
      \theta-\thetat-\eta\thetastar}}
    &\le\frac1{64}\mset(\pi_\theta)+\zeta_N(\delta;R,A),
    \label{eq:kimi-local-pairing-concentration}
    \\
    \abs*{
      \hLtkimib(\pi_\theta)-\hLtkimib(\piplust)
      -\tLtkimib(\pi_\theta)+\tLtkimib(\piplust)}
    &\le\frac1{64}\mset(\pi_\theta)+\zeta_N(\delta;R,A).
    \label{eq:kimi-local-excess-concentration}
  \end{align}
\end{lemma}

\begin{proof}[\pfref{lem:kimi-local-concentration}]
  We first apply Bernstein's inequality at a fixed parameter, then extend the
  bounds uniformly over $\nbr*{\theta-\thetat}_2\le R$.
  For each sample, write
  \begin{align}
    q_{\theta,i}\ldef{}\frac1\eta\log\frac{\pi_\theta(y_i\mid{}x_i)}{\pit(y_i\mid{}x_i)}
      -r(x_i,y_i)+\hbt(x_i).
  \end{align}
  The bound $\abs{q_{\theta,i}}\le A+\Dfeat R/\eta$ in
  \cref{eq:proof:kimi-local-residual-bounds} applies with $\bt=\hbt$.
  Write $S=A+\Dfeat R/\eta$, so $\abs{q_{\theta,i}}\le S$ whenever
  $\nbr*{\theta-\thetat}_2\le R$.
  This also holds at $\thetat+\eta\thetastar$, since $R\ge\eta\Bstar$,
  and $\pi_{\thetat+\eta\thetastar}=\piplust$ by
  \cref{lem:softmax-target-realizability}.

  Fix $\theta$ with $\nbr*{\theta-\thetat}_2\le R$ and write
  $h=\theta-\thetat-\eta\thetastar$.
  We will bound the second moments of the sample derivatives and loss
  differences in terms of $\mset(\pi_\theta)$.
  By reward linearity and the definitions of $s_t^{\pi_\theta}$
  (\cref{eq:kimi-residual}) and $\Sigmat$ (\cref{eq:kimi-local-feature-covariance}),
  \begin{align}
    s_t^{\pi_\theta}(x,y)
    &=\frac1\eta\inner*{
      \phi(x,y)-\En_{y'\sim\pit(\cdot\mid x)}[\phi(x,y')],h},
    &\mset(\pi_\theta)&=\frac1{\eta^2}h^\top\Sigmat h.
    \label{eq:proof:kimi-local-concentration-mse}
  \end{align}
  For any $v$ with $\nbr*{v-\thetat}_2\le R$, the identity
  $\nabla q_{v,i}=\eta^{-1}\rbr{\phi(x_i,y_i)
    -\En_{y\sim\pi_v(\cdot\mid{}x_i)}[\phi(x_i,y)]}$
  (\cref{lem:softmax-score-identity}) gives
  \begin{align}
    \inner*{\nabla q_{v,i},h}
    =s_t^{\pi_\theta}(x_i,y_i)
      -\En_{y\sim\pi_v(\cdot\mid x_i)}[s_t^{\pi_\theta}(x_i,y)].
  \end{align}
  Since $\En_{y\sim\pit(\cdot\mid{}x)}[s_t^{\pi_\theta}(x,y)]=0$,
  conditioning on $x_i=x$ gives
  \begin{align}
    \En_{y_i\sim\pit(\cdot\mid{}x)}\sbr*{\inner*{\nabla q_{v,i},h}^2}
    &=\mset(\pi_\theta;x)
      +\rbr*{\En_{y\sim\pi_v(\cdot\mid x)}[s_t^{\pi_\theta}(x,y)]}^2
    \\
    &=\mset(\pi_\theta;x)
      +\rbr*{\En_{y\sim\pit(\cdot\mid x)}\sbr*{
        \rbr*{\frac{\pi_v(y\mid x)}{\pit(y\mid x)}-1}
        s_t^{\pi_\theta}(x,y)}}^2
    \\
    &\le\sbr*{1+\En_{y\sim\pit(\cdot\mid x)}\sbr*{
      \rbr*{\frac{\pi_v(y\mid x)}{\pit(y\mid x)}-1}^2}}
      \mset(\pi_\theta;x)
    \\
    &=\En_{y\sim\pit(\cdot\mid x)}\sbr*{
      \rbr*{\frac{\pi_v(y\mid x)}{\pit(y\mid x)}}^2}\mset(\pi_\theta;x)
      \le e^{\Dfeat R}\mset(\pi_\theta;x).
    \label{eq:proof:kimi-local-score-second-moment}
  \end{align}
  The first inequality is Cauchy--Schwarz.
  The last line uses
  $\pi_v(y\mid x)/\pit(y\mid x)\le e^{\Dfeat R}$ from \cref{lem:softmax-policy-ratio}.

  We next bound the second moment of
  $q_{\theta,i}-q_{\thetat+\eta\thetastar,i}$.
  Let $\theta_u=(1-u)(\thetat+\eta\thetastar)+u\theta$ for $0\le u\le1$.
  Then $\nbr*{\theta_u-\thetat}_2\le R$,
  and integrating then applying Jensen's inequality, we have
  \begin{align}
    \En_{x_i\sim\mu,\,y_i\sim\pit(\cdot\mid{}x_i)}
      \sbr*{\rbr*{q_{\theta,i}-q_{\thetat+\eta\thetastar,i}}^2}
    &=\En_{x_i\sim\mu,\,y_i\sim\pit(\cdot\mid{}x_i)}\sbr*{\rbr*{\int_0^1
      \inner*{\nabla q_{\theta_u,i},h}\,du}^2}
    \\
    &\le\int_0^1\En_{x_i\sim\mu,\,y_i\sim\pit(\cdot\mid{}x_i)}
      \sbr*{\inner*{\nabla q_{\theta_u,i},h}^2}\,du
      \le e^{\Dfeat R}\mset(\pi_\theta).
    \label{eq:proof:kimi-local-residual-second-moment}
  \end{align}
  where we use \cref{eq:proof:kimi-local-score-second-moment} in the last line.

  For the gradient bound \cref{eq:kimi-local-pairing-concentration} and the
  excess loss bound \cref{eq:kimi-local-excess-concentration}, respectively, define
  \begin{align}
    Z_i^{\mathrm{grad}}(\theta)&=2q_{\theta,i}\inner*{\nabla q_{\theta,i},h},
    &Z_i^{\mathrm{loss}}(\theta)&=(q_{\theta,i}+q_{\thetat+\eta\thetastar,i})
      (q_{\theta,i}-q_{\thetat+\eta\thetastar,i}).
  \end{align}
  For $j\in\{\mathrm{grad},\mathrm{loss}\}$, define
  \begin{align}
    F^j(\theta)\ldef{}\frac1N\sum_{i=1}^N\rbr*{
      Z_i^j(\theta)-\En_{\substack{x_i\sim\mu\\
        y_i,y'_{i,1},\ldots,y'_{i,G}\sim\pit(\cdot\mid{}x_i)}}[Z_i^j(\theta)]}.
  \end{align}
  By \cref{lem:kimi-local-expected-loss},
  \begin{align}
    F^{\mathrm{grad}}(\theta)
    &=\inner*{\nabla\hLtkimib(\pi_\theta)
      -\nabla\tLtkimib(\pi_\theta),h},
    \\
    F^{\mathrm{loss}}(\theta)
    &=\hLtkimib(\pi_\theta)-\hLtkimib(\piplust)
      -\tLtkimib(\pi_\theta)+\tLtkimib(\piplust).
    \label{eq:proof:kimi-local-centered-errors}
  \end{align}
  Here $\En_{x\sim\mu}[(\sigmatG(x))^2]$ is independent of $\theta$,
  so it vanishes upon differentiation or subtraction.
  The gradient bound in \cref{lem:kimi-local-regularity} justifies
  interchanging differentiation and expectation.

  Since $\abs{q_{\theta,i}}\le S$ and
  $\abs{q_{\theta,i}+q_{\thetat+\eta\thetastar,i}}\le2S$, we obtain, for each $j$,
  \begin{align}
    \En[Z_i^j(\theta)^2]&\le4e^{\Dfeat R}S^2\mset(\pi_\theta),
    &|Z_i^j(\theta)|&\le4S^2.
    \label{eq:proof:kimi-local-concentration-moments}
  \end{align}
  The second moment bound uses \cref{eq:proof:kimi-local-score-second-moment}
  for $Z_i^{\mathrm{grad}}$ and \cref{eq:proof:kimi-local-residual-second-moment}
  for $Z_i^{\mathrm{loss}}$. For the range bound on $Z_i^{\mathrm{grad}}$, we also use
  $\nbr{h}_2\le2R$ and $\nbr{\nabla q_{\theta,i}}_2\le\Dfeat/\eta$
  from \cref{eq:proof:kimi-local-residual-gradient}.
  Thus $\Var(Z_i^j(\theta))\le4e^{\Dfeat R}S^2\mset(\pi_\theta)$ and
  $|Z_i^j(\theta)-\En[Z_i^j(\theta)]|\le8S^2$.
  Bernstein's inequality \citep[Section~2.8]{vershynin2018high} gives,
  for each $j$ and $u>0$,
  with failure probability at most $2e^{-u}$,
  \begin{align}
    |F^j(\theta)|
    &\le\sqrt{\frac{2\Var(Z_i^j(\theta))u}{N}}+\frac{16S^2u}{3N}
    \\
    &\le\sqrt{\frac{8e^{\Dfeat R}S^2\mset(\pi_\theta)u}{N}}
      +\frac{16S^2u}{3N}
      \le\frac1{64}\mset(\pi_\theta)+\frac{144e^{\Dfeat R}S^2u}{N}.
    \label{eq:proof:kimi-local-fixed-concentration}
  \end{align}
  The last inequality is Young's inequality.
  Choose a finite set $\Theta_0\subseteq\cbr*{\theta:\nbr*{\theta-\thetat}_2\le R}$
  such that every $\theta$ with $\nbr*{\theta-\thetat}_2\le R$ has some
  $\widetilde\theta\in\Theta_0$ satisfying $\nbr{\theta-\widetilde\theta}_2\le R/N$.
  Such a set exists with $|\Theta_0|\le(1+2N)^d$
  \citep[Section~4.2]{vershynin2018high}.
  Set $u=d\log(1+2N)+\log(4/\delta)$.
  A union bound makes \cref{eq:proof:kimi-local-fixed-concentration}
  hold for both $F^{\mathrm{grad}}$ and $F^{\mathrm{loss}}$ at every
  $\widetilde\theta\in\Theta_0$ with probability at least $1-\delta$.
  We condition on this event.

  We now extend \cref{eq:proof:kimi-local-fixed-concentration} to every
  $\theta$ with $\nbr*{\theta-\thetat}_2\le R$.
  Differentiating $F^{\mathrm{grad}}$ and $F^{\mathrm{loss}}$ in
  \cref{eq:proof:kimi-local-centered-errors} gives
  \begin{align}
    \nabla F^{\mathrm{grad}}(\theta)
    &=\rbr*{\nabla^2\hLtkimib(\pi_\theta)
        -\nabla^2\tLtkimib(\pi_\theta)}h
      +\nabla\hLtkimib(\pi_\theta)-\nabla\tLtkimib(\pi_\theta),
    \\
    \nabla F^{\mathrm{loss}}(\theta)
    &=\nabla\hLtkimib(\pi_\theta)-\nabla\tLtkimib(\pi_\theta).
  \end{align}
  Thus $\nbr{h}_2\le2R$ and the gradient norm bounds in \cref{lem:kimi-local-regularity} give,
  for each $j$,
  \begin{align}
    \nbr*{\nabla F^j(\theta)}_2
    &\le2R\rbr*{
      \nbr*{\nabla^2\hLtkimib(\pi_\theta)}_{\op}
      +\nbr*{\nabla^2\tLtkimib(\pi_\theta)}_{\op}}
      +\nbr*{\nabla\hLtkimib(\pi_\theta)}_2
      +\nbr*{\nabla\tLtkimib(\pi_\theta)}_2
    \\
    &\le\frac{4R\Dfeat^2}{\eta^2}\rbr*{2+\frac{\eta S}{2}}
      +\frac{4\Dfeat S}{\eta}.
    \label{eq:proof:kimi-local-error-gradient}
  \end{align}
  Since $\Var_{y\sim\pit(\cdot\mid{}x)}[v^\top\phi(x,y)]\le\Dfeat^2/4$
  for every $\nbr{v}_2=1$ by \cref{eq:proof:kimi-local-projection-variance},
  the definition of $\Sigmat$ in \cref{eq:kimi-local-feature-covariance}
  gives $\Sigmat\preceq\Dfeat^2I/4$.
  Differentiating $\mset(\pi_\theta)=h^\top\Sigmat h/\eta^2$ from
  \cref{eq:proof:kimi-local-concentration-mse} therefore yields
  \begin{align}
    \nbr*{\nabla\mset(\pi_\theta)}_2
    =\frac2{\eta^2}\nbr*{\Sigmat(\theta-\thetat-\eta\thetastar)}_2
    \le\frac{\Dfeat^2R}{\eta^2}.
    \label{eq:proof:kimi-local-mse-gradient}
  \end{align}

  For any $\theta$ with $\nbr*{\theta-\thetat}_2\le R$, choose
  $\widetilde\theta\in\Theta_0$ with
  $\nbr{\theta-\widetilde\theta}_2\le R/N$.
  The bounds on $\nbr{\nabla F^j(\theta)}_2$
  (\cref{eq:proof:kimi-local-error-gradient}) and
  $\nbr{\nabla\mset(\pi_\theta)}_2$
  (\cref{eq:proof:kimi-local-mse-gradient}), together with $\Dfeat R/\eta\le S$, give
  \begin{align}
    |F^j(\theta)-F^j(\widetilde\theta)|
    &\le\frac RN\sbr*{\frac{4R\Dfeat^2}{\eta^2}
      \rbr*{2+\frac{\eta S}{2}}+\frac{4\Dfeat S}{\eta}}
      \le\frac{(12+2\Dfeat R)S^2}{N},
    \\
    |\mset(\pi_\theta)-\mset(\pi_{\widetilde\theta})|
    &\le\frac{\Dfeat^2R^2}{\eta^2N}\le\frac{S^2}{N}.
  \end{align}
  Combining these bounds with \cref{eq:proof:kimi-local-fixed-concentration}
  at $\widetilde\theta$ yields
  \begin{align}
    |F^j(\theta)|
    &\le|F^j(\widetilde\theta)|+\frac{(12+2\Dfeat R)S^2}{N}
    \\
    &\le\frac1{64}\mset(\pi_{\widetilde\theta})
      +\frac{144e^{\Dfeat R}S^2u}{N}
      +\frac{(12+2\Dfeat R)S^2}{N}
    \\
    &\le\frac1{64}\mset(\pi_\theta)
      +\frac{144e^{\Dfeat R}S^2u}{N}
      +\frac{(12+2\Dfeat R+1/64)S^2}{N}
    \\
    &\le\frac1{64}\mset(\pi_\theta)+\frac{160e^{\Dfeat R}S^2u}{N}.
  \end{align}
  The last inequality uses $u\ge1$ and
  $12+2\Dfeat R+1/64\le13e^{\Dfeat R}$.
  Substituting $S=A+\Dfeat R/\eta$ and
  $u=d\log(1+2N)+\log(4/\delta)$ gives
  \cref{eq:kimi-local-pairing-concentration,eq:kimi-local-excess-concentration}
  simultaneously for every $\theta$ in the stated range.
\end{proof}

\begin{lemma}
  \label{lem:kimi-local-progress}
  Under the assumptions of \cref{lem:kimi-local-concentration},
  let $a\ge0$ satisfy $\tbt(x)-r(x,\pit)\le a$ for every $x$.
  On the event in that lemma, every $\theta$ with
  $\nbr*{\theta-\thetat}_2\le R$ satisfies
  \begin{align}
    \inner*{\nabla\hLtkimib(\pi_\theta),
      \theta-\thetat-\eta\thetastar}
    &\ge\hLtkimib(\pi_\theta)-\hLtkimib(\piplust)
    \\
    &\quad+\rbr*{\frac{31}{32}-\eta a e^{\Dfeat R}}\mset(\pi_\theta)
      -2\zeta_N(\delta;R,A),
    \label{eq:kimi-local-empirical-progress}
  \end{align}
  where $\zeta_N(\delta;R,A)$ is defined in
  \cref{eq:kimi-local-statistical-error}.
\end{lemma}

\begin{proof}[\pfref{lem:kimi-local-progress}]
  Condition on the event in \cref{lem:kimi-local-concentration}.
  Fix $\theta$ with $\nbr*{\theta-\thetat}_2\le R$ and write
  $h=\theta-\thetat-\eta\thetastar$.
  To bound $\inner{\nabla\tLtkimib(\pi_\theta),h}$, we use
  \cref{lem:kimi-local-curvature}, which gives
  \begin{align}
    \nabla_v^2\tLtkimib(\pi_v)
    \succeq\frac2{\eta^2}\rbr*{1-\eta a e^{\Dfeat R}}\Sigmat,
    \qquad \nbr*{v-\thetat}_2\le R.
    \label{eq:proof:kimi-local-progress-curvature}
  \end{align}
  We have $\piplust=\pi_{\thetat+\eta\thetastar}$
  (\cref{lem:softmax-target-realizability}).
  For $v=(1-u)\theta+u(\thetat+\eta\thetastar)$ with $0\le u\le1$,
  \begin{align}
    \nbr*{v-\thetat}_2\le(1-u)\nbr*{\theta-\thetat}_2+u\eta\Bstar\le R.
  \end{align}
  Thus Taylor's theorem at $\theta$ gives, for some such $v$,
  \begin{align}
    \tLtkimib(\piplust)
    &=\tLtkimib(\pi_\theta)-\inner*{\nabla\tLtkimib(\pi_\theta),h}
      +\frac12h^\top\nabla_v^2\tLtkimib(\pi_v)h
    \\
    &\ge\tLtkimib(\pi_\theta)-\inner*{\nabla\tLtkimib(\pi_\theta),h}
      +\rbr*{1-\eta a e^{\Dfeat R}}\mset(\pi_\theta).
    \label{eq:proof:kimi-local-population-progress}
  \end{align}
  The inequality uses \cref{eq:proof:kimi-local-progress-curvature} and
  $h^\top\Sigmat h/\eta^2=\mset(\pi_\theta)$ from
  \cref{eq:proof:kimi-local-concentration-mse}.
  Applying the two concentration bounds in \cref{lem:kimi-local-concentration}
  and the preceding inequality yields
  \begin{align}
    \inner*{\nabla\hLtkimib(\pi_\theta),h}
    &\ge\inner*{\nabla\tLtkimib(\pi_\theta),h}
      -\frac1{64}\mset(\pi_\theta)-\zeta_N(\delta;R,A)
    \\
    &\ge\tLtkimib(\pi_\theta)-\tLtkimib(\piplust)
      +\rbr*{\frac{63}{64}-\eta a e^{\Dfeat R}}\mset(\pi_\theta)
      -\zeta_N(\delta;R,A)
    \\
    &\ge\hLtkimib(\pi_\theta)-\hLtkimib(\piplust)
      +\rbr*{\frac{31}{32}-\eta a e^{\Dfeat R}}\mset(\pi_\theta)
      -2\zeta_N(\delta;R,A).
  \end{align}
  The first inequality uses the concentration result in \cref{eq:kimi-local-pairing-concentration} of \cref{lem:kimi-local-concentration},
  the second uses \cref{eq:proof:kimi-local-population-progress} above,
  and the third uses the concentration result in
  \cref{eq:kimi-local-excess-concentration} of \cref{lem:kimi-local-concentration}.
\end{proof}

\section{\kimipg: Global Convergence (Proof of \cref*{cor:mean-kimi-main})}\label{app:kimi}
This section proves the convergence guarantees for \kimipg (\cref{alg:kimi-main})
with the mean baseline, and is organized as follows.
\begin{itemize}
  \item \cref{app:mean-kimi-geometric-lemmas} states and proves supporting lemmas.
  \item \cref{app:mean-kimi-geometric-inner} establishes convergence of the inner optimization loop for \kimipg (\cref{alg:kimi-main}) in \cref{thm:mean-kimi-geometric-inner}.
  \item \cref{app:mean-kimi-geometric-global} combines the inner loop guarantee with the mean-squared error criterion in \cref{thm:mse-regret-main} to prove \cref{cor:mean-kimi-geometric-regret}.
\end{itemize}
\para{Notation}
Throughout this section, we use the same notation as \cref{sec:kimi}.
Recall that \kimipg uses the baseline function $\bt(x) = r(x,\pit)$,
and at an outer iteration $t$ with $\pit$, the objective is
\begin{align}
  \Lkimi(\pi)
  =
  \En_{\pit}\sbr*{ \rbr*{\frac{1}{\eta} \log \frac{\pi(y\mid{}x)}{\pit(y\mid{}x)} - r(x,y) + r(x, \pit)}^2 }.
\end{align}
We also follow the notation of \kimipg in \cref{alg:kimi-main}.
To compute the empirical objective,
for each sampled prompt $x_i\sim\mu$
recall that \kimipg independently draws one response $y_i$ to estimate the objective
and a group of $G$ additional responses $y'_{i,1},\ldots,y'_{i,G}$ to estimate the baseline
from $\pit(\cdot\mid x_i)$.
The baseline estimate is
\begin{align}
  \bar r_i =  \frac{1}{G} \sum_{j=1}^G r(x_i, y'_{i,j}),
\end{align}
which is a consistent and unbiased estimator of the baseline function,
$\En\sbr*{\bar{r}_i\mid x_i}=r(x_i,\pit)$.
The empirical objective is then computed as
\begin{align}
  \hLkimi(\pi)
  =\frac1N\sum_{i=1}^N
  \rbr*{\frac1\eta\log\frac{\pi(y_i\mid x_i)}{\pit(y_i\mid x_i)}-r_i+\bar{r}_i}^2.
\end{align}

\subsection{Supporting Lemmas}
\label{app:mean-kimi-geometric-lemmas}

We now state and prove the supporting results for the inner convergence guarantee in \cref{app:mean-kimi-geometric-inner}.
\begin{itemize}
  \item \Cref{lem:mean-kimi-objective-decomposition} decomposes the population
  objective into mean-squared error and a squared KL divergence term.
  \item \Cref{lem:mean-kimi-baseline-gap} bounds the difference between
  the mean baseline $r(x,\pit)$ and the normalizer baseline $\btstar(x)$.
  \item \Cref{lem:mean-kimi-global-curvature} establishes global convexity by
  specializing \cref{lem:kimi-local-curvature}.
  \item \Cref{lem:mean-kimi-target-gradient} bounds the gradient at the exact update,
  which controls the bias when converting excess loss to mean-squared error.
  \item \Cref{lem:mean-kimi-geometric-loss-comparison} lower-bounds excess loss
  in terms of mean-squared error.
\end{itemize}

\para{Notation}
Fix an outer iteration $t$ and adopt the notation of
\cref{app:kimi-local-supporting-lemmas}.
For the mean baseline and every integer $G\ge1$, we have
\begin{align}
  \hbt(x)=\frac1G\sum_{j=1}^G r(x,y_j'),
  \qquad
  \tbt(x)=\bt(x)=r(x,\pit).
  \label{eq:mean-kimi-geometric-baseline}
\end{align}
The sample mean satisfies
$\min_y r(x,y)\le\hbt(x)\le\max_y r(x,y)$.
Recall the empirical objective is $\hLkimi$,
and the auxiliary objective defined in \cref{eq:kimi-common-auxiliary-loss} of \cref{app:kimi-local-supporting-lemmas} is the same as the population objective, $\tLtkimib=\Lkimi$.
We use the feature covariance $\Sigmat$ defined in
\cref{eq:kimi-local-feature-covariance}.

\begin{lemma}
  \label{lem:mean-kimi-objective-decomposition}
  Every policy $\pi>0$ satisfies
  \begin{align}
    \Lkimi(\pi)
    =\mset(\pi)
      +\En_{x\sim\mu}\sbr*{
        \rbr*{\frac1\eta\Dkl{\pit(\cdot\mid{}x)}{\pi(\cdot\mid{}x)}}^2}.
    \label{eq:mean-kimi-geometric-decomposition}
  \end{align}
\end{lemma}

\begin{proof}[\pfref{lem:mean-kimi-objective-decomposition}]
  Apply \cref{eq:kimi-local-objective-decomposition} in \cref{lem:kimi-local-objectives}
  with $\tbt(x)=r(x,\pit)$ and $\tLtkimib=\Lkimi$.
\end{proof}

\begin{lemma}
  \label{lem:mean-kimi-baseline-gap}
  Let $\eta>0$ and $\pit>0$, and suppose that $r(x,y)\in[0,\Rmax]$.
  For every prompt $x$, the mean baseline satisfies
  \begin{align}
    0\le\btstar(x)-r(x,\pit)\le\frac{\eta\Rmax^2}{8}.
    \label{eq:mean-kimi-normalization-gap}
  \end{align}
  Under \cref{asm:softmax-linear}, the bound also holds with $\Dstar$ in place of $\Rmax$.
\end{lemma}

\begin{proof}[\pfref{lem:mean-kimi-baseline-gap}]
  For every prompt $x$, Jensen's inequality gives
  \begin{align}
    \btstar(x)
    =\frac1\eta\log\En_{y\sim\pit(\cdot\mid{}x)}\sbr*{e^{\eta r(x,y)}}
    \ge\En_{y\sim\pit(\cdot\mid{}x)}\sbr*{r(x,y)}
    =r(x,\pit).
  \end{align}
  Applying Hoeffding's lemma to the centered reward $r(x,y)-r(x,\pit)$,
  whose range has length at most $\Rmax$, then yields

  \begin{align}
    0\le\btstar(x)-r(x,\pit)
    &=\frac1\eta\log\En_{y\sim\pit(\cdot\mid{}x)}
    \sbr*{e^{\eta\rbr*{r(x,y)-r(x,\pit)}}}
    \le\frac{\eta\Rmax^2}{8}.
  \end{align}
  Under \cref{asm:softmax-linear}, \cref{eq:kimi-reward-range} bounds the reward range
  by $\Dstar$, so the same argument gives the bound with $\Dstar$ in place of $\Rmax$.
\end{proof}

\begin{lemma}
  \label{lem:mean-kimi-global-curvature}
  Suppose that \cref{asm:softmax-linear} holds.
  For any $\eta>0$, $\Lkimi(\pi_\theta)$ is globally convex in $\theta$, with
  \begin{align}
    \nabla^2\Lkimi(\pi_\theta)\succeq\frac2{\eta^2}\Sigmat,
    \qquad\forall\theta\in\bbR^d.
    \label{eq:mean-kimi-global-curvature}
  \end{align}
\end{lemma}

\begin{proof}[\pfref{lem:mean-kimi-global-curvature}]
  For the mean baseline, $\bt(x)=r(x,\pit)$ and $\Ltkimib=\Lkimi$.
  The global curvature bound in \cref{eq:kimi-global-curvature} of
  \cref{lem:kimi-local-curvature} gives the result.
\end{proof}

\begin{lemma}
  \label{lem:mean-kimi-target-gradient}
  Suppose that \cref{asm:softmax-linear} holds.
  For every $\eta>0$ and $\theta\in\bbR^d$,
  \begin{align}
    \abs*{\inner*{\nabla\Lkimi(\piplust),
      \theta-\thetat-\eta\thetastar}}
    \le\frac{\eta^2\Dstar^3e^{\eta\Dstar/2}}8
      \sqrt{\mset(\pi_\theta)}.
    \label{eq:mean-kimi-target-gradient}
  \end{align}
\end{lemma}

\begin{proof}[\pfref{lem:mean-kimi-target-gradient}]
  By \cref{lem:softmax-target-realizability},
  $\piplust=\pi_{\thetat+\eta\thetastar}$.
  The exact update in \cref{eq:kimi-exact-update} gives
  \begin{align}
    \frac1\eta\log\frac{\piplust(y\mid x)}{\pit(y\mid x)}
      -r(x,y)+r(x,\pit)
    =r(x,\pit)-\btstar(x).
  \end{align}
  Differentiating $\Lkimi$ at $\theta=\thetat+\eta\thetastar$
  and using the softmax linear policy gradient in \cref{lem:softmax-score-identity} therefore gives
  \begin{align}
    &\inner*{\nabla\Lkimi(\piplust),\theta-\thetat-\eta\thetastar}
    \\
    &\quad=\frac2\eta\En_{x\sim\mu}\Big[
      \rbr*{\btstar(x)-r(x,\pit)}
      \inner*{\En_{y\sim\piplust(\cdot\mid x)}[\phi(x,y)]
        -\En_{y\sim\pit(\cdot\mid x)}[\phi(x,y)],
        \theta-\thetat-\eta\thetastar}\Big]
    \\
    &\quad=2\En_{x\sim\mu}\sbr*{
      \rbr*{\btstar(x)-r(x,\pit)}
      \En_{y\sim\piplust(\cdot\mid x)}[s_t^{\pi_\theta}(x,y)]}.
    \label{eq:proof:mean-kimi-target-gradient-identity}
  \end{align}
  For the last equality, the definition of $s_t^{\pi_\theta}$ in
  \cref{eq:kimi-residual} and reward linearity give
  \begin{align}
    s_t^{\pi_\theta}(x,y)
    =\frac1\eta\inner*{
      \phi(x,y)-\En_{y'\sim\pit(\cdot\mid x)}[\phi(x,y')],
      \theta-\thetat-\eta\thetastar}.
  \end{align}

  We bound $\En_{y\sim\piplust(\cdot\mid x)}[s_t^{\pi_\theta}(x,y)]$
  in \cref{eq:proof:mean-kimi-target-gradient-identity}
  by first converting the expectation to $\pit$.
  For $\piplust=\pi_{\thetat+\eta\thetastar}$,
  \cref{lem:kimi-local-policy-ratio-second-moment} with $R=\eta\Bstar$
  gives, since $\Dfeat\Bstar=\Dstar$,
  \begin{align}
    \En_{y\sim\pit(\cdot\mid x)}\sbr*{
      \rbr*{\frac{\piplust(y\mid x)}{\pit(y\mid x)}-1}^2}
    \le\frac{(e^{\eta\Dstar}-1)^2}{4e^{\eta\Dstar}}
    \le\frac{\eta^2\Dstar^2e^{\eta\Dstar}}4.
    \label{eq:proof:mean-kimi-target-ratio-second-moment}
  \end{align}
  Since $\En_{y\sim\pit(\cdot\mid x)}[s_t^{\pi_\theta}(x,y)]=0$,
  Cauchy--Schwarz and \cref{eq:proof:mean-kimi-target-ratio-second-moment} give
  \begin{align}
    \abs*{\En_{y\sim\piplust(\cdot\mid x)}[s_t^{\pi_\theta}(x,y)]}
    &=\abs*{\En_{y\sim\pit(\cdot\mid x)}\sbr*{
      \rbr*{\frac{\piplust(y\mid x)}{\pit(y\mid x)}-1}
      s_t^{\pi_\theta}(x,y)}}
    \\
    &\le\sqrt{\En_{y\sim\pit(\cdot\mid x)}\sbr*{
      \rbr*{\frac{\piplust(y\mid x)}{\pit(y\mid x)}-1}^2}}
      \sqrt{\mset(\pi_\theta;x)}
    \\
    &\le\frac{\eta\Dstar e^{\eta\Dstar/2}}2
      \sqrt{\mset(\pi_\theta;x)}.
    \label{eq:proof:mean-kimi-target-residual}
  \end{align}
  Returning to \cref{eq:proof:mean-kimi-target-gradient-identity}, we obtain
  \begin{align}
    \abs*{\inner*{\nabla\Lkimi(\piplust),
      \theta-\thetat-\eta\thetastar}}
    &\le2\En_{x\sim\mu}\sbr*{
      \rbr*{\btstar(x)-r(x,\pit)}
      \abs*{\En_{y\sim\piplust(\cdot\mid x)}[s_t^{\pi_\theta}(x,y)]}}
    \\
    &\le2\En_{x\sim\mu}\sbr*{
      \frac{\eta\Dstar^2}{8}\cdot
      \frac{\eta\Dstar e^{\eta\Dstar/2}}2
      \sqrt{\mset(\pi_\theta;x)}}
    \\
    &=\frac{\eta^2\Dstar^3e^{\eta\Dstar/2}}8
      \En_{x\sim\mu}\sbr*{\sqrt{\mset(\pi_\theta;x)}}
    \\
    &\le\frac{\eta^2\Dstar^3e^{\eta\Dstar/2}}8
      \sqrt{\mset(\pi_\theta)}.
  \end{align}
  Here $0\le\btstar(x)-r(x,\pit)\le\eta\Dstar^2/8$ by
  \cref{lem:mean-kimi-baseline-gap}.
  The second inequality also uses \cref{eq:proof:mean-kimi-target-residual},
  and the last is Cauchy--Schwarz over $x\sim\mu$.
\end{proof}

\begin{lemma}
  \label{lem:mean-kimi-geometric-loss-comparison}
  Let $\eta>0$ and $\pit>0$.
  If $r(x,y)\in[0,\Rmax]$, every policy $\pi>0$ satisfies
  \begin{align}
    \Lkimi(\pi)-\Lkimi(\piplust)
    \ge\frac34\mset(\pi)
      -\frac{\eta^4\Rmax^6e^{\eta\Rmax}}{64}.
    \label{eq:mean-kimi-geometric-loss-lower}
  \end{align}
  Under \cref{asm:softmax-linear}, this bound also holds with $\Dstar$ in
  place of $\Rmax$.
\end{lemma}

\begin{proof}[\pfref{lem:mean-kimi-geometric-loss-comparison}]
  By \cref{lem:kimi-local-objectives} with $\tbt(x)=r(x,\pit)$,
  \begin{align}
    \Lkimi(\pi)-\Lkimi(\piplust)
    &=\mset(\pi)+\En_{x\sim\mu}\sbr*{
      \rbr*{\frac1\eta\Dkl{\pit(\cdot\mid x)}{\pi(\cdot\mid x)}}^2
      -\rbr*{\btstar(x)-r(x,\pit)}^2}
    \\
    &\ge\mset(\pi)+2\En_{x\sim\mu}\Big[
      \rbr*{\btstar(x)-r(x,\pit)}
      \rbr*{\frac1\eta\Dkl{\pit(\cdot\mid x)}{\pi(\cdot\mid x)}
        -\btstar(x)+r(x,\pit)}\Big].
    \label{eq:proof:mean-kimi-geometric-loss-difference}
  \end{align}
  To lower bound the second term in \cref{eq:proof:mean-kimi-geometric-loss-difference},
  the definition of $s_t^\pi$ in \cref{eq:kimi-residual} gives, for each prompt $x$,
  \begin{align}
    \En_{y\sim\piplust(\cdot\mid x)}\sbr*{e^{\eta s_t^\pi(x,y)}}
    &=\En_{y\sim\pit(\cdot\mid x)}\sbr*{
      \frac{\pi(y\mid x)}{\pit(y\mid x)}
      e^{\Dkl{\pit(\cdot\mid x)}{\pi(\cdot\mid x)}-\eta\btstar(x)+\eta r(x,\pit)}}
    \\
    &=e^{\Dkl{\pit(\cdot\mid x)}{\pi(\cdot\mid x)}-\eta\btstar(x)+\eta r(x,\pit)}.
  \end{align}
  Taking logarithms and applying Jensen's inequality yields
  \begin{align}
    \frac1\eta\Dkl{\pit(\cdot\mid x)}{\pi(\cdot\mid x)}
      -\btstar(x)+r(x,\pit)
    &=\frac1\eta\log\En_{y\sim\piplust(\cdot\mid x)}
      \sbr*{e^{\eta s_t^\pi(x,y)}}
    \\
    &\ge\En_{y\sim\piplust(\cdot\mid x)}[s_t^\pi(x,y)].
    \label{eq:proof:mean-kimi-geometric-kl-lower}
  \end{align}
  Since $\btstar(x)-r(x,\pit)\ge0$ by \cref{lem:mean-kimi-baseline-gap},
  substituting \cref{eq:proof:mean-kimi-geometric-kl-lower} into
  \cref{eq:proof:mean-kimi-geometric-loss-difference} yields
  \begin{align}
    \Lkimi(\pi)-\Lkimi(\piplust)
    &\ge
    \mset(\pi)+2\En_{x\sim\mu}\sbr*{
      \rbr*{\btstar(x)-r(x,\pit)}
      \En_{y\sim\piplust(\cdot\mid x)}[s_t^\pi(x,y)]}
    \\
    &\ge
    \mset(\pi)
      -2\En_{x\sim\mu}\sbr*{
        \rbr*{\btstar(x)-r(x,\pit)}
        \abs*{\En_{y\sim\piplust(\cdot\mid x)}[s_t^\pi(x,y)]}}.
    \label{eq:proof:mean-kimi-geometric-loss-cross-term}
  \end{align}
  Next, we bound $\abr{\En_{y \sim \piplust(\cdot \mid{}x)}\sbr*{s_t^\pi(x,y)}}$ in the RHS of \cref{eq:proof:mean-kimi-geometric-loss-cross-term}
  in terms of $\mset$.
  To transfer the expectation from $\piplust$ to $\pit$, we use
  $\En_{y\sim\pit(\cdot\mid x)}[e^{\eta r(x,y)}]=e^{\eta\btstar(x)}$ to obtain
  \begin{align}
    \En_{y\sim\pit(\cdot\mid x)}\sbr*{
      \rbr*{\frac{\piplust(y\mid x)}{\pit(y\mid x)}-1}^2}
    &=\frac{\En_{y\sim\pit(\cdot\mid x)}\sbr*{
      \rbr*{e^{\eta r(x,y)}-e^{\eta\btstar(x)}}^2}}
      {e^{2\eta\btstar(x)}}
    \\
    &\le\frac{\En_{y\sim\pit(\cdot\mid x)}\sbr*{
      \rbr*{1+e^{\eta\Rmax}}e^{\eta r(x,y)}-e^{\eta\Rmax}}
      -e^{2\eta\btstar(x)}}{e^{2\eta\btstar(x)}}
    \\
    &=\frac{\rbr*{e^{\eta\Rmax}-e^{\eta\btstar(x)}}
      \rbr*{e^{\eta\btstar(x)}-1}}{e^{2\eta\btstar(x)}}
    \\
    &\le\frac{(e^{\eta\Rmax}-1)^2}{4e^{\eta\Rmax}}
      \le\frac{\eta^2\Rmax^2e^{\eta\Rmax}}4.
    \label{eq:proof:mean-kimi-general-target-ratio-second-moment}
  \end{align}
  Then Cauchy--Schwarz yields
  \begin{align}
    \abs*{\En_{y\sim\piplust(\cdot\mid x)}[s_t^\pi(x,y)]}
    &=\abs*{\En_{y\sim\pit(\cdot\mid x)}\sbr*{
      \rbr*{\frac{\piplust(y\mid x)}{\pit(y\mid x)}-1}s_t^\pi(x,y)}}
    \\
    &\le\sqrt{\En_{y\sim\pit(\cdot\mid x)}\sbr*{
      \rbr*{\frac{\piplust(y\mid x)}{\pit(y\mid x)}-1}^2}}
      \sqrt{\mset(\pi;x)}
    \\
    &\le\frac{\eta\Rmax e^{\eta\Rmax/2}}2\sqrt{\mset(\pi;x)}
    ,
    \label{eq:proof:mean-kimi-geometric-residual-mean}
  \end{align}
  where in the last inequality we have used \cref{eq:proof:mean-kimi-general-target-ratio-second-moment}.
  Substituting \cref{eq:proof:mean-kimi-geometric-residual-mean} into
  \cref{eq:proof:mean-kimi-geometric-loss-cross-term} and again applying
  Cauchy--Schwarz, we obtain
  \begin{align}
    \Lkimi(\pi)-\Lkimi(\piplust)
    &\ge\mset(\pi)
      -\eta\Rmax e^{\eta\Rmax/2}\En_{x\sim\mu}\sbr*{
        \rbr*{\btstar(x)-r(x,\pit)}\sqrt{\mset(\pi;x)}}
    \\
    &\ge\mset(\pi)
      -\eta\Rmax e^{\eta\Rmax/2}
        \sqrt{\En_{x\sim\mu}\sbr*{\rbr*{\btstar(x)-r(x,\pit)}^2}}
        \sqrt{\mset(\pi)}
    \\
    &\ge\mset(\pi)
      -\frac{\eta^2\Rmax^3e^{\eta\Rmax/2}}8
        \sqrt{\mset(\pi)}
    \\
    &\ge\frac34\mset(\pi)
      -\frac{\eta^4\Rmax^6e^{\eta\Rmax}}{64}.
  \end{align}
 The third line uses $0\le\btstar(x)-r(x,\pit)\le\eta\Rmax^2/8$ from \cref{lem:mean-kimi-baseline-gap},
  and the last uses Young's inequality.
  Under \cref{asm:softmax-linear}, the same calculation applies with $\Dstar$
  in place of $\Rmax$.
\end{proof}

\subsection{Inner Loop Convergence}
\label{app:mean-kimi-geometric-inner}

We now prove inner loop convergence for the mean baseline in \cref{alg:kimi-main}.
The proof uses empirical descent to bound the excess loss of the last iterate,
then converts this bound to one on its MSE.

\begin{theorem}[Inner convergence for the mean baseline]
  \label{thm:mean-kimi-geometric-inner}
  There is a universal constant $C>0$ such that the following holds.
  Fix an outer iteration $t$ and suppose that \cref{asm:softmax-linear} holds.
  Let $\delta\in(0,1)$, let $G\ge1$ be an integer, and let $\eta,\einsq>0$ satisfy
  \begin{align}
    \eta^4\Dstar^6\le\einsq\le\Dstar^2.
    \label{eq:mean-kimi-geometric-inner-accuracy}
  \end{align}
  Run the inner loop of \cref{alg:kimi-main} with
  \begin{align}
    0<\alpha
    &\le\frac{\eta^2}{\Dfeat^2(2+\eta\Dstar)},
    \\
    K
    &\ldef{}\max\cbr*{1,\left\lceil
      \frac{2\eta^2}{3\alpha\einsq}
      \rbr*{\Bstar^2-\frac{4\einsq}{\Dfeat^2}}
      \right\rceil}.
    \label{eq:mean-kimi-geometric-inner-parameters}
  \end{align}
  Choose an integer $N\ge1$ such that
  \begin{align}
    N\ge C\frac{\Dstar^2}{\einsq}
      \sbr*{d\log(1+2N)+\log\frac4\delta}.
    \label{eq:mean-kimi-geometric-inner-batch}
  \end{align}
  Conditional on the history preceding iteration $t$, with probability at least
  $1-\delta$ over the batch and its additional baseline responses,
  \begin{align}
    \mset(\pitk[K])\le\einsq,
    \qquad
    \max_{0\le k\le K}\nbr*{s_t^{\pitk}}_\infty
      \le\sqrt{\frac{101}{96}}\Dstar.
    \label{eq:mean-kimi-geometric-inner-conclusion}
  \end{align}
\end{theorem}

With the largest allowed learning rate in \cref{eq:mean-kimi-geometric-inner-parameters},
the theorem requires $O(\Dstar^2/\einsq)$ inner steps and a batch of size
$\wt O\rbr*{\frac{\Dstar^2}{\einsq}\rbr*{d+\log(1/\delta)}}$.
These guarantees hold for every $G\ge1$.

The proof below proceeds as follows.
We first use global convexity of $\Lkimi$ and concentration to show that
$\thetatk$ stays within a bounded ball throughout the inner loop (\cref{lem:mean-kimi-descent-first}),
and moreover the empirical loss $\hLkimi(\pitk)$ decreases monotonically (\cref{lem:mean-kimi-empirical-descent}).
We use these guarantees to show that $\pitK$ converges on the empirical loss
(\cref{eq:mean-kimi-geometric-inner-lower-telescoping}),
which we then transfer to the population excess loss $\Lkimi(\pitK)-\Lkimi(\piplust)$ and finally to $\mset(\pitK)$.
The last step uses a loss--MSE comparison showing that, although $\piplust$ need
not minimize $\Lkimi$, small population excess loss guarantees small MSE up to
an error that shrinks with $\eta$
(\cref{eq:mean-kimi-geometric-inner-square-root-lower}).

\subsubsection{Proof of \cref{thm:mean-kimi-geometric-inner}}
\begin{proof}[\pfref{thm:mean-kimi-geometric-inner}]
  Condition on the history preceding iteration $t$.
  We first establish useful convexity and concentration inequalities for our analysis.

  We apply the concentration results in \cref{lem:kimi-local-concentration} for general Kimi-style objectives
  to the mean baseline where $\tbt(x) = r(x,\pit)$ and
  $\min_y r(x,y)\le\hbt(x)\le\max_y r(x,y)$.
  We will invoke it using the theorem parameter conditions, $A=\Dstar$, and $R=13\eta\Bstar/4$,
  which specifies the parameter radius over which \cref{lem:kimi-local-concentration} holds.
  With $N$ as in \cref{eq:mean-kimi-geometric-inner-batch} and
  $\eta\Dstar\le1$ from \cref{eq:mean-kimi-geometric-inner-accuracy},
  the statistical error in \cref{eq:kimi-local-statistical-error} satisfies,
  for sufficiently large $C$,
  \begin{align}
    \zeta_N(\delta;R,\Dstar)\le\frac{\einsq}{128}\ldef{}\zeta.
    \label{eq:mean-kimi-geometric-inner-statistical-budget}
  \end{align}

  Then \cref{lem:kimi-local-concentration} states that the following inequality holds
  with probability at least $1-\delta$,
  simultaneously for every $\theta$ such that $\nbr*{\theta-\thetat}_2\le13\eta\Bstar/4$,
  \begin{align}
    \abs*{\hLkimi(\pi_\theta)-\hLkimi(\piplust)
      -\Lkimi(\pi_\theta)+\Lkimi(\piplust)}
    \le\frac1{64}\mset(\pi_\theta)+\zeta.
    \label{eq:mean-kimi-geometric-inner-excess-event-proposed}
  \end{align}

  Since $\tbt(x)=r(x,\pit)$, we apply the empirical first-order convexity bound in
  \cref{eq:kimi-local-empirical-progress} of \cref{lem:kimi-local-progress}
  with $a=0$, $A=\Dstar$, and $R=13\eta\Bstar/4$.
  Together with \cref{eq:mean-kimi-geometric-inner-statistical-budget}, this gives on the same event,
  for every $\theta$ with $\nbr*{\theta-\thetat}_2\le13\eta\Bstar/4$,
  \begin{align}
    \inner*{\nabla\hLkimi(\pi_\theta),
      \theta-\thetat-\eta\thetastar}
    &\ge\hLkimi(\pi_\theta)-\hLkimi(\piplust)
      +\frac{31}{32}\mset(\pi_\theta)-2\zeta.
    \label{eq:mean-kimi-geometric-inner-lower-alignment-proposed}
  \end{align}
  We condition on \cref{eq:mean-kimi-geometric-inner-excess-event-proposed,eq:mean-kimi-geometric-inner-lower-alignment-proposed}
  holding for the remainder of the proof.

  \cref{eq:mean-kimi-geometric-inner-lower-alignment-proposed} holds whenever
  $\nbr*{\theta-\thetat}_2\le13\eta\Bstar/4$.
  The following lemma, proved at the end of this section,
  shows that all inner iterates satisfy this condition.

  \begin{lemma}[Localization for the mean baseline]
    \label{lem:mean-kimi-descent-first}

    Under the assumptions and parameter choices of
    \cref{thm:mean-kimi-geometric-inner},
    suppose that \cref{eq:mean-kimi-geometric-inner-excess-event-proposed,eq:mean-kimi-geometric-inner-lower-alignment-proposed} hold. Then
    \begin{align}
      \max_{0\le k\le K}
        \nbr*{\thetatk-\thetat-\eta\thetastar}_2^2
      &\le\frac{101}{96}\eta^2\Bstar^2,
      &
      \max_{0\le k\le K}\nbr*{s_t^{\pitk}}_\infty
      &\le\sqrt{\frac{101}{96}}\Dstar.
      \label{eq:mean-kimi-lower-localization-residual}
    \end{align}
  \end{lemma}

  The bound on $\nbr{s_t^{\pitK}}_\infty$ in the theorem statement follows directly from \cref{eq:mean-kimi-lower-localization-residual}.
  In addition, by the triangle inequality, \cref{lem:mean-kimi-descent-first} gives
  \begin{align}
    \nbr*{\thetatk-\thetat}_2
    \le\frac94\eta\Bstar,
    \qquad 0\le k\le K.
    \label{eq:mean-kimi-geometric-inner-parameter-bound}
  \end{align}

  \para{Convergence of the empirical loss}
  We next bound $\hLkimi(\pitk[K])-\hLkimi(\piplust)$.
  We apply \cref{lem:mean-kimi-empirical-descent}, stated and proved later in this section,
  with the learning rate in the theorem preconditions (\cref{eq:mean-kimi-geometric-inner-parameters}).
  For every realization of the batch and its additional baseline responses,
  \cref{lem:mean-kimi-empirical-descent} gives, for every $0\le k<K$,
  \begin{align}
    \hLkimi(\pitk)-\hLkimi(\pitkp)
    &\ge\frac\alpha2\nbr*{\nabla\hLkimi(\pitk)}_2^2,
    \label{eq:mean-kimi-lower-empirical-descent-relocated}
  \end{align}
  Expanding the update in \cref{alg:kimi-main} gives, for $0\le k<K$,
  \begin{align}
    &\nbr*{\thetatkp-\thetat-\eta\thetastar}_2^2
      -\nbr*{\thetatk-\thetat-\eta\thetastar}_2^2
    \\
    &\quad=-2\alpha\inner*{\nabla\hLkimi(\pitk),
      \thetatk-\thetat-\eta\thetastar}
      +\alpha^2\nbr*{\nabla\hLkimi(\pitk)}_2^2
    \\
    &\quad\le-2\alpha\sbr*{\hLkimi(\pitk)-\hLkimi(\piplust)
        +\frac{31}{32}\mset(\pitk)-2\zeta}
      +\alpha^2\nbr*{\nabla\hLkimi(\pitk)}_2^2
    \\
    &\quad\le-2\alpha\sbr*{\hLkimi(\pitkp)-\hLkimi(\piplust)
        +\frac{31}{32}\mset(\pitk)-2\zeta}.
    \label{eq:mean-kimi-geometric-inner-lower-progress}
  \end{align}
  The first inequality applies the empirical first-order convexity bound we proved in \cref{eq:mean-kimi-geometric-inner-lower-alignment-proposed},
  which holds for $\pitk$ due to the parameter norm bound in \cref{eq:mean-kimi-geometric-inner-parameter-bound}.
  The second uses the empirical descent bound in
  \cref{eq:mean-kimi-lower-empirical-descent-relocated} above.

  The empirical descent bound in \cref{eq:mean-kimi-lower-empirical-descent-relocated} also gives
  $\hLkimi(\pitk[K])\le\hLkimi(\pitkp)$ for every $0\le k<K$.
  Hence
  \begin{align}
    \hLkimi(\pitk[K])-\hLkimi(\piplust)
    &\le\frac1K\sum_{k=0}^{K-1}
      \rbr*{\hLkimi(\pitkp)-\hLkimi(\piplust)}
    \\
    &\le\frac1K\sum_{k=0}^{K-1}\sbr*{
      \frac{\nbr*{\thetatk-\thetat-\eta\thetastar}_2^2
        -\nbr*{\thetatkp-\thetat-\eta\thetastar}_2^2}{2\alpha}
      -\frac{31}{32}\mset(\pitk)+2\zeta}
    \\
    &=\frac{\eta^2\Bstar^2
      -\nbr*{\thetatk[K]-\thetat-\eta\thetastar}_2^2}{2\alpha K}
      -\frac{31}{32K}\sum_{k=0}^{K-1}\mset(\pitk)+2\zeta.
    \label{eq:mean-kimi-geometric-inner-lower-telescoping}
  \end{align}
  The second inequality substitutes \cref{eq:mean-kimi-geometric-inner-lower-progress}
  at each $k$.
  The last line follows by telescoping and using $\thetatk[0]=\thetat$
  and $\nbr*{\thetastar}_2=\Bstar$.

  \para{Excess empirical loss to MSE}
  Finally, we translate the empirical excess loss of $\pitK$ to $\mset$,
  first going through the population excess loss.
  Recall that $\piplust=\pi_{\thetat+\eta\thetastar}$ (\cref{lem:softmax-target-realizability}).
  For every $\theta\in\bbR^d$, Taylor's theorem at $\thetat+\eta\thetastar$
  gives, for some $\theta'\in\bbR^d$,
  \begin{align}
    \Lkimi(\pi_\theta)-\Lkimi(\piplust)
    &=\inner*{\nabla\Lkimi(\piplust),
        \theta-\thetat-\eta\thetastar}
    \\
    &\quad+\frac12(\theta-\thetat-\eta\thetastar)^\top
      \nabla^2\Lkimi(\pi_{\theta'})
      (\theta-\thetat-\eta\thetastar)
    \\
    &\ge\mset(\pi_\theta)
      +\inner*{\nabla\Lkimi(\piplust),
        \theta-\thetat-\eta\thetastar}
    \\
    &\ge\mset(\pi_\theta)
      -\frac{\eta^2\Dstar^3e^{\eta\Dstar/2}}8
        \sqrt{\mset(\pi_\theta)}.
    \label{eq:mean-kimi-geometric-inner-square-root-lower}
  \end{align}
  The first inequality uses $\nabla^2\Lkimi(\pi_{\theta'})\succeq2\Sigmat/\eta^2$ from
  \cref{lem:mean-kimi-global-curvature} (\cref{eq:mean-kimi-global-curvature})
  and the quadratic form of the MSE (\cref{eq:proof:kimi-local-concentration-mse});
  the second uses \cref{lem:mean-kimi-target-gradient}.

  Next, we convert the above to the empirical loss.
  Since $\thetatk[K]$ satisfies the parameter norm bound in \cref{eq:mean-kimi-geometric-inner-parameter-bound},
  we apply the concentration inequality in \cref{eq:mean-kimi-geometric-inner-excess-event-proposed} above.
  Combined with \cref{eq:mean-kimi-geometric-inner-square-root-lower} this gives
  \begin{align}
    \hLkimi(\pitk[K])-\hLkimi(\piplust)
    &\ge\Lkimi(\pitk[K])-\Lkimi(\piplust)
      -\frac1{64}\mset(\pitk[K])-\zeta
    \\
    &\ge\frac{63}{64}\mset(\pitk[K])
      -\frac{\eta^2\Dstar^3e^{\eta\Dstar/2}}8\sqrt{\mset(\pitk[K])}
      -\zeta.
  \end{align}
  Rearranging this lower bound and then substituting the empirical upper bound
  \cref{eq:mean-kimi-geometric-inner-lower-telescoping} gives
  \begin{align}
    \frac{63}{64}\mset(\pitk[K])
    &\le\hLkimi(\pitk[K])-\hLkimi(\piplust)
      +\frac{\eta^2\Dstar^3e^{\eta\Dstar/2}}8\sqrt{\mset(\pitk[K])}+\zeta
    \\
    &\le\frac{\eta^2\Bstar^2
      -\nbr*{\thetatk[K]-\thetat-\eta\thetastar}_2^2}{2\alpha K}
      -\frac{31}{32K}\sum_{k=0}^{K-1}\mset(\pitk)
    \\
    &\qquad+\frac{\eta^2\Dstar^3e^{\eta\Dstar/2}}8
      \sqrt{\mset(\pitk[K])}+3\zeta
    \\
    &\le\frac{\eta^2\Bstar^2
      -\nbr*{\thetatk[K]-\thetat-\eta\thetastar}_2^2}{2\alpha K}
      +\frac{\eta^2\Dstar^3e^{\eta\Dstar/2}}8
        \sqrt{\mset(\pitk[K])}+3\zeta.
    \label{eq:mean-kimi-geometric-inner-terminal-bound}
  \end{align}

  To lower bound $\nbr*{\thetatk[K]-\thetat-\eta\thetastar}_2^2$
  in \cref{eq:mean-kimi-geometric-inner-terminal-bound},
  we use the quadratic form of the MSE in \cref{eq:proof:kimi-local-concentration-mse}
  at $\thetatk[K]$ to obtain
  \begin{align}
    \mset(\pitk[K])
    &=\frac1{\eta^2}
      (\thetatk[K]-\thetat-\eta\thetastar)^\top\Sigmat
      (\thetatk[K]-\thetat-\eta\thetastar)
    \le\frac{\Dfeat^2}{4\eta^2}
      \nbr*{\thetatk[K]-\thetat-\eta\thetastar}_2^2.
    \label{eq:mean-kimi-geometric-inner-mse-distance}
  \end{align}
  Here $\Sigmat\preceq\Dfeat^2I/4$, since
  $\Var_{y\sim\pit(\cdot\mid x)}[h^\top\phi(x,y)]
    \le\Dfeat^2\nbr*{h}_2^2/4$ for every $x$ and $h\in\bbR^d$.
  Substituting \cref{eq:mean-kimi-geometric-inner-mse-distance} into
  \cref{eq:mean-kimi-geometric-inner-terminal-bound} and rearranging gives
  \begin{align}
    \rbr*{\frac{63}{64}+\frac{2\eta^2}{\alpha K\Dfeat^2}}\mset(\pitk[K])
    \le\frac{\eta^2\Bstar^2}{2\alpha K}
      +\frac{\eta^2\Dstar^3e^{\eta\Dstar/2}}8\sqrt{\mset(\pitk[K])}
      +3\zeta.
    \label{eq:mean-kimi-geometric-inner-mse-substitution}
  \end{align}
  By Young's inequality,
  \begin{align}
    \frac{\eta^2\Dstar^3e^{\eta\Dstar/2}}8\sqrt{\mset(\pitk[K])}+3\zeta
    &\le\frac18\mset(\pitk[K])
      +\frac{\eta^4\Dstar^6e^{\eta\Dstar}}{32}+3\zeta
    \\
    &\le\frac18\mset(\pitk[K])+\frac7{64}\einsq.
  \end{align}
  The second inequality uses the range of $\einsq$ in \cref{eq:mean-kimi-geometric-inner-accuracy} from the theorem preconditions,
  $e^{\eta\Dstar}\le e<11/4$,
  and $\zeta=\einsq/128$ from \cref{eq:mean-kimi-geometric-inner-statistical-budget}.
  Substituting the above panel into \cref{eq:mean-kimi-geometric-inner-mse-substitution}
  and rearranging gives
  \begin{align}
    \rbr*{\frac{55}{64}+\frac{2\eta^2}{\alpha K\Dfeat^2}}\mset(\pitk[K])
    \le\frac{\eta^2\Bstar^2}{2\alpha K}+\frac7{64}\einsq
    \le\rbr*{\frac{55}{64}+\frac{2\eta^2}{\alpha K\Dfeat^2}}\einsq.
    \label{eq:mean-kimi-geometric-inner-terminal-mse-term}
  \end{align}
  The last inequality uses the choice of $K$ in
  \cref{eq:mean-kimi-geometric-inner-parameters}.
  Thus $\mset(\pitk[K])\le\einsq$.
\end{proof}

\subsubsection{Empirical Descent}
\begin{lemma}[Empirical descent for the mean baseline]
  \label{lem:mean-kimi-empirical-descent}
  Fix an outer iteration $t$ and suppose that \cref{asm:softmax-linear} holds.
  Let $\eta>0$ and let $N,G,K\ge1$ be integers.
  Run the inner loop of \cref{alg:kimi-main} with learning rate
  \begin{align}
    0<\alpha\le\frac{\eta^2}{\Dfeat^2(2+\eta\Dstar)}.
    \label{eq:mean-kimi-descent-first-stepsize}
  \end{align}
  For every realization of the batch and its additional baseline responses,
  $\hLkimi(\pitk)\le\Dstar^2$ for $0\le k\le K$, and
  \begin{align}
    \hLkimi(\pitk)-\hLkimi(\pitkp)
    &\ge\frac\alpha2\nbr*{\nabla\hLkimi(\pitk)}_2^2,
    \label{eq:mean-kimi-descent-first-descent}
    \\
    \nbr*{\thetatkp-\thetatk}_2
    &\le\frac{2\alpha\Dfeat\Dstar}{\eta}\le\eta\Bstar,
    \qquad 0\le k<K.
    \label{eq:mean-kimi-descent-first-step}
  \end{align}
\end{lemma}

\begin{proof}[\pfref{lem:mean-kimi-empirical-descent}]
  We first bound the gradient and Hessian in terms of the empirical loss.
  For each sample, write
  \begin{align}
    q_{\theta,i}
    \ldef{}\frac1\eta\log\frac{\pi_\theta(y_i\mid x_i)}{\pit(y_i\mid x_i)}
      -r_i+\bar r_i,
    \qquad
    \hLkimi(\pi_\theta)=\frac1N\sum_{i=1}^N q_{\theta,i}^2.
  \end{align}
  By \cref{lem:softmax-score-bound,lem:softmax-log-hessian},
  \begin{align}
    \nbr*{\nabla q_{\theta,i}}_2\le\frac{\Dfeat}{\eta},
    \qquad
    \nbr*{\nabla^2q_{\theta,i}}_{\op}\le\frac{\Dfeat^2}{4\eta},
    \qquad\theta\in\bbR^d.
  \end{align}
  The second bound uses
  $\Var_{y\sim\pi_\theta(\cdot\mid x)}[h^\top\phi(x,y)]\le\Dfeat^2/4$
  for $\nbr*{h}_2=1$, which follows from the feature diameter.
  Differentiating the squared loss therefore gives
  \begin{align}
    \nbr*{\nabla\hLkimi(\pi_\theta)}_2
    &=\nbr*{\frac2N\sum_{i=1}^N q_{\theta,i}\nabla q_{\theta,i}}_2
    \le\frac{2\Dfeat}{\eta}\sqrt{\hLkimi(\pi_\theta)},
    \label{eq:mean-kimi-descent-first-gradient}
    \\
    \nbr*{\nabla^2\hLkimi(\pi_\theta)}_{\op}
    &=\nbr*{\frac2N\sum_{i=1}^N\sbr*{
      \nabla q_{\theta,i}\nabla q_{\theta,i}^\top
      +q_{\theta,i}\nabla^2q_{\theta,i}}}_{\op}
    \\
    &\le\frac{2\Dfeat^2}{\eta^2}
      +\frac{\Dfeat^2}{2\eta}\sqrt{\hLkimi(\pi_\theta)}.
    \label{eq:mean-kimi-descent-first-hessian}
  \end{align}
  Both inequalities use
  $N^{-1}\sum_i|q_{\theta,i}|\le\sqrt{\hLkimi(\pi_\theta)}$.

  The gradient bound on $q_{\theta,i}$ gives
  $|q_{\theta',i}-q_{\theta,i}|\le(\Dfeat/\eta)\nbr*{\theta'-\theta}_2$.
  Thus the triangle inequality yields
  \begin{align}
    \sqrt{\hLkimi(\pi_{\theta'})}
    &\le\sqrt{\hLkimi(\pi_\theta)}
      +\sqrt{\frac1N\sum_{i=1}^N\rbr*{q_{\theta',i}-q_{\theta,i}}^2}
    \le\sqrt{\hLkimi(\pi_\theta)}+\frac{\Dfeat}{\eta}\nbr*{\theta'-\theta}_2.
    \label{eq:mean-kimi-descent-first-loss-continuity}
  \end{align}

  We now prove descent and $\hLkimi(\pitk)\le\Dstar^2$ by induction.
  Initially, $\hLkimi(\pit)\le\Dstar^2$ because
  $|\bar r_i-r_i|\le\Dstar$ by \cref{eq:kimi-reward-range}.
  Suppose $\hLkimi(\pitk)\le\Dstar^2$ for some $k<K$.
  The gradient bound \cref{eq:mean-kimi-descent-first-gradient} and the
  learning rate in \cref{eq:mean-kimi-descent-first-stepsize} give
  \begin{align}
    \nbr*{\thetatkp-\thetatk}_2
    =\alpha\nbr*{\nabla\hLkimi(\pitk)}_2
    \le\frac{2\alpha\Dfeat\Dstar}{\eta}\le\eta\Bstar.
  \end{align}
  To apply Taylor's theorem to this update, we use
  \cref{eq:mean-kimi-descent-first-loss-continuity,eq:mean-kimi-descent-first-hessian}
  to obtain, for every $\theta'$ with
  $\nbr*{\theta'-\thetatk}_2\le\eta\Bstar$,
  \begin{align}
    \sqrt{\hLkimi(\pi_{\theta'})}
    &\le\Dstar+\frac{\Dfeat}{\eta}\cdot\eta\Bstar=2\Dstar,
    &\nbr*{\nabla^2\hLkimi(\pi_{\theta'})}_{\op}
    &\le\frac{\Dfeat^2}{\eta^2}(2+\eta\Dstar).
  \end{align}
  This Hessian bound holds at every convex combination of
  $\thetatk$ and $\thetatkp$. Taylor's theorem therefore gives
  \begin{align}
    \hLkimi(\pitk)-\hLkimi(\pitkp)
    &\ge-\inner*{\nabla\hLkimi(\pitk),\thetatkp-\thetatk}
      -\frac{\Dfeat^2(2+\eta\Dstar)}{2\eta^2}\nbr*{\thetatkp-\thetatk}_2^2
    \\
    &=\alpha\sbr*{1-\frac{\alpha\Dfeat^2(2+\eta\Dstar)}{2\eta^2}}
      \nbr*{\nabla\hLkimi(\pitk)}_2^2
    \ge\frac\alpha2\nbr*{\nabla\hLkimi(\pitk)}_2^2.
  \end{align}
  The learning rate bound in \cref{eq:mean-kimi-descent-first-stepsize} gives the
  last inequality. Since $\hLkimi(\pitkp)\le\hLkimi(\pitk)\le\Dstar^2$,
  the induction is complete.
\end{proof}

\subsubsection{Proof of \cref{lem:mean-kimi-descent-first}}
\begin{proof}[\pfref{lem:mean-kimi-descent-first}]
  We condition on the inequalities in
  \cref{eq:mean-kimi-geometric-inner-excess-event-proposed,eq:mean-kimi-geometric-inner-lower-alignment-proposed}
  from the theorem proof holding,
  and use $\zeta=\einsq/128$ from
  \cref{eq:mean-kimi-geometric-inner-statistical-budget}.
  We will prove by induction that
  \begin{align}
    \nbr*{\thetatk-\thetat-\eta\thetastar}_2^2
    \le\eta^2\Bstar^2+10k\alpha\zeta,
    \qquad 0\le k\le K.
    \label{eq:mean-kimi-descent-first-distance}
  \end{align}
  The choice of $K$ in \cref{eq:mean-kimi-geometric-inner-parameters} gives
  \begin{align}
    \alpha K\zeta
    &\le\max\cbr*{\alpha\zeta,
      \frac{\eta^2\Bstar^2}{192}
      -\rbr*{\frac{8\eta^2}{3\Dfeat^2}-\alpha}\zeta}
    \le\frac{\eta^2\Bstar^2}{192}.
    \label{eq:mean-kimi-descent-first-horizon}
  \end{align}
  Here $\alpha\le\eta^2/(2\Dfeat^2)$ and $\einsq\le\Dstar^2$ give
  $\alpha\zeta\le\eta^2\Bstar^2/256$.
  Thus \cref{eq:mean-kimi-descent-first-distance} will imply the parameter
  bound in \cref{eq:mean-kimi-lower-localization-residual}.

  The initial bound holds because $\thetatk[0]=\thetat$.
  Suppose \cref{eq:mean-kimi-descent-first-distance} holds at $k<K$.
  By \cref{eq:mean-kimi-descent-first-horizon} and the update bound
  \cref{eq:mean-kimi-descent-first-step} from
  \cref{lem:mean-kimi-empirical-descent},
  \begin{align}
    \nbr*{\thetatk-\thetat}_2
    &\le\eta\Bstar+\sqrt{\eta^2\Bstar^2+10k\alpha\zeta}
      <\frac94\eta\Bstar,
    \\
    \nbr*{\thetatkp-\thetat}_2
    &\le\nbr*{\thetatk-\thetat}_2+\nbr*{\thetatkp-\thetatk}_2
      <\frac{13}{4}\eta\Bstar.
  \end{align}
  Hence the empirical first-order bound
  \cref{eq:mean-kimi-geometric-inner-lower-alignment-proposed} applies at $\thetatk$,
  and the excess loss concentration bound
  \cref{eq:mean-kimi-geometric-inner-excess-event-proposed} applies at $\thetatkp$.

  We now bound the increase in the squared distance in
  \cref{eq:mean-kimi-descent-first-distance}.
  Expanding the gradient update and using
  \cref{eq:mean-kimi-geometric-inner-lower-alignment-proposed} and empirical descent
  from \cref{lem:mean-kimi-empirical-descent} gives
  \begin{align}
    &\nbr*{\thetatkp-\thetat-\eta\thetastar}_2^2
      -\nbr*{\thetatk-\thetat-\eta\thetastar}_2^2
    \\
    &\quad=-2\alpha\inner*{\nabla\hLkimi(\pitk),
      \thetatk-\thetat-\eta\thetastar}
      +\alpha^2\nbr*{\nabla\hLkimi(\pitk)}_2^2
    \\
    &\quad\le-2\alpha\sbr*{\hLkimi(\pitk)-\hLkimi(\piplust)
        +\frac{31}{32}\mset(\pitk)-2\zeta}
      +2\alpha\rbr*{\hLkimi(\pitk)-\hLkimi(\pitkp)}
    \\
    &\quad=-2\alpha\sbr*{\hLkimi(\pitkp)-\hLkimi(\piplust)
        +\frac{31}{32}\mset(\pitk)-2\zeta}.
    \label{eq:mean-kimi-descent-first-progress}
  \end{align}
  We now lower bound $\hLkimi(\pitkp)-\hLkimi(\piplust)$ in
  \cref{eq:mean-kimi-descent-first-progress}.
  Applying \cref{eq:mean-kimi-geometric-inner-excess-event-proposed} at $\thetatkp$
  and then the population lower bound
  \cref{eq:mean-kimi-geometric-inner-square-root-lower} from the theorem proof gives
  \begin{align}
    \hLkimi(\pitkp)-\hLkimi(\piplust)
    &\ge\Lkimi(\pitkp)-\Lkimi(\piplust)
      -\frac1{64}\mset(\pitkp)-\zeta
    \\
    &\ge\frac{63}{64}\mset(\pitkp)
      -\frac{\eta^2\Dstar^3e^{\eta\Dstar/2}}8\sqrt{\mset(\pitkp)}-\zeta
    \label{eq:mean-kimi-descent-first-mse-lower}
    \\
    &\ge\frac{15}{64}\mset(\pitkp)
      -\frac{\eta^4\Dstar^6e^{\eta\Dstar}}{192}-\zeta
      \ge-3\zeta.
    \label{eq:mean-kimi-descent-first-excess-lower}
  \end{align}
  The last line uses Young's inequality,
  $\eta^4\Dstar^6\le\einsq$, $e^{\eta\Dstar}<3$, and $\zeta=\einsq/128$.
  Substituting \cref{eq:mean-kimi-descent-first-excess-lower} into
  \cref{eq:mean-kimi-descent-first-progress} yields
  \begin{align}
    \nbr*{\thetatkp-\thetat-\eta\thetastar}_2^2
    \le\nbr*{\thetatk-\thetat-\eta\thetastar}_2^2+10\alpha\zeta
    \le\eta^2\Bstar^2+10(k+1)\alpha\zeta.
  \end{align}
  This completes the induction. Together with
  \cref{eq:mean-kimi-descent-first-horizon}, it gives
  \begin{align}
    \max_{0\le k\le K}\nbr*{\thetatk-\thetat-\eta\thetastar}_2^2
    \le\eta^2\Bstar^2+10K\alpha\zeta
    \le\frac{101}{96}\eta^2\Bstar^2.
  \end{align}
  For the residual bound, the definition of $s_t^{\pitk}$ in
  \cref{eq:kimi-residual} gives, under reward linearity,
  \begin{align}
    |s_t^{\pitk}(x,y)|
    &=\frac1\eta\abs*{\inner*{
      \phi(x,y)-\En_{y'\sim\pit(\cdot\mid x)}[\phi(x,y')],
      \thetatk-\thetat-\eta\thetastar}}
    \\
    &\le\frac{\Dfeat}{\eta}\nbr*{\thetatk-\thetat-\eta\thetastar}_2
    \le\sqrt{\frac{101}{96}}\Dstar.
  \end{align}
  The first inequality uses the feature diameter.
  Taking the supremum over $x,y$ proves the residual bound for every
  $0\le k\le K$.
\end{proof}

\subsection{Proof of \cref{cor:mean-kimi-main}}
\label{app:mean-kimi-geometric-global}

We combine the inner convergence guarantee in \cref{thm:mean-kimi-geometric-inner}
with the MSE criterion in \cref{thm:mse-regret-main}.
The corollary uses $G=1$, so each sampled prompt requires one additional response
to estimate the mean baseline.

\begin{corollary}[\kimipg is \covpres]
  \label{cor:mean-kimi-geometric-regret}
  Suppose that \cref{asm:softmax-linear} holds.
  Fix $\veps\in(0,\Rmax]$ and $\delta\in(0,1)$.
  If $\veps\ge\Dstar$, take $T=0$ and return $\pifin=\piref$ without any updates or samples.
  Otherwise, run
  \cref{alg:kimi-main} with $\pit[0]=\piref$, $G=1$, and
  \begin{align}
    T&=\left\lceil
      2\Cresp^{1/4}\rbr*{\frac{\Dstar}{\veps}}^{3/2}
      \log^{3/2}(2\Cresp)\,\log^{1/4}(e\Cresp\Dstar/\veps)
      \right\rceil,
    &\eta&=\frac{\log(2\Cresp)}{\veps T}.
    \label{eq:mean-kimi-geometric-regret-outer}
  \end{align}
  Set
  \begin{align}
    \einsq
    &=\frac{1}{4\eta^2T^2
      \rbr*{\Cresp\log(eT)+e^{\eta\Dstar\sqrt{101/96}}}},
    \label{eq:mean-kimi-geometric-regret-tolerance}
    \\
    \alpha&=\frac{\eta^2}{\Dfeat^2(2+\eta\Dstar)},
    &K&=\max\cbr*{1,\left\lceil
      \frac{2(2+\eta\Dstar)}3\rbr*{\frac{\Dstar^2}{\einsq}-4}
      \right\rceil},
    \label{eq:mean-kimi-geometric-regret-inner}
  \end{align}
  and let $N$ be the smallest positive integer satisfying
  \begin{align}
    N\ge\frac{C\Dstar^2}{\einsq}
      \sbr*{d\log(1+2N)+\log\frac{4T}{\delta}},
    \label{eq:mean-kimi-geometric-regret-batch}
  \end{align}
  where $C$ is the universal constant in \cref{thm:mean-kimi-geometric-inner}.
  With probability at least $1-\delta$ over the sampled training data,
  the returned policy $\pifin=\pit[T]$ satisfies
  \begin{align}
    \Jstar-J(\pifin)\le\veps.
    \label{eq:mean-kimi-geometric-regret-conclusion}
  \end{align}
  On the same event, when $\veps<\Dstar$, with probability at least
  $1-(2+\sqrt2)/(4\sqrt{\log(eT)})$ over $x\sim\mu$,
  \begin{align}
    \pit(\Ystarx\mid x)\ge\frac12\piref(\Ystarx\mid x),
    \qquad 0\le t\le T.
    \label{eq:mean-kimi-geometric-regret-coverage}
  \end{align}
  In this case, the numbers of gradient steps and sampled responses satisfy
  \begin{align}
    \Nstep(\veps,\delta)=KT
    &=\wt O\rbr*{\frac{\Cresp^{5/4}\Dstar^{7/2}}{\veps^{7/2}}},
    \\
    \Nsample(\veps,\delta)=2NT
    &=\wt O\rbr*{\frac{\Cresp^{5/4}\Dstar^{7/2}}{\veps^{7/2}}
      \rbr*{d+\log(1/\delta)}}.
    \label{eq:mean-kimi-geometric-regret-complexity}
  \end{align}
\end{corollary}

The outer parameters in \cref{eq:mean-kimi-geometric-regret-outer} satisfy
$\eta=\wt\Theta\rbr*{\sqrt{\veps}/(\Cresp^{1/4}\Dstar^{3/2})}$.
Since $\Dstar\le2\Bfeat\Bstar$, \cref{cor:mean-kimi-geometric-regret}
also gives the complexity bounds in \cref{cor:mean-kimi-main} with
$B=\Bfeat\Bstar$.

\begin{proof}[\pfref{cor:mean-kimi-geometric-regret}]
  If $\veps\ge\Dstar$, then
  $\Jstar-J(\piref)\le\Dstar\le\veps$ by \cref{eq:kimi-reward-range}.
  It remains to consider $\veps<\Dstar$.

  We first verify that \cref{thm:mean-kimi-geometric-inner} applies at each
  outer iteration with failure probability $\delta/T$.
  The choices of $\alpha$, $K$, and $N$ in
  \cref{eq:mean-kimi-geometric-regret-inner,eq:mean-kimi-geometric-regret-batch}
  satisfy \cref{eq:mean-kimi-geometric-inner-parameters,eq:mean-kimi-geometric-inner-batch},
  so it remains to check $\eta^4\Dstar^6\le\einsq\le\Dstar^2$.
  The outer parameters in \cref{eq:mean-kimi-geometric-regret-outer} give
  \begin{align}
    T&\le4\Cresp^{1/4}\rbr*{\frac{\Dstar}{\veps}}^{3/2}
      \log^{3/2}(2\Cresp)\,\log^{1/4}(e\Cresp\Dstar/\veps),
    \label{eq:mean-kimi-geometric-regret-horizon}
    \\
    \log(eT)&\le2\log(e\Cresp\Dstar/\veps),
    &\eta\Dstar&\le\frac{1}{2\sqrt{\log2}},
    &e^{\eta\Dstar\sqrt{101/96}}&<2.
    \label{eq:mean-kimi-geometric-regret-scalar-bounds}
  \end{align}
  Here the bound on $T$ uses $\lceil u\rceil\le2u$ for $u\ge1$,
  and the bound on $\log(eT)$ follows from $\log u\le u-1$.
  Substituting \cref{eq:mean-kimi-geometric-regret-outer,eq:mean-kimi-geometric-regret-scalar-bounds}
  into the tolerance in \cref{eq:mean-kimi-geometric-regret-tolerance} yields
  \begin{align}
    \eta^4\Dstar^6
    &=\frac{\log^4(2\Cresp)\Dstar^6}{\veps^4T^4}
      \le\frac{\veps^2}{16\Cresp\log^2(2\Cresp)\log(e\Cresp\Dstar/\veps)}
    \\
    &\le\frac{\veps^2}{4\log^2(2\Cresp)
      \rbr*{\Cresp\log(eT)+e^{\eta\Dstar\sqrt{101/96}}}}
      =\einsq
      \le\veps^2<\Dstar^2.
    \label{eq:mean-kimi-geometric-regret-accuracy}
  \end{align}

  We may therefore apply \cref{thm:mean-kimi-geometric-inner} conditional on the history
  preceding each outer iteration. A union bound gives, with
  probability at least $1-\delta$ over the sampled training data,
  \begin{align}
    \mset(\pitp)&\le\einsq,
    &\nbr*{s_t^{\pitp}}_\infty&\le\sqrt{\frac{101}{96}}\Dstar,
    \qquad 0\le t<T.
    \label{eq:mean-kimi-geometric-regret-event}
  \end{align}
  We condition on this event for the remainder of the proof.

  We now use \cref{thm:mse-regret-main} to bound $\Jstar-J(\pifin)$.
  By \cref{eq:kimi-residual,eq:mean-kimi-geometric-regret-event},
  $S=\max_{0\le t<T}\nbr*{s_t^{\pitp}}_\infty\le\sqrt{101/96}\Dstar$.
  The choice of $\einsq$ in \cref{eq:mean-kimi-geometric-regret-tolerance}
  therefore gives the required MSE bound,
  \begin{align}
    \mset(\pitp)\le\einsq
    \le\frac{1}{4\eta^2T^2\rbr*{\Cresp\log(eT)+e^{\eta S}}},
    \qquad 0\le t<T.
    \label{eq:mean-kimi-geometric-regret-mse-tolerance}
  \end{align}
  Since $\pit[0]=\piref$ and softmax policies are positive,
  applying \cref{thm:mse-regret-main} and substituting
  \cref{eq:mean-kimi-geometric-regret-outer} gives
  \begin{align}
    \Jstar-J(\pifin)
    \le\frac{\log(2\Cresp)}{\eta T}=\veps.
  \end{align}

  For coverage, we apply \cref{thm:mse-accum-regret} with $c=1/2$
  and $s_t=s_t^{\pitp}$, as defined in
  \cref{eq:kimi-residual,eq:mse-accum-residual}.
  Substituting \cref{eq:mean-kimi-geometric-regret-mse-tolerance}
  into its coverage bound gives
  \begin{align}
    &\Pr_{x\sim\mu}\rbr*{\exists\,0\le t\le T:\;
      \pit(\Ystarx\mid x)<\tfrac12\piref(\Ystarx\mid x)}
    \\
    &\quad\le(2+\sqrt2)\rbr*{
      \frac{e^{\eta S}}{16T\rbr*{\Cresp\log(eT)+e^{\eta S}}}
      +\frac14\sqrt{\frac{\Cresp}{\Cresp\log(eT)+e^{\eta S}}}}
    \\
    &\quad\le\frac{2+\sqrt2}{8\sqrt{\log(eT)}}\rbr*{
      \frac{e^{\eta S}}{\Cresp\log(eT)+e^{\eta S}}
      +1+\frac{\Cresp\log(eT)}{\Cresp\log(eT)+e^{\eta S}}}
      =\frac{2+\sqrt2}{4\sqrt{\log(eT)}}.
  \end{align}
  The second inequality uses $\sqrt u\le(1+u)/2$ and
  $\sqrt{\log(eT)}\le2T$.

  Finally, we bound the numbers of gradient steps and sampled responses.
  By \cref{eq:mean-kimi-geometric-regret-horizon,eq:mean-kimi-geometric-regret-accuracy},
  \begin{align}
    T&=\wt O\rbr*{
      \frac{\Cresp^{1/4}\Dstar^{3/2}}{\veps^{3/2}}},
    &\frac{\Dstar^2}{\einsq}
      &\le\frac{16\Cresp\Dstar^2\log^2(2\Cresp)\log(e\Cresp\Dstar/\veps)}{\veps^2}.
  \end{align}
  The choices of $K$ and $N$ in
  \cref{eq:mean-kimi-geometric-regret-inner,eq:mean-kimi-geometric-regret-batch}
  then give
  \begin{align}
    K&=O\rbr*{\frac{\Dstar^2}{\einsq}},
    &N&=\wt O\rbr*{\frac{\Dstar^2}{\einsq}
      \rbr*{d+\log(T/\delta)}}.
  \end{align}
  Each outer iteration uses $K$ gradient steps and, since $G=1$, $2N$
  sampled responses. Multiplying these bounds by $T$ proves
  \cref{eq:mean-kimi-geometric-regret-complexity}.
\end{proof}

\section{Additional Result: \kiminorm with a Normalizer Baseline}\label{app:kimi-norm}
\begin{algorithm}[!t]
  \caption{\kiminorm}
  \label{alg:kimi-norm}
  \begin{algorithmic}[1]
    \Require \begingroup\raggedright
    \mbox{Base model $\piref$},
    \mbox{outer iterations $T$}, \mbox{regularization parameter $\eta>0$},
    \mbox{prompt batch size $N$}, \mbox{response group size $G$},
    \mbox{inner steps $K$}, \mbox{and learning rate $\alpha$}.\par
    \endgroup
    \State Initialize $\pit[0]=\piref=\pi_{\theta_0}$.
    \For{$t=0,1,\ldots,T-1$}
      \State Draw a batch of $N$ prompts $x_i\overset{\mathrm{iid}}{\sim}\mu$.
      \State \label{line:kimi-norm-sampling} For each prompt $x_i$, draw $y_i,y'_{i,1},\ldots,y'_{i,G}\sim\pit(\cdot\mid{}x_i)$ independently.
      \State \label{line:kimi-norm-baseline} Set $r_i=r(x_i,y_i)$ and estimate the normalizer baseline by
      \[
        \wh{b}_{t,i}=\frac1\eta\log\rbr*{\frac1G\sum_{j=1}^G e^{\eta r(x_i,y'_{i,j})}}.
      \]
      \State Initialize $\thetatk[0]=\thetat$ and $\pitk[0]=\pit$.
      \For{$k=0,1,\ldots,K-1$}
        \State \label{line:kimi-norm-gradient} Compute the empirical gradient (e.g., via backpropagation)
        
        \[
          \nabla\hLkiminorm(\pitk)
          =\frac{2}{\eta N}\sum_{i=1}^N
          \rbr*{\frac1\eta\log\frac{\pitk(y_i\mid{}x_i)}{\pit(y_i\mid{}x_i)}-r_i+\wh{b}_{t,i}}
          \nabla\log\pitk(y_i\mid{}x_i).
        \]
        
        \State \label{line:kimi-norm-update} Update $\thetatkp=\thetatk-\alpha\nabla\hLkiminorm(\pitk)$ and $\pitkp=\pi_{\thetatkp}$.
      \EndFor
      \State Set $\thetatp=\thetatk[K]$ and $\pitp=\pitk[K]$.
    \EndFor
    \State \Return $\pifin=\pit[T]$.
  \end{algorithmic}
\end{algorithm}

This section introduces \kiminorm, which estimates the normalizer baseline $\btstar$
instead of the mean reward used by \kimipg in \cref{sec:kimi,app:kimi}.
We use the superscript $\mathrm{norm}$ to distinguish its objectives from those of \kimipg.
\begin{itemize}
  \item \cref{app:kimi-norm-setup} defines the objectives and algorithm,
  and states the global convergence guarantee in \cref{cor:kimi-norm-regret}.
  \item \cref{app:kimi-norm-shared-lemmas} bounds the baseline estimation bias
  and develops the supporting lemmas for the inner optimization loop.
  \item \cref{app:kimi-norm-shared-inner} proves the inner loop convergence guarantee.
  \item \cref{app:kimi-norm-global} combines this guarantee with \cref{thm:mse-regret-main}
  to prove \cref{cor:kimi-norm-regret}.
\end{itemize}

We use the notation and mean-squared error from \cref{app:kimi-shared},
and assume the softmax linear parameterization and linear rewards in
\cref{asm:softmax-linear}, with $r(x,y)\in[0,\Rmax]$.

\subsection{\kiminorm Algorithm and the Normalizer Baseline}
\label{app:kimi-norm-setup}

\kiminorm uses the Kimi-style objective with the normalizer baseline
$\btstar(x)=\eta^{-1}\log\En_{y\sim\pit(\cdot\mid{}x)}\sbr*{e^{\eta r(x,y)}}$,
\begin{align}
  \Lkiminorm(\pi_\theta)
  \ldef{}\En_{x\sim\mu,\,y\sim\pit(\cdot\mid{}x)}
  \sbr*{\rbr*{
    \frac1\eta\log\frac{\pi_\theta(y\mid{}x)}{\pit(y\mid{}x)}
    -r(x,y)+\btstar(x)
  }^2}.
  \label{eq:kimi-norm-population-loss}
\end{align}
By \cref{eq:kkt}, the \framework target $\piplust$ has zero loss
and minimizes \cref{eq:kimi-norm-population-loss}.

As in prior Kimi-style algorithms
\citep{du2025kimi,cohen2025soft,ritter2026llms,chang2026karl},
\kiminorm replaces the unknown $\btstar(x)$ with a plug-in estimate
from $G$ additional responses,
\begin{align}
  \wh{b}_t(x)
  \ldef{}\frac1\eta\log\rbr*{\frac1G\sum_{j=1}^G e^{\eta r(x,y'_j)}},
  \qquad y'_1,\ldots,y'_G\overset{\mathrm{iid}}{\sim}\pit(\cdot\mid{}x).
  \label{eq:kimi-norm-baseline}
\end{align}
The logarithm makes $\wh{b}_t(x)$ biased relative to $\btstar$;
in particular, $\wh{b}_t(x)$ underestimates $\btstar$ in expectation.
\Cref{lem:kimi-norm-estimation-bias} shows that this bias is $O(1/G)$,
\begin{align}
  0\le\btstar(x)-\En\sbr*{\wh{b}_t(x)\mid x}
  \le\frac{\rbr*{e^{\eta\Dstar}-1}^2}{8\eta G},
  \qquad\forall x\in\cX,
  \label{eq:kimi-norm-bias-overview}
\end{align}
where $\Dstar$ bounds the reward difference between responses to the same prompt
(\cref{eq:kimi-reward-range}).
The analysis must therefore account for how much the gap between $\wh{b}_t$ and $\btstar$
biases the objective solution away from $\piplust$,
and controlling this error will require a larger response group size $G$.

\para{Algorithm}
At each outer iteration $t$, \kiminorm (\cref{alg:kimi-norm}) draws
$N$ prompts $x_i\sim\mu$ and $G+1$ independent responses from $\pit$ for each prompt
(Line~\ref{line:kimi-norm-sampling}).
It uses $G$ responses to form $\wh{b}_{t,i}$ according to \cref{eq:kimi-norm-baseline},
and the remaining response $y_i$, with reward $r_i=r(x_i,y_i)$,
to form the empirical objective
\begin{align}
  \hLkiminorm(\pi_\theta)
  \ldef{}\frac1N\sum_{i=1}^N
  \rbr*{
    \frac1\eta\log\frac{\pi_\theta(y_i\mid{}x_i)}{\pit(y_i\mid{}x_i)}
    -r_i+\wh{b}_{t,i}
  }^2.
  \label{eq:kimi-norm-empirical-loss}
\end{align}
Starting from $\thetatk[0]=\thetat$, the inner loop takes $K$ full-batch gradient descent steps
on \cref{eq:kimi-norm-empirical-loss} with learning rate $\alpha$
(Line~\ref{line:kimi-norm-update}).
The batch, baseline estimates, and reference policy $\pit$ remain fixed during these steps.
The last inner iterate becomes $\pitp=\pitk[K]$, and the procedure repeats for $T$ outer iterations.

The following result shows that \kiminorm is \covpres under \cref{asm:softmax-linear}.
The proof and exact parameter choices are given in \cref{app:kimi-norm-global}.

\begin{corollary}[\kiminorm is \covpres]
  \label{cor:kimi-norm-regret}
  Suppose that \cref{asm:softmax-linear} holds.
  For any $\veps\in(0,\Rmax]$ and $\delta\in(0,1)$,
  run \cref{alg:kimi-norm} with $\pit[0]=\piref$ and $\eta=1/(3\Rmax+4\Dstar)$.
  There exist choices of $\alpha$, $G$, $N$, $K$, and $T$
  such that the returned policy $\pifin$ satisfies
  \begin{align}
    \Jstar-J(\pifin)\le\veps
    \label{eq:kimi-regret-conclusion}
  \end{align}
  with probability at least $1-\delta$, using
  \begin{align}
    \Nstep(\veps,\delta)=KT
    &=\wt O\rbr*{
      \frac{\Cresp\Dstar^2}{\eta\veps^3}
    },
    \\
    \Nsample(\veps,\delta)=N(G+1)T
    &=\wt O\rbr*{
      \frac{\Cresp\Dstar^2\rbr*{d+\log(1/\delta)}}{\eta\veps^3}
      \rbr*{1+\frac{\eta\sqrt{\Cresp}\Dstar^2}{\veps}}
    }.
    \label{eq:kimi-regret-complexity}
  \end{align}
\end{corollary}

\para{Parameter choices}
The regularization parameter $\eta$ in \cref{cor:kimi-norm-regret}
is independent of the target accuracy $\veps$.
With problem parameters fixed, the choices in the proof satisfy
$T=\wt\Theta(1/\veps)$ and $N,K=\wt\Theta(1/\veps^2)$ as $\veps\to0$.
Thus \kiminorm uses $KT=\wt O(1/\veps^3)$ gradient steps,
compared with $\wt O(1/\veps^{7/2})$ for \kimipg (\cref{cor:mean-kimi-main}).
The improvement comes from keeping $\eta$ fixed, whereas \kimipg decreases $\eta$
to control the bias of the mean baseline and consequently needs more outer iterations.

\para{Responses for baseline estimation}
To obtain the guarantee in \cref{cor:kimi-norm-regret}, we choose
\begin{align}
  G=\wt\Theta\rbr*{1+\frac{\eta\sqrt{\Cresp}\Dstar^2}{\veps}}.
\end{align}
This choice makes the squared bias in \cref{eq:kimi-norm-bias-overview}
small enough to meet the mean-squared error tolerance required by \cref{thm:mse-regret-main}.
For fixed problem parameters, $G=\wt\Theta(1/\veps)$,
so the total number of sampled responses is $N(G+1)T=\wt O(1/\veps^4)$.
Thus \kiminorm uses fewer gradient steps but more responses than
\kimipg, which uses $G=1$ and $\wt O(1/\veps^{7/2})$ responses.
\Cref{tab:kimi-baseline-comparison} gives the full dependence on the problem parameters.

\subsection{Supporting Lemmas}
\label{app:kimi-norm-shared-lemmas}

We instantiate the supporting lemmas from the general baseline results in \cref{app:kimi-local-supporting-lemmas} to the normalizer baseline.
As an overview,
\begin{itemize}
  \item \cref{lem:kimi-norm-estimate-range} and \cref{lem:kimi-norm-estimation-bias}
  bound the range and bias of the estimated baseline, the latter as a function of $G$.
  \item \cref{lem:kimi-norm-comparator-loss} bounds loss of $\piplust$ on the auxiliary loss $\tLkiminorm$ used in our proofs, defined below.
\end{itemize}

\para{Notation}
We adopt the notation of \cref{app:kimi-local-supporting-lemmas}.
Given $G\ge1$,
for the baseline estimate in \cref{eq:kimi-norm-baseline} and its expectation, respectively, we write
\begin{align}
  \hbt(x) = \frac1\eta\log\rbr*{\frac1G\sum_{j=1}^G e^{\eta r(x,y_j')}},
  \qquad
  \tbt(x) = \En_{y_1',\ldots,y_G'\sim\pit(\cdot\mid{}x)}\sbr*{\hbt(x)}.
\end{align}
Recall the empirical \kiminorm objective $\hLkiminorm$ from \cref{eq:kimi-norm-empirical-loss}.
Following \cref{eq:kimi-common-auxiliary-loss},
in our proofs we use the auxiliary population objective defined using $\tbt$ rather than the true estimation target $\btstar$,
\begin{align}
  \tLkiminorm(\pi)
  \ldef{}
  \En_{\pit}\sbr*{\rbr*{ \frac1\eta\log\frac{\pi(y\mid{}x)}{\pit(y\mid{}x)} -r(x,y)+\tbt(x) }^2}.
\end{align}
Below, we state and prove the supporting lemmas for the inner loop convergence result and proof in \cref{app:kimi-norm-shared-inner}.

\begin{lemma}
  \label{lem:kimi-norm-estimate-range}
  For $\eta>0$ and every integer $G\ge1$, the estimated normalizer baseline satisfies
  \begin{align}
    \min_y r(x,y)\le\hbt(x)\le\max_y r(x,y),
    \qquad x\in\cX.
    \label{eq:kimi-norm-estimate-range}
  \end{align}
\end{lemma}

\begin{proof}[\pfref{lem:kimi-norm-estimate-range}]
  For every prompt $x$, each sampled reward lies between $\min_y r(x,y)$ and $\max_y r(x,y)$.
  Hence
  \begin{align}
    e^{\eta\min_y r(x,y)}
    \le\frac1G\sum_{j=1}^G e^{\eta r(x,y_j')}
    \le e^{\eta\max_y r(x,y)}.
  \end{align}
  The claim follows by taking the logarithm of the above and dividing by $\eta$.
\end{proof}

\begin{lemma}
  \label{lem:kimi-norm-estimation-bias}
  Suppose that \cref{asm:softmax-linear} holds.
  For $\eta>0$ and every integer $G\ge1$,
  \begin{align}
    0\le\btstar(x)-\tbt(x)
    \le\frac{\rbr*{e^{\eta\Dstar}-1}^2}{8\eta G},
    \qquad x\in\cX.
    \label{eq:kimi-norm-estimation-bias}
  \end{align}
\end{lemma}

\begin{proof}[\pfref{lem:kimi-norm-estimation-bias}]
  Fix $x$, with expectations over the group of $G$ additional responses sampled for $x$.
  Subtracting $\min_y r(x,y)$ from the rewards shifts $\btstar(x)$,
  $\hbt(x)$, and $\tbt(x)$ by the same amount, so the bias is unchanged.
  By \cref{asm:softmax-linear}, the shifted rewards lie in $\sbr{0,\Dstar}$.
  We use the same notation for these shifted quantities.

  From the definition of $\hbt(x)$ (\cref{eq:kimi-norm-baseline}), we have
  $\En_{\pit}\sbr*{e^{\eta\hbt(x)}}=e^{\eta\btstar(x)}$.
  Jensen's inequality therefore yields
  \begin{align}
    \tbt(x)=\En_{\pit}\sbr*{\hbt(x)}
    \le\frac1\eta\log\En_{\pit}\sbr*{e^{\eta\hbt(x)}}
    =\btstar(x),
  \end{align}
  proving the lower bound in the lemma statement.

  Next, both $e^{\eta\hbt(x)}$ and $e^{\eta\btstar(x)}$ lie in
  $\sbr{1,e^{\eta\Dstar}}$ by \cref{lem:kimi-norm-estimate-range} and the identity above.
  Since the second derivative of $-\log$ is at most $1$ on this interval,
  Taylor's theorem around $e^{\eta\btstar(x)}$ gives
  \begin{align}
    \eta\rbr*{\btstar(x)-\hbt(x)}
    \le 1-e^{\eta\rbr*{\hbt(x)-\btstar(x)}}
      +\frac12\rbr*{e^{\eta\hbt(x)}-e^{\eta\btstar(x)}}^2.
  \end{align}
  Taking expectations and again using the fact that
  $\En_{\pit}\sbr{e^{\eta\hbt(x)}}=e^{\eta\btstar(x)}$ removes the linear term.
  Since $\tbt(x)=\En_{\pit}\sbr*{\hbt(x)}$, we obtain
  \begin{align}
    \eta\rbr*{\btstar(x)-\tbt(x)}
    \le\frac12\En_{\pit}\sbr*{\rbr*{e^{\eta\hbt(x)}-e^{\eta\btstar(x)}}^2}
    =\frac12\Var_{\pit}\sbr*{e^{\eta\hbt(x)}}.
    \label{eq:proof:kimi-norm-bias-variance}
  \end{align}
  Finally, $e^{\eta\hbt(x)}$ is the average of $G$ independent copies of
  $e^{\eta r(x,y)}$ sampled from $\pit(\cdot \mid{} x)$.
  By independence,
  \begin{align}
    \Var_{\pit}\sbr*{e^{\eta\hbt(x)}}
    =\frac1G\Var_{\pit}\sbr*{e^{\eta r(x,y)}}.
  \end{align}
  Since the mean minimizes expected squared distance,
  using the midpoint of $\sbr{1,e^{\eta\Dstar}}$ we obtain
  \begin{align}
    \Var_{\pit}\sbr*{e^{\eta r(x,y)}}
    \le\En_{\pit}\sbr*{\rbr*{e^{\eta r(x,y)}-\frac{1+e^{\eta\Dstar}}2}^2}
    \le\frac{\rbr*{e^{\eta\Dstar}-1}^2}{4}.
  \end{align}
  The last inequality holds because every value of $e^{\eta r(x,y)}$
  lies within $\rbr{e^{\eta\Dstar}-1}/2$ of this midpoint.
  Hence $\Var_{\pit}\sbr*{e^{\eta\hbt(x)}}\le\rbr{e^{\eta\Dstar}-1}^2/(4G)$.
  Substituting this variance bound into \cref{eq:proof:kimi-norm-bias-variance}
  and dividing by $\eta$ proves the upper bound.
\end{proof}

\begin{lemma}
  \label{lem:kimi-norm-comparator-loss}
  Suppose that \cref{asm:softmax-linear} holds.
  For $\eta>0$ and every integer $G\ge1$,
  \begin{align}
    \tLkiminorm(\piplust)
    \le\frac{\rbr*{e^{\eta\Dstar}-1}^4}{64\eta^2G^2}.
    \label{eq:kimi-norm-comparator-loss}
  \end{align}
\end{lemma}

\begin{proof}[\pfref{lem:kimi-norm-comparator-loss}]
  Applying \cref{eq:kimi-local-comparator-loss} in \cref{lem:kimi-local-objectives}
  to the auxiliary objective $\tLkiminorm$ gives
  \begin{align}
    \tLkiminorm(\piplust)
    =\En_{x\sim\mu}\sbr*{\rbr*{\btstar(x)-\tbt(x)}^2}.
    \label{eq:proof:kimi-norm-comparator-bias-identity}
  \end{align}
  By \cref{lem:kimi-norm-estimation-bias}, for every $x\in\cX$,
  \begin{align}
    0\le\btstar(x)-\tbt(x)
    \le\frac{\rbr*{e^{\eta\Dstar}-1}^2}{8\eta G}.
  \end{align}
  Squaring the above inequality and taking expectation over $x \sim \mu$ yields
  \begin{align}
    \En_{x\sim\mu}\sbr*{\rbr*{\btstar(x)-\tbt(x)}^2}
    \le\rbr*{\frac{\rbr*{e^{\eta\Dstar}-1}^2}{8\eta G}}^2
    =\frac{\rbr*{e^{\eta\Dstar}-1}^4}{64\eta^2G^2}.
  \end{align}
  Substituting the above into \cref{eq:proof:kimi-norm-comparator-bias-identity}
  proves the claim.
\end{proof}

\subsection{Inner Loop Convergence}
\label{app:kimi-norm-shared-inner}

Using the supporting lemmas in \cref{app:kimi-norm-shared-lemmas}
with the general supporting lemmas in \cref{app:kimi-local-supporting-lemmas},
instantiated for the normalizer baseline,
we now prove convergence of the inner optimization loop of \kiminorm (\cref{alg:kimi-norm}).

\begin{theorem}[Inner loop convergence of \kiminorm]
  \label{thm:kimi-norm-shared-inner}
  Suppose that \cref{asm:softmax-linear} holds.
  Fix an outer iteration $0\le t<T$, $\eta\in(0,1/(3\Dstar)]$,
  and $\einsq\in(0,\Dstar^2]$. Let the response group size satisfy
  \begin{align}
    G&\ge\frac{\rbr*{e^{\eta\Dstar}-1}^2}{4\eta\sqrt{\einsq}}.
    \label{eq:kimi-norm-shared-group-size}
  \end{align}
  There is a universal constant $C>0$ such that, for any $\delta\in(0,1)$,
  if the inner loop hyperparameters satisfy
  \begin{align}
    N&\ge\frac{C\Dstar^2}{\einsq}
      \rbr*{d\log(1+8N)+\log\frac{4(d+1)}{\delta}},
    &
    K&=\left\lceil\frac{\eta^2\Bstar^2}{\alpha\einsq}\right\rceil,
    &
    0<\alpha&\le\frac{\eta^2}{8\Dfeat^2},
    \label{eq:kimi-norm-shared-inner-parameters}
  \end{align}
  then the resulting policy $\pitp=\pitk[K]$ satisfies,
  with probability at least $1-\delta$,
  \begin{align}
    \mset(\pitp)&\le\einsq,
    &\nbr*{s_t^{\pitp}}_\infty&\le2\Dstar.
    \label{eq:kimi-norm-shared-inner-conclusion}
  \end{align}
\end{theorem}

The inner convergence complexity and the results in \cref{eq:kimi-norm-shared-inner-conclusion} can then be combined with \cref{thm:mse-regret-main} to establish global convergence,
which we proceed to do in \cref{app:kimi-norm-shared-global}.

The proof below proceeds as follows.
First, we show jointly that throughout the inner loop,
the parameters $\thetatk$ stay within a ball of constant radius;
moreover, the empirical objective $\hLkiminorm(\pitk)$ decreases (\cref{lem:kimi-norm-localization-descent}).
Using this, we prove that $\pitk[K]$ converges on $\hLkiminorm$ (\cref{eq:proof:kimi-norm-shared-empirical-excess}),
which we then transfer to the auxiliary population objective $\tLkiminorm$ (\cref{eq:proof:kimi-norm-shared-auxiliary-bound})
and finally to $\mset$ (\cref{eq:proof:kimi-norm-shared-mse-bound}).

\subsubsection{Proof of \cref{thm:kimi-norm-shared-inner}}
\begin{proof}[\pfref{thm:kimi-norm-shared-inner}]
  By \cref{lem:kimi-norm-estimate-range} and the reward bound in
  \cref{eq:kimi-reward-range},
  \begin{align}
    \abs*{\hbt(x)-r(x,y)}
    \le\max_{y'}r(x,y')-\min_{y'}r(x,y')\le\Dstar,
    \qquad x\in\cX,\ y\in\cY.
    \label{eq:kimi-norm-estimate-residual-bound}
  \end{align}
  Applying \cref{lem:kimi-local-concentration} with $A=\Dstar$ and
  $R=4\eta\Bstar$, with probability at least $1-\delta$,
  the following inequalities hold simultaneously for every $\theta$ with
  $\nbr*{\theta-\thetat}_2\le4\eta\Bstar$,
  \begin{align}
    \begin{split}
      \abs*{\inner*{
        \nabla\hLkiminorm(\pi_\theta)
        -\nabla\tLkiminorm(\pi_\theta),
        \theta-\thetat-\eta\thetastar}}
      &\le\frac1{64}\mset(\pi_\theta)+\zeta_N(\delta;4\eta\Bstar,\Dstar),
      \\
      \abs*{
        \hLkiminorm(\pi_\theta)-\hLkiminorm(\piplust)
        -\tLkiminorm(\pi_\theta)+\tLkiminorm(\piplust)}
      &\le\frac1{64}\mset(\pi_\theta)+\zeta_N(\delta;4\eta\Bstar,\Dstar).
      \label{eq:kimi-norm-concentration}
    \end{split}
  \end{align}
  Here $\zeta_N$ is the statistical error defined in
  \cref{eq:kimi-local-statistical-error}.

  For the remainder of this proof we condition on \cref{eq:kimi-norm-concentration} holding.
  By \cref{lem:kimi-norm-estimation-bias,lem:mean-kimi-baseline-gap},
  \begin{align}
    \tbt(x)-r(x,\pit)
    \le\btstar(x)-r(x,\pit)\le\frac{\eta\Dstar^2}{8},
    \qquad x\in\cX.
  \end{align}
  We apply \cref{eq:kimi-local-empirical-progress} in \cref{lem:kimi-local-progress}
  with $A=\Dstar$, $a=\eta\Dstar^2/8$, and $R=4\eta\Bstar$.
  Then simultaneously for every $\theta$ with $\nbr*{\theta-\thetat}_2\le4\eta\Bstar$,
  \begin{align}
    \inner*{\nabla\hLkiminorm(\pi_\theta),
      \theta-\thetat-\eta\thetastar}
    &\ge
    \hLkiminorm(\pi_\theta)-\hLkiminorm(\piplust)
    + \rbr*{\frac{31}{32} - \frac{e^{4 / 3}}{72}}\mset(\pi_\theta)
      -2\zeta_N(\delta;4\eta\Bstar,\Dstar).
    \\
    &\ge
    \hLkiminorm(\pi_\theta)-\hLkiminorm(\piplust)
      -2\zeta_N(\delta;4\eta\Bstar,\Dstar).
    \label{eq:kimi-norm-empirical-gradient-bound}
 \end{align}
  Under the above inequalities, we have the following lemma,
  which is proved at the end of this section.
  It says that, throughout the inner loop,
  the inner policy iterates stay within a parameter ball of constant radius,
  and have monotonically decreasing empirical loss $\hLkiminorm$.\loose
  \begin{lemma}[Localization and empirical descent for the normalizer baseline]
    \label{lem:kimi-norm-localization-descent}
    Under the conditions on $\eta$, $G$, $N$, $\alpha$, and $K$ in \cref{thm:kimi-norm-shared-inner},
    the concentration event \cref{eq:kimi-norm-concentration} implies
    \begin{align}
      \max_{0\le k\le K}\nbr*{\frac{\thetatk-\thetat}{\eta}-\thetastar}_2
      \le2\Bstar,
      \qquad
      \max_{0\le k\le K}\nbr*{s_t^{\pitk}}_\infty\le2\Dstar.
      \label{eq:kimi-norm-localization-residual}
    \end{align}
    Moreover, for every $0\le k<K$,
    \begin{align}
      \hLkiminorm(\pitk)-\hLkiminorm(\pitkp)
      \ge\frac\alpha2\nbr*{\nabla\hLkiminorm(\pitk)}_2^2.
      \label{eq:kimi-norm-empirical-descent}
    \end{align}
  \end{lemma}

  The parameter bound in \cref{eq:kimi-norm-localization-residual} of \cref{lem:kimi-norm-localization-descent} therefore gives
  \begin{align}
    \nbr*{\thetatk-\thetat}_2\le3\eta\Bstar,
    \qquad 0\le k\le K.
    \label{eq:kimi-norm-inner-parameter-bound}
  \end{align}
  
  \para{Convergence on empirical loss}
  Using this and \cref{lem:kimi-norm-localization-descent},
  we next prove that $\pitk[K]$ converges on the empirical loss $\hLkiminorm$.
  Applying the empirical first-order convexity bound in
  \cref{eq:kimi-norm-empirical-gradient-bound} at $\thetatk$, then expanding
  the squared distance after the gradient update, gives, for every $0\le k<K$,
  \begin{align}
    \hLkiminorm(\pitk)-\hLkiminorm(\piplust)
    &\le\inner*{\nabla\hLkiminorm(\pitk),
      \thetatk-\thetat-\eta\thetastar}
      +2\zeta_N(\delta;4\eta\Bstar,\Dstar)
    \\
    &=\frac{
      \nbr*{\thetatk-\thetat-\eta\thetastar}_2^2
      -\nbr*{\thetatkp-\thetat-\eta\thetastar}_2^2}{2\alpha}
    \\
    &\qquad+\frac\alpha2\nbr*{\nabla\hLkiminorm(\pitk)}_2^2
    +2\zeta_N(\delta;4\eta\Bstar,\Dstar).
    \label{eq:kimi-norm-one-step}
  \end{align}
  Using this result,
  below we bound the excess empirical loss at $\pitk[k+1]$ in terms of the difference in parameter norm between $\pitk$ and $\pitk[k+1]$.
  \begin{align}
    \hLkiminorm(\pitkp)-\hLkiminorm(\piplust)
    &=
    \hLkiminorm(\pitk)-\hLkiminorm(\piplust)
      -\rbr*{\hLkiminorm(\pitk)-\hLkiminorm(\pitkp)}
    \\
    &\le\hLkiminorm(\pitk)-\hLkiminorm(\piplust)
      -\frac\alpha2\nbr*{\nabla\hLkiminorm(\pitk)}_2^2
    \\
    &\le
    \frac{
      \nbr*{\thetatk-\thetat-\eta\thetastar}_2^2
      -\nbr*{\thetatkp-\thetat-\eta\thetastar}_2^2}{2\alpha}
      +2\zeta_N(\delta;4\eta\Bstar,\Dstar).
      \label{eq:kimi-norm-descent}
  \end{align}
  The second line uses \cref{eq:kimi-norm-empirical-descent} from \cref{lem:kimi-norm-localization-descent},
  and we substitute in \cref{eq:kimi-norm-one-step} to obtain the last line.

  We now invoke \cref{eq:kimi-norm-descent} to prove that $\pitk[K]$ has small excess empirical loss,
  \begin{align}
    \hLkiminorm(\pitk[K])-\hLkiminorm(\piplust)
    &\le\frac1K\sum_{k=0}^{K-1}
      \rbr*{\hLkiminorm(\pitkp)-\hLkiminorm(\piplust)}
    \\
    &\le\frac1{2\alpha K}\sum_{k=0}^{K-1}
      \sbr*{\nbr*{\thetatk-\thetat-\eta\thetastar}_2^2
        -\nbr*{\thetatkp-\thetat-\eta\thetastar}_2^2}
      +2\zeta_N(\delta;4\eta\Bstar,\Dstar)
    \\
    &=\frac{
      \nbr*{\thetatk[0]-\thetat-\eta\thetastar}_2^2
      -\nbr*{\thetatk[K]-\thetat-\eta\thetastar}_2^2}{2\alpha K}
      +2\zeta_N(\delta;4\eta\Bstar,\Dstar)
    .
  \end{align}
  The first line uses $\hLkiminorm(\pitk) - \hLkiminorm(\pitk[k+1]) \ge 0$ from \cref{eq:kimi-norm-empirical-descent} in \cref{lem:kimi-norm-localization-descent},
  and the second line uses \cref{eq:kimi-norm-descent}.
  Finally, using $\thetatk[0]=\thetat$ and $\nbr*{\thetastar}_2=\Bstar$, we obtain
  \begin{align}
    \hLkiminorm(\pitk[K])-\hLkiminorm(\piplust)
    &\le
      \frac{\nbr*{\thetatk[0]-\thetat-\eta\thetastar}_2^2}{2\alpha K}
      +2\zeta_N(\delta;4\eta\Bstar,\Dstar)
    \\
    &\le\frac{\eta^2\Bstar^2}{2\alpha K}
      +2\zeta_N(\delta;4\eta\Bstar,\Dstar)
    .
    \label{eq:proof:kimi-norm-shared-empirical-excess}
  \end{align}

  \para{Small mean-squared error from empirical loss}
  We now use \cref{eq:proof:kimi-norm-shared-empirical-excess} to bound
  $\mset(\pitk[K])$.
  By \cref{eq:kimi-local-objective-decomposition} in
  \cref{lem:kimi-local-objectives},
  \begin{align}
    \mset(\pitk[K])\le\tLkiminorm(\pitk[K]).
    \label{eq:mse-kimi-norm}
  \end{align}
  It therefore suffices to bound $\tLkiminorm(\pitk[K])$ through $\hLkiminorm(\pitK)$.
  Applying the second inequality in \cref{eq:kimi-norm-concentration},
  which relates $\tLkiminorm$ to $\hLkiminorm$,
  at $\thetatk[K]$ and then substituting the empirical convergence result in
  \cref{eq:proof:kimi-norm-shared-empirical-excess} gives\loose
  \begin{align}
    \tLkiminorm(\pitk[K])-\tLkiminorm(\piplust)
    &\le\hLkiminorm(\pitk[K])-\hLkiminorm(\piplust)
      +\frac1{64}\mset(\pitk[K])+\zeta_N(\delta;4\eta\Bstar,\Dstar)
    \\
    &\le\frac{\eta^2\Bstar^2}{2\alpha K}
      +\frac1{64}\mset(\pitk[K])+3\zeta_N(\delta;4\eta\Bstar,\Dstar)
    \\
    &\le\frac{\eta^2\Bstar^2}{2\alpha K}
      +\frac1{64}\tLkiminorm(\pitk[K])+3\zeta_N(\delta;4\eta\Bstar,\Dstar),
    \label{eq:proof:kimi-norm-shared-excess-concentration}
  \end{align}
  where we use \cref{eq:mse-kimi-norm} in the last line above.
  Rearranging \cref{eq:proof:kimi-norm-shared-excess-concentration}, we obtain
  \begin{align}
    \frac{63}{64}\tLkiminorm(\pitk[K])
    &\le\frac{\eta^2\Bstar^2}{2\alpha K}
     +\tLkiminorm(\piplust)+3\zeta_N(\delta;4\eta\Bstar,\Dstar)
    \\
    &\le
    \einsq/2 +\tLkiminorm(\piplust)+3\zeta_N(\delta;4\eta\Bstar,\Dstar).
    \label{eq:proof:kimi-norm-shared-auxiliary-bound}
  \end{align}
  The last line above uses the choice of $K$ in \cref{eq:kimi-norm-shared-inner-parameters},
  which gives $\eta^2\Bstar^2/(2\alpha K)\le\einsq/2$.
  We bound the remaining terms on the RHS of \cref{eq:proof:kimi-norm-shared-auxiliary-bound}.
  Invoking \cref{lem:kimi-norm-comparator-loss}, which upper bounds $\tLkiminorm(\piplust)$ in terms of $G$,
  and combining with the theorem condition on $G$ (\cref{eq:kimi-norm-shared-group-size}),
  we obtain
  \begin{align}
    \tLkiminorm(\piplust)
    \le\frac{\rbr*{e^{\eta\Dstar}-1}^4}{64\eta^2G^2}
    \le\frac{\einsq}{4}.
    \label{eq:proof:kimi-norm-shared-comparator}
  \end{align}
  Finally, since $\eta\Dstar\le1/3$, the definition of $\zeta_N$ in
  \cref{eq:kimi-local-statistical-error} at $R=4\eta\Bstar$ and the bound
  on $N$ in \cref{eq:kimi-norm-shared-inner-parameters} give
  \begin{align}
    \zeta_N(\delta;4\eta\Bstar,\Dstar)\le\einsq/128
    \label{eq:proof:kimi-norm-shared-statistical-error}
  \end{align}
  for sufficiently large $C$.
  Substituting the previous two panels into \cref{eq:proof:kimi-norm-shared-auxiliary-bound} yields
  \begin{align}
    \mset(\pitk[K])\le\tLkiminorm(\pitk[K])
    \le\frac{64}{63}\rbr*{\frac12+\frac14+\frac3{128}}\einsq
    \le\einsq.
    \label{eq:proof:kimi-norm-shared-mse-bound}
  \end{align}
  This proves the first result in \cref{eq:kimi-norm-shared-inner-conclusion} of \cref{thm:kimi-norm-shared-inner}.

  Finally, \cref{eq:kimi-norm-localization-residual} from \cref{lem:kimi-norm-localization-descent} directly gives
  $\nbr*{s_t^{\pitp}}_\infty = \nbr{s_t^{\pitk[K]}}_\infty\le2\Dstar$, the other result in \cref{thm:kimi-norm-shared-inner}.
  This completes the proof.
\end{proof}

\subsubsection{Proof of \cref{lem:kimi-norm-localization-descent}}
\begin{proof}[\pfref{lem:kimi-norm-localization-descent}]
  Throughout this proof,
  we condition on the event that the concentration inequalities in \cref{eq:kimi-norm-concentration} hold.

  Under this event,
  we will prove by induction that all policy iterates $\pitk$ satisfy \cref{eq:kimi-norm-empirical-descent}
  and stay within the parameter ball given by \cref{eq:kimi-norm-localization-residual}.
  Initially, $\thetatk[0]=\thetat$ and
  $\nbr*{\thetatk[0]-\thetat-\eta\thetastar}_2^2=\eta^2\Bstar^2$.
  Suppose $\nbr*{\thetatk[j]-\thetat-\eta\thetastar}_2\le2\eta\Bstar$
  for every $0\le j\le k$, where $k<K$,
  which implies that $\nbr*{\thetatk-\thetat}_2\le3\eta\Bstar$.
  We will then show for the $k+1$th inner step that
  $\nbr*{\thetatk[k+1] - \thetat - \eta\thetastar}_2 \le 2\eta\Bstar$ 
  and that $\hLkiminorm(\pitk) - \hLkiminorm(\pitk[k+1]) \ge \frac{\alpha}{2} \nbr{\nabla \hLkiminorm(\pitk)}_2^2$.

  \para{Proving empirical descent (\cref{eq:kimi-norm-empirical-descent})}
  We first prove the latter statement.
  The estimator bound \cref{eq:kimi-norm-estimate-residual-bound} allows us
  to apply \cref{eq:kimi-local-gradient-bound} in \cref{lem:kimi-local-regularity}
  with $A=\Dstar$ and $R=3\eta\Bstar$, giving
  \begin{align}
    \nbr*{\nabla\hLkiminorm(\pitk)}_2
    \le\frac{2\Dfeat}{\eta^2}\rbr*{3\eta\Dstar+\eta\Dstar}
    =\frac{8\Dfeat\Dstar}{\eta}.
  \end{align}
  Then,
  for the update in Line~\ref{line:kimi-norm-update} of \cref{alg:kimi-norm},
  using the above inequalities we obtain
  \begin{align}
    \nbr*{\thetatkp-\thetat}_2
    &=
    \nbr*{\thetatk - \alpha \nabla \hLkiminorm(\pitk) -\thetat}_2
    \\
    &\le\nbr*{\thetatk-\thetat}_2
      +\alpha\nbr*{\nabla \hLkiminorm(\pitk)}_2
    \\
    &\le
    3\eta\Bstar+\frac{8\alpha\Dfeat\Dstar}{\eta}
    \\
    &\le
    4\eta\Bstar,
    \label{eq:proof:kimi-norm-shared-next-parameter-bound}
  \end{align}
  where in the last inequality we use the theorem condition on $\alpha$ (\cref{eq:kimi-norm-shared-inner-parameters}).

  We now use Taylor's theorem to bound $\hLkiminorm(\pitk)-\hLkiminorm(\pitkp)$.
  The inductive hypothesis and \cref{eq:proof:kimi-norm-shared-next-parameter-bound}
  give
  \begin{align}
    \nbr*{\thetatk-\thetat}_2\le4\eta\Bstar,
    \qquad
    \nbr*{\thetatkp-\thetat}_2\le4\eta\Bstar.
  \end{align}
  We again invoke \cref{lem:kimi-local-regularity}, but with $A=\Dstar$ and $R=4\eta\Bstar$,
  and apply the upper bound on the Hessian in \cref{eq:kimi-local-smoothness}.
  Since $\eta\Dstar\le1/3$, for every $\theta$ with $\nbr*{\theta-\thetat}_2\le4\eta\Bstar$,
  this gives,
  \begin{align}
    \nbr*{\nabla^2\hLkiminorm(\pi_\theta)}_{\op}
    \le\frac{\Dfeat^2}{\eta^2}\max\cbr*{2,1+\frac{5\eta\Dstar}{2}}
    =\frac{2\Dfeat^2}{\eta^2}.
    \label{eq:proof:kimi-norm-shared-smoothness}
  \end{align}
  By convexity of $\nbr{\cdot}_2$, the Hessian bound in
  \cref{eq:proof:kimi-norm-shared-smoothness} holds for convex combinations of $\thetatk$ and $\thetatkp$.
  Applying Taylor's theorem at $\thetatk$, we obtain
  \begin{align}
    \hLkiminorm(\pitk)-\hLkiminorm(\pitkp)
    &\ge-\inner*{\nabla\hLkiminorm(\pitk),\thetatkp-\thetatk}
      -\frac{\Dfeat^2}{\eta^2}\nbr*{\thetatkp-\thetatk}_2^2
    \\
    &=\alpha\rbr*{1-\frac{\alpha\Dfeat^2}{\eta^2}}
      \nbr*{\nabla\hLkiminorm(\pitk)}_2^2
    \\
    &\ge\frac\alpha2\nbr*{\nabla\hLkiminorm(\pitk)}_2^2.
    \label{eq:proof:kimi-norm-shared-descent}
  \end{align}
  The second line uses
  $\thetatkp-\thetatk=-\alpha\nabla\hLkiminorm(\pitk)$
  from Line~\ref{line:kimi-norm-update} of \cref{alg:kimi-norm}.
  The last inequality uses $\alpha\le\eta^2/(8\Dfeat^2)$ from
  the theorem conditions in \cref{eq:kimi-norm-shared-inner-parameters}.
  This completes the proof of \cref{eq:kimi-norm-empirical-descent} at step $k$.

  \para{Proving the parameter bound (\cref{eq:kimi-norm-localization-residual})}
  It remains to prove $\nbr*{\thetatkp-\thetat-\eta\thetastar}_2\le2\eta\Bstar$.
  The inductive hypothesis gives $\nbr*{\thetatk-\thetat}_2\le3\eta\Bstar$.
  Expanding the squared distance after the gradient update and applying the empirical first-order convexity bound in
  \cref{eq:kimi-norm-empirical-gradient-bound} at $\thetatk$, we obtain
  \begin{align}
    \nbr*{\thetatkp-\thetat-\eta\thetastar}_2^2
      -\nbr*{\thetatk-\thetat-\eta\thetastar}_2^2
    &\le-2\alpha\rbr*{\hLkiminorm(\pitk)-\hLkiminorm(\piplust)}
    \\
    &\qquad+\alpha^2\nbr*{\nabla\hLkiminorm(\pitk)}_2^2
      +4\alpha\zeta_N(\delta;4\eta\Bstar,\Dstar).
    \label{eq:proof:kimi-norm-shared-one-step-progress}
  \end{align}

  We now use \cref{eq:proof:kimi-norm-shared-descent} to bound
  $\alpha^2\nbr*{\nabla\hLkiminorm(\pitk)}_2^2$
  in \cref{eq:proof:kimi-norm-shared-one-step-progress}, yielding
  \begin{align}
    \nbr*{\thetatkp-\thetat-\eta\thetastar}_2^2
      -\nbr*{\thetatk-\thetat-\eta\thetastar}_2^2
    &\le-2\alpha\rbr*{\hLkiminorm(\pitk)-\hLkiminorm(\piplust)}
    \\
    &\qquad+2\alpha\rbr*{\hLkiminorm(\pitk)-\hLkiminorm(\pitkp)}
      +4\alpha\zeta_N(\delta;4\eta\Bstar,\Dstar)
    \\
    &=-2\alpha\rbr*{\hLkiminorm(\pitkp)-\hLkiminorm(\piplust)}
      +4\alpha\zeta_N(\delta;4\eta\Bstar,\Dstar).
    \label{eq:proof:kimi-norm-shared-next-excess}
  \end{align}
  Rearranging \cref{eq:proof:kimi-norm-shared-next-excess} gives
  \begin{align}
    \nbr*{\thetatkp-\thetat-\eta\thetastar}_2^2
    \le\nbr*{\thetatk-\thetat-\eta\thetastar}_2^2
      -2\alpha\rbr*{\hLkiminorm(\pitkp)-\hLkiminorm(\piplust)}
      +4\alpha\zeta_N(\delta;4\eta\Bstar,\Dstar).
    \label{eq:proof:kimi-norm-shared-distance-excess}
  \end{align}
  We now lower bound $\hLkiminorm(\pitkp)-\hLkiminorm(\piplust)$ in
  \cref{eq:proof:kimi-norm-shared-distance-excess}.
  Since $\nbr*{\thetatkp-\thetat}_2\le4\eta\Bstar$ by
  \cref{eq:proof:kimi-norm-shared-next-parameter-bound},
  the second concentration inequality in \cref{eq:kimi-norm-concentration} applies at $\thetatkp$, giving
  \begin{align}
    \hLkiminorm(\pitkp)-\hLkiminorm(\piplust)
    &\ge\tLkiminorm(\pitkp)-\tLkiminorm(\piplust)
      -\frac1{64}\mset(\pitkp)-\zeta_N(\delta;4\eta\Bstar,\Dstar)
    \\
    &\ge\frac{63}{64}\tLkiminorm(\pitkp)
      -\tLkiminorm(\piplust)-\zeta_N(\delta;4\eta\Bstar,\Dstar)
    \\
    &\ge-\tLkiminorm(\piplust)-\zeta_N(\delta;4\eta\Bstar,\Dstar).
    \label{eq:proof:kimi-norm-shared-excess-lower-bound}
  \end{align}
  The second inequality uses $\mset(\pitkp)\le\tLkiminorm(\pitkp)$ from \cref{eq:mse-kimi-norm}.
  Substituting \cref{eq:proof:kimi-norm-shared-excess-lower-bound} into \cref{eq:proof:kimi-norm-shared-distance-excess} now gives
  \begin{align}
    \nbr*{\thetatkp-\thetat-\eta\thetastar}_2^2
    \le\nbr*{\thetatk-\thetat-\eta\thetastar}_2^2
      +2\alpha\rbr*{\tLkiminorm(\piplust)+3\zeta_N(\delta;4\eta\Bstar,\Dstar)}.
    \label{eq:proof:kimi-norm-shared-distance-auxiliary}
  \end{align}
  By \cref{eq:proof:kimi-norm-shared-comparator,eq:proof:kimi-norm-shared-statistical-error}
  in the proof of \cref{thm:kimi-norm-shared-inner},
  \begin{align}
    \tLkiminorm(\piplust)&\le\frac{\einsq}{4},
    &\zeta_N(\delta;4\eta\Bstar,\Dstar)&\le\frac{\einsq}{128}.
  \end{align}
  Substituting these bounds into \cref{eq:proof:kimi-norm-shared-distance-auxiliary} yields
  \begin{align}
    \nbr*{\thetatkp-\thetat-\eta\thetastar}_2^2
    &\le\nbr*{\thetatk-\thetat-\eta\thetastar}_2^2
      +2\alpha\rbr*{\frac{\einsq}{4}+\frac{3\einsq}{128}}
    \\
    &=\nbr*{\thetatk-\thetat-\eta\thetastar}_2^2
      +\frac{35\alpha\einsq}{64}.
    \label{eq:proof:kimi-norm-shared-distance-recursion}
  \end{align}
  The choices of $K$ and $\alpha$ in \cref{eq:kimi-norm-shared-inner-parameters}
  and $\einsq\le\Dstar^2$ give
  \begin{align}
    \alpha K\einsq
    \le\eta^2\Bstar^2+\alpha\einsq
    \le\frac98\eta^2\Bstar^2.
    \label{eq:proof:kimi-norm-shared-alpha-k-bound}
  \end{align}
  Because the inductive hypothesis assumes $\nbr*{\thetatk[j]-\thetat-\eta\thetastar}_2\le2\eta\Bstar$
  for every $0\le j\le k$,
  \cref{eq:proof:kimi-norm-shared-distance-recursion} also holds for every $0 \le j \le k$.
  Then summing \cref{eq:proof:kimi-norm-shared-distance-recursion} over these steps gives
  \begin{align}
    \nbr*{\thetatkp-\thetat-\eta\thetastar}_2^2
    &=\nbr*{\thetatk[0]-\thetat-\eta\thetastar}_2^2
      +\sum_{j=0}^k\sbr*{
        \nbr*{\thetatk[j+1]-\thetat-\eta\thetastar}_2^2
        -\nbr*{\thetatk[j]-\thetat-\eta\thetastar}_2^2}
    \\
    &\le\eta^2\Bstar^2
      +\frac{35\alpha(k+1)\einsq}{64}
    \\
    &\le\eta^2\Bstar^2+\frac{35\alpha K\einsq}{64}
    <4\eta^2\Bstar^2.
    \label{eq:proof:kimi-norm-shared-accumulated-distance}
  \end{align}
  The first inequality uses \cref{eq:proof:kimi-norm-shared-distance-recursion}
  and $\thetatk[0]=\thetat$,
  and the last line uses \cref{eq:proof:kimi-norm-shared-alpha-k-bound}.
  This completes the induction and proves the parameter bound in
  \cref{eq:kimi-norm-localization-residual},
  \begin{align}
    \max_{0\le k\le K}
    \nbr*{\frac{\thetatk-\thetat}{\eta}-\thetastar}_2\le2\Bstar.
    \label{eq:proof:kimi-norm-shared-localization}
  \end{align}

  Finally, we use \cref{eq:proof:kimi-norm-shared-localization} to prove
  $\nbr{s_t^{\pitk}}_\infty\le2\Dstar$.
  The definition of $s_t^\pi$ in \cref{eq:kimi-residual} and
  \cref{asm:softmax-linear} give, for every $0\le k\le K$,
  \begin{align}
    s_t^{\pitk}(x,y)
    &=\frac1\eta\log\frac{\pitk(y\mid x)}{\pit(y\mid x)}-r(x,y)
      -\En_{y'\sim\pit(\cdot\mid x)}\sbr*{
        \frac1\eta\log\frac{\pitk(y'\mid x)}{\pit(y'\mid x)}-r(x,y')}
    \\
    &=\inner*{
      \phi(x,y)-\En_{y'\sim\pit(\cdot\mid x)}\sbr*{\phi(x,y')},
      \frac{\thetatk-\thetat}{\eta}-\thetastar}.
  \end{align}
  By the definition of $\Dfeat$,
  $\nbr*{\phi(x,y)-\En_{y'\sim\pit(\cdot\mid x)}\sbr*{\phi(x,y')}}_2\le\Dfeat$.
  Cauchy--Schwarz and \cref{eq:proof:kimi-norm-shared-localization} therefore give
  \begin{align}
    \nbr*{s_t^{\pitk}}_\infty
    \le\Dfeat\nbr*{\frac{\thetatk-\thetat}{\eta}-\thetastar}_2
    \le2\Dstar,
    \qquad 0\le k\le K.
  \end{align}
  This proves \cref{eq:kimi-norm-localization-residual}.
\end{proof}

\subsection{Proof of \cref{cor:kimi-norm-regret}}\label{app:kimi-norm-shared-global}
\label{app:kimi-norm-global}

\begin{corollary}[\kiminorm is \covpres (full version of \protect\cref{cor:kimi-norm-regret})]
  \label{cor:kimi-norm-regret-full}
  Suppose that \cref{asm:softmax-linear} holds.
  Fix $\veps\in(0,\Rmax]$ and $\delta\in(0,1)$.
  Run \cref{alg:kimi-norm} with $\pit[0]=\piref$ and
  $\eta=1/(3\Rmax+4\Dstar)$.
  If $\veps\ge\Dstar$, take $T=0$, with arbitrary $\alpha>0$ and integers
  $G,N,K\ge1$.
  Otherwise, set
  \begin{align}
    T=\left\lceil\frac{\log(2\Cresp)}{\eta\veps}\right\rceil,
    \label{eq:kimi-regret-outer-parameters}
  \end{align}
  and choose the inner loop parameters as
  \begin{align}
    \einsq&=\frac{1}{12\Cresp\eta^2T^2\log(eT)},
    &\alpha&=\frac{\eta^2}{8\Dfeat^2},
    &K&=\left\lceil\frac{8\Dstar^2}{\einsq}\right\rceil,
    \label{eq:kimi-regret-inner-parameters}
  \end{align}
  and let $G,N$ be the smallest positive integers satisfying
  \begin{align}
    G&\ge\frac{\rbr*{e^{\eta\Dstar}-1}^2}{4\eta\sqrt{\einsq}},
    &N&\ge\frac{C\Dstar^2}{\einsq}
      \rbr*{d\log(1+8N)+\log\frac{4(d+1)T}{\delta}},
    \label{eq:kimi-regret-sample-parameters}
  \end{align}
  where $C$ is the universal constant in \cref{thm:kimi-norm-shared-inner}.
  Then the returned policy $\pifin=\pit[T]$ satisfies
  \begin{align}
    \Jstar-J(\pifin)\le\veps
    \label{eq:kimi-norm-full-regret-conclusion}
  \end{align}
  with probability at least $1-\delta$ over the sampled training data.
  The numbers of gradient steps and sampled responses are
  \begin{align}
    \Nstep(\veps,\delta)=KT
    &=\wt O\rbr*{\frac{\Cresp\Dstar^2}{\eta\veps^3}},
    \\
    \Nsample(\veps,\delta)=N(G+1)T
    &=\wt O\rbr*{
      \frac{\Cresp\Dstar^2\rbr*{d+\log(1/\delta)}}{\eta\veps^3}
      \rbr*{1+\frac{\eta\sqrt{\Cresp}\Dstar^2}{\veps}}}.
    \label{eq:kimi-norm-full-regret-complexity}
  \end{align}
\end{corollary}

\begin{proof}[\pfref{cor:kimi-norm-regret-full}]
  We prove the result supposing $\veps<\Dstar$
  as otherwise we immediately have $\Jstar-J(\piref)\le\Dstar\le\veps$ from \cref{eq:kimi-reward-range}.

  We show that the hyperparameter choices in \cref{cor:kimi-norm-regret-full} satisfy the conditions for \cref{thm:mse-regret-main}.
  For each outer iteration $t$,
  we invoke \cref{thm:kimi-norm-shared-inner}, the inner convergence guarantee,
  with failure probability $\delta/T$.
  We check that the corollary parameters match the theorem requirements.
  First, the bounds on $G$ and $N$ in \cref{eq:kimi-regret-sample-parameters} in \cref{cor:kimi-norm-regret-full}
  match those in \cref{thm:kimi-norm-shared-inner} (\cref{eq:kimi-norm-shared-group-size,eq:kimi-norm-shared-inner-parameters}) with $\delta / T$.
  The inner learning rate and number of steps in \cref{eq:kimi-regret-inner-parameters},
  \begin{align}
    \alpha=\frac{\eta^2}{8\Dfeat^2},
    \qquad
    K=\left\lceil\frac{8\Dstar^2}{\einsq}\right\rceil
      =\left\lceil\frac{\eta^2\Bstar^2}{\alpha\einsq}\right\rceil,
  \end{align}
  also satisfy \cref{eq:kimi-norm-shared-inner-parameters}.
  Finally, $\eta\Dstar\le1/4$, and
  \cref{eq:kimi-regret-outer-parameters,eq:kimi-regret-inner-parameters}
  give $0<\einsq\le\veps^2<\Dstar^2$.
  Applying \cref{thm:kimi-norm-shared-inner} conditional on the preceding history at each
  iteration and taking a union bound over the $T$ iterations gives, with probability at least
  $1-\delta$ over the sampled training data,
  \begin{align}
    \mset(\pitp)&\le\einsq,
    &\nbr*{s_t^{\pitp}}_\infty&\le2\Dstar,
    \qquad 0\le t<T.
    \label{eq:kimi-regret-inner-event}
  \end{align}
  We condition on this event for the remainder of the proof.

  We next verify \cref{eq:framework-mse-tolerance} in \cref{thm:mse-regret-main}.
  By \cref{eq:kimi-residual,eq:kimi-regret-inner-event},
  $S=\max_{0\le t<T}\nbr*{s_t^{\pitp}}_\infty\le2\Dstar$,
  so $e^{\eta S}\le e^{1/2}<2$.
  Using \cref{eq:kimi-regret-inner-event,eq:kimi-regret-inner-parameters},
  \begin{align}
    \mset(\pitp)
    \le\einsq
    =\frac{1}{12\Cresp\eta^2T^2\log(eT)}
    \le\frac{1}{4\eta^2T^2\rbr*{\Cresp\log(eT)+e^{\eta S}}},
    \qquad 0\le t<T.
  \end{align}
  The last inequality uses
  $\Cresp\log(eT)+e^{\eta S}\le3\Cresp\log(eT)$, since $\Cresp\log(eT)\ge1$.

  Next, under \cref{asm:softmax-linear} we have $\pit > 0$ for all $0 \le t \le T$,
  and $\pit[0]=\piref$.
  Applying \cref{thm:mse-regret-main} and substituting \cref{eq:kimi-regret-outer-parameters} therefore gives
  \begin{align}
    \Jstar-J(\pifin)
    =\Jstar-J(\pit[T])
    \le\frac{\log(2\Cresp)}{\eta T}
    \le\veps.
  \end{align}
  This proves \cref{eq:kimi-norm-full-regret-conclusion} with probability at least $1-\delta$.

  Finally, we bound the numbers of gradient steps and sampled responses.
  Since $\eta\veps\le\eta\Rmax\le1/3$ and $\Cresp\ge1$,
  the choice of $T$ gives
  \begin{align}
    \frac{\log(2\Cresp)}{\eta\veps}
    \le T\le\frac{2\log(2\Cresp)}{\eta\veps}.
    \label{eq:proof:kimi-outer-bounds}
  \end{align}
  Substituting \cref{eq:proof:kimi-outer-bounds} into the definition of $\einsq$
  in \cref{eq:kimi-regret-inner-parameters} gives
  \begin{align}
    \frac{1}{\einsq}
    =12\Cresp\eta^2T^2\log(eT)
    =\wt O\rbr*{\frac{\Cresp}{\veps^2}}.
    \label{eq:proof:kimi-tolerance-scaling}
  \end{align}
  By \cref{eq:proof:kimi-tolerance-scaling}, the choice of $K$ in
  \cref{eq:kimi-regret-inner-parameters} and the smallest admissible $N$ in
  \cref{eq:kimi-regret-sample-parameters} satisfy
  \begin{align}
    K&=\wt O\rbr*{\frac{\Cresp\Dstar^2}{\veps^2}},
    &N&=\wt O\rbr*{\frac{\Cresp\Dstar^2}{\veps^2}
      \rbr*{d+\log(T/\delta)}}.
    \label{eq:proof:kimi-inner-cost-scaling}
  \end{align}
  The $\log(1+8N)$ term in \cref{eq:kimi-regret-sample-parameters}
  contributes a logarithmic factor to this bound on $N$.
  Since $e^{\eta\Dstar}-1\le2\eta\Dstar$ for $\eta\Dstar\le1/4$,
  the choice of $G$ in \cref{eq:kimi-regret-sample-parameters} gives
  \begin{align}
    G\le1+\frac{\eta\Dstar^2}{\sqrt{\einsq}}
    =\wt O\rbr*{1+\frac{\eta\sqrt{\Cresp}\Dstar^2}{\veps}},
    \label{eq:proof:kimi-baseline-sample-count}
  \end{align}
  where the last bound uses \cref{eq:proof:kimi-tolerance-scaling}.
  Each outer iteration uses $K$ gradient steps and samples $N(G+1)$ responses.
  Substituting \cref{eq:proof:kimi-outer-bounds,eq:proof:kimi-inner-cost-scaling,eq:proof:kimi-baseline-sample-count}
  into $KT$ and $N(G+1)T$ gives
  \begin{align}
    \Nstep(\veps,\delta)=KT
    &=\wt O\rbr*{\frac{\Cresp\Dstar^2}{\eta\veps^3}},
    \\
    \Nsample(\veps,\delta)=N(G+1)T
    &=\wt O\rbr*{
      \frac{\Cresp\Dstar^2\rbr*{d+\log(1/\delta)}}{\eta\veps^3}
      \rbr*{1+\frac{\eta\sqrt{\Cresp}\Dstar^2}{\veps}}}.
  \end{align}
\end{proof}

\section{\rebelon: Global Convergence (Proof of \cref*{cor:rebel-main})}\label{app:rebel}
This section gives the algorithm and convergence analysis for \rebelon,
and is organized as follows.
\begin{itemize}
  \item \cref{app:rebel-overview} describes the full \rebelon algorithm (\cref{alg:dpo}),
  and \cref{cor:dpo-regret} is the formal global convergence result summarized in \cref{cor:rebel-main}.
  \item \cref{app:rebel-lemmas} contains the supporting lemmas for the proof of \cref{cor:dpo-regret}.
  \item \cref{app:rebel-inner-loop} establishes convergence of the inner optimization loop in \cref{alg:dpo}.
  \item \cref{app:rebel-global} proves \cref{cor:dpo-regret} by combining \cref{thm:mse-accum-regret} with the inner loop convergence in \cref{thm:dpo-inner}.
\end{itemize}

\para{Notation}
We assume the softmax linear policies and linear rewards in \cref{asm:softmax-linear},
with $r(x,y)\in[0,\Rmax]$, and write
\begin{align}
  \Bstar&\ldef{}\nbr*{\thetastar}_2,
  &\Dfeat&\ldef{}\max_{x,y,y'}\nbr*{\phi(x,y)-\phi(x,y')}_2,
  &\Dstar&\ldef{}\Dfeat\Bstar.
\end{align}
Thus $\abs*{r(x,y)-r(x,y')}\le\Dstar$ for responses to the same prompt.
In this section we use $\Dstar$ as the more natural bound on reward differences,
while the bounds in the main body use $B\ldef{}\Bfeat\Bstar$
to be consistent with \cref{sec:pg}.
Note that $\Dstar\le2B$.

We will also denote an error tolerance on $\mset$ as $\einsq$,
reserving $\ein$ for the KL tolerance in \cref{eq:framework-kl-tolerance}.

\subsection{\rebelon Algorithm}
\label{app:rebel-overview}

Recall the \rebelon objective in \cref{eq:rebel-objective},
\begin{align}
  \Ltrebelon(\pi)
  \ldef{}\En_{\substack{x\sim\mu,\\y,y'\overset{\mathrm{iid}}{\sim}\pit(\cdot\mid{}x)}}
  \sbr*{\rbr*{
    \frac1\eta\log\frac{\pi(y\mid{}x)}{\pit(y\mid{}x)}
    -\frac1\eta\log\frac{\pi(y'\mid{}x)}{\pit(y'\mid{}x)}
    -\rbr*{r(x,y)-r(x,y')}
  }^2}.
  \label{eq:rebel-population-loss}
\end{align}
The full \rebelon algorithm is displayed in \cref{alg:dpo}.
At outer iteration $t$, draw $N$ independent prompts $x_i\sim\mu$ and,
conditionally on each $x_i$, two independent responses $y_i,y_i'\sim\pit(\cdot\mid{}x_i)$.
Write $r_i=r(x_i,y_i)$ and $r_i'=r(x_i,y_i')$.
The empirical objective is
\begin{align}
  \hLtrebelon(\pi)
  \ldef{}\frac1N\sum_{i=1}^N
  \rbr*{
    \frac1\eta\log\frac{\pi(y_i\mid{}x_i)}{\pit(y_i\mid{}x_i)}
    -\frac1\eta\log\frac{\pi(y_i'\mid{}x_i)}{\pit(y_i'\mid{}x_i)}
    -\rbr*{r_i-r_i'}
  }^2.
  \label{eq:dpo-empirical-loss}
\end{align}
Starting from $\thetatk[0]=\thetat$, \cref{alg:dpo} takes $K$ gradient descent steps
on this objective and returns the last iterate as $\pitp$.
The batch and the policy $\pit$ in the denominator remain fixed during these steps.

\begin{algorithm}[t]
  \caption{\rebelon}
  \label{alg:dpo}
  \begin{algorithmic}[1]
    \Require \begingroup\raggedright
    \mbox{Base model $\piref$},
    \mbox{outer iterations $T$}, \mbox{regularization parameter $\eta>0$},
    \mbox{prompt batch size $N$}, \mbox{inner steps $K$}, \mbox{and learning rate $\alpha$}.\par
    \endgroup
    \State Initialize $\pit[0]=\piref=\pi_{\theta_0}$.
    \For{$t=0,1,\ldots,T-1$}
      \State Draw a batch of $N$ prompts $x_i\overset{\mathrm{iid}}{\sim}\mu$.
      \State For each prompt $x_i$, draw $y_i,y_i'\sim\pit(\cdot\mid{}x_i)$ independently,
      and set $r_i=r(x_i,y_i)$ and $r_i'=r(x_i,y_i')$.
      \State Initialize $\thetatk[0]=\thetat$ and $\pitk[0]=\pit$.
      \For{$k=0,1,\ldots,K-1$}
        \State Compute the empirical gradient (e.g., via backpropagation)
        
        \begin{equation}
          \begin{aligned}
            \nabla\hLtrebelon(\pitk)
            ={}&\frac{2}{\eta N}\sum_{i=1}^N
            \rbr*{\nabla\log\pitk(y_i\mid{}x_i)
              -\nabla\log\pitk(y_i'\mid{}x_i)}
            \\
            &\quad\times\rbr*{
              \frac1\eta\log\frac{\pitk(y_i\mid{}x_i)}{\pit(y_i\mid{}x_i)}
              -\frac1\eta\log\frac{\pitk(y_i'\mid{}x_i)}{\pit(y_i'\mid{}x_i)}
              -\rbr*{r_i-r_i'}
            }.
          \end{aligned}
          \label{eq:dpo-empirical-gradient}
        \end{equation}
        \State Update $\thetatkp=\thetatk-\alpha\nabla\hLtrebelon(\pitk)$
        and $\pitkp=\pi_{\thetatkp}$.
        \label{line:dpo-update}
      \EndFor
      \State Set $\thetatp=\thetatk[K]$ and $\pitp=\pitk[K]$.
    \EndFor
    \State \Return $\pifin=\pit[T]$.
  \end{algorithmic}
\end{algorithm}

The following corollary shows that \rebelon is \covpres,
and is the full version of the result presented in \cref{cor:rebel-main}.

\begin{corollary}[\rebelon is \covpres]
  \label{cor:dpo-regret}
  Suppose that \cref{asm:softmax-linear} holds.
  Fix $\veps\in(0,\Rmax]$ with $\veps<\Dstar$ and $\delta\in(0,1)$.
  Run \cref{alg:dpo} with $\pit[0]=\piref$ and
  \begin{align}
    \eta&=\frac{\log\rbr*{2\Cresp}}{\Dstar},
    &T&=\left\lceil\frac{\log\rbr*{2\Cresp}}{\eta\veps}\right\rceil.
    \label{eq:dpo-regret-outer-parameters}
  \end{align}
  Set
  \begin{align}
    \einsq&\ldef{}\frac{1}{12\Cresp\eta^2T^2\log(eT)},
    &\alpha&=\frac{\eta^2}{2\Dfeat^2},
    \label{eq:dpo-regret-inner-parameters}
    \\
    K&=\left\lceil\frac{2\Dstar^2}{\einsq}\right\rceil,
    &N&=\left\lceil\frac{28\Dstar^2}{3\einsq}\log\frac{2dT}{\delta}\right\rceil.
    \label{eq:dpo-regret-sample-parameters}
  \end{align}
  Then the returned policy satisfies
  \begin{align}
    \Jstar-J(\pifin)\le\veps
    \label{eq:dpo-regret-conclusion}
  \end{align}
  with probability at least $1-\delta$ over the sampled training data.
  On the same event, with probability at least
  $1-(2+\sqrt2)/(4\sqrt{\log(eT)})$ over $x\sim\mu$,
  \begin{align}
    \pit(\Ystarx\mid x)\ge\frac12\piref(\Ystarx\mid x),
    \qquad\forall 0\le t\le T.
    \label{eq:dpo-regret-coverage}
  \end{align}
  The numbers of gradient steps and sampled responses are
  \begin{align}
    \Nstep(\veps,\delta)=KT
    &=\wt O\rbr*{\frac{\Cresp\Dstar^3}{\veps^3}},
    \\
    \Nsample(\veps,\delta)=2NT
    &=\wt O\rbr*{\frac{\Cresp\Dstar^3}{\veps^3}\log\frac{2d}{\delta}}.
    \label{eq:dpo-regret-complexity}
  \end{align}
\end{corollary}

\para{Parameter choices}
The inner loop convergence analysis permits any $\eta>0$.
The choice in \cref{eq:dpo-regret-outer-parameters} balances the number of outer iterations
against the exponential factor in \cref{thm:mse-accum-regret}, giving
$e^{\eta\Dstar}=2\Cresp$.
For fixed problem parameters, both the number of gradient steps and the number of sampled responses are
$\wt O(1/\veps^3)$, and the response count depends only logarithmically on $d$ and $1/\delta$.

\subsection{Supporting Lemmas}
\label{app:rebel-lemmas}

The following lemmas connect optimization of the empirical objective to the
mean-squared error needed for the global convergence guarantee.
\begin{itemize}
  \item \cref{lem:rebel-mse} shows that $\Ltrebelon(\pi)=2\mset(\pi)$,
  so minimizing the population loss directly minimizes the mean-squared error.
  \item \cref{lem:dpo-convexity} shows that the population and empirical losses
  are convex quadratics and identifies the exact update $\piplust$ as a common
  minimizer with zero loss.
  \item \cref{lem:dpo-smoothness} bounds the smoothness of both losses.
  \item \cref{lem:dpo-population-transfer} bounds the population loss by the
  empirical loss and an error proportional to the squared parameter distance
  from the exact update, uniformly over $\theta$.
\end{itemize}
Fix an outer iteration $t$ and $\eta>0$ throughout this subsection.
Probabilities and expectations over the batch are conditional on $\pit$.
Recall that for any $\pi > 0$, the residual and mean-squared error are defined as
\begin{align}
  s_t^\pi(x,y)
  &\ldef{}\frac1\eta\log\frac{\pi(y\mid{}x)}{\pit(y\mid{}x)}
  -\En_{y'\sim\pit(\cdot\mid{}x)}\sbr*{\frac1\eta\log\frac{\pi(y'\mid{}x)}{\pit(y'\mid{}x)}}
  -\rbr*{r(x,y)-r(x,\pit)},
  \label{eq:rebel-residual}
  \\
  \mset(\pi)
  &\ldef{}\En_{x\sim\mu,\,y\sim\pit(\cdot\mid{}x)}\sbr*{s_t^\pi(x,y)^2}.
  \label{eq:rebel-mse-definition}
\end{align}

\begin{lemma}
  \label{lem:rebel-mse}
  For every policy $\pi > 0$,
  the \rebelon loss \cref{eq:rebel-population-loss} satisfies
  \begin{align}
    \mset(\pi)=\frac12\Ltrebelon(\pi).
    \label{eq:rebel-mse}
  \end{align}
\end{lemma}

\begin{proof}[\pfref{lem:rebel-mse}]
  For a fixed prompt $x$, the definition of $s_t^\pi$ in \cref{eq:rebel-residual} gives
  for a pair of responses $(y,y')$,
  \begin{align}
    s_t^\pi(x,y)-s_t^\pi(x,y')
    =\frac1\eta\log\frac{\pi(y\mid{}x)}{\pit(y\mid{}x)}
    -\frac1\eta\log\frac{\pi(y'\mid{}x)}{\pit(y'\mid{}x)}
    -r(x,y)+r(x,y').
    \label{eq:proof:rebel-residual-difference}
  \end{align}
  Thus,
  since $\En_{y\sim\pit(\cdot\mid{}x)}\sbr*{s_t^\pi(x,y)}=0$
  and the two responses are conditionally independent given $x$,
  \begin{align}
    \Ltrebelon(\pi)
    &=\En_{\substack{x\sim\mu,\\y,y'\overset{\mathrm{iid}}{\sim}\pit(\cdot\mid{}x)}}
      \sbr*{\rbr*{s_t^\pi(x,y)-s_t^\pi(x,y')}^2}
    \\
    &=2\En_{x\sim\mu,\,y\sim\pit(\cdot\mid{}x)}\sbr*{s_t^\pi(x,y)^2}
    =2\mset(\pi).
  \end{align}
\end{proof}

For the remaining lemmas, define the population and empirical covariance
matrices of the feature differences by
\begin{align}
  \Sigma_t^{\sfon}
  &\ldef{}\En_{\substack{x\sim\mu,\\y,y'\overset{\mathrm{iid}}{\sim}\pit(\cdot\mid{}x)}}
  \sbr*{\rbr*{\phi(x,y)-\phi(x,y')}\rbr*{\phi(x,y)-\phi(x,y')}^\top},
  \label{eq:dpo-population-covariance}
  \\
  \wh\Sigma_t^{\sfon}
  &\ldef{}\frac1N\sum_{i=1}^N
  \rbr*{\phi(x_i,y_i)-\phi(x_i,y_i')}
  \rbr*{\phi(x_i,y_i)-\phi(x_i,y_i')}^\top,
  \label{eq:dpo-covariances}
\end{align}
where the batch is sampled as in \cref{alg:dpo}.

\begin{lemma}
  \label{lem:dpo-convexity}
  Suppose that \cref{asm:softmax-linear} holds.
  For every $\theta\in\bbR^d$, the population and empirical losses satisfy
  \begin{align}
    \Ltrebelon(\pi_\theta)
    &=\rbr*{\frac{\theta-\thetat}{\eta}-\thetastar}^\top
    \Sigma_t^{\sfon}
    \rbr*{\frac{\theta-\thetat}{\eta}-\thetastar},
    \label{eq:dpo-population-quadratic}
    \\
    \hLtrebelon(\pi_\theta)
    &=\rbr*{\frac{\theta-\thetat}{\eta}-\thetastar}^\top
    \wh\Sigma_t^{\sfon}
    \rbr*{\frac{\theta-\thetat}{\eta}-\thetastar}.
    \label{eq:dpo-quadratic}
  \end{align}
  Both losses are convex as functions of $\theta$,
  and $\piplust=\pi_{\thetat+\eta\thetastar}$ minimizes both with zero loss.
  The empirical claims hold for every realization of the batch.
\end{lemma}

\begin{proof}[\pfref{lem:dpo-convexity}]
  For every prompt $x$ and pair of responses $(y,y')$,
  \cref{asm:softmax-linear} gives
  \begin{align}
    &\frac1\eta\log\frac{\pi_\theta(y\mid{}x)}{\pit(y\mid{}x)}
    -\frac1\eta\log\frac{\pi_\theta(y'\mid{}x)}{\pit(y'\mid{}x)}
    -\rbr*{r(x,y)-r(x,y')}
    \\
    &\qquad=\rbr*{\phi(x,y)-\phi(x,y')}^\top
      \rbr*{\frac{\theta-\thetat}{\eta}-\thetastar}.
    \label{eq:proof:dpo-pairwise-linear-error}
  \end{align}
  Squaring \cref{eq:proof:dpo-pairwise-linear-error} and taking the population
  expectation gives
  \begin{align}
    \Ltrebelon(\pi_\theta)
    &=\En_{\substack{x\sim\mu,\\y,y'\overset{\mathrm{iid}}{\sim}\pit(\cdot\mid{}x)}}
      \sbr*{\rbr*{\rbr*{\phi(x,y)-\phi(x,y')}^\top
        \rbr*{\frac{\theta-\thetat}{\eta}-\thetastar}}^2}
    \\
    &=\rbr*{\frac{\theta-\thetat}{\eta}-\thetastar}^\top
      \En_{\substack{x\sim\mu,\\y,y'\overset{\mathrm{iid}}{\sim}\pit(\cdot\mid{}x)}}
      \sbr*{\rbr*{\phi(x,y)-\phi(x,y')}\rbr*{\phi(x,y)-\phi(x,y')}^\top}
      \rbr*{\frac{\theta-\thetat}{\eta}-\thetastar}
    \\
    &=\rbr*{\frac{\theta-\thetat}{\eta}-\thetastar}^\top
      \Sigma_t^{\sfon}\rbr*{\frac{\theta-\thetat}{\eta}-\thetastar},
  \end{align}
  where the last equality uses the definition of $\Sigma_t^{\sfon}$ in
  \cref{eq:dpo-population-covariance}.
  Averaging over the batch instead gives
  \begin{align}
    \hLtrebelon(\pi_\theta)
    &=\frac1N\sum_{i=1}^N
      \rbr*{\rbr*{\phi(x_i,y_i)-\phi(x_i,y_i')}^\top
        \rbr*{\frac{\theta-\thetat}{\eta}-\thetastar}}^2
    \\
    &=\rbr*{\frac{\theta-\thetat}{\eta}-\thetastar}^\top
      \wh\Sigma_t^{\sfon}\rbr*{\frac{\theta-\thetat}{\eta}-\thetastar},
  \end{align}
  by the definition of $\wh\Sigma_t^{\sfon}$ in \cref{eq:dpo-covariances}.
  Since $\Sigma_t^{\sfon}\succeq0$ and $\wh\Sigma_t^{\sfon}\succeq0$,
  both losses are convex in $\theta$.
  The exact update $\piplust=\pi_{\thetat+\eta\thetastar}$ from
  \cref{lem:exact-apmd-closed-form} has zero loss under both objectives.
\end{proof}

\begin{lemma}
  \label{lem:dpo-smoothness}
  Suppose that \cref{asm:softmax-linear} holds.
  The population loss $\Ltrebelon(\pi_\theta)$ and empirical loss
  $\hLtrebelon(\pi_\theta)$ are $2\Dfeat^2/\eta^2$-smooth as functions of
  $\theta\in\bbR^d$.
  This holds for every $\hLtrebelon$, regardless of the batch size or samples.
\end{lemma}

\begin{proof}[\pfref{lem:dpo-smoothness}]
  Differentiating the quadratic formulas in \cref{lem:dpo-convexity} twice gives
  \begin{align}
    \nabla^2\Ltrebelon(\pi_\theta)
    &=\frac2{\eta^2}\Sigma_t^{\sfon},
    \\
    \nabla^2\hLtrebelon(\pi_\theta)
    &=\frac2{\eta^2}\wh\Sigma_t^{\sfon}.
    \label{eq:dpo-hessians}
  \end{align}
  To bound their operator norms, for every $v\in\bbR^d$,
  Cauchy--Schwarz and the definition of $\Dfeat$ give
  \begin{align}
    0\le v^\top\Sigma_t^{\sfon}v
    &=\En_{\substack{x\sim\mu,\\y,y'\overset{\mathrm{iid}}{\sim}\pit(\cdot\mid{}x)}}
      \sbr*{\rbr*{v^\top\rbr*{\phi(x,y)-\phi(x,y')}}^2}
    \le\Dfeat^2\nbr*{v}_2^2,
    \label{eq:dpo-population-covariance-bound}
    \\
    0\le v^\top\wh\Sigma_t^{\sfon}v
    &=\frac1N\sum_{i=1}^N
    \rbr*{v^\top\rbr*{\phi(x_i,y_i)-\phi(x_i,y_i')}}^2
    \le\Dfeat^2\nbr*{v}_2^2.
    \label{eq:dpo-covariance-bounds}
  \end{align}
  Hence both Hessians in \cref{eq:dpo-hessians} have operator norm at most
  $2\Dfeat^2/\eta^2$, proving the stated smoothness bound.
\end{proof}

\begin{lemma}
  \label{lem:dpo-population-transfer}
  Suppose that \cref{asm:softmax-linear} holds.
  Fix $\delta\in(0,1)$ and draw a batch as in \cref{alg:dpo}, with
  $N\ge\frac{28}{3}\log\frac{2d}{\delta}$ prompts and two responses per prompt.
  With probability at least $1-\delta$ over the batch, simultaneously for every
  $\theta\in\bbR^d$,
  \begin{align}
    \Ltrebelon(\pi_\theta)
    \le2\hLtrebelon(\pi_\theta)
    +\frac{28\Dfeat^2}{3\eta^2N}\log\frac{2d}{\delta}
    \nbr*{\theta-\thetat-\eta\thetastar}_2^2.
    \label{eq:dpo-population-transfer}
  \end{align}
\end{lemma}

\begin{proof}[\pfref{lem:dpo-population-transfer}]
  By \cref{lem:dpo-convexity}, both losses are quadratic forms in
  $(\theta-\thetat)/\eta-\thetastar$.
  We first prove that, with probability at least $1-\delta$,
  \begin{align}
    \Sigma_t^{\sfon}\preceq2\wh\Sigma_t^{\sfon}+\lambda I,
    \qquad
    \lambda\ldef{}\frac{28\Dfeat^2}{3N}\log\frac{2d}{\delta}.
    \label{eq:dpo-transfer-matrix}
  \end{align}
  We will obtain \cref{eq:dpo-transfer-matrix} by bounding the operator norm of
  \begin{align}
    \rbr*{\Sigma_t^{\sfon}+\lambda I}^{-1/2}
    \rbr*{\wh\Sigma_t^{\sfon}-\Sigma_t^{\sfon}}
    \rbr*{\Sigma_t^{\sfon}+\lambda I}^{-1/2}
    =\frac1N\sum_{i=1}^N\rbr*{Y_i-\En[Y_i]},
    \label{eq:proof:dpo-normalized-covariance-difference}
  \end{align}
  where
  \begin{align}
    Y_i\ldef{}\rbr*{\Sigma_t^{\sfon}+\lambda I}^{-1/2}
    \rbr*{\phi(x_i,y_i)-\phi(x_i,y_i')}
    \rbr*{\phi(x_i,y_i)-\phi(x_i,y_i')}^\top
    \rbr*{\Sigma_t^{\sfon}+\lambda I}^{-1/2}.
  \end{align}
  To apply matrix Bernstein, we bound the norm and variance of $Y_i-\En[Y_i]$.
  The definition of $\Dfeat$ and
  $(\Sigma_t^{\sfon}+\lambda I)^{-1}\preceq\lambda^{-1}I$ give
  \begin{align}
    0\preceq Y_i&\preceq\frac{\Dfeat^2}{\lambda}I,
    \\
    0\preceq\En[Y_i]
    &=\rbr*{\Sigma_t^{\sfon}+\lambda I}^{-1/2}
    \Sigma_t^{\sfon}
    \rbr*{\Sigma_t^{\sfon}+\lambda I}^{-1/2}
    \preceq I.
  \end{align}
  Since the sample size condition ensures $0<\lambda\le\Dfeat^2$, it follows that
  \begin{align}
    \nbr*{Y_i-\En[Y_i]}_{\op}&\le\frac{\Dfeat^2}{\lambda},
    \\
    \En\sbr*{\rbr*{Y_i-\En[Y_i]}^2}
    &=\En[Y_i^2]-\rbr*{\En[Y_i]}^2
    \preceq\En[Y_i^2]
    \preceq\frac{\Dfeat^2}{\lambda}\En[Y_i]
    \preceq\frac{\Dfeat^2}{\lambda}I.
  \end{align}
  The matrices $Y_i-\En[Y_i]$ are independent and have mean zero.
  Matrix Bernstein \citep[Theorem~6.1.1]{tropp2015matrix} therefore gives
  \begin{align}
    \Pr\rbr*{\nbr*{\frac1N\sum_{i=1}^N\rbr*{Y_i-\En[Y_i]}}_{\op}\ge\frac12}
    &\le2d\exp\rbr*{-\frac{N/4}{2\Dfeat^2/\lambda+\Dfeat^2/(3\lambda)}}
    \\
    &=2d\exp\rbr*{-\frac{3N\lambda}{28\Dfeat^2}}
    =\delta.
  \end{align}
  Thus, with probability at least $1-\delta$,
  $\nbr*{\frac1N\sum_{i=1}^N\rbr*{Y_i-\En[Y_i]}}_{\op}<\tfrac12$.
  On this event, \cref{eq:proof:dpo-normalized-covariance-difference} gives
  \begin{align}
    \rbr*{\Sigma_t^{\sfon}+\lambda I}^{-1/2}
    \rbr*{\wh\Sigma_t^{\sfon}-\Sigma_t^{\sfon}}
    \rbr*{\Sigma_t^{\sfon}+\lambda I}^{-1/2}
    \succeq-\frac12I.
    \label{eq:proof:dpo-normalized-covariance-lower}
  \end{align}
  Multiplying both sides of \cref{eq:proof:dpo-normalized-covariance-lower}
  on the left and right by
  $(\Sigma_t^{\sfon}+\lambda I)^{1/2}$ gives
  \begin{align}
    \wh\Sigma_t^{\sfon}-\Sigma_t^{\sfon}
    \succeq-\frac12\rbr*{\Sigma_t^{\sfon}+\lambda I},
  \end{align}
  which rearranges to \cref{eq:dpo-transfer-matrix}.
  Using \cref{eq:dpo-transfer-matrix} in
  \cref{eq:dpo-population-quadratic,eq:dpo-quadratic} gives
  \begin{align}
    \Ltrebelon(\pi_\theta)
    &\le2\hLtrebelon(\pi_\theta)
      +\frac\lambda{\eta^2}\nbr*{\theta-\thetat-\eta\thetastar}_2^2
    \\
    &=2\hLtrebelon(\pi_\theta)
      +\frac{28\Dfeat^2}{3\eta^2N}\log\frac{2d}{\delta}
        \nbr*{\theta-\thetat-\eta\thetastar}_2^2.
  \end{align}
  The event depends only on the batch, so this bound holds simultaneously
  for every $\theta\in\bbR^d$.
\end{proof}

\subsection{Inner Loop Convergence}
\label{app:rebel-inner-loop}

We next prove the convergence of the inner optimization loop in \cref{alg:dpo}.

\begin{theorem}[Inner loop convergence of \rebelon]
  \label{thm:dpo-inner}
  \label{lem:dpo-empirical-convergence}
  Fix an outer iteration $t$, any $\eta>0$, $\einsq\in(0,\Dstar^2]$, and $\delta\in(0,1)$.
  Run the inner loop of \cref{alg:dpo} with
  \begin{align}
    N&\ge\frac{28\Dstar^2}{3\einsq}\log\frac{2d}{\delta},
    &K&\ge\frac{2\Dstar^2}{\einsq},
    &\alpha&=\frac{\eta^2}{2\Dfeat^2}.
    \label{eq:dpo-inner-conditions}
  \end{align}
  With probability at least $1-\delta$, the returned policy satisfies
  \begin{align}
    \mset(\pitk[K])&\le\einsq,
    &\nbr*{s_t^{\pitk[K]}}_\infty&\le\Dstar.
    \label{eq:dpo-inner-conclusion}
  \end{align}
\end{theorem}

Because every empirical realization of $\hLtrebelon$ is convex and is minimized by $\piplust$ (\cref{lem:dpo-convexity}),
the proof below first proves convergence of gradient descent on the empirical objective.
It then transfers this to the population objective by \cref{lem:dpo-population-transfer}.
Because the empirical covariance $\wh\Sigma_t^{\sfon}$ is the only estimated term in $\hLtrebelon$,
regardless of the parameter $\theta$,
the concentration inequality in \cref{lem:dpo-population-transfer} scales with $\log d$ and not $d$.

\begin{proof}[\pfref{thm:dpo-inner}]
  For every realization of the batch, \cref{lem:dpo-convexity,lem:dpo-smoothness}
  show that $\hLtrebelon(\pi_\theta)$ is convex and
  $2\Dfeat^2/\eta^2$-smooth, with minimum value zero at
  $\theta=\thetat+\eta\thetastar$.
  Applying \cref{thm:population-gd-convergence} from $\thetatk[0]=\thetat$
  with learning rate $\alpha=\eta^2/(2\Dfeat^2)$ gives
  \begin{align}
    \hLtrebelon(\pitk[K])
    &\le\frac{\nbr*{\thetat-(\thetat+\eta\thetastar)}_2^2}{2\alpha K}
    =\frac{\Dstar^2}{K}.
    \label{eq:dpo-empirical-rate}
  \end{align}

  We next bound $\mset(\pitk[K])$ in terms of $\hLtrebelon$ using \cref{lem:dpo-population-transfer}, which holds uniformly over $\theta$ and therefore applies to $\thetatk[K]$.
  The lemma's requirement on $N$ follows from $\einsq\le\Dstar^2$ and theorem conditions in
  \cref{eq:dpo-inner-conditions}.
  Together with \cref{lem:rebel-mse}, this gives, with probability at least $1-\delta$,
  \begin{align}
    \mset(\pitk[K])
    &=\frac12\Ltrebelon(\pitk[K])
    \\
    &\le\hLtrebelon(\pitk[K])
      +\frac{14\Dfeat^2}{3\eta^2N}\log\frac{2d}{\delta}
        \nbr*{\thetatk[K]-\thetat-\eta\thetastar}_2^2.
    \label{eq:proof:dpo-inner-mse-comparison}
  \end{align}
  We condition on this event for the remainder of the proof.

  We now bound $\nbr*{\thetatk[K]-\thetat-\eta\thetastar}_2$ in
  \cref{eq:proof:dpo-inner-mse-comparison}.
  Differentiating \cref{eq:dpo-quadratic} gives
  \begin{align}
    \nabla\hLtrebelon(\pi_\theta)
    =\frac2{\eta^2}\wh\Sigma_t^{\sfon}(\theta-\thetat-\eta\thetastar).
    \label{eq:dpo-gradients}
  \end{align}
  Substituting into the gradient descent update in Line~\ref{line:dpo-update}
  of \cref{alg:dpo}
  with $\alpha=\eta^2/(2\Dfeat^2)$ gives
  \begin{align}
    \thetatkp-\thetat-\eta\thetastar
    =\rbr*{I-\frac{\wh\Sigma_t^{\sfon}}{\Dfeat^2}}
    \rbr*{\thetatk-\thetat-\eta\thetastar}.
    \label{eq:dpo-error-recursion}
  \end{align}
  By \cref{eq:dpo-covariance-bounds} in the proof of \cref{lem:dpo-smoothness},
  $0\preceq\wh\Sigma_t^{\sfon}/\Dfeat^2\preceq I$, so
  $\nbr*{I-\wh\Sigma_t^{\sfon}/\Dfeat^2}_{\op}\le1$.
  Iterating \cref{eq:dpo-error-recursion} from $\thetatk[0]=\thetat$ therefore gives
  \begin{align}
    \nbr*{\thetatk-\thetat-\eta\thetastar}_2
    \le\eta\nbr*{\thetastar}_2=\eta\Bstar,
    \qquad 0\le k\le K.
    \label{eq:dpo-empirical-localization}
  \end{align}
  Substituting \cref{eq:dpo-empirical-rate,eq:dpo-empirical-localization} into
  \cref{eq:proof:dpo-inner-mse-comparison} gives
  \begin{align}
    \mset(\pitk[K])
    \le\frac{\Dstar^2}{K}+\frac{14\Dstar^2}{3N}\log\frac{2d}{\delta}
    \le\einsq,
  \end{align}
  where the last inequality uses the choices of $K$ and $N$ in
  \cref{eq:dpo-inner-conditions}.

  Finally, we bound the residual $s_t^{\pitk}$
  and rewrite it under \cref{asm:softmax-linear} as
  \begin{align}
    s_t^{\pitk}(x,y)
    =\rbr*{\phi(x,y)-\En_{y'\sim\pit(\cdot\mid{}x)}\sbr*{\phi(x,y')}}^\top
    \rbr*{\frac{\thetatk-\thetat}{\eta}-\thetastar}.
    \label{eq:dpo-centered-error}
  \end{align}
  For every prompt $x$ and response $y$, the definition of $\Dfeat$ gives
  \begin{align}
    \nbr*{\phi(x,y)-\En_{y'\sim\pit(\cdot\mid{}x)}\sbr*{\phi(x,y')}}_2
    &\le\En_{y'\sim\pit(\cdot\mid{}x)}
      \sbr*{\nbr*{\phi(x,y)-\phi(x,y')}_2}
    \le\Dfeat.
  \end{align}
  Applying Cauchy--Schwarz to \cref{eq:dpo-centered-error}
  and using the preceding bound together with \cref{eq:dpo-empirical-localization} gives
  \begin{align}
    \nbr*{s_t^{\pitk}}_\infty
    &\le\sup_{x,y}
      \nbr*{\phi(x,y)-\En_{y'\sim\pit(\cdot\mid{}x)}\sbr*{\phi(x,y')}}_2
      \nbr*{\frac{\thetatk-\thetat}{\eta}-\thetastar}_2
    \\
    &\le\Dfeat\Bstar=\Dstar,
    \qquad 0\le k\le K.
    \label{eq:dpo-empirical-range}
  \end{align}

\end{proof}

\subsection{Proof of \cref{cor:rebel-main}}
\label{app:rebel-global}
\label{app:rebel-proof}

\begin{proof}[\pfref{cor:dpo-regret}]
  The parameter choices in \cref{eq:dpo-regret-outer-parameters,eq:dpo-regret-inner-parameters}
  give $0<\einsq\le\veps^2<\Dstar^2$.
  At each outer iteration $t$, condition on the batches sampled in iterations
  $0,\ldots,t-1$.
  The parameter choices in \cref{eq:dpo-regret-inner-parameters,eq:dpo-regret-sample-parameters}
  satisfy the conditions of \cref{thm:dpo-inner} with failure probability $\delta/T$.
  A union bound therefore gives, with probability at least $1-\delta$,
  \begin{align}
    \mset(\pitp)&\le\einsq,
    &\nbr*{s_t^{\pitp}}_\infty&\le\Dstar,
    \qquad \forall\,0\le t<T.
    \label{eq:dpo-regret-inner-event}
  \end{align}
  We condition on the event in \cref{eq:dpo-regret-inner-event} for the remainder of the proof.

  Using \cref{eq:dpo-regret-inner-event} in \cref{thm:mse-accum-regret}
  with $S=\Dstar$, together with
  $\En_{x\sim\mu}[\log(1/\piref(\Ystarx\mid x))]\le\log\Cresp$
  by Jensen's inequality and $\sum_{t=1}^T1/t\le\log(eT)$, gives
  \begin{align}
    \Jstar-J(\pifin)
    &\le\frac{\log\Cresp}{\eta T}
      +\frac{\eta e^{\eta\Dstar}T}{2}\einsq
      +\sqrt{\Cresp\log(eT)\einsq}.
    \label{eq:proof:dpo-regret-suboptimality}
  \end{align}
  Since $e^{\eta\Dstar}=2\Cresp$, substituting $\einsq$ from
  \cref{eq:dpo-regret-inner-parameters} into \cref{eq:proof:dpo-regret-suboptimality} yields
  \begin{align}
    \Jstar-J(\pifin)
    &\le\frac{\log\Cresp+1/12+1/\sqrt{12}}{\eta T}
      \le\frac{\log(2\Cresp)}{\eta T}
      \le\veps,
  \end{align}
  where the last inequality uses the choice of $T$ in \cref{eq:dpo-regret-outer-parameters}.
  On the same event, the coverage bound in \cref{thm:mse-accum-regret} with
  $c=1/2$ and the same parameter choices gives
  \begin{align}
    &\Pr_{x\sim\mu}\rbr*{
      \exists\,0\le t\le T:\;
      \pit(\Ystarx\mid x)<\tfrac12\piref(\Ystarx\mid x)}
    \\
    &\quad\le(2+\sqrt2)\rbr*{
      \frac{\eta^2e^{\eta\Dstar}T\einsq}{4}
      +\frac{\eta T\sqrt{\Cresp\einsq}}2}
    \label{eq:proof:dpo-regret-coverage}
    \\
    &\quad\le(2+\sqrt2)\rbr*{
      \frac{1}{24T\log(eT)}+\frac{1}{4\sqrt{3\log(eT)}}}
    \\
    &\quad\le\frac{2+\sqrt2}{4\sqrt{\log(eT)}}.
  \end{align}
  This proves \cref{eq:dpo-regret-conclusion,eq:dpo-regret-coverage}.

  Finally, \cref{alg:dpo} uses one batch of $2N$ responses for $K$ gradient steps
  at each outer iteration. The choices of $K$ and $N$ in
  \cref{eq:dpo-regret-sample-parameters} therefore give
  \begin{align}
    \Nstep(\veps,\delta)=KT
    &=O\rbr*{\frac{T\Dstar^2}{\einsq}},
    \\
    \Nsample(\veps,\delta)=2NT
    &=O\rbr*{\frac{T\Dstar^2}{\einsq}\log\frac{2dT}{\delta}}.
    \label{eq:proof:dpo-regret-counts}
  \end{align}
  Since $\veps<\Dstar$, the choices in
  \cref{eq:dpo-regret-outer-parameters,eq:dpo-regret-inner-parameters} give
  \begin{align}
    \frac{\Dstar}{\veps}&\le T\le\frac{2\Dstar}{\veps},
    \\
    \frac{\Dstar^2}{\einsq}
    &=12\Cresp\Dstar^2\eta^2T^2\log(eT)
    \le\frac{48\Cresp\Dstar^2\log^2(2\Cresp)\log(eT)}{\veps^2}.
    \label{eq:proof:dpo-regret-parameter-bounds}
  \end{align}
  Substituting \cref{eq:proof:dpo-regret-parameter-bounds} into
  \cref{eq:proof:dpo-regret-counts} yields
  \begin{align}
    KT&=O\rbr*{
      \frac{\Cresp\Dstar^3\log^2(2\Cresp)\log(eT)}{\veps^3}},
    \\
    2NT&=O\rbr*{
      \frac{\Cresp\Dstar^3\log^2(2\Cresp)\log(eT)}{\veps^3}
      \log\frac{4d\Dstar}{\veps\delta}}.
  \end{align}
  Writing these bounds in $\wt O$ notation gives \cref{eq:dpo-regret-complexity}.
\end{proof}

To obtain \cref{cor:rebel-main}, let $B=\Bfeat\Bstar$ and choose
\begin{align}
  \eta=\frac{\log\rbr*{2\Cresp}}{2B},
  \label{eq:dpo-regret-main-parameters}
\end{align}
with the remaining parameters defined as in \cref{cor:dpo-regret} using this value of $\eta$.
Since $\Dstar\le2B$, we have $e^{\eta\Dstar}\le2\Cresp$, while
the definition of $T$ still gives $\eta T\ge\log(2\Cresp)/\veps$.
The same substitutions in \cref{eq:proof:dpo-regret-suboptimality,eq:proof:dpo-regret-coverage}
therefore give the reward and coverage guarantees in
\cref{eq:dpo-regret-conclusion,eq:dpo-regret-coverage}.

Moreover, $\veps\le\Rmax\le B$ implies
$T=\lceil2B/\veps\rceil=O(B/\veps)$ and
$\eta T=O(\log(2\Cresp)/\veps)$.
Using these bounds, $\Dstar\le2B$, and the definition of $\einsq$ in
\cref{eq:dpo-regret-inner-parameters}, the complexities in \cref{eq:proof:dpo-regret-counts} give
\begin{align}
  \Nstep(\veps,\delta)&=\wt O\rbr*{\frac{\Cresp B^3}{\veps^3}},
  &\Nsample(\veps,\delta)&=\wt O\rbr*{
    \frac{\Cresp B^3}{\veps^3}\log\frac{2d}{\delta}
  }.
  \label{eq:dpo-regret-main-complexity}
\end{align}

\section{Discussion: Connection to Natural Policy Gradient}\label{app:npg}
\label{sec:npg}
    Under \cref{asm:softmax-linear}, the objectives of \cref{sec:regression} are closely related to
  natural policy gradient (NPG) \citep{kakade2001naturalpg}.
  Write a candidate policy as $\pi_\theta$ with $\theta=\thetat+\eta g$,
  and let $\bar\phi_t(x)\ldef{}\En_{y\sim\pit(\cdot\mid{}x)}\sbr{\phi(x,y)}$.
  Then
  \begin{align}
    \frac1\eta\log\frac{\pi_\theta(y\mid{}x)}{\pit(y\mid{}x)}
    =\phi(x,y)^\top g
    -\frac1\eta\log\En_{y'\sim\pit(\cdot\mid{}x)}\sbr*{e^{\eta\phi(x,y')^\top g}},
  \end{align}
  so the idealized objective \cref{eq:squared-loss-example}
  is a linear regression of $r-\btstar$ onto the features,
  plus a nonlinear term from the policy's normalizer.
  Both this term and $\btstar(x)$ are log-mean-exponentials,
  and replacing each by its first-order approximation in $\eta$
  ($\bar\phi_t(x)^\top g$ and $r(x,\pit)$, respectively)
  turns the objective into centered linear regression,
  \begin{align}
    \min_g\;\En_{\pit}\sbr*{
      \rbr*{(\phi(x,y)-\bar\phi_t(x))^\top g-(r(x,y)-r(x,\pit))}^2}.
    \label{eq:npg-regression}
  \end{align}
  Since centering removes the normalizer, the objective in \cref{eq:npg-regression}
  is exactly $\mset(\pi_\theta)$, and its minimizers satisfy $F_tg=\nabla J(\pit)$,
  where $F_t\ldef{}\En_{\pit}\sbr*{\rbr*{\phi-\bar\phi_t}\rbr*{\phi-\bar\phi_t}^\top}$
  is the Fisher information matrix.
  Reward realizability gives $\nabla J(\pit)=F_t\thetastar$, so $g=\thetastar$
  is a minimizer and yields the \npg update \citep{kakade2001naturalpg,agarwal2020theory},
  \begin{align}
    \thetatp=\thetat+\eta\thetastar.
  \end{align}
  Exact minimization of the population \rebelon objective therefore implements an \npg step.
  The mean-baseline objective of \kimipg adds a nonlinear correction; we emphasize however that this connection is only precise under \cref{asm:softmax-linear} and the softmax linear parameterization.\loose

\clearpage

\end{document}